\providecommand{\SDBVersion}{preprint}
\documentclass[11pt]{article}
\expandafter\usepackage\expandafter[\SDBVersion]{acl}

\makeatletter
\@ifpackageloaded{lineno}{%
  \IfFormatAtLeastTF{2025-06-01}{%
    \RemoveFromHook{build/column/before}[lineno]%
    \AddToHook{build/column/after}[aclfix/lineno]{%
      \setbox\@outputbox\vbox{%
        \boxmaxdepth\@maxdepth
        \protected@write\@auxout{}{%
          \string\def\string\@LN@column{\if@firstcolumn1\else2\fi}}%
        \box\@outputbox}}%
  }{}%
}{}
\makeatother

\usepackage{times}
\usepackage{latexsym}
\usepackage[T1]{fontenc}
\usepackage[utf8]{inputenc}
\usepackage{microtype}
\usepackage{inconsolata}
\usepackage{graphicx}
\usepackage{upquote}

\usepackage{amsmath}
\usepackage{amssymb}
\usepackage{booktabs}
\usepackage{array}
\usepackage{enumitem}
\usepackage{placeins}

\newcommand{\LunaLabel}{\mbox{GPT-5.6-Luna}}
\newcommand{\TerraLabel}{\mbox{GPT-5.6-Terra}}
\newcommand{\AstraLabel}{\mbox{GPT-6-Astra}}

\newcommand{\ReleasesPerScenario}{60}                                   % data/lite/v1/manifest.json:episodes.*.steps
\newcommand{\ReleaseIntervalSec}{2}                                     % data/lite/v1/*.json:tick_seconds
\newcommand{\NumStates}{480}                                            % data/lite/v1/manifest.json:episodes.*.steps (sum)
\newcommand{\DatasetHashShort}{fdfdd55d}                                % data/lite/v1/manifest.json:dataset_hash (first 8 hex digits)
\newcommand{\QuestionsMin}{6}                                           % data/lite/v1/*.json:questions (count, minimum over scenarios)
\newcommand{\QuestionsMax}{7}                                           % data/lite/v1/*.json:questions (count, maximum over scenarios)
\newcommand{\AnswersPerPass}{3120}                                      % data/lite/v1/*.json:questions x steps (summed)
\newcommand{\RouteOptionsMin}{4}                                        % data/lite/v1/*.json:option_semantics.route (count, minimum)
\newcommand{\RouteOptionsMax}{7}                                        % data/lite/v1/*.json:option_semantics.route (count, maximum)
\newcommand{\OptionsMin}{2}                                             % data/lite/v1/*.json:option_semantics.* (options per question, minimum)
\newcommand{\OptionsMax}{12}                                            % data/lite/v1/*.json:option_semantics.* (options per question, maximum)
\newcommand{\TransitionsTotal}{177}                                     % data/lite/v1/manifest.json:episodes.*.decision_transitions (sum)
\newcommand{\TransitionsMin}{20}                                        % data/lite/v1/manifest.json:episodes.*.decision_transitions (minimum)
\newcommand{\ConsumedAnswersMin}{2}                                     % data/lite/v1/*.json:steps[].gold composed by decision_spec (answers, minimum)
\newcommand{\ConsumedAnswersMax}{5}                                     % data/lite/v1/*.json:steps[].gold composed by decision_spec (answers, maximum)
\newcommand{\ReferenceAgreementStates}{480}                             % src/streamdecisionbench/lite/tasks/{debugging,assembly,support,presenter}.py:reference(state) == data/lite/v1/*.json:steps[].gold (states agreeing, of 480)
\newcommand{\DebugATitle}{Refund watch loop}                            % data/lite/v1/lite_debugging_a.json:title
\newcommand{\DebugAQuestions}{7}                                        % data/lite/v1/lite_debugging_a.json:questions (count)
\newcommand{\DebugARouteOptions}{6}                                     % data/lite/v1/lite_debugging_a.json:option_semantics.route (count)
\newcommand{\DebugAOptionsMin}{2}                                       % data/lite/v1/lite_debugging_a.json:option_semantics (options per question, minimum)
\newcommand{\DebugAOptionsMax}{6}                                       % data/lite/v1/lite_debugging_a.json:option_semantics (options per question, maximum)
\newcommand{\DebugATransitions}{20}                                     % data/lite/v1/manifest.json:episodes[0].decision_transitions
\newcommand{\DebugBTitle}{Settlement integration debug session}         % data/lite/v1/lite_debugging_b.json:title
\newcommand{\DebugBQuestions}{7}                                        % data/lite/v1/lite_debugging_b.json:questions (count)
\newcommand{\DebugBRouteOptions}{6}                                     % data/lite/v1/lite_debugging_b.json:option_semantics.route (count)
\newcommand{\DebugBOptionsMin}{2}                                       % data/lite/v1/lite_debugging_b.json:option_semantics (options per question, minimum)
\newcommand{\DebugBOptionsMax}{6}                                       % data/lite/v1/lite_debugging_b.json:option_semantics (options per question, maximum)
\newcommand{\DebugBTransitions}{21}                                     % data/lite/v1/manifest.json:episodes[1].decision_transitions
\newcommand{\AssemblyATitle}{Gearbox cover: mismatched part, joint rework and quality hold} % data/lite/v1/lite_assembly_a.json:title
\newcommand{\AssemblyAQuestions}{6}                                     % data/lite/v1/lite_assembly_a.json:questions (count)
\newcommand{\AssemblyARouteOptions}{7}                                  % data/lite/v1/lite_assembly_a.json:option_semantics.route (count)
\newcommand{\AssemblyAOptionsMin}{5}                                    % data/lite/v1/lite_assembly_a.json:option_semantics (options per question, minimum)
\newcommand{\AssemblyAOptionsMax}{7}                                    % data/lite/v1/lite_assembly_a.json:option_semantics (options per question, maximum)
\newcommand{\AssemblyATransitions}{21}                                  % data/lite/v1/manifest.json:episodes[2].decision_transitions
\newcommand{\AssemblyBTitle}{Cable module: orientation, crimp repair and stale electrical checks} % data/lite/v1/lite_assembly_b.json:title
\newcommand{\AssemblyBQuestions}{6}                                     % data/lite/v1/lite_assembly_b.json:questions (count)
\newcommand{\AssemblyBRouteOptions}{7}                                  % data/lite/v1/lite_assembly_b.json:option_semantics.route (count)
\newcommand{\AssemblyBOptionsMin}{5}                                    % data/lite/v1/lite_assembly_b.json:option_semantics (options per question, minimum)
\newcommand{\AssemblyBOptionsMax}{9}                                    % data/lite/v1/lite_assembly_b.json:option_semantics (options per question, maximum)
\newcommand{\AssemblyBTransitions}{23}                                  % data/lite/v1/manifest.json:episodes[3].decision_transitions
\newcommand{\SupportATitle}{Payment, reconnection and recorder control} % data/lite/v1/lite_support_a.json:title
\newcommand{\SupportAQuestions}{7}                                      % data/lite/v1/lite_support_a.json:questions (count)
\newcommand{\SupportARouteOptions}{4}                                   % data/lite/v1/lite_support_a.json:option_semantics.route (count)
\newcommand{\SupportAOptionsMin}{3}                                     % data/lite/v1/lite_support_a.json:option_semantics (options per question, minimum)
\newcommand{\SupportAOptionsMax}{7}                                     % data/lite/v1/lite_support_a.json:option_semantics (options per question, maximum)
\newcommand{\SupportATransitions}{22}                                   % data/lite/v1/manifest.json:episodes[4].decision_transitions
\newcommand{\SupportBTitle}{Intercom delivery, cancellation and repair guidance} % data/lite/v1/lite_support_b.json:title
\newcommand{\SupportBQuestions}{7}                                      % data/lite/v1/lite_support_b.json:questions (count)
\newcommand{\SupportBRouteOptions}{6}                                   % data/lite/v1/lite_support_b.json:option_semantics.route (count)
\newcommand{\SupportBOptionsMin}{3}                                     % data/lite/v1/lite_support_b.json:option_semantics (options per question, minimum)
\newcommand{\SupportBOptionsMax}{8}                                     % data/lite/v1/lite_support_b.json:option_semantics (options per question, maximum)
\newcommand{\SupportBTransitions}{24}                                   % data/lite/v1/manifest.json:episodes[5].decision_transitions
\newcommand{\PresenterATitle}{Conference talk with a demo clip and floor questions} % data/lite/v1/lite_presenter_a.json:title
\newcommand{\PresenterAQuestions}{6}                                    % data/lite/v1/lite_presenter_a.json:questions (count)
\newcommand{\PresenterARouteOptions}{4}                                 % data/lite/v1/lite_presenter_a.json:option_semantics.mode (count)
\newcommand{\PresenterAOptionsMin}{3}                                   % data/lite/v1/lite_presenter_a.json:option_semantics (options per question, minimum)
\newcommand{\PresenterAOptionsMax}{12}                                  % data/lite/v1/lite_presenter_a.json:option_semantics (options per question, maximum)
\newcommand{\PresenterATransitions}{24}                                 % data/lite/v1/manifest.json:episodes[6].decision_transitions
\newcommand{\PresenterBTitle}{Flood-barrier briefing with a clip and a mid-talk question} % data/lite/v1/lite_presenter_b.json:title
\newcommand{\PresenterBQuestions}{6}                                    % data/lite/v1/lite_presenter_b.json:questions (count)
\newcommand{\PresenterBRouteOptions}{4}                                 % data/lite/v1/lite_presenter_b.json:option_semantics.mode (count)
\newcommand{\PresenterBOptionsMin}{3}                                   % data/lite/v1/lite_presenter_b.json:option_semantics (options per question, minimum)
\newcommand{\PresenterBOptionsMax}{12}                                  % data/lite/v1/lite_presenter_b.json:option_semantics (options per question, maximum)
\newcommand{\PresenterBTransitions}{22}                                 % data/lite/v1/manifest.json:episodes[7].decision_transitions
\newcommand{\BaselineBestConstant}{14.8}                                % data/lite/v1/*.json:steps[].gold composed by decision_spec (share of the most frequent decision, mean over scenarios)
\newcommand{\BaselineBestConstantMin}{10.0}                             % data/lite/v1/*.json:steps[].gold (most frequent decision share, minimum over scenarios)
\newcommand{\BaselineBestConstantMax}{25.0}                             % data/lite/v1/*.json:steps[].gold (most frequent decision share, maximum over scenarios)
\newcommand{\BaselineUniformRoute}{19.2}                                % data/lite/v1/*.json:option_semantics (uniform route choice, mean over scenarios)
\newcommand{\BaselineUniform}{0.59}                                     % data/lite/v1/*.json:option_semantics, steps[].gold (uniform answers to the consumed questions, mean over scenarios)
\newcommand{\NumFamiliesWord}{four}                                     % \NumFamilies spelled out
\newcommand{\ScenariosPerFamilyWord}{two}                               % \ScenariosPerFamily spelled out
\newcommand{\NumScenariosWord}{eight}                                   % \NumScenarios spelled out
\newcommand{\ExampleStateStep}{33}                                      % data/lite/v1/lite_support_a.json:steps[33].state.clock.now (time steps)
\newcommand{\ExampleStateSec}{66}                                       % data/lite/v1/lite_support_a.json:steps[33].state.clock.now x tick_seconds
\newcommand{\ExampleHoldSinceStep}{29}                                  % data/lite/v1/lite_support_a.json:steps[33].state.telephony.since (time steps; also the agent's claim, transcript[].at)
\newcommand{\SupportAHoldThresholdSteps}{4}                             % data/lite/v1/lite_support_a.json:steps[33].state.prepared.hold_check_ticks
\newcommand{\AssemblyRouteOptions}{7}                                   % data/lite/v1/{lite_assembly_a,lite_assembly_b}.json (RouteOptions, equal in both scenarios)
\newcommand{\LunaUntimed}{89.9}                                         % equal mean across passes: docs/lite/results/four-family/gpt-5.6-luna-low/analysis.json:scores.overall.untimed_decision_accuracy; docs/lite/results/hosted-api-repeats-20261003/pass2/gpt-5.6-luna-low/analysis.json:scores.overall.untimed_decision_accuracy; docs/lite/results/hosted-api-repeats-20261003/pass3/gpt-5.6-luna-low/analysis.json:scores.overall.untimed_decision_accuracy
\newcommand{\LunaInForce}{47.9}                                         % equal mean across passes: docs/lite/results/four-family/gpt-5.6-luna-low/analysis.json:scores.overall.time_accuracy; docs/lite/results/hosted-api-repeats-20261003/pass2/gpt-5.6-luna-low/analysis.json:scores.overall.time_accuracy; docs/lite/results/hosted-api-repeats-20261003/pass3/gpt-5.6-luna-low/analysis.json:scores.overall.time_accuracy
\newcommand{\LunaInForceSeg}{39.1}                                      % equal mean across passes: docs/lite/results/four-family/gpt-5.6-luna-low/analysis.json:scores.overall.segment_time_accuracy; docs/lite/results/hosted-api-repeats-20261003/pass2/gpt-5.6-luna-low/analysis.json:scores.overall.segment_time_accuracy; docs/lite/results/hosted-api-repeats-20261003/pass3/gpt-5.6-luna-low/analysis.json:scores.overall.segment_time_accuracy
\newcommand{\LunaNetRemovedSeg}{49.9}                                   % equal mean across passes: docs/lite/results/four-family/gpt-5.6-luna-low/analysis.json:network_adjustment.scores.estimate.overall.segment_time_accuracy; docs/lite/results/hosted-api-repeats-20261003/pass2/gpt-5.6-luna-low/analysis.json:network_adjustment.scores.estimate.overall.segment_time_accuracy; docs/lite/results/hosted-api-repeats-20261003/pass3/gpt-5.6-luna-low/analysis.json:network_adjustment.scores.estimate.overall.segment_time_accuracy
\newcommand{\LunaNetSec}{0.69}                                          % equal mean across passes: docs/lite/results/four-family/gpt-5.6-luna-low/analysis.json:network_adjustment.network_s.estimate; docs/lite/results/hosted-api-repeats-20261003/pass2/gpt-5.6-luna-low/analysis.json:network_adjustment.network_s.estimate; docs/lite/results/hosted-api-repeats-20261003/pass3/gpt-5.6-luna-low/analysis.json:network_adjustment.network_s.estimate
\newcommand{\LunaNetRemovedSegLow}{46.8}                                % equal mean across passes: docs/lite/results/four-family/gpt-5.6-luna-low/analysis.json:network_adjustment.scores.low.overall.segment_time_accuracy; docs/lite/results/hosted-api-repeats-20261003/pass2/gpt-5.6-luna-low/analysis.json:network_adjustment.scores.low.overall.segment_time_accuracy; docs/lite/results/hosted-api-repeats-20261003/pass3/gpt-5.6-luna-low/analysis.json:network_adjustment.scores.low.overall.segment_time_accuracy
\newcommand{\LunaNetSecLow}{0.50}                                       % equal mean across passes: docs/lite/results/four-family/gpt-5.6-luna-low/analysis.json:network_adjustment.network_s.low; docs/lite/results/hosted-api-repeats-20261003/pass2/gpt-5.6-luna-low/analysis.json:network_adjustment.network_s.low; docs/lite/results/hosted-api-repeats-20261003/pass3/gpt-5.6-luna-low/analysis.json:network_adjustment.network_s.low
\newcommand{\LunaNetRemovedSegHigh}{51.6}                               % equal mean across passes: docs/lite/results/four-family/gpt-5.6-luna-low/analysis.json:network_adjustment.scores.high.overall.segment_time_accuracy; docs/lite/results/hosted-api-repeats-20261003/pass2/gpt-5.6-luna-low/analysis.json:network_adjustment.scores.high.overall.segment_time_accuracy; docs/lite/results/hosted-api-repeats-20261003/pass3/gpt-5.6-luna-low/analysis.json:network_adjustment.scores.high.overall.segment_time_accuracy
\newcommand{\LunaNetSecHigh}{0.79}                                      % equal mean across passes: docs/lite/results/four-family/gpt-5.6-luna-low/analysis.json:network_adjustment.network_s.high; docs/lite/results/hosted-api-repeats-20261003/pass2/gpt-5.6-luna-low/analysis.json:network_adjustment.network_s.high; docs/lite/results/hosted-api-repeats-20261003/pass3/gpt-5.6-luna-low/analysis.json:network_adjustment.network_s.high
\newcommand{\LunaPrefillMsPerKTok}{33.1}                                % equal mean across passes: docs/lite/results/four-family/gpt-5.6-luna-low/analysis.json:network_adjustment.prefill_s_per_1k_input_tokens (x 1000); docs/lite/results/hosted-api-repeats-20261003/pass2/gpt-5.6-luna-low/analysis.json:network_adjustment.prefill_s_per_1k_input_tokens (x 1000); docs/lite/results/hosted-api-repeats-20261003/pass3/gpt-5.6-luna-low/analysis.json:network_adjustment.prefill_s_per_1k_input_tokens (x 1000)
\newcommand{\LunaDecodeMsPerTok}{8.6}                                   % equal mean across passes: docs/lite/results/four-family/gpt-5.6-luna-low/analysis.json:network_adjustment.decode_s_per_output_token (x 1000); docs/lite/results/hosted-api-repeats-20261003/pass2/gpt-5.6-luna-low/analysis.json:network_adjustment.decode_s_per_output_token (x 1000); docs/lite/results/hosted-api-repeats-20261003/pass3/gpt-5.6-luna-low/analysis.json:network_adjustment.decode_s_per_output_token (x 1000)
\newcommand{\LunaLatencyFloor}{1.43}                                    % minimum across passes: docs/lite/results/four-family/gpt-5.6-luna-low/analysis.json:network_adjustment.latency_floor_s; docs/lite/results/hosted-api-repeats-20261003/pass2/gpt-5.6-luna-low/analysis.json:network_adjustment.latency_floor_s; docs/lite/results/hosted-api-repeats-20261003/pass3/gpt-5.6-luna-low/analysis.json:network_adjustment.latency_floor_s
\newcommand{\LunaLatencyMedian}{2.44}                                   % equal mean across passes: docs/lite/results/four-family/gpt-5.6-luna-low/analysis.json:latency_s.p50; docs/lite/results/hosted-api-repeats-20261003/pass2/gpt-5.6-luna-low/analysis.json:latency_s.p50; docs/lite/results/hosted-api-repeats-20261003/pass3/gpt-5.6-luna-low/analysis.json:latency_s.p50
\newcommand{\LunaLatencyPNinetyFive}{3.98}                              % equal mean across passes: docs/lite/results/four-family/gpt-5.6-luna-low/analysis.json:latency_s.p95; docs/lite/results/hosted-api-repeats-20261003/pass2/gpt-5.6-luna-low/analysis.json:latency_s.p95; docs/lite/results/hosted-api-repeats-20261003/pass3/gpt-5.6-luna-low/analysis.json:latency_s.p95
\newcommand{\LunaLatencyMax}{7.50}                                      % maximum across passes: docs/lite/results/four-family/gpt-5.6-luna-low/analysis.json:latency_s.max; docs/lite/results/hosted-api-repeats-20261003/pass2/gpt-5.6-luna-low/analysis.json:latency_s.max; docs/lite/results/hosted-api-repeats-20261003/pass3/gpt-5.6-luna-low/analysis.json:latency_s.max
\newcommand{\LunaLatencyMean}{2.63}                                     % equal mean across passes: runs/lite-v1-gpt-5.6-luna-low-four-family-retry-v1/events.jsonl:response.attempts[-1].completed_s - started_s (mean over 480 responses); runs/hosted-api-repeats-20261003/pass2/gpt-5.6-luna-low/events.jsonl:response.attempts[-1].completed_s - started_s (mean over 480 responses); runs/hosted-api-repeats-20261003/pass3/gpt-5.6-luna-low/events.jsonl:response.attempts[-1].completed_s - started_s (mean over 480 responses)
\newcommand{\LunaInputTokMedian}{3034}                                  % equal mean across passes: runs/lite-v1-gpt-5.6-luna-low-four-family-retry-v1/events.jsonl:response.usage.input_tokens (median over 480 responses, half-up); runs/hosted-api-repeats-20261003/pass2/gpt-5.6-luna-low/events.jsonl:response.usage.input_tokens (median over 480 responses, half-up); runs/hosted-api-repeats-20261003/pass3/gpt-5.6-luna-low/events.jsonl:response.usage.input_tokens (median over 480 responses, half-up)
\newcommand{\LunaInputTokTotal}{4{,}429{,}467}                          % sum across passes: docs/lite/results/four-family/gpt-5.6-luna-low/analysis.json:usage.input_tokens; docs/lite/results/hosted-api-repeats-20261003/pass2/gpt-5.6-luna-low/analysis.json:usage.input_tokens; docs/lite/results/hosted-api-repeats-20261003/pass3/gpt-5.6-luna-low/analysis.json:usage.input_tokens
\newcommand{\LunaOutputTokMedian}{149}                                  % equal mean across passes: runs/lite-v1-gpt-5.6-luna-low-four-family-retry-v1/events.jsonl:response.usage.output_tokens (median over 480 responses, half-up); runs/hosted-api-repeats-20261003/pass2/gpt-5.6-luna-low/events.jsonl:response.usage.output_tokens (median over 480 responses, half-up); runs/hosted-api-repeats-20261003/pass3/gpt-5.6-luna-low/events.jsonl:response.usage.output_tokens (median over 480 responses, half-up)
\newcommand{\LunaOutputTokTotal}{239{,}795}                             % sum across passes: docs/lite/results/four-family/gpt-5.6-luna-low/analysis.json:usage.output_tokens; docs/lite/results/hosted-api-repeats-20261003/pass2/gpt-5.6-luna-low/analysis.json:usage.output_tokens; docs/lite/results/hosted-api-repeats-20261003/pass3/gpt-5.6-luna-low/analysis.json:usage.output_tokens
\newcommand{\LunaReasoningTokMedian}{101}                               % equal mean across passes: runs/lite-v1-gpt-5.6-luna-low-four-family-retry-v1/events.jsonl:response.usage.reasoning_tokens (median over 480 responses, half-up); runs/hosted-api-repeats-20261003/pass2/gpt-5.6-luna-low/events.jsonl:response.usage.reasoning_tokens (median over 480 responses, half-up); runs/hosted-api-repeats-20261003/pass3/gpt-5.6-luna-low/events.jsonl:response.usage.reasoning_tokens (median over 480 responses, half-up)
\newcommand{\LunaCachedTokTotal}{0}                                     % sum across passes: docs/lite/results/four-family/gpt-5.6-luna-low/analysis.json:usage.cached_tokens; docs/lite/results/hosted-api-repeats-20261003/pass2/gpt-5.6-luna-low/analysis.json:usage.cached_tokens; docs/lite/results/hosted-api-repeats-20261003/pass3/gpt-5.6-luna-low/analysis.json:usage.cached_tokens
\newcommand{\LunaCostInputUSD}{0.886}                                   % sum across passes: docs/lite/results/four-family/gpt-5.6-luna-low/analysis.json:usage at developers.openai.com/api/docs/pricing (gpt-5.6-luna, Standard, short context) list prices retrieved 2026-09-29 (input: uncached x input price + cached x cached price); docs/lite/results/hosted-api-repeats-20261003/pass2/gpt-5.6-luna-low/analysis.json:usage at developers.openai.com/api/docs/pricing (gpt-5.6-luna, Standard, short context) list prices retrieved 2026-09-29 (input: uncached x input price + cached x cached price); docs/lite/results/hosted-api-repeats-20261003/pass3/gpt-5.6-luna-low/analysis.json:usage at developers.openai.com/api/docs/pricing (gpt-5.6-luna, Standard, short context) list prices retrieved 2026-09-29 (input: uncached x input price + cached x cached price)
\newcommand{\LunaCostOutputUSD}{0.288}                                  % sum across passes: docs/lite/results/four-family/gpt-5.6-luna-low/analysis.json:usage at developers.openai.com/api/docs/pricing (gpt-5.6-luna, Standard, short context) list prices retrieved 2026-09-29 (output tokens, reasoning included, x output price); docs/lite/results/hosted-api-repeats-20261003/pass2/gpt-5.6-luna-low/analysis.json:usage at developers.openai.com/api/docs/pricing (gpt-5.6-luna, Standard, short context) list prices retrieved 2026-09-29 (output tokens, reasoning included, x output price); docs/lite/results/hosted-api-repeats-20261003/pass3/gpt-5.6-luna-low/analysis.json:usage at developers.openai.com/api/docs/pricing (gpt-5.6-luna, Standard, short context) list prices retrieved 2026-09-29 (output tokens, reasoning included, x output price)
\newcommand{\LunaCostUSD}{1.174}                                        % sum across passes: docs/lite/results/four-family/gpt-5.6-luna-low/analysis.json:usage at developers.openai.com/api/docs/pricing (gpt-5.6-luna, Standard, short context) list prices retrieved 2026-09-29 (input + output); docs/lite/results/hosted-api-repeats-20261003/pass2/gpt-5.6-luna-low/analysis.json:usage at developers.openai.com/api/docs/pricing (gpt-5.6-luna, Standard, short context) list prices retrieved 2026-09-29 (input + output); docs/lite/results/hosted-api-repeats-20261003/pass3/gpt-5.6-luna-low/analysis.json:usage at developers.openai.com/api/docs/pricing (gpt-5.6-luna, Standard, short context) list prices retrieved 2026-09-29 (input + output)
\newcommand{\LunaPriceInput}{0.200}                                     % equal mean across passes: developers.openai.com/api/docs/pricing (gpt-5.6-luna, Standard, short context): USD per 1M input tokens; developers.openai.com/api/docs/pricing (gpt-5.6-luna, Standard, short context): USD per 1M input tokens; developers.openai.com/api/docs/pricing (gpt-5.6-luna, Standard, short context): USD per 1M input tokens
\newcommand{\LunaPriceOutput}{1.20}                                     % equal mean across passes: developers.openai.com/api/docs/pricing (gpt-5.6-luna, Standard, short context): USD per 1M output tokens; developers.openai.com/api/docs/pricing (gpt-5.6-luna, Standard, short context): USD per 1M output tokens; developers.openai.com/api/docs/pricing (gpt-5.6-luna, Standard, short context): USD per 1M output tokens
\newcommand{\LunaErrNoDecisionPct}{2.1}                                 % equal mean across passes: docs/lite/results/four-family/gpt-5.6-luna-low/analysis.json:error_seconds.no_decision (scenario error fraction, then equal family mean); docs/lite/results/hosted-api-repeats-20261003/pass2/gpt-5.6-luna-low/analysis.json:error_seconds.no_decision (scenario error fraction, then equal family mean); docs/lite/results/hosted-api-repeats-20261003/pass3/gpt-5.6-luna-low/analysis.json:error_seconds.no_decision (scenario error fraction, then equal family mean)
\newcommand{\LunaErrStalePct}{41.2}                                     % equal mean across passes: docs/lite/results/four-family/gpt-5.6-luna-low/analysis.json:error_seconds.source_correct (scenario error fraction, then equal family mean); docs/lite/results/hosted-api-repeats-20261003/pass2/gpt-5.6-luna-low/analysis.json:error_seconds.source_correct (scenario error fraction, then equal family mean); docs/lite/results/hosted-api-repeats-20261003/pass3/gpt-5.6-luna-low/analysis.json:error_seconds.source_correct (scenario error fraction, then equal family mean)
\newcommand{\LunaInForceRecorded}{47.9}                                 % equal mean across passes: docs/lite/results/four-family/gpt-5.6-luna-low/analysis.json:scores.overall.time_accuracy; docs/lite/results/hosted-api-repeats-20261003/pass2/gpt-5.6-luna-low/analysis.json:scores.overall.time_accuracy; docs/lite/results/hosted-api-repeats-20261003/pass3/gpt-5.6-luna-low/analysis.json:scores.overall.time_accuracy
\newcommand{\LunaInForcePhysical}{47.9}                                 % equal mean across passes: docs/lite/results/four-family/gpt-5.6-luna-low/analysis.json:raw_wallclock_scores.overall.time_accuracy (physical wall-clock trace); docs/lite/results/hosted-api-repeats-20261003/pass2/gpt-5.6-luna-low/analysis.json:raw_wallclock_scores.overall.time_accuracy (physical wall-clock trace); docs/lite/results/hosted-api-repeats-20261003/pass3/gpt-5.6-luna-low/analysis.json:raw_wallclock_scores.overall.time_accuracy (physical wall-clock trace)
\newcommand{\LunaExcludedDispatchSec}{0.24}                             % equal mean across passes: docs/lite/results/four-family/gpt-5.6-luna-low/analysis.json:retry_reliability.excluded_dispatch_s (summed over the pass); docs/lite/results/hosted-api-repeats-20261003/pass2/gpt-5.6-luna-low/analysis.json:retry_reliability.excluded_dispatch_s (summed over the pass); docs/lite/results/hosted-api-repeats-20261003/pass3/gpt-5.6-luna-low/analysis.json:retry_reliability.excluded_dispatch_s (summed over the pass)
\newcommand{\LunaEventsHash}{61379abdedb29aea}                          % pass 1: docs/lite/results/four-family/gpt-5.6-luna-low/analysis.json:events_sha256 (first 16 hex digits; equals sha256 of runs/lite-v1-gpt-5.6-luna-low-four-family-retry-v1/events.jsonl)
\newcommand{\LunaAcceptedUpdates}{1404}                                 % sum across passes: docs/lite/results/four-family/gpt-5.6-luna-low/analysis.json:accepted_updates; docs/lite/results/hosted-api-repeats-20261003/pass2/gpt-5.6-luna-low/analysis.json:accepted_updates; docs/lite/results/hosted-api-repeats-20261003/pass3/gpt-5.6-luna-low/analysis.json:accepted_updates
\newcommand{\LunaDiscardedAfterHorizon}{23}                             % sum across passes: docs/lite/results/four-family/gpt-5.6-luna-low/analysis.json:discarded_updates.after_horizon (absent = 0); docs/lite/results/hosted-api-repeats-20261003/pass2/gpt-5.6-luna-low/analysis.json:discarded_updates.after_horizon (absent = 0); docs/lite/results/hosted-api-repeats-20261003/pass3/gpt-5.6-luna-low/analysis.json:discarded_updates.after_horizon (absent = 0)
\newcommand{\LunaDiscardedOlder}{13}                                    % sum across passes: docs/lite/results/four-family/gpt-5.6-luna-low/analysis.json:discarded_updates.older_than_active (absent = 0); docs/lite/results/hosted-api-repeats-20261003/pass2/gpt-5.6-luna-low/analysis.json:discarded_updates.older_than_active (absent = 0); docs/lite/results/hosted-api-repeats-20261003/pass3/gpt-5.6-luna-low/analysis.json:discarded_updates.older_than_active (absent = 0)
\newcommand{\LunaValidResponses}{1440}                                  % sum across passes: docs/lite/results/four-family/gpt-5.6-luna-low/analysis.json:retry_reliability.successful_logical_requests; docs/lite/results/hosted-api-repeats-20261003/pass2/gpt-5.6-luna-low/analysis.json:retry_reliability.successful_logical_requests; docs/lite/results/hosted-api-repeats-20261003/pass3/gpt-5.6-luna-low/analysis.json:retry_reliability.successful_logical_requests
\newcommand{\LunaAttempts}{1440}                                        % sum across passes: docs/lite/results/four-family/gpt-5.6-luna-low/analysis.json:retry_reliability.attempts; docs/lite/results/hosted-api-repeats-20261003/pass2/gpt-5.6-luna-low/analysis.json:retry_reliability.attempts; docs/lite/results/hosted-api-repeats-20261003/pass3/gpt-5.6-luna-low/analysis.json:retry_reliability.attempts
\newcommand{\LunaFailedAttempts}{0}                                     % sum across passes: docs/lite/results/four-family/gpt-5.6-luna-low/analysis.json:retry_reliability.failed_attempts; docs/lite/results/hosted-api-repeats-20261003/pass2/gpt-5.6-luna-low/analysis.json:retry_reliability.failed_attempts; docs/lite/results/hosted-api-repeats-20261003/pass3/gpt-5.6-luna-low/analysis.json:retry_reliability.failed_attempts
\newcommand{\LunaRetriedRequests}{0}                                    % sum across passes: docs/lite/results/four-family/gpt-5.6-luna-low/analysis.json:retry_reliability.retried_logical_requests; docs/lite/results/hosted-api-repeats-20261003/pass2/gpt-5.6-luna-low/analysis.json:retry_reliability.retried_logical_requests; docs/lite/results/hosted-api-repeats-20261003/pass3/gpt-5.6-luna-low/analysis.json:retry_reliability.retried_logical_requests
\newcommand{\LunaDebugUntimed}{88.9}                                    % equal mean across passes: docs/lite/results/four-family/gpt-5.6-luna-low/analysis.json:scores.by_family.live_debugging.untimed_decision_accuracy; docs/lite/results/hosted-api-repeats-20261003/pass2/gpt-5.6-luna-low/analysis.json:scores.by_family.live_debugging.untimed_decision_accuracy; docs/lite/results/hosted-api-repeats-20261003/pass3/gpt-5.6-luna-low/analysis.json:scores.by_family.live_debugging.untimed_decision_accuracy
\newcommand{\LunaDebugInForce}{52.7}                                    % equal mean across passes: docs/lite/results/four-family/gpt-5.6-luna-low/analysis.json:scores.by_family.live_debugging.time_accuracy; docs/lite/results/hosted-api-repeats-20261003/pass2/gpt-5.6-luna-low/analysis.json:scores.by_family.live_debugging.time_accuracy; docs/lite/results/hosted-api-repeats-20261003/pass3/gpt-5.6-luna-low/analysis.json:scores.by_family.live_debugging.time_accuracy
\newcommand{\LunaDebugNetRemoved}{62.0}                                 % equal mean across passes: docs/lite/results/four-family/gpt-5.6-luna-low/analysis.json:network_adjustment.scores.estimate.by_family.live_debugging.time_accuracy; docs/lite/results/hosted-api-repeats-20261003/pass2/gpt-5.6-luna-low/analysis.json:network_adjustment.scores.estimate.by_family.live_debugging.time_accuracy; docs/lite/results/hosted-api-repeats-20261003/pass3/gpt-5.6-luna-low/analysis.json:network_adjustment.scores.estimate.by_family.live_debugging.time_accuracy
\newcommand{\LunaDebugNetRemovedLow}{59.4}                              % equal mean across passes: docs/lite/results/four-family/gpt-5.6-luna-low/analysis.json:network_adjustment.scores.low.by_family.live_debugging.time_accuracy; docs/lite/results/hosted-api-repeats-20261003/pass2/gpt-5.6-luna-low/analysis.json:network_adjustment.scores.low.by_family.live_debugging.time_accuracy; docs/lite/results/hosted-api-repeats-20261003/pass3/gpt-5.6-luna-low/analysis.json:network_adjustment.scores.low.by_family.live_debugging.time_accuracy
\newcommand{\LunaDebugNetRemovedHigh}{63.4}                             % equal mean across passes: docs/lite/results/four-family/gpt-5.6-luna-low/analysis.json:network_adjustment.scores.high.by_family.live_debugging.time_accuracy; docs/lite/results/hosted-api-repeats-20261003/pass2/gpt-5.6-luna-low/analysis.json:network_adjustment.scores.high.by_family.live_debugging.time_accuracy; docs/lite/results/hosted-api-repeats-20261003/pass3/gpt-5.6-luna-low/analysis.json:network_adjustment.scores.high.by_family.live_debugging.time_accuracy
\newcommand{\LunaAssemblyUntimed}{86.9}                                 % equal mean across passes: docs/lite/results/four-family/gpt-5.6-luna-low/analysis.json:scores.by_family.procedural_coaching.untimed_decision_accuracy; docs/lite/results/hosted-api-repeats-20261003/pass2/gpt-5.6-luna-low/analysis.json:scores.by_family.procedural_coaching.untimed_decision_accuracy; docs/lite/results/hosted-api-repeats-20261003/pass3/gpt-5.6-luna-low/analysis.json:scores.by_family.procedural_coaching.untimed_decision_accuracy
\newcommand{\LunaAssemblyInForce}{45.7}                                 % equal mean across passes: docs/lite/results/four-family/gpt-5.6-luna-low/analysis.json:scores.by_family.procedural_coaching.time_accuracy; docs/lite/results/hosted-api-repeats-20261003/pass2/gpt-5.6-luna-low/analysis.json:scores.by_family.procedural_coaching.time_accuracy; docs/lite/results/hosted-api-repeats-20261003/pass3/gpt-5.6-luna-low/analysis.json:scores.by_family.procedural_coaching.time_accuracy
\newcommand{\LunaAssemblyNetRemoved}{55.5}                              % equal mean across passes: docs/lite/results/four-family/gpt-5.6-luna-low/analysis.json:network_adjustment.scores.estimate.by_family.procedural_coaching.time_accuracy; docs/lite/results/hosted-api-repeats-20261003/pass2/gpt-5.6-luna-low/analysis.json:network_adjustment.scores.estimate.by_family.procedural_coaching.time_accuracy; docs/lite/results/hosted-api-repeats-20261003/pass3/gpt-5.6-luna-low/analysis.json:network_adjustment.scores.estimate.by_family.procedural_coaching.time_accuracy
\newcommand{\LunaAssemblyNetRemovedLow}{52.7}                           % equal mean across passes: docs/lite/results/four-family/gpt-5.6-luna-low/analysis.json:network_adjustment.scores.low.by_family.procedural_coaching.time_accuracy; docs/lite/results/hosted-api-repeats-20261003/pass2/gpt-5.6-luna-low/analysis.json:network_adjustment.scores.low.by_family.procedural_coaching.time_accuracy; docs/lite/results/hosted-api-repeats-20261003/pass3/gpt-5.6-luna-low/analysis.json:network_adjustment.scores.low.by_family.procedural_coaching.time_accuracy
\newcommand{\LunaAssemblyNetRemovedHigh}{57.1}                          % equal mean across passes: docs/lite/results/four-family/gpt-5.6-luna-low/analysis.json:network_adjustment.scores.high.by_family.procedural_coaching.time_accuracy; docs/lite/results/hosted-api-repeats-20261003/pass2/gpt-5.6-luna-low/analysis.json:network_adjustment.scores.high.by_family.procedural_coaching.time_accuracy; docs/lite/results/hosted-api-repeats-20261003/pass3/gpt-5.6-luna-low/analysis.json:network_adjustment.scores.high.by_family.procedural_coaching.time_accuracy
\newcommand{\LunaSupportUntimed}{87.2}                                  % equal mean across passes: docs/lite/results/four-family/gpt-5.6-luna-low/analysis.json:scores.by_family.support_call_assist.untimed_decision_accuracy; docs/lite/results/hosted-api-repeats-20261003/pass2/gpt-5.6-luna-low/analysis.json:scores.by_family.support_call_assist.untimed_decision_accuracy; docs/lite/results/hosted-api-repeats-20261003/pass3/gpt-5.6-luna-low/analysis.json:scores.by_family.support_call_assist.untimed_decision_accuracy
\newcommand{\LunaSupportInForce}{48.4}                                  % equal mean across passes: docs/lite/results/four-family/gpt-5.6-luna-low/analysis.json:scores.by_family.support_call_assist.time_accuracy; docs/lite/results/hosted-api-repeats-20261003/pass2/gpt-5.6-luna-low/analysis.json:scores.by_family.support_call_assist.time_accuracy; docs/lite/results/hosted-api-repeats-20261003/pass3/gpt-5.6-luna-low/analysis.json:scores.by_family.support_call_assist.time_accuracy
\newcommand{\LunaSupportNetRemoved}{59.1}                               % equal mean across passes: docs/lite/results/four-family/gpt-5.6-luna-low/analysis.json:network_adjustment.scores.estimate.by_family.support_call_assist.time_accuracy; docs/lite/results/hosted-api-repeats-20261003/pass2/gpt-5.6-luna-low/analysis.json:network_adjustment.scores.estimate.by_family.support_call_assist.time_accuracy; docs/lite/results/hosted-api-repeats-20261003/pass3/gpt-5.6-luna-low/analysis.json:network_adjustment.scores.estimate.by_family.support_call_assist.time_accuracy
\newcommand{\LunaSupportNetRemovedLow}{56.2}                            % equal mean across passes: docs/lite/results/four-family/gpt-5.6-luna-low/analysis.json:network_adjustment.scores.low.by_family.support_call_assist.time_accuracy; docs/lite/results/hosted-api-repeats-20261003/pass2/gpt-5.6-luna-low/analysis.json:network_adjustment.scores.low.by_family.support_call_assist.time_accuracy; docs/lite/results/hosted-api-repeats-20261003/pass3/gpt-5.6-luna-low/analysis.json:network_adjustment.scores.low.by_family.support_call_assist.time_accuracy
\newcommand{\LunaSupportNetRemovedHigh}{60.7}                           % equal mean across passes: docs/lite/results/four-family/gpt-5.6-luna-low/analysis.json:network_adjustment.scores.high.by_family.support_call_assist.time_accuracy; docs/lite/results/hosted-api-repeats-20261003/pass2/gpt-5.6-luna-low/analysis.json:network_adjustment.scores.high.by_family.support_call_assist.time_accuracy; docs/lite/results/hosted-api-repeats-20261003/pass3/gpt-5.6-luna-low/analysis.json:network_adjustment.scores.high.by_family.support_call_assist.time_accuracy
\newcommand{\LunaPresenterUntimed}{96.7}                                % equal mean across passes: docs/lite/results/four-family/gpt-5.6-luna-low/analysis.json:scores.by_family.presenter_voice_control.untimed_decision_accuracy; docs/lite/results/hosted-api-repeats-20261003/pass2/gpt-5.6-luna-low/analysis.json:scores.by_family.presenter_voice_control.untimed_decision_accuracy; docs/lite/results/hosted-api-repeats-20261003/pass3/gpt-5.6-luna-low/analysis.json:scores.by_family.presenter_voice_control.untimed_decision_accuracy
\newcommand{\LunaPresenterInForce}{44.9}                                % equal mean across passes: docs/lite/results/four-family/gpt-5.6-luna-low/analysis.json:scores.by_family.presenter_voice_control.time_accuracy; docs/lite/results/hosted-api-repeats-20261003/pass2/gpt-5.6-luna-low/analysis.json:scores.by_family.presenter_voice_control.time_accuracy; docs/lite/results/hosted-api-repeats-20261003/pass3/gpt-5.6-luna-low/analysis.json:scores.by_family.presenter_voice_control.time_accuracy
\newcommand{\LunaPresenterNetRemoved}{54.8}                             % equal mean across passes: docs/lite/results/four-family/gpt-5.6-luna-low/analysis.json:network_adjustment.scores.estimate.by_family.presenter_voice_control.time_accuracy; docs/lite/results/hosted-api-repeats-20261003/pass2/gpt-5.6-luna-low/analysis.json:network_adjustment.scores.estimate.by_family.presenter_voice_control.time_accuracy; docs/lite/results/hosted-api-repeats-20261003/pass3/gpt-5.6-luna-low/analysis.json:network_adjustment.scores.estimate.by_family.presenter_voice_control.time_accuracy
\newcommand{\LunaPresenterNetRemovedLow}{52.0}                          % equal mean across passes: docs/lite/results/four-family/gpt-5.6-luna-low/analysis.json:network_adjustment.scores.low.by_family.presenter_voice_control.time_accuracy; docs/lite/results/hosted-api-repeats-20261003/pass2/gpt-5.6-luna-low/analysis.json:network_adjustment.scores.low.by_family.presenter_voice_control.time_accuracy; docs/lite/results/hosted-api-repeats-20261003/pass3/gpt-5.6-luna-low/analysis.json:network_adjustment.scores.low.by_family.presenter_voice_control.time_accuracy
\newcommand{\LunaPresenterNetRemovedHigh}{56.4}                         % equal mean across passes: docs/lite/results/four-family/gpt-5.6-luna-low/analysis.json:network_adjustment.scores.high.by_family.presenter_voice_control.time_accuracy; docs/lite/results/hosted-api-repeats-20261003/pass2/gpt-5.6-luna-low/analysis.json:network_adjustment.scores.high.by_family.presenter_voice_control.time_accuracy; docs/lite/results/hosted-api-repeats-20261003/pass3/gpt-5.6-luna-low/analysis.json:network_adjustment.scores.high.by_family.presenter_voice_control.time_accuracy
\newcommand{\LunaDebugAUntimed}{87.8}                                   % equal mean across passes: docs/lite/results/four-family/gpt-5.6-luna-low/analysis.json:scores.per_episode[0].untimed_decision_accuracy (lite_debugging_a); docs/lite/results/hosted-api-repeats-20261003/pass2/gpt-5.6-luna-low/analysis.json:scores.per_episode[0].untimed_decision_accuracy (lite_debugging_a); docs/lite/results/hosted-api-repeats-20261003/pass3/gpt-5.6-luna-low/analysis.json:scores.per_episode[0].untimed_decision_accuracy (lite_debugging_a)
\newcommand{\LunaDebugAInForce}{57.9}                                   % equal mean across passes: docs/lite/results/four-family/gpt-5.6-luna-low/analysis.json:scores.per_episode[0].time_accuracy (lite_debugging_a); docs/lite/results/hosted-api-repeats-20261003/pass2/gpt-5.6-luna-low/analysis.json:scores.per_episode[0].time_accuracy (lite_debugging_a); docs/lite/results/hosted-api-repeats-20261003/pass3/gpt-5.6-luna-low/analysis.json:scores.per_episode[0].time_accuracy (lite_debugging_a)
\newcommand{\LunaDebugAInForceSeg}{54.8}                                % equal mean across passes: docs/lite/results/four-family/gpt-5.6-luna-low/analysis.json:scores.per_episode[0].segment_time_accuracy (lite_debugging_a); docs/lite/results/hosted-api-repeats-20261003/pass2/gpt-5.6-luna-low/analysis.json:scores.per_episode[0].segment_time_accuracy (lite_debugging_a); docs/lite/results/hosted-api-repeats-20261003/pass3/gpt-5.6-luna-low/analysis.json:scores.per_episode[0].segment_time_accuracy (lite_debugging_a)
\newcommand{\LunaDebugAErrNoDecisionSec}{2.5}                           % equal mean across passes: docs/lite/results/four-family/gpt-5.6-luna-low/analysis.json:scores.per_episode[0].error_seconds.no_decision (lite_debugging_a); docs/lite/results/hosted-api-repeats-20261003/pass2/gpt-5.6-luna-low/analysis.json:scores.per_episode[0].error_seconds.no_decision (lite_debugging_a); docs/lite/results/hosted-api-repeats-20261003/pass3/gpt-5.6-luna-low/analysis.json:scores.per_episode[0].error_seconds.no_decision (lite_debugging_a)
\newcommand{\LunaDebugAErrStaleSec}{39.7}                               % equal mean across passes: docs/lite/results/four-family/gpt-5.6-luna-low/analysis.json:scores.per_episode[0].error_seconds.source_correct (lite_debugging_a); docs/lite/results/hosted-api-repeats-20261003/pass2/gpt-5.6-luna-low/analysis.json:scores.per_episode[0].error_seconds.source_correct (lite_debugging_a); docs/lite/results/hosted-api-repeats-20261003/pass3/gpt-5.6-luna-low/analysis.json:scores.per_episode[0].error_seconds.source_correct (lite_debugging_a)
\newcommand{\LunaDebugALatencyMedian}{2.20}                             % equal mean across passes: docs/lite/results/four-family/gpt-5.6-luna-low/analysis.json:scores.per_episode[0].latency_s_p50 (lite_debugging_a); docs/lite/results/hosted-api-repeats-20261003/pass2/gpt-5.6-luna-low/analysis.json:scores.per_episode[0].latency_s_p50 (lite_debugging_a); docs/lite/results/hosted-api-repeats-20261003/pass3/gpt-5.6-luna-low/analysis.json:scores.per_episode[0].latency_s_p50 (lite_debugging_a)
\newcommand{\LunaDebugBUntimed}{90.0}                                   % equal mean across passes: docs/lite/results/four-family/gpt-5.6-luna-low/analysis.json:scores.per_episode[1].untimed_decision_accuracy (lite_debugging_b); docs/lite/results/hosted-api-repeats-20261003/pass2/gpt-5.6-luna-low/analysis.json:scores.per_episode[1].untimed_decision_accuracy (lite_debugging_b); docs/lite/results/hosted-api-repeats-20261003/pass3/gpt-5.6-luna-low/analysis.json:scores.per_episode[1].untimed_decision_accuracy (lite_debugging_b)
\newcommand{\LunaDebugBInForce}{47.4}                                   % equal mean across passes: docs/lite/results/four-family/gpt-5.6-luna-low/analysis.json:scores.per_episode[1].time_accuracy (lite_debugging_b); docs/lite/results/hosted-api-repeats-20261003/pass2/gpt-5.6-luna-low/analysis.json:scores.per_episode[1].time_accuracy (lite_debugging_b); docs/lite/results/hosted-api-repeats-20261003/pass3/gpt-5.6-luna-low/analysis.json:scores.per_episode[1].time_accuracy (lite_debugging_b)
\newcommand{\LunaDebugBInForceSeg}{40.4}                                % equal mean across passes: docs/lite/results/four-family/gpt-5.6-luna-low/analysis.json:scores.per_episode[1].segment_time_accuracy (lite_debugging_b); docs/lite/results/hosted-api-repeats-20261003/pass2/gpt-5.6-luna-low/analysis.json:scores.per_episode[1].segment_time_accuracy (lite_debugging_b); docs/lite/results/hosted-api-repeats-20261003/pass3/gpt-5.6-luna-low/analysis.json:scores.per_episode[1].segment_time_accuracy (lite_debugging_b)
\newcommand{\LunaDebugBErrNoDecisionSec}{3.4}                           % equal mean across passes: docs/lite/results/four-family/gpt-5.6-luna-low/analysis.json:scores.per_episode[1].error_seconds.no_decision (lite_debugging_b); docs/lite/results/hosted-api-repeats-20261003/pass2/gpt-5.6-luna-low/analysis.json:scores.per_episode[1].error_seconds.no_decision (lite_debugging_b); docs/lite/results/hosted-api-repeats-20261003/pass3/gpt-5.6-luna-low/analysis.json:scores.per_episode[1].error_seconds.no_decision (lite_debugging_b)
\newcommand{\LunaDebugBErrStaleSec}{50.0}                               % equal mean across passes: docs/lite/results/four-family/gpt-5.6-luna-low/analysis.json:scores.per_episode[1].error_seconds.source_correct (lite_debugging_b); docs/lite/results/hosted-api-repeats-20261003/pass2/gpt-5.6-luna-low/analysis.json:scores.per_episode[1].error_seconds.source_correct (lite_debugging_b); docs/lite/results/hosted-api-repeats-20261003/pass3/gpt-5.6-luna-low/analysis.json:scores.per_episode[1].error_seconds.source_correct (lite_debugging_b)
\newcommand{\LunaDebugBLatencyMedian}{2.57}                             % equal mean across passes: docs/lite/results/four-family/gpt-5.6-luna-low/analysis.json:scores.per_episode[1].latency_s_p50 (lite_debugging_b); docs/lite/results/hosted-api-repeats-20261003/pass2/gpt-5.6-luna-low/analysis.json:scores.per_episode[1].latency_s_p50 (lite_debugging_b); docs/lite/results/hosted-api-repeats-20261003/pass3/gpt-5.6-luna-low/analysis.json:scores.per_episode[1].latency_s_p50 (lite_debugging_b)
\newcommand{\LunaAssemblyAUntimed}{90.6}                                % equal mean across passes: docs/lite/results/four-family/gpt-5.6-luna-low/analysis.json:scores.per_episode[2].untimed_decision_accuracy (lite_assembly_a); docs/lite/results/hosted-api-repeats-20261003/pass2/gpt-5.6-luna-low/analysis.json:scores.per_episode[2].untimed_decision_accuracy (lite_assembly_a); docs/lite/results/hosted-api-repeats-20261003/pass3/gpt-5.6-luna-low/analysis.json:scores.per_episode[2].untimed_decision_accuracy (lite_assembly_a)
\newcommand{\LunaAssemblyAInForce}{50.0}                                % equal mean across passes: docs/lite/results/four-family/gpt-5.6-luna-low/analysis.json:scores.per_episode[2].time_accuracy (lite_assembly_a); docs/lite/results/hosted-api-repeats-20261003/pass2/gpt-5.6-luna-low/analysis.json:scores.per_episode[2].time_accuracy (lite_assembly_a); docs/lite/results/hosted-api-repeats-20261003/pass3/gpt-5.6-luna-low/analysis.json:scores.per_episode[2].time_accuracy (lite_assembly_a)
\newcommand{\LunaAssemblyAInForceSeg}{40.9}                             % equal mean across passes: docs/lite/results/four-family/gpt-5.6-luna-low/analysis.json:scores.per_episode[2].segment_time_accuracy (lite_assembly_a); docs/lite/results/hosted-api-repeats-20261003/pass2/gpt-5.6-luna-low/analysis.json:scores.per_episode[2].segment_time_accuracy (lite_assembly_a); docs/lite/results/hosted-api-repeats-20261003/pass3/gpt-5.6-luna-low/analysis.json:scores.per_episode[2].segment_time_accuracy (lite_assembly_a)
\newcommand{\LunaAssemblyAErrNoDecisionSec}{2.4}                        % equal mean across passes: docs/lite/results/four-family/gpt-5.6-luna-low/analysis.json:scores.per_episode[2].error_seconds.no_decision (lite_assembly_a); docs/lite/results/hosted-api-repeats-20261003/pass2/gpt-5.6-luna-low/analysis.json:scores.per_episode[2].error_seconds.no_decision (lite_assembly_a); docs/lite/results/hosted-api-repeats-20261003/pass3/gpt-5.6-luna-low/analysis.json:scores.per_episode[2].error_seconds.no_decision (lite_assembly_a)
\newcommand{\LunaAssemblyAErrStaleSec}{46.9}                            % equal mean across passes: docs/lite/results/four-family/gpt-5.6-luna-low/analysis.json:scores.per_episode[2].error_seconds.source_correct (lite_assembly_a); docs/lite/results/hosted-api-repeats-20261003/pass2/gpt-5.6-luna-low/analysis.json:scores.per_episode[2].error_seconds.source_correct (lite_assembly_a); docs/lite/results/hosted-api-repeats-20261003/pass3/gpt-5.6-luna-low/analysis.json:scores.per_episode[2].error_seconds.source_correct (lite_assembly_a)
\newcommand{\LunaAssemblyALatencyMedian}{2.45}                          % equal mean across passes: docs/lite/results/four-family/gpt-5.6-luna-low/analysis.json:scores.per_episode[2].latency_s_p50 (lite_assembly_a); docs/lite/results/hosted-api-repeats-20261003/pass2/gpt-5.6-luna-low/analysis.json:scores.per_episode[2].latency_s_p50 (lite_assembly_a); docs/lite/results/hosted-api-repeats-20261003/pass3/gpt-5.6-luna-low/analysis.json:scores.per_episode[2].latency_s_p50 (lite_assembly_a)
\newcommand{\LunaAssemblyBUntimed}{83.3}                                % equal mean across passes: docs/lite/results/four-family/gpt-5.6-luna-low/analysis.json:scores.per_episode[3].untimed_decision_accuracy (lite_assembly_b); docs/lite/results/hosted-api-repeats-20261003/pass2/gpt-5.6-luna-low/analysis.json:scores.per_episode[3].untimed_decision_accuracy (lite_assembly_b); docs/lite/results/hosted-api-repeats-20261003/pass3/gpt-5.6-luna-low/analysis.json:scores.per_episode[3].untimed_decision_accuracy (lite_assembly_b)
\newcommand{\LunaAssemblyBInForce}{41.3}                                % equal mean across passes: docs/lite/results/four-family/gpt-5.6-luna-low/analysis.json:scores.per_episode[3].time_accuracy (lite_assembly_b); docs/lite/results/hosted-api-repeats-20261003/pass2/gpt-5.6-luna-low/analysis.json:scores.per_episode[3].time_accuracy (lite_assembly_b); docs/lite/results/hosted-api-repeats-20261003/pass3/gpt-5.6-luna-low/analysis.json:scores.per_episode[3].time_accuracy (lite_assembly_b)
\newcommand{\LunaAssemblyBInForceSeg}{29.5}                             % equal mean across passes: docs/lite/results/four-family/gpt-5.6-luna-low/analysis.json:scores.per_episode[3].segment_time_accuracy (lite_assembly_b); docs/lite/results/hosted-api-repeats-20261003/pass2/gpt-5.6-luna-low/analysis.json:scores.per_episode[3].segment_time_accuracy (lite_assembly_b); docs/lite/results/hosted-api-repeats-20261003/pass3/gpt-5.6-luna-low/analysis.json:scores.per_episode[3].segment_time_accuracy (lite_assembly_b)
\newcommand{\LunaAssemblyBErrNoDecisionSec}{2.1}                        % equal mean across passes: docs/lite/results/four-family/gpt-5.6-luna-low/analysis.json:scores.per_episode[3].error_seconds.no_decision (lite_assembly_b); docs/lite/results/hosted-api-repeats-20261003/pass2/gpt-5.6-luna-low/analysis.json:scores.per_episode[3].error_seconds.no_decision (lite_assembly_b); docs/lite/results/hosted-api-repeats-20261003/pass3/gpt-5.6-luna-low/analysis.json:scores.per_episode[3].error_seconds.no_decision (lite_assembly_b)
\newcommand{\LunaAssemblyBErrStaleSec}{53.1}                            % equal mean across passes: docs/lite/results/four-family/gpt-5.6-luna-low/analysis.json:scores.per_episode[3].error_seconds.source_correct (lite_assembly_b); docs/lite/results/hosted-api-repeats-20261003/pass2/gpt-5.6-luna-low/analysis.json:scores.per_episode[3].error_seconds.source_correct (lite_assembly_b); docs/lite/results/hosted-api-repeats-20261003/pass3/gpt-5.6-luna-low/analysis.json:scores.per_episode[3].error_seconds.source_correct (lite_assembly_b)
\newcommand{\LunaAssemblyBLatencyMedian}{2.90}                          % equal mean across passes: docs/lite/results/four-family/gpt-5.6-luna-low/analysis.json:scores.per_episode[3].latency_s_p50 (lite_assembly_b); docs/lite/results/hosted-api-repeats-20261003/pass2/gpt-5.6-luna-low/analysis.json:scores.per_episode[3].latency_s_p50 (lite_assembly_b); docs/lite/results/hosted-api-repeats-20261003/pass3/gpt-5.6-luna-low/analysis.json:scores.per_episode[3].latency_s_p50 (lite_assembly_b)
\newcommand{\LunaSupportAUntimed}{82.8}                                 % equal mean across passes: docs/lite/results/four-family/gpt-5.6-luna-low/analysis.json:scores.per_episode[4].untimed_decision_accuracy (lite_support_a); docs/lite/results/hosted-api-repeats-20261003/pass2/gpt-5.6-luna-low/analysis.json:scores.per_episode[4].untimed_decision_accuracy (lite_support_a); docs/lite/results/hosted-api-repeats-20261003/pass3/gpt-5.6-luna-low/analysis.json:scores.per_episode[4].untimed_decision_accuracy (lite_support_a)
\newcommand{\LunaSupportAInForce}{46.4}                                 % equal mean across passes: docs/lite/results/four-family/gpt-5.6-luna-low/analysis.json:scores.per_episode[4].time_accuracy (lite_support_a); docs/lite/results/hosted-api-repeats-20261003/pass2/gpt-5.6-luna-low/analysis.json:scores.per_episode[4].time_accuracy (lite_support_a); docs/lite/results/hosted-api-repeats-20261003/pass3/gpt-5.6-luna-low/analysis.json:scores.per_episode[4].time_accuracy (lite_support_a)
\newcommand{\LunaSupportAInForceSeg}{37.0}                              % equal mean across passes: docs/lite/results/four-family/gpt-5.6-luna-low/analysis.json:scores.per_episode[4].segment_time_accuracy (lite_support_a); docs/lite/results/hosted-api-repeats-20261003/pass2/gpt-5.6-luna-low/analysis.json:scores.per_episode[4].segment_time_accuracy (lite_support_a); docs/lite/results/hosted-api-repeats-20261003/pass3/gpt-5.6-luna-low/analysis.json:scores.per_episode[4].segment_time_accuracy (lite_support_a)
\newcommand{\LunaSupportAErrNoDecisionSec}{2.4}                         % equal mean across passes: docs/lite/results/four-family/gpt-5.6-luna-low/analysis.json:scores.per_episode[4].error_seconds.no_decision (lite_support_a); docs/lite/results/hosted-api-repeats-20261003/pass2/gpt-5.6-luna-low/analysis.json:scores.per_episode[4].error_seconds.no_decision (lite_support_a); docs/lite/results/hosted-api-repeats-20261003/pass3/gpt-5.6-luna-low/analysis.json:scores.per_episode[4].error_seconds.no_decision (lite_support_a)
\newcommand{\LunaSupportAErrStaleSec}{39.3}                             % equal mean across passes: docs/lite/results/four-family/gpt-5.6-luna-low/analysis.json:scores.per_episode[4].error_seconds.source_correct (lite_support_a); docs/lite/results/hosted-api-repeats-20261003/pass2/gpt-5.6-luna-low/analysis.json:scores.per_episode[4].error_seconds.source_correct (lite_support_a); docs/lite/results/hosted-api-repeats-20261003/pass3/gpt-5.6-luna-low/analysis.json:scores.per_episode[4].error_seconds.source_correct (lite_support_a)
\newcommand{\LunaSupportALatencyMedian}{2.13}                           % equal mean across passes: docs/lite/results/four-family/gpt-5.6-luna-low/analysis.json:scores.per_episode[4].latency_s_p50 (lite_support_a); docs/lite/results/hosted-api-repeats-20261003/pass2/gpt-5.6-luna-low/analysis.json:scores.per_episode[4].latency_s_p50 (lite_support_a); docs/lite/results/hosted-api-repeats-20261003/pass3/gpt-5.6-luna-low/analysis.json:scores.per_episode[4].latency_s_p50 (lite_support_a)
\newcommand{\LunaSupportBUntimed}{91.7}                                 % equal mean across passes: docs/lite/results/four-family/gpt-5.6-luna-low/analysis.json:scores.per_episode[5].untimed_decision_accuracy (lite_support_b); docs/lite/results/hosted-api-repeats-20261003/pass2/gpt-5.6-luna-low/analysis.json:scores.per_episode[5].untimed_decision_accuracy (lite_support_b); docs/lite/results/hosted-api-repeats-20261003/pass3/gpt-5.6-luna-low/analysis.json:scores.per_episode[5].untimed_decision_accuracy (lite_support_b)
\newcommand{\LunaSupportBInForce}{50.4}                                 % equal mean across passes: docs/lite/results/four-family/gpt-5.6-luna-low/analysis.json:scores.per_episode[5].time_accuracy (lite_support_b); docs/lite/results/hosted-api-repeats-20261003/pass2/gpt-5.6-luna-low/analysis.json:scores.per_episode[5].time_accuracy (lite_support_b); docs/lite/results/hosted-api-repeats-20261003/pass3/gpt-5.6-luna-low/analysis.json:scores.per_episode[5].time_accuracy (lite_support_b)
\newcommand{\LunaSupportBInForceSeg}{44.0}                              % equal mean across passes: docs/lite/results/four-family/gpt-5.6-luna-low/analysis.json:scores.per_episode[5].segment_time_accuracy (lite_support_b); docs/lite/results/hosted-api-repeats-20261003/pass2/gpt-5.6-luna-low/analysis.json:scores.per_episode[5].segment_time_accuracy (lite_support_b); docs/lite/results/hosted-api-repeats-20261003/pass3/gpt-5.6-luna-low/analysis.json:scores.per_episode[5].segment_time_accuracy (lite_support_b)
\newcommand{\LunaSupportBErrNoDecisionSec}{2.7}                         % equal mean across passes: docs/lite/results/four-family/gpt-5.6-luna-low/analysis.json:scores.per_episode[5].error_seconds.no_decision (lite_support_b); docs/lite/results/hosted-api-repeats-20261003/pass2/gpt-5.6-luna-low/analysis.json:scores.per_episode[5].error_seconds.no_decision (lite_support_b); docs/lite/results/hosted-api-repeats-20261003/pass3/gpt-5.6-luna-low/analysis.json:scores.per_episode[5].error_seconds.no_decision (lite_support_b)
\newcommand{\LunaSupportBErrStaleSec}{47.9}                             % equal mean across passes: docs/lite/results/four-family/gpt-5.6-luna-low/analysis.json:scores.per_episode[5].error_seconds.source_correct (lite_support_b); docs/lite/results/hosted-api-repeats-20261003/pass2/gpt-5.6-luna-low/analysis.json:scores.per_episode[5].error_seconds.source_correct (lite_support_b); docs/lite/results/hosted-api-repeats-20261003/pass3/gpt-5.6-luna-low/analysis.json:scores.per_episode[5].error_seconds.source_correct (lite_support_b)
\newcommand{\LunaSupportBLatencyMedian}{2.23}                           % equal mean across passes: docs/lite/results/four-family/gpt-5.6-luna-low/analysis.json:scores.per_episode[5].latency_s_p50 (lite_support_b); docs/lite/results/hosted-api-repeats-20261003/pass2/gpt-5.6-luna-low/analysis.json:scores.per_episode[5].latency_s_p50 (lite_support_b); docs/lite/results/hosted-api-repeats-20261003/pass3/gpt-5.6-luna-low/analysis.json:scores.per_episode[5].latency_s_p50 (lite_support_b)
\newcommand{\LunaPresenterAUntimed}{99.4}                               % equal mean across passes: docs/lite/results/four-family/gpt-5.6-luna-low/analysis.json:scores.per_episode[6].untimed_decision_accuracy (lite_presenter_a); docs/lite/results/hosted-api-repeats-20261003/pass2/gpt-5.6-luna-low/analysis.json:scores.per_episode[6].untimed_decision_accuracy (lite_presenter_a); docs/lite/results/hosted-api-repeats-20261003/pass3/gpt-5.6-luna-low/analysis.json:scores.per_episode[6].untimed_decision_accuracy (lite_presenter_a)
\newcommand{\LunaPresenterAInForce}{47.0}                               % equal mean across passes: docs/lite/results/four-family/gpt-5.6-luna-low/analysis.json:scores.per_episode[6].time_accuracy (lite_presenter_a); docs/lite/results/hosted-api-repeats-20261003/pass2/gpt-5.6-luna-low/analysis.json:scores.per_episode[6].time_accuracy (lite_presenter_a); docs/lite/results/hosted-api-repeats-20261003/pass3/gpt-5.6-luna-low/analysis.json:scores.per_episode[6].time_accuracy (lite_presenter_a)
\newcommand{\LunaPresenterAInForceSeg}{33.3}                            % equal mean across passes: docs/lite/results/four-family/gpt-5.6-luna-low/analysis.json:scores.per_episode[6].segment_time_accuracy (lite_presenter_a); docs/lite/results/hosted-api-repeats-20261003/pass2/gpt-5.6-luna-low/analysis.json:scores.per_episode[6].segment_time_accuracy (lite_presenter_a); docs/lite/results/hosted-api-repeats-20261003/pass3/gpt-5.6-luna-low/analysis.json:scores.per_episode[6].segment_time_accuracy (lite_presenter_a)
\newcommand{\LunaPresenterAErrNoDecisionSec}{2.3}                       % equal mean across passes: docs/lite/results/four-family/gpt-5.6-luna-low/analysis.json:scores.per_episode[6].error_seconds.no_decision (lite_presenter_a); docs/lite/results/hosted-api-repeats-20261003/pass2/gpt-5.6-luna-low/analysis.json:scores.per_episode[6].error_seconds.no_decision (lite_presenter_a); docs/lite/results/hosted-api-repeats-20261003/pass3/gpt-5.6-luna-low/analysis.json:scores.per_episode[6].error_seconds.no_decision (lite_presenter_a)
\newcommand{\LunaPresenterAErrStaleSec}{60.5}                           % equal mean across passes: docs/lite/results/four-family/gpt-5.6-luna-low/analysis.json:scores.per_episode[6].error_seconds.source_correct (lite_presenter_a); docs/lite/results/hosted-api-repeats-20261003/pass2/gpt-5.6-luna-low/analysis.json:scores.per_episode[6].error_seconds.source_correct (lite_presenter_a); docs/lite/results/hosted-api-repeats-20261003/pass3/gpt-5.6-luna-low/analysis.json:scores.per_episode[6].error_seconds.source_correct (lite_presenter_a)
\newcommand{\LunaPresenterALatencyMedian}{2.71}                         % equal mean across passes: docs/lite/results/four-family/gpt-5.6-luna-low/analysis.json:scores.per_episode[6].latency_s_p50 (lite_presenter_a); docs/lite/results/hosted-api-repeats-20261003/pass2/gpt-5.6-luna-low/analysis.json:scores.per_episode[6].latency_s_p50 (lite_presenter_a); docs/lite/results/hosted-api-repeats-20261003/pass3/gpt-5.6-luna-low/analysis.json:scores.per_episode[6].latency_s_p50 (lite_presenter_a)
\newcommand{\LunaPresenterBUntimed}{93.9}                               % equal mean across passes: docs/lite/results/four-family/gpt-5.6-luna-low/analysis.json:scores.per_episode[7].untimed_decision_accuracy (lite_presenter_b); docs/lite/results/hosted-api-repeats-20261003/pass2/gpt-5.6-luna-low/analysis.json:scores.per_episode[7].untimed_decision_accuracy (lite_presenter_b); docs/lite/results/hosted-api-repeats-20261003/pass3/gpt-5.6-luna-low/analysis.json:scores.per_episode[7].untimed_decision_accuracy (lite_presenter_b)
\newcommand{\LunaPresenterBInForce}{42.8}                               % equal mean across passes: docs/lite/results/four-family/gpt-5.6-luna-low/analysis.json:scores.per_episode[7].time_accuracy (lite_presenter_b); docs/lite/results/hosted-api-repeats-20261003/pass2/gpt-5.6-luna-low/analysis.json:scores.per_episode[7].time_accuracy (lite_presenter_b); docs/lite/results/hosted-api-repeats-20261003/pass3/gpt-5.6-luna-low/analysis.json:scores.per_episode[7].time_accuracy (lite_presenter_b)
\newcommand{\LunaPresenterBInForceSeg}{32.7}                            % equal mean across passes: docs/lite/results/four-family/gpt-5.6-luna-low/analysis.json:scores.per_episode[7].segment_time_accuracy (lite_presenter_b); docs/lite/results/hosted-api-repeats-20261003/pass2/gpt-5.6-luna-low/analysis.json:scores.per_episode[7].segment_time_accuracy (lite_presenter_b); docs/lite/results/hosted-api-repeats-20261003/pass3/gpt-5.6-luna-low/analysis.json:scores.per_episode[7].segment_time_accuracy (lite_presenter_b)
\newcommand{\LunaPresenterBErrNoDecisionSec}{2.2}                       % equal mean across passes: docs/lite/results/four-family/gpt-5.6-luna-low/analysis.json:scores.per_episode[7].error_seconds.no_decision (lite_presenter_b); docs/lite/results/hosted-api-repeats-20261003/pass2/gpt-5.6-luna-low/analysis.json:scores.per_episode[7].error_seconds.no_decision (lite_presenter_b); docs/lite/results/hosted-api-repeats-20261003/pass3/gpt-5.6-luna-low/analysis.json:scores.per_episode[7].error_seconds.no_decision (lite_presenter_b)
\newcommand{\LunaPresenterBErrStaleSec}{58.6}                           % equal mean across passes: docs/lite/results/four-family/gpt-5.6-luna-low/analysis.json:scores.per_episode[7].error_seconds.source_correct (lite_presenter_b); docs/lite/results/hosted-api-repeats-20261003/pass2/gpt-5.6-luna-low/analysis.json:scores.per_episode[7].error_seconds.source_correct (lite_presenter_b); docs/lite/results/hosted-api-repeats-20261003/pass3/gpt-5.6-luna-low/analysis.json:scores.per_episode[7].error_seconds.source_correct (lite_presenter_b)
\newcommand{\LunaPresenterBLatencyMedian}{3.09}                         % equal mean across passes: docs/lite/results/four-family/gpt-5.6-luna-low/analysis.json:scores.per_episode[7].latency_s_p50 (lite_presenter_b); docs/lite/results/hosted-api-repeats-20261003/pass2/gpt-5.6-luna-low/analysis.json:scores.per_episode[7].latency_s_p50 (lite_presenter_b); docs/lite/results/hosted-api-repeats-20261003/pass3/gpt-5.6-luna-low/analysis.json:scores.per_episode[7].latency_s_p50 (lite_presenter_b)
\newcommand{\LunaNoneUntimed}{43.5}                                     % equal mean across passes: docs/lite/results/four-family/gpt-5.6-luna-none/analysis.json:scores.overall.untimed_decision_accuracy; docs/lite/results/hosted-api-repeats-20261003/pass2/gpt-5.6-luna-none/analysis.json:scores.overall.untimed_decision_accuracy; docs/lite/results/hosted-api-repeats-20261003/pass3/gpt-5.6-luna-none/analysis.json:scores.overall.untimed_decision_accuracy
\newcommand{\LunaNoneInForce}{33.5}                                     % equal mean across passes: docs/lite/results/four-family/gpt-5.6-luna-none/analysis.json:scores.overall.time_accuracy; docs/lite/results/hosted-api-repeats-20261003/pass2/gpt-5.6-luna-none/analysis.json:scores.overall.time_accuracy; docs/lite/results/hosted-api-repeats-20261003/pass3/gpt-5.6-luna-none/analysis.json:scores.overall.time_accuracy
\newcommand{\LunaNoneInForceSeg}{30.4}                                  % equal mean across passes: docs/lite/results/four-family/gpt-5.6-luna-none/analysis.json:scores.overall.segment_time_accuracy; docs/lite/results/hosted-api-repeats-20261003/pass2/gpt-5.6-luna-none/analysis.json:scores.overall.segment_time_accuracy; docs/lite/results/hosted-api-repeats-20261003/pass3/gpt-5.6-luna-none/analysis.json:scores.overall.segment_time_accuracy
\newcommand{\LunaNoneNetRemovedSeg}{39.5}                               % equal mean across passes: docs/lite/results/four-family/gpt-5.6-luna-none/analysis.json:network_adjustment.scores.estimate.overall.segment_time_accuracy; docs/lite/results/hosted-api-repeats-20261003/pass2/gpt-5.6-luna-none/analysis.json:network_adjustment.scores.estimate.overall.segment_time_accuracy; docs/lite/results/hosted-api-repeats-20261003/pass3/gpt-5.6-luna-none/analysis.json:network_adjustment.scores.estimate.overall.segment_time_accuracy
\newcommand{\LunaNoneNetSec}{1.01}                                      % equal mean across passes: docs/lite/results/four-family/gpt-5.6-luna-none/analysis.json:network_adjustment.network_s.estimate; docs/lite/results/hosted-api-repeats-20261003/pass2/gpt-5.6-luna-none/analysis.json:network_adjustment.network_s.estimate; docs/lite/results/hosted-api-repeats-20261003/pass3/gpt-5.6-luna-none/analysis.json:network_adjustment.network_s.estimate
\newcommand{\LunaNoneNetRemovedSegLow}{36.9}                            % equal mean across passes: docs/lite/results/four-family/gpt-5.6-luna-none/analysis.json:network_adjustment.scores.low.overall.segment_time_accuracy; docs/lite/results/hosted-api-repeats-20261003/pass2/gpt-5.6-luna-none/analysis.json:network_adjustment.scores.low.overall.segment_time_accuracy; docs/lite/results/hosted-api-repeats-20261003/pass3/gpt-5.6-luna-none/analysis.json:network_adjustment.scores.low.overall.segment_time_accuracy
\newcommand{\LunaNoneNetSecLow}{0.72}                                   % equal mean across passes: docs/lite/results/four-family/gpt-5.6-luna-none/analysis.json:network_adjustment.network_s.low; docs/lite/results/hosted-api-repeats-20261003/pass2/gpt-5.6-luna-none/analysis.json:network_adjustment.network_s.low; docs/lite/results/hosted-api-repeats-20261003/pass3/gpt-5.6-luna-none/analysis.json:network_adjustment.network_s.low
\newcommand{\LunaNoneNetRemovedSegHigh}{40.3}                           % equal mean across passes: docs/lite/results/four-family/gpt-5.6-luna-none/analysis.json:network_adjustment.scores.high.overall.segment_time_accuracy; docs/lite/results/hosted-api-repeats-20261003/pass2/gpt-5.6-luna-none/analysis.json:network_adjustment.scores.high.overall.segment_time_accuracy; docs/lite/results/hosted-api-repeats-20261003/pass3/gpt-5.6-luna-none/analysis.json:network_adjustment.scores.high.overall.segment_time_accuracy
\newcommand{\LunaNoneNetSecHigh}{1.10}                                  % equal mean across passes: docs/lite/results/four-family/gpt-5.6-luna-none/analysis.json:network_adjustment.network_s.high; docs/lite/results/hosted-api-repeats-20261003/pass2/gpt-5.6-luna-none/analysis.json:network_adjustment.network_s.high; docs/lite/results/hosted-api-repeats-20261003/pass3/gpt-5.6-luna-none/analysis.json:network_adjustment.network_s.high
\newcommand{\LunaNonePrefillMsPerKTok}{0.0}                             % equal mean across passes: docs/lite/results/four-family/gpt-5.6-luna-none/analysis.json:network_adjustment.prefill_s_per_1k_input_tokens (x 1000); docs/lite/results/hosted-api-repeats-20261003/pass2/gpt-5.6-luna-none/analysis.json:network_adjustment.prefill_s_per_1k_input_tokens (x 1000); docs/lite/results/hosted-api-repeats-20261003/pass3/gpt-5.6-luna-none/analysis.json:network_adjustment.prefill_s_per_1k_input_tokens (x 1000)
\newcommand{\LunaNoneDecodeMsPerTok}{1.6}                               % equal mean across passes: docs/lite/results/four-family/gpt-5.6-luna-none/analysis.json:network_adjustment.decode_s_per_output_token (x 1000); docs/lite/results/hosted-api-repeats-20261003/pass2/gpt-5.6-luna-none/analysis.json:network_adjustment.decode_s_per_output_token (x 1000); docs/lite/results/hosted-api-repeats-20261003/pass3/gpt-5.6-luna-none/analysis.json:network_adjustment.decode_s_per_output_token (x 1000)
\newcommand{\LunaNoneLatencyFloor}{0.91}                                % minimum across passes: docs/lite/results/four-family/gpt-5.6-luna-none/analysis.json:network_adjustment.latency_floor_s; docs/lite/results/hosted-api-repeats-20261003/pass2/gpt-5.6-luna-none/analysis.json:network_adjustment.latency_floor_s; docs/lite/results/hosted-api-repeats-20261003/pass3/gpt-5.6-luna-none/analysis.json:network_adjustment.latency_floor_s
\newcommand{\LunaNoneClampedRequests}{176}                              % sum across passes: docs/lite/results/four-family/gpt-5.6-luna-none/analysis.json:network_adjustment.scores.{estimate,low,high}.clamped_requests (maximum); docs/lite/results/hosted-api-repeats-20261003/pass2/gpt-5.6-luna-none/analysis.json:network_adjustment.scores.{estimate,low,high}.clamped_requests (maximum); docs/lite/results/hosted-api-repeats-20261003/pass3/gpt-5.6-luna-none/analysis.json:network_adjustment.scores.{estimate,low,high}.clamped_requests (maximum)
\newcommand{\LunaNoneLatencyMedian}{1.31}                               % equal mean across passes: docs/lite/results/four-family/gpt-5.6-luna-none/analysis.json:latency_s.p50; docs/lite/results/hosted-api-repeats-20261003/pass2/gpt-5.6-luna-none/analysis.json:latency_s.p50; docs/lite/results/hosted-api-repeats-20261003/pass3/gpt-5.6-luna-none/analysis.json:latency_s.p50
\newcommand{\LunaNoneLatencyPNinetyFive}{1.78}                          % equal mean across passes: docs/lite/results/four-family/gpt-5.6-luna-none/analysis.json:latency_s.p95; docs/lite/results/hosted-api-repeats-20261003/pass2/gpt-5.6-luna-none/analysis.json:latency_s.p95; docs/lite/results/hosted-api-repeats-20261003/pass3/gpt-5.6-luna-none/analysis.json:latency_s.p95
\newcommand{\LunaNoneLatencyMax}{9.99}                                  % maximum across passes: docs/lite/results/four-family/gpt-5.6-luna-none/analysis.json:latency_s.max; docs/lite/results/hosted-api-repeats-20261003/pass2/gpt-5.6-luna-none/analysis.json:latency_s.max; docs/lite/results/hosted-api-repeats-20261003/pass3/gpt-5.6-luna-none/analysis.json:latency_s.max
\newcommand{\LunaNoneLatencyMean}{1.37}                                 % equal mean across passes: runs/lite-v1-gpt-5.6-luna-none-four-family-retry-v1/events.jsonl:response.attempts[-1].completed_s - started_s (mean over 480 responses); runs/hosted-api-repeats-20261003/pass2/gpt-5.6-luna-none/events.jsonl:response.attempts[-1].completed_s - started_s (mean over 480 responses); runs/hosted-api-repeats-20261003/pass3/gpt-5.6-luna-none/events.jsonl:response.attempts[-1].completed_s - started_s (mean over 480 responses)
\newcommand{\LunaNoneInputTokMedian}{3034}                              % equal mean across passes: runs/lite-v1-gpt-5.6-luna-none-four-family-retry-v1/events.jsonl:response.usage.input_tokens (median over 480 responses, half-up); runs/hosted-api-repeats-20261003/pass2/gpt-5.6-luna-none/events.jsonl:response.usage.input_tokens (median over 480 responses, half-up); runs/hosted-api-repeats-20261003/pass3/gpt-5.6-luna-none/events.jsonl:response.usage.input_tokens (median over 480 responses, half-up)
\newcommand{\LunaNoneInputTokTotal}{4{,}429{,}467}                      % sum across passes: docs/lite/results/four-family/gpt-5.6-luna-none/analysis.json:usage.input_tokens; docs/lite/results/hosted-api-repeats-20261003/pass2/gpt-5.6-luna-none/analysis.json:usage.input_tokens; docs/lite/results/hosted-api-repeats-20261003/pass3/gpt-5.6-luna-none/analysis.json:usage.input_tokens
\newcommand{\LunaNoneOutputTokMedian}{46}                               % equal mean across passes: runs/lite-v1-gpt-5.6-luna-none-four-family-retry-v1/events.jsonl:response.usage.output_tokens (median over 480 responses, half-up); runs/hosted-api-repeats-20261003/pass2/gpt-5.6-luna-none/events.jsonl:response.usage.output_tokens (median over 480 responses, half-up); runs/hosted-api-repeats-20261003/pass3/gpt-5.6-luna-none/events.jsonl:response.usage.output_tokens (median over 480 responses, half-up)
\newcommand{\LunaNoneOutputTokTotal}{64{,}440}                          % sum across passes: docs/lite/results/four-family/gpt-5.6-luna-none/analysis.json:usage.output_tokens; docs/lite/results/hosted-api-repeats-20261003/pass2/gpt-5.6-luna-none/analysis.json:usage.output_tokens; docs/lite/results/hosted-api-repeats-20261003/pass3/gpt-5.6-luna-none/analysis.json:usage.output_tokens
\newcommand{\LunaNoneReasoningTokMedian}{0}                             % equal mean across passes: runs/lite-v1-gpt-5.6-luna-none-four-family-retry-v1/events.jsonl:response.usage.reasoning_tokens (median over 480 responses, half-up); runs/hosted-api-repeats-20261003/pass2/gpt-5.6-luna-none/events.jsonl:response.usage.reasoning_tokens (median over 480 responses, half-up); runs/hosted-api-repeats-20261003/pass3/gpt-5.6-luna-none/events.jsonl:response.usage.reasoning_tokens (median over 480 responses, half-up)
\newcommand{\LunaNoneCachedTokTotal}{0}                                 % sum across passes: docs/lite/results/four-family/gpt-5.6-luna-none/analysis.json:usage.cached_tokens; docs/lite/results/hosted-api-repeats-20261003/pass2/gpt-5.6-luna-none/analysis.json:usage.cached_tokens; docs/lite/results/hosted-api-repeats-20261003/pass3/gpt-5.6-luna-none/analysis.json:usage.cached_tokens
\newcommand{\LunaNoneCostInputUSD}{0.886}                               % sum across passes: docs/lite/results/four-family/gpt-5.6-luna-none/analysis.json:usage at developers.openai.com/api/docs/pricing (gpt-5.6-luna, Standard, short context) list prices retrieved 2026-09-29 (input: uncached x input price + cached x cached price); docs/lite/results/hosted-api-repeats-20261003/pass2/gpt-5.6-luna-none/analysis.json:usage at developers.openai.com/api/docs/pricing (gpt-5.6-luna, Standard, short context) list prices retrieved 2026-09-29 (input: uncached x input price + cached x cached price); docs/lite/results/hosted-api-repeats-20261003/pass3/gpt-5.6-luna-none/analysis.json:usage at developers.openai.com/api/docs/pricing (gpt-5.6-luna, Standard, short context) list prices retrieved 2026-09-29 (input: uncached x input price + cached x cached price)
\newcommand{\LunaNoneCostOutputUSD}{0.077}                              % sum across passes: docs/lite/results/four-family/gpt-5.6-luna-none/analysis.json:usage at developers.openai.com/api/docs/pricing (gpt-5.6-luna, Standard, short context) list prices retrieved 2026-09-29 (output tokens, reasoning included, x output price); docs/lite/results/hosted-api-repeats-20261003/pass2/gpt-5.6-luna-none/analysis.json:usage at developers.openai.com/api/docs/pricing (gpt-5.6-luna, Standard, short context) list prices retrieved 2026-09-29 (output tokens, reasoning included, x output price); docs/lite/results/hosted-api-repeats-20261003/pass3/gpt-5.6-luna-none/analysis.json:usage at developers.openai.com/api/docs/pricing (gpt-5.6-luna, Standard, short context) list prices retrieved 2026-09-29 (output tokens, reasoning included, x output price)
\newcommand{\LunaNoneCostUSD}{0.963}                                    % sum across passes: docs/lite/results/four-family/gpt-5.6-luna-none/analysis.json:usage at developers.openai.com/api/docs/pricing (gpt-5.6-luna, Standard, short context) list prices retrieved 2026-09-29 (input + output); docs/lite/results/hosted-api-repeats-20261003/pass2/gpt-5.6-luna-none/analysis.json:usage at developers.openai.com/api/docs/pricing (gpt-5.6-luna, Standard, short context) list prices retrieved 2026-09-29 (input + output); docs/lite/results/hosted-api-repeats-20261003/pass3/gpt-5.6-luna-none/analysis.json:usage at developers.openai.com/api/docs/pricing (gpt-5.6-luna, Standard, short context) list prices retrieved 2026-09-29 (input + output)
\newcommand{\LunaNonePriceInput}{0.200}                                 % equal mean across passes: developers.openai.com/api/docs/pricing (gpt-5.6-luna, Standard, short context): USD per 1M input tokens; developers.openai.com/api/docs/pricing (gpt-5.6-luna, Standard, short context): USD per 1M input tokens; developers.openai.com/api/docs/pricing (gpt-5.6-luna, Standard, short context): USD per 1M input tokens
\newcommand{\LunaNonePriceOutput}{1.20}                                 % equal mean across passes: developers.openai.com/api/docs/pricing (gpt-5.6-luna, Standard, short context): USD per 1M output tokens; developers.openai.com/api/docs/pricing (gpt-5.6-luna, Standard, short context): USD per 1M output tokens; developers.openai.com/api/docs/pricing (gpt-5.6-luna, Standard, short context): USD per 1M output tokens
\newcommand{\LunaNoneErrNoDecisionPct}{1.2}                             % equal mean across passes: docs/lite/results/four-family/gpt-5.6-luna-none/analysis.json:error_seconds.no_decision (scenario error fraction, then equal family mean); docs/lite/results/hosted-api-repeats-20261003/pass2/gpt-5.6-luna-none/analysis.json:error_seconds.no_decision (scenario error fraction, then equal family mean); docs/lite/results/hosted-api-repeats-20261003/pass3/gpt-5.6-luna-none/analysis.json:error_seconds.no_decision (scenario error fraction, then equal family mean)
\newcommand{\LunaNoneErrStalePct}{10.7}                                 % equal mean across passes: docs/lite/results/four-family/gpt-5.6-luna-none/analysis.json:error_seconds.source_correct (scenario error fraction, then equal family mean); docs/lite/results/hosted-api-repeats-20261003/pass2/gpt-5.6-luna-none/analysis.json:error_seconds.source_correct (scenario error fraction, then equal family mean); docs/lite/results/hosted-api-repeats-20261003/pass3/gpt-5.6-luna-none/analysis.json:error_seconds.source_correct (scenario error fraction, then equal family mean)
\newcommand{\LunaNoneInForceRecorded}{33.5}                             % equal mean across passes: docs/lite/results/four-family/gpt-5.6-luna-none/analysis.json:scores.overall.time_accuracy; docs/lite/results/hosted-api-repeats-20261003/pass2/gpt-5.6-luna-none/analysis.json:scores.overall.time_accuracy; docs/lite/results/hosted-api-repeats-20261003/pass3/gpt-5.6-luna-none/analysis.json:scores.overall.time_accuracy
\newcommand{\LunaNoneInForcePhysical}{33.5}                             % equal mean across passes: docs/lite/results/four-family/gpt-5.6-luna-none/analysis.json:raw_wallclock_scores.overall.time_accuracy (physical wall-clock trace); docs/lite/results/hosted-api-repeats-20261003/pass2/gpt-5.6-luna-none/analysis.json:raw_wallclock_scores.overall.time_accuracy (physical wall-clock trace); docs/lite/results/hosted-api-repeats-20261003/pass3/gpt-5.6-luna-none/analysis.json:raw_wallclock_scores.overall.time_accuracy (physical wall-clock trace)
\newcommand{\LunaNoneExcludedDispatchSec}{0.24}                         % equal mean across passes: docs/lite/results/four-family/gpt-5.6-luna-none/analysis.json:retry_reliability.excluded_dispatch_s (summed over the pass); docs/lite/results/hosted-api-repeats-20261003/pass2/gpt-5.6-luna-none/analysis.json:retry_reliability.excluded_dispatch_s (summed over the pass); docs/lite/results/hosted-api-repeats-20261003/pass3/gpt-5.6-luna-none/analysis.json:retry_reliability.excluded_dispatch_s (summed over the pass)
\newcommand{\LunaNoneEventsHash}{67913b2242cd8535}                      % pass 1: docs/lite/results/four-family/gpt-5.6-luna-none/analysis.json:events_sha256 (first 16 hex digits; equals sha256 of runs/lite-v1-gpt-5.6-luna-none-four-family-retry-v1/events.jsonl)
\newcommand{\LunaNoneAcceptedUpdates}{1434}                             % sum across passes: docs/lite/results/four-family/gpt-5.6-luna-none/analysis.json:accepted_updates; docs/lite/results/hosted-api-repeats-20261003/pass2/gpt-5.6-luna-none/analysis.json:accepted_updates; docs/lite/results/hosted-api-repeats-20261003/pass3/gpt-5.6-luna-none/analysis.json:accepted_updates
\newcommand{\LunaNoneDiscardedAfterHorizon}{0}                          % sum across passes: docs/lite/results/four-family/gpt-5.6-luna-none/analysis.json:discarded_updates.after_horizon (absent = 0); docs/lite/results/hosted-api-repeats-20261003/pass2/gpt-5.6-luna-none/analysis.json:discarded_updates.after_horizon (absent = 0); docs/lite/results/hosted-api-repeats-20261003/pass3/gpt-5.6-luna-none/analysis.json:discarded_updates.after_horizon (absent = 0)
\newcommand{\LunaNoneDiscardedOlder}{6}                                 % sum across passes: docs/lite/results/four-family/gpt-5.6-luna-none/analysis.json:discarded_updates.older_than_active (absent = 0); docs/lite/results/hosted-api-repeats-20261003/pass2/gpt-5.6-luna-none/analysis.json:discarded_updates.older_than_active (absent = 0); docs/lite/results/hosted-api-repeats-20261003/pass3/gpt-5.6-luna-none/analysis.json:discarded_updates.older_than_active (absent = 0)
\newcommand{\LunaNoneValidResponses}{1440}                              % sum across passes: docs/lite/results/four-family/gpt-5.6-luna-none/analysis.json:retry_reliability.successful_logical_requests; docs/lite/results/hosted-api-repeats-20261003/pass2/gpt-5.6-luna-none/analysis.json:retry_reliability.successful_logical_requests; docs/lite/results/hosted-api-repeats-20261003/pass3/gpt-5.6-luna-none/analysis.json:retry_reliability.successful_logical_requests
\newcommand{\LunaNoneAttempts}{1440}                                    % sum across passes: docs/lite/results/four-family/gpt-5.6-luna-none/analysis.json:retry_reliability.attempts; docs/lite/results/hosted-api-repeats-20261003/pass2/gpt-5.6-luna-none/analysis.json:retry_reliability.attempts; docs/lite/results/hosted-api-repeats-20261003/pass3/gpt-5.6-luna-none/analysis.json:retry_reliability.attempts
\newcommand{\LunaNoneFailedAttempts}{0}                                 % sum across passes: docs/lite/results/four-family/gpt-5.6-luna-none/analysis.json:retry_reliability.failed_attempts; docs/lite/results/hosted-api-repeats-20261003/pass2/gpt-5.6-luna-none/analysis.json:retry_reliability.failed_attempts; docs/lite/results/hosted-api-repeats-20261003/pass3/gpt-5.6-luna-none/analysis.json:retry_reliability.failed_attempts
\newcommand{\LunaNoneRetriedRequests}{0}                                % sum across passes: docs/lite/results/four-family/gpt-5.6-luna-none/analysis.json:retry_reliability.retried_logical_requests; docs/lite/results/hosted-api-repeats-20261003/pass2/gpt-5.6-luna-none/analysis.json:retry_reliability.retried_logical_requests; docs/lite/results/hosted-api-repeats-20261003/pass3/gpt-5.6-luna-none/analysis.json:retry_reliability.retried_logical_requests
\newcommand{\LunaNoneDebugUntimed}{17.8}                                % equal mean across passes: docs/lite/results/four-family/gpt-5.6-luna-none/analysis.json:scores.by_family.live_debugging.untimed_decision_accuracy; docs/lite/results/hosted-api-repeats-20261003/pass2/gpt-5.6-luna-none/analysis.json:scores.by_family.live_debugging.untimed_decision_accuracy; docs/lite/results/hosted-api-repeats-20261003/pass3/gpt-5.6-luna-none/analysis.json:scores.by_family.live_debugging.untimed_decision_accuracy
\newcommand{\LunaNoneDebugInForce}{14.4}                                % equal mean across passes: docs/lite/results/four-family/gpt-5.6-luna-none/analysis.json:scores.by_family.live_debugging.time_accuracy; docs/lite/results/hosted-api-repeats-20261003/pass2/gpt-5.6-luna-none/analysis.json:scores.by_family.live_debugging.time_accuracy; docs/lite/results/hosted-api-repeats-20261003/pass3/gpt-5.6-luna-none/analysis.json:scores.by_family.live_debugging.time_accuracy
\newcommand{\LunaNoneDebugNetRemoved}{16.8}                             % equal mean across passes: docs/lite/results/four-family/gpt-5.6-luna-none/analysis.json:network_adjustment.scores.estimate.by_family.live_debugging.time_accuracy; docs/lite/results/hosted-api-repeats-20261003/pass2/gpt-5.6-luna-none/analysis.json:network_adjustment.scores.estimate.by_family.live_debugging.time_accuracy; docs/lite/results/hosted-api-repeats-20261003/pass3/gpt-5.6-luna-none/analysis.json:network_adjustment.scores.estimate.by_family.live_debugging.time_accuracy
\newcommand{\LunaNoneDebugNetRemovedLow}{16.2}                          % equal mean across passes: docs/lite/results/four-family/gpt-5.6-luna-none/analysis.json:network_adjustment.scores.low.by_family.live_debugging.time_accuracy; docs/lite/results/hosted-api-repeats-20261003/pass2/gpt-5.6-luna-none/analysis.json:network_adjustment.scores.low.by_family.live_debugging.time_accuracy; docs/lite/results/hosted-api-repeats-20261003/pass3/gpt-5.6-luna-none/analysis.json:network_adjustment.scores.low.by_family.live_debugging.time_accuracy
\newcommand{\LunaNoneDebugNetRemovedHigh}{17.0}                         % equal mean across passes: docs/lite/results/four-family/gpt-5.6-luna-none/analysis.json:network_adjustment.scores.high.by_family.live_debugging.time_accuracy; docs/lite/results/hosted-api-repeats-20261003/pass2/gpt-5.6-luna-none/analysis.json:network_adjustment.scores.high.by_family.live_debugging.time_accuracy; docs/lite/results/hosted-api-repeats-20261003/pass3/gpt-5.6-luna-none/analysis.json:network_adjustment.scores.high.by_family.live_debugging.time_accuracy
\newcommand{\LunaNoneAssemblyUntimed}{28.9}                             % equal mean across passes: docs/lite/results/four-family/gpt-5.6-luna-none/analysis.json:scores.by_family.procedural_coaching.untimed_decision_accuracy; docs/lite/results/hosted-api-repeats-20261003/pass2/gpt-5.6-luna-none/analysis.json:scores.by_family.procedural_coaching.untimed_decision_accuracy; docs/lite/results/hosted-api-repeats-20261003/pass3/gpt-5.6-luna-none/analysis.json:scores.by_family.procedural_coaching.untimed_decision_accuracy
\newcommand{\LunaNoneAssemblyInForce}{21.4}                             % equal mean across passes: docs/lite/results/four-family/gpt-5.6-luna-none/analysis.json:scores.by_family.procedural_coaching.time_accuracy; docs/lite/results/hosted-api-repeats-20261003/pass2/gpt-5.6-luna-none/analysis.json:scores.by_family.procedural_coaching.time_accuracy; docs/lite/results/hosted-api-repeats-20261003/pass3/gpt-5.6-luna-none/analysis.json:scores.by_family.procedural_coaching.time_accuracy
\newcommand{\LunaNoneAssemblyNetRemoved}{26.6}                          % equal mean across passes: docs/lite/results/four-family/gpt-5.6-luna-none/analysis.json:network_adjustment.scores.estimate.by_family.procedural_coaching.time_accuracy; docs/lite/results/hosted-api-repeats-20261003/pass2/gpt-5.6-luna-none/analysis.json:network_adjustment.scores.estimate.by_family.procedural_coaching.time_accuracy; docs/lite/results/hosted-api-repeats-20261003/pass3/gpt-5.6-luna-none/analysis.json:network_adjustment.scores.estimate.by_family.procedural_coaching.time_accuracy
\newcommand{\LunaNoneAssemblyNetRemovedLow}{25.1}                       % equal mean across passes: docs/lite/results/four-family/gpt-5.6-luna-none/analysis.json:network_adjustment.scores.low.by_family.procedural_coaching.time_accuracy; docs/lite/results/hosted-api-repeats-20261003/pass2/gpt-5.6-luna-none/analysis.json:network_adjustment.scores.low.by_family.procedural_coaching.time_accuracy; docs/lite/results/hosted-api-repeats-20261003/pass3/gpt-5.6-luna-none/analysis.json:network_adjustment.scores.low.by_family.procedural_coaching.time_accuracy
\newcommand{\LunaNoneAssemblyNetRemovedHigh}{27.1}                      % equal mean across passes: docs/lite/results/four-family/gpt-5.6-luna-none/analysis.json:network_adjustment.scores.high.by_family.procedural_coaching.time_accuracy; docs/lite/results/hosted-api-repeats-20261003/pass2/gpt-5.6-luna-none/analysis.json:network_adjustment.scores.high.by_family.procedural_coaching.time_accuracy; docs/lite/results/hosted-api-repeats-20261003/pass3/gpt-5.6-luna-none/analysis.json:network_adjustment.scores.high.by_family.procedural_coaching.time_accuracy
\newcommand{\LunaNoneSupportUntimed}{65.0}                              % equal mean across passes: docs/lite/results/four-family/gpt-5.6-luna-none/analysis.json:scores.by_family.support_call_assist.untimed_decision_accuracy; docs/lite/results/hosted-api-repeats-20261003/pass2/gpt-5.6-luna-none/analysis.json:scores.by_family.support_call_assist.untimed_decision_accuracy; docs/lite/results/hosted-api-repeats-20261003/pass3/gpt-5.6-luna-none/analysis.json:scores.by_family.support_call_assist.untimed_decision_accuracy
\newcommand{\LunaNoneSupportInForce}{48.0}                              % equal mean across passes: docs/lite/results/four-family/gpt-5.6-luna-none/analysis.json:scores.by_family.support_call_assist.time_accuracy; docs/lite/results/hosted-api-repeats-20261003/pass2/gpt-5.6-luna-none/analysis.json:scores.by_family.support_call_assist.time_accuracy; docs/lite/results/hosted-api-repeats-20261003/pass3/gpt-5.6-luna-none/analysis.json:scores.by_family.support_call_assist.time_accuracy
\newcommand{\LunaNoneSupportNetRemoved}{59.9}                           % equal mean across passes: docs/lite/results/four-family/gpt-5.6-luna-none/analysis.json:network_adjustment.scores.estimate.by_family.support_call_assist.time_accuracy; docs/lite/results/hosted-api-repeats-20261003/pass2/gpt-5.6-luna-none/analysis.json:network_adjustment.scores.estimate.by_family.support_call_assist.time_accuracy; docs/lite/results/hosted-api-repeats-20261003/pass3/gpt-5.6-luna-none/analysis.json:network_adjustment.scores.estimate.by_family.support_call_assist.time_accuracy
\newcommand{\LunaNoneSupportNetRemovedLow}{56.4}                        % equal mean across passes: docs/lite/results/four-family/gpt-5.6-luna-none/analysis.json:network_adjustment.scores.low.by_family.support_call_assist.time_accuracy; docs/lite/results/hosted-api-repeats-20261003/pass2/gpt-5.6-luna-none/analysis.json:network_adjustment.scores.low.by_family.support_call_assist.time_accuracy; docs/lite/results/hosted-api-repeats-20261003/pass3/gpt-5.6-luna-none/analysis.json:network_adjustment.scores.low.by_family.support_call_assist.time_accuracy
\newcommand{\LunaNoneSupportNetRemovedHigh}{60.9}                       % equal mean across passes: docs/lite/results/four-family/gpt-5.6-luna-none/analysis.json:network_adjustment.scores.high.by_family.support_call_assist.time_accuracy; docs/lite/results/hosted-api-repeats-20261003/pass2/gpt-5.6-luna-none/analysis.json:network_adjustment.scores.high.by_family.support_call_assist.time_accuracy; docs/lite/results/hosted-api-repeats-20261003/pass3/gpt-5.6-luna-none/analysis.json:network_adjustment.scores.high.by_family.support_call_assist.time_accuracy
\newcommand{\LunaNonePresenterUntimed}{62.5}                            % equal mean across passes: docs/lite/results/four-family/gpt-5.6-luna-none/analysis.json:scores.by_family.presenter_voice_control.untimed_decision_accuracy; docs/lite/results/hosted-api-repeats-20261003/pass2/gpt-5.6-luna-none/analysis.json:scores.by_family.presenter_voice_control.untimed_decision_accuracy; docs/lite/results/hosted-api-repeats-20261003/pass3/gpt-5.6-luna-none/analysis.json:scores.by_family.presenter_voice_control.untimed_decision_accuracy
\newcommand{\LunaNonePresenterInForce}{50.1}                            % equal mean across passes: docs/lite/results/four-family/gpt-5.6-luna-none/analysis.json:scores.by_family.presenter_voice_control.time_accuracy; docs/lite/results/hosted-api-repeats-20261003/pass2/gpt-5.6-luna-none/analysis.json:scores.by_family.presenter_voice_control.time_accuracy; docs/lite/results/hosted-api-repeats-20261003/pass3/gpt-5.6-luna-none/analysis.json:scores.by_family.presenter_voice_control.time_accuracy
\newcommand{\LunaNonePresenterNetRemoved}{59.6}                         % equal mean across passes: docs/lite/results/four-family/gpt-5.6-luna-none/analysis.json:network_adjustment.scores.estimate.by_family.presenter_voice_control.time_accuracy; docs/lite/results/hosted-api-repeats-20261003/pass2/gpt-5.6-luna-none/analysis.json:network_adjustment.scores.estimate.by_family.presenter_voice_control.time_accuracy; docs/lite/results/hosted-api-repeats-20261003/pass3/gpt-5.6-luna-none/analysis.json:network_adjustment.scores.estimate.by_family.presenter_voice_control.time_accuracy
\newcommand{\LunaNonePresenterNetRemovedLow}{56.8}                      % equal mean across passes: docs/lite/results/four-family/gpt-5.6-luna-none/analysis.json:network_adjustment.scores.low.by_family.presenter_voice_control.time_accuracy; docs/lite/results/hosted-api-repeats-20261003/pass2/gpt-5.6-luna-none/analysis.json:network_adjustment.scores.low.by_family.presenter_voice_control.time_accuracy; docs/lite/results/hosted-api-repeats-20261003/pass3/gpt-5.6-luna-none/analysis.json:network_adjustment.scores.low.by_family.presenter_voice_control.time_accuracy
\newcommand{\LunaNonePresenterNetRemovedHigh}{60.4}                     % equal mean across passes: docs/lite/results/four-family/gpt-5.6-luna-none/analysis.json:network_adjustment.scores.high.by_family.presenter_voice_control.time_accuracy; docs/lite/results/hosted-api-repeats-20261003/pass2/gpt-5.6-luna-none/analysis.json:network_adjustment.scores.high.by_family.presenter_voice_control.time_accuracy; docs/lite/results/hosted-api-repeats-20261003/pass3/gpt-5.6-luna-none/analysis.json:network_adjustment.scores.high.by_family.presenter_voice_control.time_accuracy
\newcommand{\LunaNoneDebugAUntimed}{15.0}                               % equal mean across passes: docs/lite/results/four-family/gpt-5.6-luna-none/analysis.json:scores.per_episode[0].untimed_decision_accuracy (lite_debugging_a); docs/lite/results/hosted-api-repeats-20261003/pass2/gpt-5.6-luna-none/analysis.json:scores.per_episode[0].untimed_decision_accuracy (lite_debugging_a); docs/lite/results/hosted-api-repeats-20261003/pass3/gpt-5.6-luna-none/analysis.json:scores.per_episode[0].untimed_decision_accuracy (lite_debugging_a)
\newcommand{\LunaNoneDebugAInForce}{11.4}                               % equal mean across passes: docs/lite/results/four-family/gpt-5.6-luna-none/analysis.json:scores.per_episode[0].time_accuracy (lite_debugging_a); docs/lite/results/hosted-api-repeats-20261003/pass2/gpt-5.6-luna-none/analysis.json:scores.per_episode[0].time_accuracy (lite_debugging_a); docs/lite/results/hosted-api-repeats-20261003/pass3/gpt-5.6-luna-none/analysis.json:scores.per_episode[0].time_accuracy (lite_debugging_a)
\newcommand{\LunaNoneDebugAInForceSeg}{11.2}                            % equal mean across passes: docs/lite/results/four-family/gpt-5.6-luna-none/analysis.json:scores.per_episode[0].segment_time_accuracy (lite_debugging_a); docs/lite/results/hosted-api-repeats-20261003/pass2/gpt-5.6-luna-none/analysis.json:scores.per_episode[0].segment_time_accuracy (lite_debugging_a); docs/lite/results/hosted-api-repeats-20261003/pass3/gpt-5.6-luna-none/analysis.json:scores.per_episode[0].segment_time_accuracy (lite_debugging_a)
\newcommand{\LunaNoneDebugAErrNoDecisionSec}{1.9}                       % equal mean across passes: docs/lite/results/four-family/gpt-5.6-luna-none/analysis.json:scores.per_episode[0].error_seconds.no_decision (lite_debugging_a); docs/lite/results/hosted-api-repeats-20261003/pass2/gpt-5.6-luna-none/analysis.json:scores.per_episode[0].error_seconds.no_decision (lite_debugging_a); docs/lite/results/hosted-api-repeats-20261003/pass3/gpt-5.6-luna-none/analysis.json:scores.per_episode[0].error_seconds.no_decision (lite_debugging_a)
\newcommand{\LunaNoneDebugAErrStaleSec}{3.7}                            % equal mean across passes: docs/lite/results/four-family/gpt-5.6-luna-none/analysis.json:scores.per_episode[0].error_seconds.source_correct (lite_debugging_a); docs/lite/results/hosted-api-repeats-20261003/pass2/gpt-5.6-luna-none/analysis.json:scores.per_episode[0].error_seconds.source_correct (lite_debugging_a); docs/lite/results/hosted-api-repeats-20261003/pass3/gpt-5.6-luna-none/analysis.json:scores.per_episode[0].error_seconds.source_correct (lite_debugging_a)
\newcommand{\LunaNoneDebugALatencyMedian}{1.26}                         % equal mean across passes: docs/lite/results/four-family/gpt-5.6-luna-none/analysis.json:scores.per_episode[0].latency_s_p50 (lite_debugging_a); docs/lite/results/hosted-api-repeats-20261003/pass2/gpt-5.6-luna-none/analysis.json:scores.per_episode[0].latency_s_p50 (lite_debugging_a); docs/lite/results/hosted-api-repeats-20261003/pass3/gpt-5.6-luna-none/analysis.json:scores.per_episode[0].latency_s_p50 (lite_debugging_a)
\newcommand{\LunaNoneDebugBUntimed}{20.6}                               % equal mean across passes: docs/lite/results/four-family/gpt-5.6-luna-none/analysis.json:scores.per_episode[1].untimed_decision_accuracy (lite_debugging_b); docs/lite/results/hosted-api-repeats-20261003/pass2/gpt-5.6-luna-none/analysis.json:scores.per_episode[1].untimed_decision_accuracy (lite_debugging_b); docs/lite/results/hosted-api-repeats-20261003/pass3/gpt-5.6-luna-none/analysis.json:scores.per_episode[1].untimed_decision_accuracy (lite_debugging_b)
\newcommand{\LunaNoneDebugBInForce}{17.5}                               % equal mean across passes: docs/lite/results/four-family/gpt-5.6-luna-none/analysis.json:scores.per_episode[1].time_accuracy (lite_debugging_b); docs/lite/results/hosted-api-repeats-20261003/pass2/gpt-5.6-luna-none/analysis.json:scores.per_episode[1].time_accuracy (lite_debugging_b); docs/lite/results/hosted-api-repeats-20261003/pass3/gpt-5.6-luna-none/analysis.json:scores.per_episode[1].time_accuracy (lite_debugging_b)
\newcommand{\LunaNoneDebugBInForceSeg}{14.4}                            % equal mean across passes: docs/lite/results/four-family/gpt-5.6-luna-none/analysis.json:scores.per_episode[1].segment_time_accuracy (lite_debugging_b); docs/lite/results/hosted-api-repeats-20261003/pass2/gpt-5.6-luna-none/analysis.json:scores.per_episode[1].segment_time_accuracy (lite_debugging_b); docs/lite/results/hosted-api-repeats-20261003/pass3/gpt-5.6-luna-none/analysis.json:scores.per_episode[1].segment_time_accuracy (lite_debugging_b)
\newcommand{\LunaNoneDebugBErrNoDecisionSec}{1.4}                       % equal mean across passes: docs/lite/results/four-family/gpt-5.6-luna-none/analysis.json:scores.per_episode[1].error_seconds.no_decision (lite_debugging_b); docs/lite/results/hosted-api-repeats-20261003/pass2/gpt-5.6-luna-none/analysis.json:scores.per_episode[1].error_seconds.no_decision (lite_debugging_b); docs/lite/results/hosted-api-repeats-20261003/pass3/gpt-5.6-luna-none/analysis.json:scores.per_episode[1].error_seconds.no_decision (lite_debugging_b)
\newcommand{\LunaNoneDebugBErrStaleSec}{4.0}                            % equal mean across passes: docs/lite/results/four-family/gpt-5.6-luna-none/analysis.json:scores.per_episode[1].error_seconds.source_correct (lite_debugging_b); docs/lite/results/hosted-api-repeats-20261003/pass2/gpt-5.6-luna-none/analysis.json:scores.per_episode[1].error_seconds.source_correct (lite_debugging_b); docs/lite/results/hosted-api-repeats-20261003/pass3/gpt-5.6-luna-none/analysis.json:scores.per_episode[1].error_seconds.source_correct (lite_debugging_b)
\newcommand{\LunaNoneDebugBLatencyMedian}{1.28}                         % equal mean across passes: docs/lite/results/four-family/gpt-5.6-luna-none/analysis.json:scores.per_episode[1].latency_s_p50 (lite_debugging_b); docs/lite/results/hosted-api-repeats-20261003/pass2/gpt-5.6-luna-none/analysis.json:scores.per_episode[1].latency_s_p50 (lite_debugging_b); docs/lite/results/hosted-api-repeats-20261003/pass3/gpt-5.6-luna-none/analysis.json:scores.per_episode[1].latency_s_p50 (lite_debugging_b)
\newcommand{\LunaNoneAssemblyAUntimed}{28.3}                            % equal mean across passes: docs/lite/results/four-family/gpt-5.6-luna-none/analysis.json:scores.per_episode[2].untimed_decision_accuracy (lite_assembly_a); docs/lite/results/hosted-api-repeats-20261003/pass2/gpt-5.6-luna-none/analysis.json:scores.per_episode[2].untimed_decision_accuracy (lite_assembly_a); docs/lite/results/hosted-api-repeats-20261003/pass3/gpt-5.6-luna-none/analysis.json:scores.per_episode[2].untimed_decision_accuracy (lite_assembly_a)
\newcommand{\LunaNoneAssemblyAInForce}{21.8}                            % equal mean across passes: docs/lite/results/four-family/gpt-5.6-luna-none/analysis.json:scores.per_episode[2].time_accuracy (lite_assembly_a); docs/lite/results/hosted-api-repeats-20261003/pass2/gpt-5.6-luna-none/analysis.json:scores.per_episode[2].time_accuracy (lite_assembly_a); docs/lite/results/hosted-api-repeats-20261003/pass3/gpt-5.6-luna-none/analysis.json:scores.per_episode[2].time_accuracy (lite_assembly_a)
\newcommand{\LunaNoneAssemblyAInForceSeg}{20.5}                         % equal mean across passes: docs/lite/results/four-family/gpt-5.6-luna-none/analysis.json:scores.per_episode[2].segment_time_accuracy (lite_assembly_a); docs/lite/results/hosted-api-repeats-20261003/pass2/gpt-5.6-luna-none/analysis.json:scores.per_episode[2].segment_time_accuracy (lite_assembly_a); docs/lite/results/hosted-api-repeats-20261003/pass3/gpt-5.6-luna-none/analysis.json:scores.per_episode[2].segment_time_accuracy (lite_assembly_a)
\newcommand{\LunaNoneAssemblyAErrNoDecisionSec}{1.4}                    % equal mean across passes: docs/lite/results/four-family/gpt-5.6-luna-none/analysis.json:scores.per_episode[2].error_seconds.no_decision (lite_assembly_a); docs/lite/results/hosted-api-repeats-20261003/pass2/gpt-5.6-luna-none/analysis.json:scores.per_episode[2].error_seconds.no_decision (lite_assembly_a); docs/lite/results/hosted-api-repeats-20261003/pass3/gpt-5.6-luna-none/analysis.json:scores.per_episode[2].error_seconds.no_decision (lite_assembly_a)
\newcommand{\LunaNoneAssemblyAErrStaleSec}{9.5}                         % equal mean across passes: docs/lite/results/four-family/gpt-5.6-luna-none/analysis.json:scores.per_episode[2].error_seconds.source_correct (lite_assembly_a); docs/lite/results/hosted-api-repeats-20261003/pass2/gpt-5.6-luna-none/analysis.json:scores.per_episode[2].error_seconds.source_correct (lite_assembly_a); docs/lite/results/hosted-api-repeats-20261003/pass3/gpt-5.6-luna-none/analysis.json:scores.per_episode[2].error_seconds.source_correct (lite_assembly_a)
\newcommand{\LunaNoneAssemblyALatencyMedian}{1.39}                      % equal mean across passes: docs/lite/results/four-family/gpt-5.6-luna-none/analysis.json:scores.per_episode[2].latency_s_p50 (lite_assembly_a); docs/lite/results/hosted-api-repeats-20261003/pass2/gpt-5.6-luna-none/analysis.json:scores.per_episode[2].latency_s_p50 (lite_assembly_a); docs/lite/results/hosted-api-repeats-20261003/pass3/gpt-5.6-luna-none/analysis.json:scores.per_episode[2].latency_s_p50 (lite_assembly_a)
\newcommand{\LunaNoneAssemblyBUntimed}{29.4}                            % equal mean across passes: docs/lite/results/four-family/gpt-5.6-luna-none/analysis.json:scores.per_episode[3].untimed_decision_accuracy (lite_assembly_b); docs/lite/results/hosted-api-repeats-20261003/pass2/gpt-5.6-luna-none/analysis.json:scores.per_episode[3].untimed_decision_accuracy (lite_assembly_b); docs/lite/results/hosted-api-repeats-20261003/pass3/gpt-5.6-luna-none/analysis.json:scores.per_episode[3].untimed_decision_accuracy (lite_assembly_b)
\newcommand{\LunaNoneAssemblyBInForce}{21.0}                            % equal mean across passes: docs/lite/results/four-family/gpt-5.6-luna-none/analysis.json:scores.per_episode[3].time_accuracy (lite_assembly_b); docs/lite/results/hosted-api-repeats-20261003/pass2/gpt-5.6-luna-none/analysis.json:scores.per_episode[3].time_accuracy (lite_assembly_b); docs/lite/results/hosted-api-repeats-20261003/pass3/gpt-5.6-luna-none/analysis.json:scores.per_episode[3].time_accuracy (lite_assembly_b)
\newcommand{\LunaNoneAssemblyBInForceSeg}{21.3}                         % equal mean across passes: docs/lite/results/four-family/gpt-5.6-luna-none/analysis.json:scores.per_episode[3].segment_time_accuracy (lite_assembly_b); docs/lite/results/hosted-api-repeats-20261003/pass2/gpt-5.6-luna-none/analysis.json:scores.per_episode[3].segment_time_accuracy (lite_assembly_b); docs/lite/results/hosted-api-repeats-20261003/pass3/gpt-5.6-luna-none/analysis.json:scores.per_episode[3].segment_time_accuracy (lite_assembly_b)
\newcommand{\LunaNoneAssemblyBErrNoDecisionSec}{1.4}                    % equal mean across passes: docs/lite/results/four-family/gpt-5.6-luna-none/analysis.json:scores.per_episode[3].error_seconds.no_decision (lite_assembly_b); docs/lite/results/hosted-api-repeats-20261003/pass2/gpt-5.6-luna-none/analysis.json:scores.per_episode[3].error_seconds.no_decision (lite_assembly_b); docs/lite/results/hosted-api-repeats-20261003/pass3/gpt-5.6-luna-none/analysis.json:scores.per_episode[3].error_seconds.no_decision (lite_assembly_b)
\newcommand{\LunaNoneAssemblyBErrStaleSec}{9.8}                         % equal mean across passes: docs/lite/results/four-family/gpt-5.6-luna-none/analysis.json:scores.per_episode[3].error_seconds.source_correct (lite_assembly_b); docs/lite/results/hosted-api-repeats-20261003/pass2/gpt-5.6-luna-none/analysis.json:scores.per_episode[3].error_seconds.source_correct (lite_assembly_b); docs/lite/results/hosted-api-repeats-20261003/pass3/gpt-5.6-luna-none/analysis.json:scores.per_episode[3].error_seconds.source_correct (lite_assembly_b)
\newcommand{\LunaNoneAssemblyBLatencyMedian}{1.33}                      % equal mean across passes: docs/lite/results/four-family/gpt-5.6-luna-none/analysis.json:scores.per_episode[3].latency_s_p50 (lite_assembly_b); docs/lite/results/hosted-api-repeats-20261003/pass2/gpt-5.6-luna-none/analysis.json:scores.per_episode[3].latency_s_p50 (lite_assembly_b); docs/lite/results/hosted-api-repeats-20261003/pass3/gpt-5.6-luna-none/analysis.json:scores.per_episode[3].latency_s_p50 (lite_assembly_b)
\newcommand{\LunaNoneSupportAUntimed}{59.4}                             % equal mean across passes: docs/lite/results/four-family/gpt-5.6-luna-none/analysis.json:scores.per_episode[4].untimed_decision_accuracy (lite_support_a); docs/lite/results/hosted-api-repeats-20261003/pass2/gpt-5.6-luna-none/analysis.json:scores.per_episode[4].untimed_decision_accuracy (lite_support_a); docs/lite/results/hosted-api-repeats-20261003/pass3/gpt-5.6-luna-none/analysis.json:scores.per_episode[4].untimed_decision_accuracy (lite_support_a)
\newcommand{\LunaNoneSupportAInForce}{42.8}                             % equal mean across passes: docs/lite/results/four-family/gpt-5.6-luna-none/analysis.json:scores.per_episode[4].time_accuracy (lite_support_a); docs/lite/results/hosted-api-repeats-20261003/pass2/gpt-5.6-luna-none/analysis.json:scores.per_episode[4].time_accuracy (lite_support_a); docs/lite/results/hosted-api-repeats-20261003/pass3/gpt-5.6-luna-none/analysis.json:scores.per_episode[4].time_accuracy (lite_support_a)
\newcommand{\LunaNoneSupportAInForceSeg}{37.3}                          % equal mean across passes: docs/lite/results/four-family/gpt-5.6-luna-none/analysis.json:scores.per_episode[4].segment_time_accuracy (lite_support_a); docs/lite/results/hosted-api-repeats-20261003/pass2/gpt-5.6-luna-none/analysis.json:scores.per_episode[4].segment_time_accuracy (lite_support_a); docs/lite/results/hosted-api-repeats-20261003/pass3/gpt-5.6-luna-none/analysis.json:scores.per_episode[4].segment_time_accuracy (lite_support_a)
\newcommand{\LunaNoneSupportAErrNoDecisionSec}{1.6}                     % equal mean across passes: docs/lite/results/four-family/gpt-5.6-luna-none/analysis.json:scores.per_episode[4].error_seconds.no_decision (lite_support_a); docs/lite/results/hosted-api-repeats-20261003/pass2/gpt-5.6-luna-none/analysis.json:scores.per_episode[4].error_seconds.no_decision (lite_support_a); docs/lite/results/hosted-api-repeats-20261003/pass3/gpt-5.6-luna-none/analysis.json:scores.per_episode[4].error_seconds.no_decision (lite_support_a)
\newcommand{\LunaNoneSupportAErrStaleSec}{18.1}                         % equal mean across passes: docs/lite/results/four-family/gpt-5.6-luna-none/analysis.json:scores.per_episode[4].error_seconds.source_correct (lite_support_a); docs/lite/results/hosted-api-repeats-20261003/pass2/gpt-5.6-luna-none/analysis.json:scores.per_episode[4].error_seconds.source_correct (lite_support_a); docs/lite/results/hosted-api-repeats-20261003/pass3/gpt-5.6-luna-none/analysis.json:scores.per_episode[4].error_seconds.source_correct (lite_support_a)
\newcommand{\LunaNoneSupportALatencyMedian}{1.43}                       % equal mean across passes: docs/lite/results/four-family/gpt-5.6-luna-none/analysis.json:scores.per_episode[4].latency_s_p50 (lite_support_a); docs/lite/results/hosted-api-repeats-20261003/pass2/gpt-5.6-luna-none/analysis.json:scores.per_episode[4].latency_s_p50 (lite_support_a); docs/lite/results/hosted-api-repeats-20261003/pass3/gpt-5.6-luna-none/analysis.json:scores.per_episode[4].latency_s_p50 (lite_support_a)
\newcommand{\LunaNoneSupportBUntimed}{70.6}                             % equal mean across passes: docs/lite/results/four-family/gpt-5.6-luna-none/analysis.json:scores.per_episode[5].untimed_decision_accuracy (lite_support_b); docs/lite/results/hosted-api-repeats-20261003/pass2/gpt-5.6-luna-none/analysis.json:scores.per_episode[5].untimed_decision_accuracy (lite_support_b); docs/lite/results/hosted-api-repeats-20261003/pass3/gpt-5.6-luna-none/analysis.json:scores.per_episode[5].untimed_decision_accuracy (lite_support_b)
\newcommand{\LunaNoneSupportBInForce}{53.2}                             % equal mean across passes: docs/lite/results/four-family/gpt-5.6-luna-none/analysis.json:scores.per_episode[5].time_accuracy (lite_support_b); docs/lite/results/hosted-api-repeats-20261003/pass2/gpt-5.6-luna-none/analysis.json:scores.per_episode[5].time_accuracy (lite_support_b); docs/lite/results/hosted-api-repeats-20261003/pass3/gpt-5.6-luna-none/analysis.json:scores.per_episode[5].time_accuracy (lite_support_b)
\newcommand{\LunaNoneSupportBInForceSeg}{51.5}                          % equal mean across passes: docs/lite/results/four-family/gpt-5.6-luna-none/analysis.json:scores.per_episode[5].segment_time_accuracy (lite_support_b); docs/lite/results/hosted-api-repeats-20261003/pass2/gpt-5.6-luna-none/analysis.json:scores.per_episode[5].segment_time_accuracy (lite_support_b); docs/lite/results/hosted-api-repeats-20261003/pass3/gpt-5.6-luna-none/analysis.json:scores.per_episode[5].segment_time_accuracy (lite_support_b)
\newcommand{\LunaNoneSupportBErrNoDecisionSec}{1.5}                     % equal mean across passes: docs/lite/results/four-family/gpt-5.6-luna-none/analysis.json:scores.per_episode[5].error_seconds.no_decision (lite_support_b); docs/lite/results/hosted-api-repeats-20261003/pass2/gpt-5.6-luna-none/analysis.json:scores.per_episode[5].error_seconds.no_decision (lite_support_b); docs/lite/results/hosted-api-repeats-20261003/pass3/gpt-5.6-luna-none/analysis.json:scores.per_episode[5].error_seconds.no_decision (lite_support_b)
\newcommand{\LunaNoneSupportBErrStaleSec}{22.9}                         % equal mean across passes: docs/lite/results/four-family/gpt-5.6-luna-none/analysis.json:scores.per_episode[5].error_seconds.source_correct (lite_support_b); docs/lite/results/hosted-api-repeats-20261003/pass2/gpt-5.6-luna-none/analysis.json:scores.per_episode[5].error_seconds.source_correct (lite_support_b); docs/lite/results/hosted-api-repeats-20261003/pass3/gpt-5.6-luna-none/analysis.json:scores.per_episode[5].error_seconds.source_correct (lite_support_b)
\newcommand{\LunaNoneSupportBLatencyMedian}{1.38}                       % equal mean across passes: docs/lite/results/four-family/gpt-5.6-luna-none/analysis.json:scores.per_episode[5].latency_s_p50 (lite_support_b); docs/lite/results/hosted-api-repeats-20261003/pass2/gpt-5.6-luna-none/analysis.json:scores.per_episode[5].latency_s_p50 (lite_support_b); docs/lite/results/hosted-api-repeats-20261003/pass3/gpt-5.6-luna-none/analysis.json:scores.per_episode[5].latency_s_p50 (lite_support_b)
\newcommand{\LunaNonePresenterAUntimed}{73.3}                           % equal mean across passes: docs/lite/results/four-family/gpt-5.6-luna-none/analysis.json:scores.per_episode[6].untimed_decision_accuracy (lite_presenter_a); docs/lite/results/hosted-api-repeats-20261003/pass2/gpt-5.6-luna-none/analysis.json:scores.per_episode[6].untimed_decision_accuracy (lite_presenter_a); docs/lite/results/hosted-api-repeats-20261003/pass3/gpt-5.6-luna-none/analysis.json:scores.per_episode[6].untimed_decision_accuracy (lite_presenter_a)
\newcommand{\LunaNonePresenterAInForce}{59.2}                           % equal mean across passes: docs/lite/results/four-family/gpt-5.6-luna-none/analysis.json:scores.per_episode[6].time_accuracy (lite_presenter_a); docs/lite/results/hosted-api-repeats-20261003/pass2/gpt-5.6-luna-none/analysis.json:scores.per_episode[6].time_accuracy (lite_presenter_a); docs/lite/results/hosted-api-repeats-20261003/pass3/gpt-5.6-luna-none/analysis.json:scores.per_episode[6].time_accuracy (lite_presenter_a)
\newcommand{\LunaNonePresenterAInForceSeg}{48.7}                        % equal mean across passes: docs/lite/results/four-family/gpt-5.6-luna-none/analysis.json:scores.per_episode[6].segment_time_accuracy (lite_presenter_a); docs/lite/results/hosted-api-repeats-20261003/pass2/gpt-5.6-luna-none/analysis.json:scores.per_episode[6].segment_time_accuracy (lite_presenter_a); docs/lite/results/hosted-api-repeats-20261003/pass3/gpt-5.6-luna-none/analysis.json:scores.per_episode[6].segment_time_accuracy (lite_presenter_a)
\newcommand{\LunaNonePresenterAErrNoDecisionSec}{1.4}                   % equal mean across passes: docs/lite/results/four-family/gpt-5.6-luna-none/analysis.json:scores.per_episode[6].error_seconds.no_decision (lite_presenter_a); docs/lite/results/hosted-api-repeats-20261003/pass2/gpt-5.6-luna-none/analysis.json:scores.per_episode[6].error_seconds.no_decision (lite_presenter_a); docs/lite/results/hosted-api-repeats-20261003/pass3/gpt-5.6-luna-none/analysis.json:scores.per_episode[6].error_seconds.no_decision (lite_presenter_a)
\newcommand{\LunaNonePresenterAErrStaleSec}{20.7}                       % equal mean across passes: docs/lite/results/four-family/gpt-5.6-luna-none/analysis.json:scores.per_episode[6].error_seconds.source_correct (lite_presenter_a); docs/lite/results/hosted-api-repeats-20261003/pass2/gpt-5.6-luna-none/analysis.json:scores.per_episode[6].error_seconds.source_correct (lite_presenter_a); docs/lite/results/hosted-api-repeats-20261003/pass3/gpt-5.6-luna-none/analysis.json:scores.per_episode[6].error_seconds.source_correct (lite_presenter_a)
\newcommand{\LunaNonePresenterALatencyMedian}{1.20}                     % equal mean across passes: docs/lite/results/four-family/gpt-5.6-luna-none/analysis.json:scores.per_episode[6].latency_s_p50 (lite_presenter_a); docs/lite/results/hosted-api-repeats-20261003/pass2/gpt-5.6-luna-none/analysis.json:scores.per_episode[6].latency_s_p50 (lite_presenter_a); docs/lite/results/hosted-api-repeats-20261003/pass3/gpt-5.6-luna-none/analysis.json:scores.per_episode[6].latency_s_p50 (lite_presenter_a)
\newcommand{\LunaNonePresenterBUntimed}{51.7}                           % equal mean across passes: docs/lite/results/four-family/gpt-5.6-luna-none/analysis.json:scores.per_episode[7].untimed_decision_accuracy (lite_presenter_b); docs/lite/results/hosted-api-repeats-20261003/pass2/gpt-5.6-luna-none/analysis.json:scores.per_episode[7].untimed_decision_accuracy (lite_presenter_b); docs/lite/results/hosted-api-repeats-20261003/pass3/gpt-5.6-luna-none/analysis.json:scores.per_episode[7].untimed_decision_accuracy (lite_presenter_b)
\newcommand{\LunaNonePresenterBInForce}{40.9}                           % equal mean across passes: docs/lite/results/four-family/gpt-5.6-luna-none/analysis.json:scores.per_episode[7].time_accuracy (lite_presenter_b); docs/lite/results/hosted-api-repeats-20261003/pass2/gpt-5.6-luna-none/analysis.json:scores.per_episode[7].time_accuracy (lite_presenter_b); docs/lite/results/hosted-api-repeats-20261003/pass3/gpt-5.6-luna-none/analysis.json:scores.per_episode[7].time_accuracy (lite_presenter_b)
\newcommand{\LunaNonePresenterBInForceSeg}{38.7}                        % equal mean across passes: docs/lite/results/four-family/gpt-5.6-luna-none/analysis.json:scores.per_episode[7].segment_time_accuracy (lite_presenter_b); docs/lite/results/hosted-api-repeats-20261003/pass2/gpt-5.6-luna-none/analysis.json:scores.per_episode[7].segment_time_accuracy (lite_presenter_b); docs/lite/results/hosted-api-repeats-20261003/pass3/gpt-5.6-luna-none/analysis.json:scores.per_episode[7].segment_time_accuracy (lite_presenter_b)
\newcommand{\LunaNonePresenterBErrNoDecisionSec}{1.4}                   % equal mean across passes: docs/lite/results/four-family/gpt-5.6-luna-none/analysis.json:scores.per_episode[7].error_seconds.no_decision (lite_presenter_b); docs/lite/results/hosted-api-repeats-20261003/pass2/gpt-5.6-luna-none/analysis.json:scores.per_episode[7].error_seconds.no_decision (lite_presenter_b); docs/lite/results/hosted-api-repeats-20261003/pass3/gpt-5.6-luna-none/analysis.json:scores.per_episode[7].error_seconds.no_decision (lite_presenter_b)
\newcommand{\LunaNonePresenterBErrStaleSec}{14.3}                       % equal mean across passes: docs/lite/results/four-family/gpt-5.6-luna-none/analysis.json:scores.per_episode[7].error_seconds.source_correct (lite_presenter_b); docs/lite/results/hosted-api-repeats-20261003/pass2/gpt-5.6-luna-none/analysis.json:scores.per_episode[7].error_seconds.source_correct (lite_presenter_b); docs/lite/results/hosted-api-repeats-20261003/pass3/gpt-5.6-luna-none/analysis.json:scores.per_episode[7].error_seconds.source_correct (lite_presenter_b)
\newcommand{\LunaNonePresenterBLatencyMedian}{1.24}                     % equal mean across passes: docs/lite/results/four-family/gpt-5.6-luna-none/analysis.json:scores.per_episode[7].latency_s_p50 (lite_presenter_b); docs/lite/results/hosted-api-repeats-20261003/pass2/gpt-5.6-luna-none/analysis.json:scores.per_episode[7].latency_s_p50 (lite_presenter_b); docs/lite/results/hosted-api-repeats-20261003/pass3/gpt-5.6-luna-none/analysis.json:scores.per_episode[7].latency_s_p50 (lite_presenter_b)
\newcommand{\TerraUntimed}{95.8}                                        % equal mean across passes: docs/lite/results/four-family/gpt-5.6-terra-low/analysis.json:scores.overall.untimed_decision_accuracy; docs/lite/results/hosted-api-repeats-20261003/pass2/gpt-5.6-terra-low/analysis.json:scores.overall.untimed_decision_accuracy; docs/lite/results/hosted-api-repeats-20261003/pass3/gpt-5.6-terra-low/analysis.json:scores.overall.untimed_decision_accuracy
\newcommand{\TerraInForce}{55.9}                                        % equal mean across passes: docs/lite/results/four-family/gpt-5.6-terra-low/analysis.json:scores.overall.time_accuracy; docs/lite/results/hosted-api-repeats-20261003/pass2/gpt-5.6-terra-low/analysis.json:scores.overall.time_accuracy; docs/lite/results/hosted-api-repeats-20261003/pass3/gpt-5.6-terra-low/analysis.json:scores.overall.time_accuracy
\newcommand{\TerraInForceSeg}{47.2}                                     % equal mean across passes: docs/lite/results/four-family/gpt-5.6-terra-low/analysis.json:scores.overall.segment_time_accuracy; docs/lite/results/hosted-api-repeats-20261003/pass2/gpt-5.6-terra-low/analysis.json:scores.overall.segment_time_accuracy; docs/lite/results/hosted-api-repeats-20261003/pass3/gpt-5.6-terra-low/analysis.json:scores.overall.segment_time_accuracy
\newcommand{\TerraNetRemovedSeg}{59.9}                                  % equal mean across passes: docs/lite/results/four-family/gpt-5.6-terra-low/analysis.json:network_adjustment.scores.estimate.overall.segment_time_accuracy; docs/lite/results/hosted-api-repeats-20261003/pass2/gpt-5.6-terra-low/analysis.json:network_adjustment.scores.estimate.overall.segment_time_accuracy; docs/lite/results/hosted-api-repeats-20261003/pass3/gpt-5.6-terra-low/analysis.json:network_adjustment.scores.estimate.overall.segment_time_accuracy
\newcommand{\TerraNetSec}{0.68}                                         % equal mean across passes: docs/lite/results/four-family/gpt-5.6-terra-low/analysis.json:network_adjustment.network_s.estimate; docs/lite/results/hosted-api-repeats-20261003/pass2/gpt-5.6-terra-low/analysis.json:network_adjustment.network_s.estimate; docs/lite/results/hosted-api-repeats-20261003/pass3/gpt-5.6-terra-low/analysis.json:network_adjustment.network_s.estimate
\newcommand{\TerraNetRemovedSegLow}{56.9}                               % equal mean across passes: docs/lite/results/four-family/gpt-5.6-terra-low/analysis.json:network_adjustment.scores.low.overall.segment_time_accuracy; docs/lite/results/hosted-api-repeats-20261003/pass2/gpt-5.6-terra-low/analysis.json:network_adjustment.scores.low.overall.segment_time_accuracy; docs/lite/results/hosted-api-repeats-20261003/pass3/gpt-5.6-terra-low/analysis.json:network_adjustment.scores.low.overall.segment_time_accuracy
\newcommand{\TerraNetSecLow}{0.52}                                      % equal mean across passes: docs/lite/results/four-family/gpt-5.6-terra-low/analysis.json:network_adjustment.network_s.low; docs/lite/results/hosted-api-repeats-20261003/pass2/gpt-5.6-terra-low/analysis.json:network_adjustment.network_s.low; docs/lite/results/hosted-api-repeats-20261003/pass3/gpt-5.6-terra-low/analysis.json:network_adjustment.network_s.low
\newcommand{\TerraNetRemovedSegHigh}{61.5}                              % equal mean across passes: docs/lite/results/four-family/gpt-5.6-terra-low/analysis.json:network_adjustment.scores.high.overall.segment_time_accuracy; docs/lite/results/hosted-api-repeats-20261003/pass2/gpt-5.6-terra-low/analysis.json:network_adjustment.scores.high.overall.segment_time_accuracy; docs/lite/results/hosted-api-repeats-20261003/pass3/gpt-5.6-terra-low/analysis.json:network_adjustment.scores.high.overall.segment_time_accuracy
\newcommand{\TerraNetSecHigh}{0.76}                                     % equal mean across passes: docs/lite/results/four-family/gpt-5.6-terra-low/analysis.json:network_adjustment.network_s.high; docs/lite/results/hosted-api-repeats-20261003/pass2/gpt-5.6-terra-low/analysis.json:network_adjustment.network_s.high; docs/lite/results/hosted-api-repeats-20261003/pass3/gpt-5.6-terra-low/analysis.json:network_adjustment.network_s.high
\newcommand{\TerraPrefillMsPerKTok}{0.0}                                % equal mean across passes: docs/lite/results/four-family/gpt-5.6-terra-low/analysis.json:network_adjustment.prefill_s_per_1k_input_tokens (x 1000); docs/lite/results/hosted-api-repeats-20261003/pass2/gpt-5.6-terra-low/analysis.json:network_adjustment.prefill_s_per_1k_input_tokens (x 1000); docs/lite/results/hosted-api-repeats-20261003/pass3/gpt-5.6-terra-low/analysis.json:network_adjustment.prefill_s_per_1k_input_tokens (x 1000)
\newcommand{\TerraDecodeMsPerTok}{10.1}                                 % equal mean across passes: docs/lite/results/four-family/gpt-5.6-terra-low/analysis.json:network_adjustment.decode_s_per_output_token (x 1000); docs/lite/results/hosted-api-repeats-20261003/pass2/gpt-5.6-terra-low/analysis.json:network_adjustment.decode_s_per_output_token (x 1000); docs/lite/results/hosted-api-repeats-20261003/pass3/gpt-5.6-terra-low/analysis.json:network_adjustment.decode_s_per_output_token (x 1000)
\newcommand{\TerraLatencyFloor}{0.98}                                   % minimum across passes: docs/lite/results/four-family/gpt-5.6-terra-low/analysis.json:network_adjustment.latency_floor_s; docs/lite/results/hosted-api-repeats-20261003/pass2/gpt-5.6-terra-low/analysis.json:network_adjustment.latency_floor_s; docs/lite/results/hosted-api-repeats-20261003/pass3/gpt-5.6-terra-low/analysis.json:network_adjustment.latency_floor_s
\newcommand{\TerraLatencyMedian}{2.13}                                  % equal mean across passes: docs/lite/results/four-family/gpt-5.6-terra-low/analysis.json:latency_s.p50; docs/lite/results/hosted-api-repeats-20261003/pass2/gpt-5.6-terra-low/analysis.json:latency_s.p50; docs/lite/results/hosted-api-repeats-20261003/pass3/gpt-5.6-terra-low/analysis.json:latency_s.p50
\newcommand{\TerraLatencyPNinetyFive}{3.25}                             % equal mean across passes: docs/lite/results/four-family/gpt-5.6-terra-low/analysis.json:latency_s.p95; docs/lite/results/hosted-api-repeats-20261003/pass2/gpt-5.6-terra-low/analysis.json:latency_s.p95; docs/lite/results/hosted-api-repeats-20261003/pass3/gpt-5.6-terra-low/analysis.json:latency_s.p95
\newcommand{\TerraLatencyMax}{10.71}                                    % maximum across passes: docs/lite/results/four-family/gpt-5.6-terra-low/analysis.json:latency_s.max; docs/lite/results/hosted-api-repeats-20261003/pass2/gpt-5.6-terra-low/analysis.json:latency_s.max; docs/lite/results/hosted-api-repeats-20261003/pass3/gpt-5.6-terra-low/analysis.json:latency_s.max
\newcommand{\TerraLatencyMean}{2.22}                                    % equal mean across passes: runs/lite-v1-gpt-5.6-terra-low-four-family-retry-v1/events.jsonl:response.attempts[-1].completed_s - started_s (mean over 480 responses); runs/hosted-api-repeats-20261003/pass2/gpt-5.6-terra-low/events.jsonl:response.attempts[-1].completed_s - started_s (mean over 480 responses); runs/hosted-api-repeats-20261003/pass3/gpt-5.6-terra-low/events.jsonl:response.attempts[-1].completed_s - started_s (mean over 480 responses)
\newcommand{\TerraInputTokMedian}{3034}                                 % equal mean across passes: runs/lite-v1-gpt-5.6-terra-low-four-family-retry-v1/events.jsonl:response.usage.input_tokens (median over 480 responses, half-up); runs/hosted-api-repeats-20261003/pass2/gpt-5.6-terra-low/events.jsonl:response.usage.input_tokens (median over 480 responses, half-up); runs/hosted-api-repeats-20261003/pass3/gpt-5.6-terra-low/events.jsonl:response.usage.input_tokens (median over 480 responses, half-up)
\newcommand{\TerraInputTokTotal}{4{,}429{,}467}                         % sum across passes: docs/lite/results/four-family/gpt-5.6-terra-low/analysis.json:usage.input_tokens; docs/lite/results/hosted-api-repeats-20261003/pass2/gpt-5.6-terra-low/analysis.json:usage.input_tokens; docs/lite/results/hosted-api-repeats-20261003/pass3/gpt-5.6-terra-low/analysis.json:usage.input_tokens
\newcommand{\TerraOutputTokMedian}{116}                                 % equal mean across passes: runs/lite-v1-gpt-5.6-terra-low-four-family-retry-v1/events.jsonl:response.usage.output_tokens (median over 480 responses, half-up); runs/hosted-api-repeats-20261003/pass2/gpt-5.6-terra-low/events.jsonl:response.usage.output_tokens (median over 480 responses, half-up); runs/hosted-api-repeats-20261003/pass3/gpt-5.6-terra-low/events.jsonl:response.usage.output_tokens (median over 480 responses, half-up)
\newcommand{\TerraOutputTokTotal}{173{,}080}                            % sum across passes: docs/lite/results/four-family/gpt-5.6-terra-low/analysis.json:usage.output_tokens; docs/lite/results/hosted-api-repeats-20261003/pass2/gpt-5.6-terra-low/analysis.json:usage.output_tokens; docs/lite/results/hosted-api-repeats-20261003/pass3/gpt-5.6-terra-low/analysis.json:usage.output_tokens
\newcommand{\TerraReasoningTokMedian}{68}                               % equal mean across passes: runs/lite-v1-gpt-5.6-terra-low-four-family-retry-v1/events.jsonl:response.usage.reasoning_tokens (median over 480 responses, half-up); runs/hosted-api-repeats-20261003/pass2/gpt-5.6-terra-low/events.jsonl:response.usage.reasoning_tokens (median over 480 responses, half-up); runs/hosted-api-repeats-20261003/pass3/gpt-5.6-terra-low/events.jsonl:response.usage.reasoning_tokens (median over 480 responses, half-up)
\newcommand{\TerraCachedTokTotal}{0}                                    % sum across passes: docs/lite/results/four-family/gpt-5.6-terra-low/analysis.json:usage.cached_tokens; docs/lite/results/hosted-api-repeats-20261003/pass2/gpt-5.6-terra-low/analysis.json:usage.cached_tokens; docs/lite/results/hosted-api-repeats-20261003/pass3/gpt-5.6-terra-low/analysis.json:usage.cached_tokens
\newcommand{\TerraCostInputUSD}{8.859}                                  % sum across passes: docs/lite/results/four-family/gpt-5.6-terra-low/analysis.json:usage at developers.openai.com/api/docs/pricing (gpt-5.6-terra, Standard, short context) list prices retrieved 2026-09-29 (input: uncached x input price + cached x cached price); docs/lite/results/hosted-api-repeats-20261003/pass2/gpt-5.6-terra-low/analysis.json:usage at developers.openai.com/api/docs/pricing (gpt-5.6-terra, Standard, short context) list prices retrieved 2026-09-29 (input: uncached x input price + cached x cached price); docs/lite/results/hosted-api-repeats-20261003/pass3/gpt-5.6-terra-low/analysis.json:usage at developers.openai.com/api/docs/pricing (gpt-5.6-terra, Standard, short context) list prices retrieved 2026-09-29 (input: uncached x input price + cached x cached price)
\newcommand{\TerraCostOutputUSD}{2.077}                                 % sum across passes: docs/lite/results/four-family/gpt-5.6-terra-low/analysis.json:usage at developers.openai.com/api/docs/pricing (gpt-5.6-terra, Standard, short context) list prices retrieved 2026-09-29 (output tokens, reasoning included, x output price); docs/lite/results/hosted-api-repeats-20261003/pass2/gpt-5.6-terra-low/analysis.json:usage at developers.openai.com/api/docs/pricing (gpt-5.6-terra, Standard, short context) list prices retrieved 2026-09-29 (output tokens, reasoning included, x output price); docs/lite/results/hosted-api-repeats-20261003/pass3/gpt-5.6-terra-low/analysis.json:usage at developers.openai.com/api/docs/pricing (gpt-5.6-terra, Standard, short context) list prices retrieved 2026-09-29 (output tokens, reasoning included, x output price)
\newcommand{\TerraCostUSD}{10.936}                                      % sum across passes: docs/lite/results/four-family/gpt-5.6-terra-low/analysis.json:usage at developers.openai.com/api/docs/pricing (gpt-5.6-terra, Standard, short context) list prices retrieved 2026-09-29 (input + output); docs/lite/results/hosted-api-repeats-20261003/pass2/gpt-5.6-terra-low/analysis.json:usage at developers.openai.com/api/docs/pricing (gpt-5.6-terra, Standard, short context) list prices retrieved 2026-09-29 (input + output); docs/lite/results/hosted-api-repeats-20261003/pass3/gpt-5.6-terra-low/analysis.json:usage at developers.openai.com/api/docs/pricing (gpt-5.6-terra, Standard, short context) list prices retrieved 2026-09-29 (input + output)
\newcommand{\TerraPriceInput}{2.000}                                    % equal mean across passes: developers.openai.com/api/docs/pricing (gpt-5.6-terra, Standard, short context): USD per 1M input tokens; developers.openai.com/api/docs/pricing (gpt-5.6-terra, Standard, short context): USD per 1M input tokens; developers.openai.com/api/docs/pricing (gpt-5.6-terra, Standard, short context): USD per 1M input tokens
\newcommand{\TerraPriceOutput}{12.00}                                   % equal mean across passes: developers.openai.com/api/docs/pricing (gpt-5.6-terra, Standard, short context): USD per 1M output tokens; developers.openai.com/api/docs/pricing (gpt-5.6-terra, Standard, short context): USD per 1M output tokens; developers.openai.com/api/docs/pricing (gpt-5.6-terra, Standard, short context): USD per 1M output tokens
\newcommand{\TerraErrNoDecisionPct}{1.7}                                % equal mean across passes: docs/lite/results/four-family/gpt-5.6-terra-low/analysis.json:error_seconds.no_decision (scenario error fraction, then equal family mean); docs/lite/results/hosted-api-repeats-20261003/pass2/gpt-5.6-terra-low/analysis.json:error_seconds.no_decision (scenario error fraction, then equal family mean); docs/lite/results/hosted-api-repeats-20261003/pass3/gpt-5.6-terra-low/analysis.json:error_seconds.no_decision (scenario error fraction, then equal family mean)
\newcommand{\TerraErrStalePct}{38.5}                                    % equal mean across passes: docs/lite/results/four-family/gpt-5.6-terra-low/analysis.json:error_seconds.source_correct (scenario error fraction, then equal family mean); docs/lite/results/hosted-api-repeats-20261003/pass2/gpt-5.6-terra-low/analysis.json:error_seconds.source_correct (scenario error fraction, then equal family mean); docs/lite/results/hosted-api-repeats-20261003/pass3/gpt-5.6-terra-low/analysis.json:error_seconds.source_correct (scenario error fraction, then equal family mean)
\newcommand{\TerraInForceRecorded}{55.9}                                % equal mean across passes: docs/lite/results/four-family/gpt-5.6-terra-low/analysis.json:scores.overall.time_accuracy; docs/lite/results/hosted-api-repeats-20261003/pass2/gpt-5.6-terra-low/analysis.json:scores.overall.time_accuracy; docs/lite/results/hosted-api-repeats-20261003/pass3/gpt-5.6-terra-low/analysis.json:scores.overall.time_accuracy
\newcommand{\TerraInForcePhysical}{55.9}                                % equal mean across passes: docs/lite/results/four-family/gpt-5.6-terra-low/analysis.json:raw_wallclock_scores.overall.time_accuracy (physical wall-clock trace); docs/lite/results/hosted-api-repeats-20261003/pass2/gpt-5.6-terra-low/analysis.json:raw_wallclock_scores.overall.time_accuracy (physical wall-clock trace); docs/lite/results/hosted-api-repeats-20261003/pass3/gpt-5.6-terra-low/analysis.json:raw_wallclock_scores.overall.time_accuracy (physical wall-clock trace)
\newcommand{\TerraExcludedDispatchSec}{0.25}                            % equal mean across passes: docs/lite/results/four-family/gpt-5.6-terra-low/analysis.json:retry_reliability.excluded_dispatch_s (summed over the pass); docs/lite/results/hosted-api-repeats-20261003/pass2/gpt-5.6-terra-low/analysis.json:retry_reliability.excluded_dispatch_s (summed over the pass); docs/lite/results/hosted-api-repeats-20261003/pass3/gpt-5.6-terra-low/analysis.json:retry_reliability.excluded_dispatch_s (summed over the pass)
\newcommand{\TerraEventsHash}{770f5d2195a80fd5}                         % pass 1: docs/lite/results/four-family/gpt-5.6-terra-low/analysis.json:events_sha256 (first 16 hex digits; equals sha256 of runs/lite-v1-gpt-5.6-terra-low-four-family-retry-v1/events.jsonl)
\newcommand{\TerraAcceptedUpdates}{1416}                                % sum across passes: docs/lite/results/four-family/gpt-5.6-terra-low/analysis.json:accepted_updates; docs/lite/results/hosted-api-repeats-20261003/pass2/gpt-5.6-terra-low/analysis.json:accepted_updates; docs/lite/results/hosted-api-repeats-20261003/pass3/gpt-5.6-terra-low/analysis.json:accepted_updates
\newcommand{\TerraDiscardedAfterHorizon}{14}                            % sum across passes: docs/lite/results/four-family/gpt-5.6-terra-low/analysis.json:discarded_updates.after_horizon (absent = 0); docs/lite/results/hosted-api-repeats-20261003/pass2/gpt-5.6-terra-low/analysis.json:discarded_updates.after_horizon (absent = 0); docs/lite/results/hosted-api-repeats-20261003/pass3/gpt-5.6-terra-low/analysis.json:discarded_updates.after_horizon (absent = 0)
\newcommand{\TerraDiscardedOlder}{10}                                   % sum across passes: docs/lite/results/four-family/gpt-5.6-terra-low/analysis.json:discarded_updates.older_than_active (absent = 0); docs/lite/results/hosted-api-repeats-20261003/pass2/gpt-5.6-terra-low/analysis.json:discarded_updates.older_than_active (absent = 0); docs/lite/results/hosted-api-repeats-20261003/pass3/gpt-5.6-terra-low/analysis.json:discarded_updates.older_than_active (absent = 0)
\newcommand{\TerraValidResponses}{1440}                                 % sum across passes: docs/lite/results/four-family/gpt-5.6-terra-low/analysis.json:retry_reliability.successful_logical_requests; docs/lite/results/hosted-api-repeats-20261003/pass2/gpt-5.6-terra-low/analysis.json:retry_reliability.successful_logical_requests; docs/lite/results/hosted-api-repeats-20261003/pass3/gpt-5.6-terra-low/analysis.json:retry_reliability.successful_logical_requests
\newcommand{\TerraAttempts}{1440}                                       % sum across passes: docs/lite/results/four-family/gpt-5.6-terra-low/analysis.json:retry_reliability.attempts; docs/lite/results/hosted-api-repeats-20261003/pass2/gpt-5.6-terra-low/analysis.json:retry_reliability.attempts; docs/lite/results/hosted-api-repeats-20261003/pass3/gpt-5.6-terra-low/analysis.json:retry_reliability.attempts
\newcommand{\TerraFailedAttempts}{0}                                    % sum across passes: docs/lite/results/four-family/gpt-5.6-terra-low/analysis.json:retry_reliability.failed_attempts; docs/lite/results/hosted-api-repeats-20261003/pass2/gpt-5.6-terra-low/analysis.json:retry_reliability.failed_attempts; docs/lite/results/hosted-api-repeats-20261003/pass3/gpt-5.6-terra-low/analysis.json:retry_reliability.failed_attempts
\newcommand{\TerraRetriedRequests}{0}                                   % sum across passes: docs/lite/results/four-family/gpt-5.6-terra-low/analysis.json:retry_reliability.retried_logical_requests; docs/lite/results/hosted-api-repeats-20261003/pass2/gpt-5.6-terra-low/analysis.json:retry_reliability.retried_logical_requests; docs/lite/results/hosted-api-repeats-20261003/pass3/gpt-5.6-terra-low/analysis.json:retry_reliability.retried_logical_requests
\newcommand{\TerraDebugUntimed}{99.2}                                   % equal mean across passes: docs/lite/results/four-family/gpt-5.6-terra-low/analysis.json:scores.by_family.live_debugging.untimed_decision_accuracy; docs/lite/results/hosted-api-repeats-20261003/pass2/gpt-5.6-terra-low/analysis.json:scores.by_family.live_debugging.untimed_decision_accuracy; docs/lite/results/hosted-api-repeats-20261003/pass3/gpt-5.6-terra-low/analysis.json:scores.by_family.live_debugging.untimed_decision_accuracy
\newcommand{\TerraDebugInForce}{57.5}                                   % equal mean across passes: docs/lite/results/four-family/gpt-5.6-terra-low/analysis.json:scores.by_family.live_debugging.time_accuracy; docs/lite/results/hosted-api-repeats-20261003/pass2/gpt-5.6-terra-low/analysis.json:scores.by_family.live_debugging.time_accuracy; docs/lite/results/hosted-api-repeats-20261003/pass3/gpt-5.6-terra-low/analysis.json:scores.by_family.live_debugging.time_accuracy
\newcommand{\TerraDebugNetRemoved}{69.2}                                % equal mean across passes: docs/lite/results/four-family/gpt-5.6-terra-low/analysis.json:network_adjustment.scores.estimate.by_family.live_debugging.time_accuracy; docs/lite/results/hosted-api-repeats-20261003/pass2/gpt-5.6-terra-low/analysis.json:network_adjustment.scores.estimate.by_family.live_debugging.time_accuracy; docs/lite/results/hosted-api-repeats-20261003/pass3/gpt-5.6-terra-low/analysis.json:network_adjustment.scores.estimate.by_family.live_debugging.time_accuracy
\newcommand{\TerraDebugNetRemovedLow}{66.5}                             % equal mean across passes: docs/lite/results/four-family/gpt-5.6-terra-low/analysis.json:network_adjustment.scores.low.by_family.live_debugging.time_accuracy; docs/lite/results/hosted-api-repeats-20261003/pass2/gpt-5.6-terra-low/analysis.json:network_adjustment.scores.low.by_family.live_debugging.time_accuracy; docs/lite/results/hosted-api-repeats-20261003/pass3/gpt-5.6-terra-low/analysis.json:network_adjustment.scores.low.by_family.live_debugging.time_accuracy
\newcommand{\TerraDebugNetRemovedHigh}{70.6}                            % equal mean across passes: docs/lite/results/four-family/gpt-5.6-terra-low/analysis.json:network_adjustment.scores.high.by_family.live_debugging.time_accuracy; docs/lite/results/hosted-api-repeats-20261003/pass2/gpt-5.6-terra-low/analysis.json:network_adjustment.scores.high.by_family.live_debugging.time_accuracy; docs/lite/results/hosted-api-repeats-20261003/pass3/gpt-5.6-terra-low/analysis.json:network_adjustment.scores.high.by_family.live_debugging.time_accuracy
\newcommand{\TerraAssemblyUntimed}{91.4}                                % equal mean across passes: docs/lite/results/four-family/gpt-5.6-terra-low/analysis.json:scores.by_family.procedural_coaching.untimed_decision_accuracy; docs/lite/results/hosted-api-repeats-20261003/pass2/gpt-5.6-terra-low/analysis.json:scores.by_family.procedural_coaching.untimed_decision_accuracy; docs/lite/results/hosted-api-repeats-20261003/pass3/gpt-5.6-terra-low/analysis.json:scores.by_family.procedural_coaching.untimed_decision_accuracy
\newcommand{\TerraAssemblyInForce}{52.1}                                % equal mean across passes: docs/lite/results/four-family/gpt-5.6-terra-low/analysis.json:scores.by_family.procedural_coaching.time_accuracy; docs/lite/results/hosted-api-repeats-20261003/pass2/gpt-5.6-terra-low/analysis.json:scores.by_family.procedural_coaching.time_accuracy; docs/lite/results/hosted-api-repeats-20261003/pass3/gpt-5.6-terra-low/analysis.json:scores.by_family.procedural_coaching.time_accuracy
\newcommand{\TerraAssemblyNetRemoved}{63.1}                             % equal mean across passes: docs/lite/results/four-family/gpt-5.6-terra-low/analysis.json:network_adjustment.scores.estimate.by_family.procedural_coaching.time_accuracy; docs/lite/results/hosted-api-repeats-20261003/pass2/gpt-5.6-terra-low/analysis.json:network_adjustment.scores.estimate.by_family.procedural_coaching.time_accuracy; docs/lite/results/hosted-api-repeats-20261003/pass3/gpt-5.6-terra-low/analysis.json:network_adjustment.scores.estimate.by_family.procedural_coaching.time_accuracy
\newcommand{\TerraAssemblyNetRemovedLow}{60.7}                          % equal mean across passes: docs/lite/results/four-family/gpt-5.6-terra-low/analysis.json:network_adjustment.scores.low.by_family.procedural_coaching.time_accuracy; docs/lite/results/hosted-api-repeats-20261003/pass2/gpt-5.6-terra-low/analysis.json:network_adjustment.scores.low.by_family.procedural_coaching.time_accuracy; docs/lite/results/hosted-api-repeats-20261003/pass3/gpt-5.6-terra-low/analysis.json:network_adjustment.scores.low.by_family.procedural_coaching.time_accuracy
\newcommand{\TerraAssemblyNetRemovedHigh}{64.5}                         % equal mean across passes: docs/lite/results/four-family/gpt-5.6-terra-low/analysis.json:network_adjustment.scores.high.by_family.procedural_coaching.time_accuracy; docs/lite/results/hosted-api-repeats-20261003/pass2/gpt-5.6-terra-low/analysis.json:network_adjustment.scores.high.by_family.procedural_coaching.time_accuracy; docs/lite/results/hosted-api-repeats-20261003/pass3/gpt-5.6-terra-low/analysis.json:network_adjustment.scores.high.by_family.procedural_coaching.time_accuracy
\newcommand{\TerraSupportUntimed}{95.0}                                 % equal mean across passes: docs/lite/results/four-family/gpt-5.6-terra-low/analysis.json:scores.by_family.support_call_assist.untimed_decision_accuracy; docs/lite/results/hosted-api-repeats-20261003/pass2/gpt-5.6-terra-low/analysis.json:scores.by_family.support_call_assist.untimed_decision_accuracy; docs/lite/results/hosted-api-repeats-20261003/pass3/gpt-5.6-terra-low/analysis.json:scores.by_family.support_call_assist.untimed_decision_accuracy
\newcommand{\TerraSupportInForce}{60.2}                                 % equal mean across passes: docs/lite/results/four-family/gpt-5.6-terra-low/analysis.json:scores.by_family.support_call_assist.time_accuracy; docs/lite/results/hosted-api-repeats-20261003/pass2/gpt-5.6-terra-low/analysis.json:scores.by_family.support_call_assist.time_accuracy; docs/lite/results/hosted-api-repeats-20261003/pass3/gpt-5.6-terra-low/analysis.json:scores.by_family.support_call_assist.time_accuracy
\newcommand{\TerraSupportNetRemoved}{71.9}                              % equal mean across passes: docs/lite/results/four-family/gpt-5.6-terra-low/analysis.json:network_adjustment.scores.estimate.by_family.support_call_assist.time_accuracy; docs/lite/results/hosted-api-repeats-20261003/pass2/gpt-5.6-terra-low/analysis.json:network_adjustment.scores.estimate.by_family.support_call_assist.time_accuracy; docs/lite/results/hosted-api-repeats-20261003/pass3/gpt-5.6-terra-low/analysis.json:network_adjustment.scores.estimate.by_family.support_call_assist.time_accuracy
\newcommand{\TerraSupportNetRemovedLow}{69.2}                           % equal mean across passes: docs/lite/results/four-family/gpt-5.6-terra-low/analysis.json:network_adjustment.scores.low.by_family.support_call_assist.time_accuracy; docs/lite/results/hosted-api-repeats-20261003/pass2/gpt-5.6-terra-low/analysis.json:network_adjustment.scores.low.by_family.support_call_assist.time_accuracy; docs/lite/results/hosted-api-repeats-20261003/pass3/gpt-5.6-terra-low/analysis.json:network_adjustment.scores.low.by_family.support_call_assist.time_accuracy
\newcommand{\TerraSupportNetRemovedHigh}{73.3}                          % equal mean across passes: docs/lite/results/four-family/gpt-5.6-terra-low/analysis.json:network_adjustment.scores.high.by_family.support_call_assist.time_accuracy; docs/lite/results/hosted-api-repeats-20261003/pass2/gpt-5.6-terra-low/analysis.json:network_adjustment.scores.high.by_family.support_call_assist.time_accuracy; docs/lite/results/hosted-api-repeats-20261003/pass3/gpt-5.6-terra-low/analysis.json:network_adjustment.scores.high.by_family.support_call_assist.time_accuracy
\newcommand{\TerraPresenterUntimed}{97.5}                               % equal mean across passes: docs/lite/results/four-family/gpt-5.6-terra-low/analysis.json:scores.by_family.presenter_voice_control.untimed_decision_accuracy; docs/lite/results/hosted-api-repeats-20261003/pass2/gpt-5.6-terra-low/analysis.json:scores.by_family.presenter_voice_control.untimed_decision_accuracy; docs/lite/results/hosted-api-repeats-20261003/pass3/gpt-5.6-terra-low/analysis.json:scores.by_family.presenter_voice_control.untimed_decision_accuracy
\newcommand{\TerraPresenterInForce}{54.0}                               % equal mean across passes: docs/lite/results/four-family/gpt-5.6-terra-low/analysis.json:scores.by_family.presenter_voice_control.time_accuracy; docs/lite/results/hosted-api-repeats-20261003/pass2/gpt-5.6-terra-low/analysis.json:scores.by_family.presenter_voice_control.time_accuracy; docs/lite/results/hosted-api-repeats-20261003/pass3/gpt-5.6-terra-low/analysis.json:scores.by_family.presenter_voice_control.time_accuracy
\newcommand{\TerraPresenterNetRemoved}{64.6}                            % equal mean across passes: docs/lite/results/four-family/gpt-5.6-terra-low/analysis.json:network_adjustment.scores.estimate.by_family.presenter_voice_control.time_accuracy; docs/lite/results/hosted-api-repeats-20261003/pass2/gpt-5.6-terra-low/analysis.json:network_adjustment.scores.estimate.by_family.presenter_voice_control.time_accuracy; docs/lite/results/hosted-api-repeats-20261003/pass3/gpt-5.6-terra-low/analysis.json:network_adjustment.scores.estimate.by_family.presenter_voice_control.time_accuracy
\newcommand{\TerraPresenterNetRemovedLow}{62.1}                         % equal mean across passes: docs/lite/results/four-family/gpt-5.6-terra-low/analysis.json:network_adjustment.scores.low.by_family.presenter_voice_control.time_accuracy; docs/lite/results/hosted-api-repeats-20261003/pass2/gpt-5.6-terra-low/analysis.json:network_adjustment.scores.low.by_family.presenter_voice_control.time_accuracy; docs/lite/results/hosted-api-repeats-20261003/pass3/gpt-5.6-terra-low/analysis.json:network_adjustment.scores.low.by_family.presenter_voice_control.time_accuracy
\newcommand{\TerraPresenterNetRemovedHigh}{66.0}                        % equal mean across passes: docs/lite/results/four-family/gpt-5.6-terra-low/analysis.json:network_adjustment.scores.high.by_family.presenter_voice_control.time_accuracy; docs/lite/results/hosted-api-repeats-20261003/pass2/gpt-5.6-terra-low/analysis.json:network_adjustment.scores.high.by_family.presenter_voice_control.time_accuracy; docs/lite/results/hosted-api-repeats-20261003/pass3/gpt-5.6-terra-low/analysis.json:network_adjustment.scores.high.by_family.presenter_voice_control.time_accuracy
\newcommand{\TerraDebugAUntimed}{100.0}                                 % equal mean across passes: docs/lite/results/four-family/gpt-5.6-terra-low/analysis.json:scores.per_episode[0].untimed_decision_accuracy (lite_debugging_a); docs/lite/results/hosted-api-repeats-20261003/pass2/gpt-5.6-terra-low/analysis.json:scores.per_episode[0].untimed_decision_accuracy (lite_debugging_a); docs/lite/results/hosted-api-repeats-20261003/pass3/gpt-5.6-terra-low/analysis.json:scores.per_episode[0].untimed_decision_accuracy (lite_debugging_a)
\newcommand{\TerraDebugAInForce}{58.6}                                  % equal mean across passes: docs/lite/results/four-family/gpt-5.6-terra-low/analysis.json:scores.per_episode[0].time_accuracy (lite_debugging_a); docs/lite/results/hosted-api-repeats-20261003/pass2/gpt-5.6-terra-low/analysis.json:scores.per_episode[0].time_accuracy (lite_debugging_a); docs/lite/results/hosted-api-repeats-20261003/pass3/gpt-5.6-terra-low/analysis.json:scores.per_episode[0].time_accuracy (lite_debugging_a)
\newcommand{\TerraDebugAInForceSeg}{54.5}                               % equal mean across passes: docs/lite/results/four-family/gpt-5.6-terra-low/analysis.json:scores.per_episode[0].segment_time_accuracy (lite_debugging_a); docs/lite/results/hosted-api-repeats-20261003/pass2/gpt-5.6-terra-low/analysis.json:scores.per_episode[0].segment_time_accuracy (lite_debugging_a); docs/lite/results/hosted-api-repeats-20261003/pass3/gpt-5.6-terra-low/analysis.json:scores.per_episode[0].segment_time_accuracy (lite_debugging_a)
\newcommand{\TerraDebugAErrNoDecisionSec}{2.5}                          % equal mean across passes: docs/lite/results/four-family/gpt-5.6-terra-low/analysis.json:scores.per_episode[0].error_seconds.no_decision (lite_debugging_a); docs/lite/results/hosted-api-repeats-20261003/pass2/gpt-5.6-terra-low/analysis.json:scores.per_episode[0].error_seconds.no_decision (lite_debugging_a); docs/lite/results/hosted-api-repeats-20261003/pass3/gpt-5.6-terra-low/analysis.json:scores.per_episode[0].error_seconds.no_decision (lite_debugging_a)
\newcommand{\TerraDebugAErrStaleSec}{47.2}                              % equal mean across passes: docs/lite/results/four-family/gpt-5.6-terra-low/analysis.json:scores.per_episode[0].error_seconds.source_correct (lite_debugging_a); docs/lite/results/hosted-api-repeats-20261003/pass2/gpt-5.6-terra-low/analysis.json:scores.per_episode[0].error_seconds.source_correct (lite_debugging_a); docs/lite/results/hosted-api-repeats-20261003/pass3/gpt-5.6-terra-low/analysis.json:scores.per_episode[0].error_seconds.source_correct (lite_debugging_a)
\newcommand{\TerraDebugALatencyMedian}{2.16}                            % equal mean across passes: docs/lite/results/four-family/gpt-5.6-terra-low/analysis.json:scores.per_episode[0].latency_s_p50 (lite_debugging_a); docs/lite/results/hosted-api-repeats-20261003/pass2/gpt-5.6-terra-low/analysis.json:scores.per_episode[0].latency_s_p50 (lite_debugging_a); docs/lite/results/hosted-api-repeats-20261003/pass3/gpt-5.6-terra-low/analysis.json:scores.per_episode[0].latency_s_p50 (lite_debugging_a)
\newcommand{\TerraDebugBUntimed}{98.3}                                  % equal mean across passes: docs/lite/results/four-family/gpt-5.6-terra-low/analysis.json:scores.per_episode[1].untimed_decision_accuracy (lite_debugging_b); docs/lite/results/hosted-api-repeats-20261003/pass2/gpt-5.6-terra-low/analysis.json:scores.per_episode[1].untimed_decision_accuracy (lite_debugging_b); docs/lite/results/hosted-api-repeats-20261003/pass3/gpt-5.6-terra-low/analysis.json:scores.per_episode[1].untimed_decision_accuracy (lite_debugging_b)
\newcommand{\TerraDebugBInForce}{56.4}                                  % equal mean across passes: docs/lite/results/four-family/gpt-5.6-terra-low/analysis.json:scores.per_episode[1].time_accuracy (lite_debugging_b); docs/lite/results/hosted-api-repeats-20261003/pass2/gpt-5.6-terra-low/analysis.json:scores.per_episode[1].time_accuracy (lite_debugging_b); docs/lite/results/hosted-api-repeats-20261003/pass3/gpt-5.6-terra-low/analysis.json:scores.per_episode[1].time_accuracy (lite_debugging_b)
\newcommand{\TerraDebugBInForceSeg}{48.7}                               % equal mean across passes: docs/lite/results/four-family/gpt-5.6-terra-low/analysis.json:scores.per_episode[1].segment_time_accuracy (lite_debugging_b); docs/lite/results/hosted-api-repeats-20261003/pass2/gpt-5.6-terra-low/analysis.json:scores.per_episode[1].segment_time_accuracy (lite_debugging_b); docs/lite/results/hosted-api-repeats-20261003/pass3/gpt-5.6-terra-low/analysis.json:scores.per_episode[1].segment_time_accuracy (lite_debugging_b)
\newcommand{\TerraDebugBErrNoDecisionSec}{2.5}                          % equal mean across passes: docs/lite/results/four-family/gpt-5.6-terra-low/analysis.json:scores.per_episode[1].error_seconds.no_decision (lite_debugging_b); docs/lite/results/hosted-api-repeats-20261003/pass2/gpt-5.6-terra-low/analysis.json:scores.per_episode[1].error_seconds.no_decision (lite_debugging_b); docs/lite/results/hosted-api-repeats-20261003/pass3/gpt-5.6-terra-low/analysis.json:scores.per_episode[1].error_seconds.no_decision (lite_debugging_b)
\newcommand{\TerraDebugBErrStaleSec}{47.9}                              % equal mean across passes: docs/lite/results/four-family/gpt-5.6-terra-low/analysis.json:scores.per_episode[1].error_seconds.source_correct (lite_debugging_b); docs/lite/results/hosted-api-repeats-20261003/pass2/gpt-5.6-terra-low/analysis.json:scores.per_episode[1].error_seconds.source_correct (lite_debugging_b); docs/lite/results/hosted-api-repeats-20261003/pass3/gpt-5.6-terra-low/analysis.json:scores.per_episode[1].error_seconds.source_correct (lite_debugging_b)
\newcommand{\TerraDebugBLatencyMedian}{2.13}                            % equal mean across passes: docs/lite/results/four-family/gpt-5.6-terra-low/analysis.json:scores.per_episode[1].latency_s_p50 (lite_debugging_b); docs/lite/results/hosted-api-repeats-20261003/pass2/gpt-5.6-terra-low/analysis.json:scores.per_episode[1].latency_s_p50 (lite_debugging_b); docs/lite/results/hosted-api-repeats-20261003/pass3/gpt-5.6-terra-low/analysis.json:scores.per_episode[1].latency_s_p50 (lite_debugging_b)
\newcommand{\TerraAssemblyAUntimed}{95.0}                               % equal mean across passes: docs/lite/results/four-family/gpt-5.6-terra-low/analysis.json:scores.per_episode[2].untimed_decision_accuracy (lite_assembly_a); docs/lite/results/hosted-api-repeats-20261003/pass2/gpt-5.6-terra-low/analysis.json:scores.per_episode[2].untimed_decision_accuracy (lite_assembly_a); docs/lite/results/hosted-api-repeats-20261003/pass3/gpt-5.6-terra-low/analysis.json:scores.per_episode[2].untimed_decision_accuracy (lite_assembly_a)
\newcommand{\TerraAssemblyAInForce}{56.1}                               % equal mean across passes: docs/lite/results/four-family/gpt-5.6-terra-low/analysis.json:scores.per_episode[2].time_accuracy (lite_assembly_a); docs/lite/results/hosted-api-repeats-20261003/pass2/gpt-5.6-terra-low/analysis.json:scores.per_episode[2].time_accuracy (lite_assembly_a); docs/lite/results/hosted-api-repeats-20261003/pass3/gpt-5.6-terra-low/analysis.json:scores.per_episode[2].time_accuracy (lite_assembly_a)
\newcommand{\TerraAssemblyAInForceSeg}{47.7}                            % equal mean across passes: docs/lite/results/four-family/gpt-5.6-terra-low/analysis.json:scores.per_episode[2].segment_time_accuracy (lite_assembly_a); docs/lite/results/hosted-api-repeats-20261003/pass2/gpt-5.6-terra-low/analysis.json:scores.per_episode[2].segment_time_accuracy (lite_assembly_a); docs/lite/results/hosted-api-repeats-20261003/pass3/gpt-5.6-terra-low/analysis.json:scores.per_episode[2].segment_time_accuracy (lite_assembly_a)
\newcommand{\TerraAssemblyAErrNoDecisionSec}{2.0}                       % equal mean across passes: docs/lite/results/four-family/gpt-5.6-terra-low/analysis.json:scores.per_episode[2].error_seconds.no_decision (lite_assembly_a); docs/lite/results/hosted-api-repeats-20261003/pass2/gpt-5.6-terra-low/analysis.json:scores.per_episode[2].error_seconds.no_decision (lite_assembly_a); docs/lite/results/hosted-api-repeats-20261003/pass3/gpt-5.6-terra-low/analysis.json:scores.per_episode[2].error_seconds.no_decision (lite_assembly_a)
\newcommand{\TerraAssemblyAErrStaleSec}{45.3}                           % equal mean across passes: docs/lite/results/four-family/gpt-5.6-terra-low/analysis.json:scores.per_episode[2].error_seconds.source_correct (lite_assembly_a); docs/lite/results/hosted-api-repeats-20261003/pass2/gpt-5.6-terra-low/analysis.json:scores.per_episode[2].error_seconds.source_correct (lite_assembly_a); docs/lite/results/hosted-api-repeats-20261003/pass3/gpt-5.6-terra-low/analysis.json:scores.per_episode[2].error_seconds.source_correct (lite_assembly_a)
\newcommand{\TerraAssemblyALatencyMedian}{2.18}                         % equal mean across passes: docs/lite/results/four-family/gpt-5.6-terra-low/analysis.json:scores.per_episode[2].latency_s_p50 (lite_assembly_a); docs/lite/results/hosted-api-repeats-20261003/pass2/gpt-5.6-terra-low/analysis.json:scores.per_episode[2].latency_s_p50 (lite_assembly_a); docs/lite/results/hosted-api-repeats-20261003/pass3/gpt-5.6-terra-low/analysis.json:scores.per_episode[2].latency_s_p50 (lite_assembly_a)
\newcommand{\TerraAssemblyBUntimed}{87.8}                               % equal mean across passes: docs/lite/results/four-family/gpt-5.6-terra-low/analysis.json:scores.per_episode[3].untimed_decision_accuracy (lite_assembly_b); docs/lite/results/hosted-api-repeats-20261003/pass2/gpt-5.6-terra-low/analysis.json:scores.per_episode[3].untimed_decision_accuracy (lite_assembly_b); docs/lite/results/hosted-api-repeats-20261003/pass3/gpt-5.6-terra-low/analysis.json:scores.per_episode[3].untimed_decision_accuracy (lite_assembly_b)
\newcommand{\TerraAssemblyBInForce}{48.2}                               % equal mean across passes: docs/lite/results/four-family/gpt-5.6-terra-low/analysis.json:scores.per_episode[3].time_accuracy (lite_assembly_b); docs/lite/results/hosted-api-repeats-20261003/pass2/gpt-5.6-terra-low/analysis.json:scores.per_episode[3].time_accuracy (lite_assembly_b); docs/lite/results/hosted-api-repeats-20261003/pass3/gpt-5.6-terra-low/analysis.json:scores.per_episode[3].time_accuracy (lite_assembly_b)
\newcommand{\TerraAssemblyBInForceSeg}{37.0}                            % equal mean across passes: docs/lite/results/four-family/gpt-5.6-terra-low/analysis.json:scores.per_episode[3].segment_time_accuracy (lite_assembly_b); docs/lite/results/hosted-api-repeats-20261003/pass2/gpt-5.6-terra-low/analysis.json:scores.per_episode[3].segment_time_accuracy (lite_assembly_b); docs/lite/results/hosted-api-repeats-20261003/pass3/gpt-5.6-terra-low/analysis.json:scores.per_episode[3].segment_time_accuracy (lite_assembly_b)
\newcommand{\TerraAssemblyBErrNoDecisionSec}{2.0}                       % equal mean across passes: docs/lite/results/four-family/gpt-5.6-terra-low/analysis.json:scores.per_episode[3].error_seconds.no_decision (lite_assembly_b); docs/lite/results/hosted-api-repeats-20261003/pass2/gpt-5.6-terra-low/analysis.json:scores.per_episode[3].error_seconds.no_decision (lite_assembly_b); docs/lite/results/hosted-api-repeats-20261003/pass3/gpt-5.6-terra-low/analysis.json:scores.per_episode[3].error_seconds.no_decision (lite_assembly_b)
\newcommand{\TerraAssemblyBErrStaleSec}{48.5}                           % equal mean across passes: docs/lite/results/four-family/gpt-5.6-terra-low/analysis.json:scores.per_episode[3].error_seconds.source_correct (lite_assembly_b); docs/lite/results/hosted-api-repeats-20261003/pass2/gpt-5.6-terra-low/analysis.json:scores.per_episode[3].error_seconds.source_correct (lite_assembly_b); docs/lite/results/hosted-api-repeats-20261003/pass3/gpt-5.6-terra-low/analysis.json:scores.per_episode[3].error_seconds.source_correct (lite_assembly_b)
\newcommand{\TerraAssemblyBLatencyMedian}{2.45}                         % equal mean across passes: docs/lite/results/four-family/gpt-5.6-terra-low/analysis.json:scores.per_episode[3].latency_s_p50 (lite_assembly_b); docs/lite/results/hosted-api-repeats-20261003/pass2/gpt-5.6-terra-low/analysis.json:scores.per_episode[3].latency_s_p50 (lite_assembly_b); docs/lite/results/hosted-api-repeats-20261003/pass3/gpt-5.6-terra-low/analysis.json:scores.per_episode[3].latency_s_p50 (lite_assembly_b)
\newcommand{\TerraSupportAUntimed}{96.7}                                % equal mean across passes: docs/lite/results/four-family/gpt-5.6-terra-low/analysis.json:scores.per_episode[4].untimed_decision_accuracy (lite_support_a); docs/lite/results/hosted-api-repeats-20261003/pass2/gpt-5.6-terra-low/analysis.json:scores.per_episode[4].untimed_decision_accuracy (lite_support_a); docs/lite/results/hosted-api-repeats-20261003/pass3/gpt-5.6-terra-low/analysis.json:scores.per_episode[4].untimed_decision_accuracy (lite_support_a)
\newcommand{\TerraSupportAInForce}{60.3}                                % equal mean across passes: docs/lite/results/four-family/gpt-5.6-terra-low/analysis.json:scores.per_episode[4].time_accuracy (lite_support_a); docs/lite/results/hosted-api-repeats-20261003/pass2/gpt-5.6-terra-low/analysis.json:scores.per_episode[4].time_accuracy (lite_support_a); docs/lite/results/hosted-api-repeats-20261003/pass3/gpt-5.6-terra-low/analysis.json:scores.per_episode[4].time_accuracy (lite_support_a)
\newcommand{\TerraSupportAInForceSeg}{50.5}                             % equal mean across passes: docs/lite/results/four-family/gpt-5.6-terra-low/analysis.json:scores.per_episode[4].segment_time_accuracy (lite_support_a); docs/lite/results/hosted-api-repeats-20261003/pass2/gpt-5.6-terra-low/analysis.json:scores.per_episode[4].segment_time_accuracy (lite_support_a); docs/lite/results/hosted-api-repeats-20261003/pass3/gpt-5.6-terra-low/analysis.json:scores.per_episode[4].segment_time_accuracy (lite_support_a)
\newcommand{\TerraSupportAErrNoDecisionSec}{2.0}                        % equal mean across passes: docs/lite/results/four-family/gpt-5.6-terra-low/analysis.json:scores.per_episode[4].error_seconds.no_decision (lite_support_a); docs/lite/results/hosted-api-repeats-20261003/pass2/gpt-5.6-terra-low/analysis.json:scores.per_episode[4].error_seconds.no_decision (lite_support_a); docs/lite/results/hosted-api-repeats-20261003/pass3/gpt-5.6-terra-low/analysis.json:scores.per_episode[4].error_seconds.no_decision (lite_support_a)
\newcommand{\TerraSupportAErrStaleSec}{43.2}                            % equal mean across passes: docs/lite/results/four-family/gpt-5.6-terra-low/analysis.json:scores.per_episode[4].error_seconds.source_correct (lite_support_a); docs/lite/results/hosted-api-repeats-20261003/pass2/gpt-5.6-terra-low/analysis.json:scores.per_episode[4].error_seconds.source_correct (lite_support_a); docs/lite/results/hosted-api-repeats-20261003/pass3/gpt-5.6-terra-low/analysis.json:scores.per_episode[4].error_seconds.source_correct (lite_support_a)
\newcommand{\TerraSupportALatencyMedian}{2.00}                          % equal mean across passes: docs/lite/results/four-family/gpt-5.6-terra-low/analysis.json:scores.per_episode[4].latency_s_p50 (lite_support_a); docs/lite/results/hosted-api-repeats-20261003/pass2/gpt-5.6-terra-low/analysis.json:scores.per_episode[4].latency_s_p50 (lite_support_a); docs/lite/results/hosted-api-repeats-20261003/pass3/gpt-5.6-terra-low/analysis.json:scores.per_episode[4].latency_s_p50 (lite_support_a)
\newcommand{\TerraSupportBUntimed}{93.3}                                % equal mean across passes: docs/lite/results/four-family/gpt-5.6-terra-low/analysis.json:scores.per_episode[5].untimed_decision_accuracy (lite_support_b); docs/lite/results/hosted-api-repeats-20261003/pass2/gpt-5.6-terra-low/analysis.json:scores.per_episode[5].untimed_decision_accuracy (lite_support_b); docs/lite/results/hosted-api-repeats-20261003/pass3/gpt-5.6-terra-low/analysis.json:scores.per_episode[5].untimed_decision_accuracy (lite_support_b)
\newcommand{\TerraSupportBInForce}{60.0}                                % equal mean across passes: docs/lite/results/four-family/gpt-5.6-terra-low/analysis.json:scores.per_episode[5].time_accuracy (lite_support_b); docs/lite/results/hosted-api-repeats-20261003/pass2/gpt-5.6-terra-low/analysis.json:scores.per_episode[5].time_accuracy (lite_support_b); docs/lite/results/hosted-api-repeats-20261003/pass3/gpt-5.6-terra-low/analysis.json:scores.per_episode[5].time_accuracy (lite_support_b)
\newcommand{\TerraSupportBInForceSeg}{53.8}                             % equal mean across passes: docs/lite/results/four-family/gpt-5.6-terra-low/analysis.json:scores.per_episode[5].segment_time_accuracy (lite_support_b); docs/lite/results/hosted-api-repeats-20261003/pass2/gpt-5.6-terra-low/analysis.json:scores.per_episode[5].segment_time_accuracy (lite_support_b); docs/lite/results/hosted-api-repeats-20261003/pass3/gpt-5.6-terra-low/analysis.json:scores.per_episode[5].segment_time_accuracy (lite_support_b)
\newcommand{\TerraSupportBErrNoDecisionSec}{1.8}                        % equal mean across passes: docs/lite/results/four-family/gpt-5.6-terra-low/analysis.json:scores.per_episode[5].error_seconds.no_decision (lite_support_b); docs/lite/results/hosted-api-repeats-20261003/pass2/gpt-5.6-terra-low/analysis.json:scores.per_episode[5].error_seconds.no_decision (lite_support_b); docs/lite/results/hosted-api-repeats-20261003/pass3/gpt-5.6-terra-low/analysis.json:scores.per_episode[5].error_seconds.no_decision (lite_support_b)
\newcommand{\TerraSupportBErrStaleSec}{37.6}                            % equal mean across passes: docs/lite/results/four-family/gpt-5.6-terra-low/analysis.json:scores.per_episode[5].error_seconds.source_correct (lite_support_b); docs/lite/results/hosted-api-repeats-20261003/pass2/gpt-5.6-terra-low/analysis.json:scores.per_episode[5].error_seconds.source_correct (lite_support_b); docs/lite/results/hosted-api-repeats-20261003/pass3/gpt-5.6-terra-low/analysis.json:scores.per_episode[5].error_seconds.source_correct (lite_support_b)
\newcommand{\TerraSupportBErrIncorrectSec}{8.6}                         % equal mean across passes: docs/lite/results/four-family/gpt-5.6-terra-low/analysis.json:scores.per_episode[5].error_seconds.source_incorrect (lite_support_b); docs/lite/results/hosted-api-repeats-20261003/pass2/gpt-5.6-terra-low/analysis.json:scores.per_episode[5].error_seconds.source_incorrect (lite_support_b); docs/lite/results/hosted-api-repeats-20261003/pass3/gpt-5.6-terra-low/analysis.json:scores.per_episode[5].error_seconds.source_incorrect (lite_support_b)
\newcommand{\TerraSupportBLatencyMedian}{1.77}                          % equal mean across passes: docs/lite/results/four-family/gpt-5.6-terra-low/analysis.json:scores.per_episode[5].latency_s_p50 (lite_support_b); docs/lite/results/hosted-api-repeats-20261003/pass2/gpt-5.6-terra-low/analysis.json:scores.per_episode[5].latency_s_p50 (lite_support_b); docs/lite/results/hosted-api-repeats-20261003/pass3/gpt-5.6-terra-low/analysis.json:scores.per_episode[5].latency_s_p50 (lite_support_b)
\newcommand{\TerraPresenterAUntimed}{98.3}                              % equal mean across passes: docs/lite/results/four-family/gpt-5.6-terra-low/analysis.json:scores.per_episode[6].untimed_decision_accuracy (lite_presenter_a); docs/lite/results/hosted-api-repeats-20261003/pass2/gpt-5.6-terra-low/analysis.json:scores.per_episode[6].untimed_decision_accuracy (lite_presenter_a); docs/lite/results/hosted-api-repeats-20261003/pass3/gpt-5.6-terra-low/analysis.json:scores.per_episode[6].untimed_decision_accuracy (lite_presenter_a)
\newcommand{\TerraPresenterAInForce}{54.7}                              % equal mean across passes: docs/lite/results/four-family/gpt-5.6-terra-low/analysis.json:scores.per_episode[6].time_accuracy (lite_presenter_a); docs/lite/results/hosted-api-repeats-20261003/pass2/gpt-5.6-terra-low/analysis.json:scores.per_episode[6].time_accuracy (lite_presenter_a); docs/lite/results/hosted-api-repeats-20261003/pass3/gpt-5.6-terra-low/analysis.json:scores.per_episode[6].time_accuracy (lite_presenter_a)
\newcommand{\TerraPresenterAInForceSeg}{42.7}                           % equal mean across passes: docs/lite/results/four-family/gpt-5.6-terra-low/analysis.json:scores.per_episode[6].segment_time_accuracy (lite_presenter_a); docs/lite/results/hosted-api-repeats-20261003/pass2/gpt-5.6-terra-low/analysis.json:scores.per_episode[6].segment_time_accuracy (lite_presenter_a); docs/lite/results/hosted-api-repeats-20261003/pass3/gpt-5.6-terra-low/analysis.json:scores.per_episode[6].segment_time_accuracy (lite_presenter_a)
\newcommand{\TerraPresenterAErrNoDecisionSec}{2.3}                      % equal mean across passes: docs/lite/results/four-family/gpt-5.6-terra-low/analysis.json:scores.per_episode[6].error_seconds.no_decision (lite_presenter_a); docs/lite/results/hosted-api-repeats-20261003/pass2/gpt-5.6-terra-low/analysis.json:scores.per_episode[6].error_seconds.no_decision (lite_presenter_a); docs/lite/results/hosted-api-repeats-20261003/pass3/gpt-5.6-terra-low/analysis.json:scores.per_episode[6].error_seconds.no_decision (lite_presenter_a)
\newcommand{\TerraPresenterAErrStaleSec}{50.4}                          % equal mean across passes: docs/lite/results/four-family/gpt-5.6-terra-low/analysis.json:scores.per_episode[6].error_seconds.source_correct (lite_presenter_a); docs/lite/results/hosted-api-repeats-20261003/pass2/gpt-5.6-terra-low/analysis.json:scores.per_episode[6].error_seconds.source_correct (lite_presenter_a); docs/lite/results/hosted-api-repeats-20261003/pass3/gpt-5.6-terra-low/analysis.json:scores.per_episode[6].error_seconds.source_correct (lite_presenter_a)
\newcommand{\TerraPresenterALatencyMedian}{2.21}                        % equal mean across passes: docs/lite/results/four-family/gpt-5.6-terra-low/analysis.json:scores.per_episode[6].latency_s_p50 (lite_presenter_a); docs/lite/results/hosted-api-repeats-20261003/pass2/gpt-5.6-terra-low/analysis.json:scores.per_episode[6].latency_s_p50 (lite_presenter_a); docs/lite/results/hosted-api-repeats-20261003/pass3/gpt-5.6-terra-low/analysis.json:scores.per_episode[6].latency_s_p50 (lite_presenter_a)
\newcommand{\TerraPresenterBUntimed}{96.7}                              % equal mean across passes: docs/lite/results/four-family/gpt-5.6-terra-low/analysis.json:scores.per_episode[7].untimed_decision_accuracy (lite_presenter_b); docs/lite/results/hosted-api-repeats-20261003/pass2/gpt-5.6-terra-low/analysis.json:scores.per_episode[7].untimed_decision_accuracy (lite_presenter_b); docs/lite/results/hosted-api-repeats-20261003/pass3/gpt-5.6-terra-low/analysis.json:scores.per_episode[7].untimed_decision_accuracy (lite_presenter_b)
\newcommand{\TerraPresenterBInForce}{53.3}                              % equal mean across passes: docs/lite/results/four-family/gpt-5.6-terra-low/analysis.json:scores.per_episode[7].time_accuracy (lite_presenter_b); docs/lite/results/hosted-api-repeats-20261003/pass2/gpt-5.6-terra-low/analysis.json:scores.per_episode[7].time_accuracy (lite_presenter_b); docs/lite/results/hosted-api-repeats-20261003/pass3/gpt-5.6-terra-low/analysis.json:scores.per_episode[7].time_accuracy (lite_presenter_b)
\newcommand{\TerraPresenterBInForceSeg}{42.5}                           % equal mean across passes: docs/lite/results/four-family/gpt-5.6-terra-low/analysis.json:scores.per_episode[7].segment_time_accuracy (lite_presenter_b); docs/lite/results/hosted-api-repeats-20261003/pass2/gpt-5.6-terra-low/analysis.json:scores.per_episode[7].segment_time_accuracy (lite_presenter_b); docs/lite/results/hosted-api-repeats-20261003/pass3/gpt-5.6-terra-low/analysis.json:scores.per_episode[7].segment_time_accuracy (lite_presenter_b)
\newcommand{\TerraPresenterBErrNoDecisionSec}{1.6}                      % equal mean across passes: docs/lite/results/four-family/gpt-5.6-terra-low/analysis.json:scores.per_episode[7].error_seconds.no_decision (lite_presenter_b); docs/lite/results/hosted-api-repeats-20261003/pass2/gpt-5.6-terra-low/analysis.json:scores.per_episode[7].error_seconds.no_decision (lite_presenter_b); docs/lite/results/hosted-api-repeats-20261003/pass3/gpt-5.6-terra-low/analysis.json:scores.per_episode[7].error_seconds.no_decision (lite_presenter_b)
\newcommand{\TerraPresenterBErrStaleSec}{49.6}                          % equal mean across passes: docs/lite/results/four-family/gpt-5.6-terra-low/analysis.json:scores.per_episode[7].error_seconds.source_correct (lite_presenter_b); docs/lite/results/hosted-api-repeats-20261003/pass2/gpt-5.6-terra-low/analysis.json:scores.per_episode[7].error_seconds.source_correct (lite_presenter_b); docs/lite/results/hosted-api-repeats-20261003/pass3/gpt-5.6-terra-low/analysis.json:scores.per_episode[7].error_seconds.source_correct (lite_presenter_b)
\newcommand{\TerraPresenterBLatencyMedian}{2.52}                        % equal mean across passes: docs/lite/results/four-family/gpt-5.6-terra-low/analysis.json:scores.per_episode[7].latency_s_p50 (lite_presenter_b); docs/lite/results/hosted-api-repeats-20261003/pass2/gpt-5.6-terra-low/analysis.json:scores.per_episode[7].latency_s_p50 (lite_presenter_b); docs/lite/results/hosted-api-repeats-20261003/pass3/gpt-5.6-terra-low/analysis.json:scores.per_episode[7].latency_s_p50 (lite_presenter_b)
\newcommand{\TerraNoneUntimed}{81.9}                                    % equal mean across passes: docs/lite/results/four-family/gpt-5.6-terra-none/analysis.json:scores.overall.untimed_decision_accuracy; docs/lite/results/hosted-api-repeats-20261003/pass2/gpt-5.6-terra-none/analysis.json:scores.overall.untimed_decision_accuracy; docs/lite/results/hosted-api-repeats-20261003/pass3/gpt-5.6-terra-none/analysis.json:scores.overall.untimed_decision_accuracy
\newcommand{\TerraNoneInForce}{61.3}                                    % equal mean across passes: docs/lite/results/four-family/gpt-5.6-terra-none/analysis.json:scores.overall.time_accuracy; docs/lite/results/hosted-api-repeats-20261003/pass2/gpt-5.6-terra-none/analysis.json:scores.overall.time_accuracy; docs/lite/results/hosted-api-repeats-20261003/pass3/gpt-5.6-terra-none/analysis.json:scores.overall.time_accuracy
\newcommand{\TerraNoneInForceSeg}{55.0}                                 % equal mean across passes: docs/lite/results/four-family/gpt-5.6-terra-none/analysis.json:scores.overall.segment_time_accuracy; docs/lite/results/hosted-api-repeats-20261003/pass2/gpt-5.6-terra-none/analysis.json:scores.overall.segment_time_accuracy; docs/lite/results/hosted-api-repeats-20261003/pass3/gpt-5.6-terra-none/analysis.json:scores.overall.segment_time_accuracy
\newcommand{\TerraNoneNetRemovedSeg}{69.6}                              % equal mean across passes: docs/lite/results/four-family/gpt-5.6-terra-none/analysis.json:network_adjustment.scores.estimate.overall.segment_time_accuracy; docs/lite/results/hosted-api-repeats-20261003/pass2/gpt-5.6-terra-none/analysis.json:network_adjustment.scores.estimate.overall.segment_time_accuracy; docs/lite/results/hosted-api-repeats-20261003/pass3/gpt-5.6-terra-none/analysis.json:network_adjustment.scores.estimate.overall.segment_time_accuracy
\newcommand{\TerraNoneNetSec}{0.87}                                     % equal mean across passes: docs/lite/results/four-family/gpt-5.6-terra-none/analysis.json:network_adjustment.network_s.estimate; docs/lite/results/hosted-api-repeats-20261003/pass2/gpt-5.6-terra-none/analysis.json:network_adjustment.network_s.estimate; docs/lite/results/hosted-api-repeats-20261003/pass3/gpt-5.6-terra-none/analysis.json:network_adjustment.network_s.estimate
\newcommand{\TerraNoneNetRemovedSegLow}{66.9}                           % equal mean across passes: docs/lite/results/four-family/gpt-5.6-terra-none/analysis.json:network_adjustment.scores.low.overall.segment_time_accuracy; docs/lite/results/hosted-api-repeats-20261003/pass2/gpt-5.6-terra-none/analysis.json:network_adjustment.scores.low.overall.segment_time_accuracy; docs/lite/results/hosted-api-repeats-20261003/pass3/gpt-5.6-terra-none/analysis.json:network_adjustment.scores.low.overall.segment_time_accuracy
\newcommand{\TerraNoneNetSecLow}{0.71}                                  % equal mean across passes: docs/lite/results/four-family/gpt-5.6-terra-none/analysis.json:network_adjustment.network_s.low; docs/lite/results/hosted-api-repeats-20261003/pass2/gpt-5.6-terra-none/analysis.json:network_adjustment.network_s.low; docs/lite/results/hosted-api-repeats-20261003/pass3/gpt-5.6-terra-none/analysis.json:network_adjustment.network_s.low
\newcommand{\TerraNoneNetRemovedSegHigh}{73.1}                          % equal mean across passes: docs/lite/results/four-family/gpt-5.6-terra-none/analysis.json:network_adjustment.scores.high.overall.segment_time_accuracy; docs/lite/results/hosted-api-repeats-20261003/pass2/gpt-5.6-terra-none/analysis.json:network_adjustment.scores.high.overall.segment_time_accuracy; docs/lite/results/hosted-api-repeats-20261003/pass3/gpt-5.6-terra-none/analysis.json:network_adjustment.scores.high.overall.segment_time_accuracy
\newcommand{\TerraNoneNetSecHigh}{1.08}                                 % equal mean across passes: docs/lite/results/four-family/gpt-5.6-terra-none/analysis.json:network_adjustment.network_s.high; docs/lite/results/hosted-api-repeats-20261003/pass2/gpt-5.6-terra-none/analysis.json:network_adjustment.network_s.high; docs/lite/results/hosted-api-repeats-20261003/pass3/gpt-5.6-terra-none/analysis.json:network_adjustment.network_s.high
\newcommand{\TerraNonePrefillMsPerKTok}{0.0}                            % equal mean across passes: docs/lite/results/four-family/gpt-5.6-terra-none/analysis.json:network_adjustment.prefill_s_per_1k_input_tokens (x 1000); docs/lite/results/hosted-api-repeats-20261003/pass2/gpt-5.6-terra-none/analysis.json:network_adjustment.prefill_s_per_1k_input_tokens (x 1000); docs/lite/results/hosted-api-repeats-20261003/pass3/gpt-5.6-terra-none/analysis.json:network_adjustment.prefill_s_per_1k_input_tokens (x 1000)
\newcommand{\TerraNoneDecodeMsPerTok}{6.5}                              % equal mean across passes: docs/lite/results/four-family/gpt-5.6-terra-none/analysis.json:network_adjustment.decode_s_per_output_token (x 1000); docs/lite/results/hosted-api-repeats-20261003/pass2/gpt-5.6-terra-none/analysis.json:network_adjustment.decode_s_per_output_token (x 1000); docs/lite/results/hosted-api-repeats-20261003/pass3/gpt-5.6-terra-none/analysis.json:network_adjustment.decode_s_per_output_token (x 1000)
\newcommand{\TerraNoneLatencyFloor}{0.91}                               % minimum across passes: docs/lite/results/four-family/gpt-5.6-terra-none/analysis.json:network_adjustment.latency_floor_s; docs/lite/results/hosted-api-repeats-20261003/pass2/gpt-5.6-terra-none/analysis.json:network_adjustment.latency_floor_s; docs/lite/results/hosted-api-repeats-20261003/pass3/gpt-5.6-terra-none/analysis.json:network_adjustment.latency_floor_s
\newcommand{\TerraNoneClampedRequests}{113}                             % sum across passes: docs/lite/results/four-family/gpt-5.6-terra-none/analysis.json:network_adjustment.scores.{estimate,low,high}.clamped_requests (maximum); docs/lite/results/hosted-api-repeats-20261003/pass2/gpt-5.6-terra-none/analysis.json:network_adjustment.scores.{estimate,low,high}.clamped_requests (maximum); docs/lite/results/hosted-api-repeats-20261003/pass3/gpt-5.6-terra-none/analysis.json:network_adjustment.scores.{estimate,low,high}.clamped_requests (maximum)
\newcommand{\TerraNoneLatencyMedian}{1.36}                              % equal mean across passes: docs/lite/results/four-family/gpt-5.6-terra-none/analysis.json:latency_s.p50; docs/lite/results/hosted-api-repeats-20261003/pass2/gpt-5.6-terra-none/analysis.json:latency_s.p50; docs/lite/results/hosted-api-repeats-20261003/pass3/gpt-5.6-terra-none/analysis.json:latency_s.p50
\newcommand{\TerraNoneLatencyPNinetyFive}{1.82}                         % equal mean across passes: docs/lite/results/four-family/gpt-5.6-terra-none/analysis.json:latency_s.p95; docs/lite/results/hosted-api-repeats-20261003/pass2/gpt-5.6-terra-none/analysis.json:latency_s.p95; docs/lite/results/hosted-api-repeats-20261003/pass3/gpt-5.6-terra-none/analysis.json:latency_s.p95
\newcommand{\TerraNoneLatencyMax}{8.58}                                 % maximum across passes: docs/lite/results/four-family/gpt-5.6-terra-none/analysis.json:latency_s.max; docs/lite/results/hosted-api-repeats-20261003/pass2/gpt-5.6-terra-none/analysis.json:latency_s.max; docs/lite/results/hosted-api-repeats-20261003/pass3/gpt-5.6-terra-none/analysis.json:latency_s.max
\newcommand{\TerraNoneLatencyMean}{1.41}                                % equal mean across passes: runs/lite-v1-gpt-5.6-terra-none-four-family-retry-v1/events.jsonl:response.attempts[-1].completed_s - started_s (mean over 480 responses); runs/hosted-api-repeats-20261003/pass2/gpt-5.6-terra-none/events.jsonl:response.attempts[-1].completed_s - started_s (mean over 480 responses); runs/hosted-api-repeats-20261003/pass3/gpt-5.6-terra-none/events.jsonl:response.attempts[-1].completed_s - started_s (mean over 480 responses)
\newcommand{\TerraNoneInputTokMedian}{3034}                             % equal mean across passes: runs/lite-v1-gpt-5.6-terra-none-four-family-retry-v1/events.jsonl:response.usage.input_tokens (median over 480 responses, half-up); runs/hosted-api-repeats-20261003/pass2/gpt-5.6-terra-none/events.jsonl:response.usage.input_tokens (median over 480 responses, half-up); runs/hosted-api-repeats-20261003/pass3/gpt-5.6-terra-none/events.jsonl:response.usage.input_tokens (median over 480 responses, half-up)
\newcommand{\TerraNoneInputTokTotal}{4{,}429{,}467}                     % sum across passes: docs/lite/results/four-family/gpt-5.6-terra-none/analysis.json:usage.input_tokens; docs/lite/results/hosted-api-repeats-20261003/pass2/gpt-5.6-terra-none/analysis.json:usage.input_tokens; docs/lite/results/hosted-api-repeats-20261003/pass3/gpt-5.6-terra-none/analysis.json:usage.input_tokens
\newcommand{\TerraNoneOutputTokMedian}{46}                              % equal mean across passes: runs/lite-v1-gpt-5.6-terra-none-four-family-retry-v1/events.jsonl:response.usage.output_tokens (median over 480 responses, half-up); runs/hosted-api-repeats-20261003/pass2/gpt-5.6-terra-none/events.jsonl:response.usage.output_tokens (median over 480 responses, half-up); runs/hosted-api-repeats-20261003/pass3/gpt-5.6-terra-none/events.jsonl:response.usage.output_tokens (median over 480 responses, half-up)
\newcommand{\TerraNoneOutputTokTotal}{64{,}440}                         % sum across passes: docs/lite/results/four-family/gpt-5.6-terra-none/analysis.json:usage.output_tokens; docs/lite/results/hosted-api-repeats-20261003/pass2/gpt-5.6-terra-none/analysis.json:usage.output_tokens; docs/lite/results/hosted-api-repeats-20261003/pass3/gpt-5.6-terra-none/analysis.json:usage.output_tokens
\newcommand{\TerraNoneReasoningTokMedian}{0}                            % equal mean across passes: runs/lite-v1-gpt-5.6-terra-none-four-family-retry-v1/events.jsonl:response.usage.reasoning_tokens (median over 480 responses, half-up); runs/hosted-api-repeats-20261003/pass2/gpt-5.6-terra-none/events.jsonl:response.usage.reasoning_tokens (median over 480 responses, half-up); runs/hosted-api-repeats-20261003/pass3/gpt-5.6-terra-none/events.jsonl:response.usage.reasoning_tokens (median over 480 responses, half-up)
\newcommand{\TerraNoneCachedTokTotal}{0}                                % sum across passes: docs/lite/results/four-family/gpt-5.6-terra-none/analysis.json:usage.cached_tokens; docs/lite/results/hosted-api-repeats-20261003/pass2/gpt-5.6-terra-none/analysis.json:usage.cached_tokens; docs/lite/results/hosted-api-repeats-20261003/pass3/gpt-5.6-terra-none/analysis.json:usage.cached_tokens
\newcommand{\TerraNoneCostInputUSD}{8.859}                              % sum across passes: docs/lite/results/four-family/gpt-5.6-terra-none/analysis.json:usage at developers.openai.com/api/docs/pricing (gpt-5.6-terra, Standard, short context) list prices retrieved 2026-09-29 (input: uncached x input price + cached x cached price); docs/lite/results/hosted-api-repeats-20261003/pass2/gpt-5.6-terra-none/analysis.json:usage at developers.openai.com/api/docs/pricing (gpt-5.6-terra, Standard, short context) list prices retrieved 2026-09-29 (input: uncached x input price + cached x cached price); docs/lite/results/hosted-api-repeats-20261003/pass3/gpt-5.6-terra-none/analysis.json:usage at developers.openai.com/api/docs/pricing (gpt-5.6-terra, Standard, short context) list prices retrieved 2026-09-29 (input: uncached x input price + cached x cached price)
\newcommand{\TerraNoneCostOutputUSD}{0.773}                             % sum across passes: docs/lite/results/four-family/gpt-5.6-terra-none/analysis.json:usage at developers.openai.com/api/docs/pricing (gpt-5.6-terra, Standard, short context) list prices retrieved 2026-09-29 (output tokens, reasoning included, x output price); docs/lite/results/hosted-api-repeats-20261003/pass2/gpt-5.6-terra-none/analysis.json:usage at developers.openai.com/api/docs/pricing (gpt-5.6-terra, Standard, short context) list prices retrieved 2026-09-29 (output tokens, reasoning included, x output price); docs/lite/results/hosted-api-repeats-20261003/pass3/gpt-5.6-terra-none/analysis.json:usage at developers.openai.com/api/docs/pricing (gpt-5.6-terra, Standard, short context) list prices retrieved 2026-09-29 (output tokens, reasoning included, x output price)
\newcommand{\TerraNoneCostUSD}{9.632}                                   % sum across passes: docs/lite/results/four-family/gpt-5.6-terra-none/analysis.json:usage at developers.openai.com/api/docs/pricing (gpt-5.6-terra, Standard, short context) list prices retrieved 2026-09-29 (input + output); docs/lite/results/hosted-api-repeats-20261003/pass2/gpt-5.6-terra-none/analysis.json:usage at developers.openai.com/api/docs/pricing (gpt-5.6-terra, Standard, short context) list prices retrieved 2026-09-29 (input + output); docs/lite/results/hosted-api-repeats-20261003/pass3/gpt-5.6-terra-none/analysis.json:usage at developers.openai.com/api/docs/pricing (gpt-5.6-terra, Standard, short context) list prices retrieved 2026-09-29 (input + output)
\newcommand{\TerraNonePriceInput}{2.000}                                % equal mean across passes: developers.openai.com/api/docs/pricing (gpt-5.6-terra, Standard, short context): USD per 1M input tokens; developers.openai.com/api/docs/pricing (gpt-5.6-terra, Standard, short context): USD per 1M input tokens; developers.openai.com/api/docs/pricing (gpt-5.6-terra, Standard, short context): USD per 1M input tokens
\newcommand{\TerraNonePriceOutput}{12.00}                               % equal mean across passes: developers.openai.com/api/docs/pricing (gpt-5.6-terra, Standard, short context): USD per 1M output tokens; developers.openai.com/api/docs/pricing (gpt-5.6-terra, Standard, short context): USD per 1M output tokens; developers.openai.com/api/docs/pricing (gpt-5.6-terra, Standard, short context): USD per 1M output tokens
\newcommand{\TerraNoneErrNoDecisionPct}{1.4}                            % equal mean across passes: docs/lite/results/four-family/gpt-5.6-terra-none/analysis.json:error_seconds.no_decision (scenario error fraction, then equal family mean); docs/lite/results/hosted-api-repeats-20261003/pass2/gpt-5.6-terra-none/analysis.json:error_seconds.no_decision (scenario error fraction, then equal family mean); docs/lite/results/hosted-api-repeats-20261003/pass3/gpt-5.6-terra-none/analysis.json:error_seconds.no_decision (scenario error fraction, then equal family mean)
\newcommand{\TerraNoneErrStalePct}{21.0}                                % equal mean across passes: docs/lite/results/four-family/gpt-5.6-terra-none/analysis.json:error_seconds.source_correct (scenario error fraction, then equal family mean); docs/lite/results/hosted-api-repeats-20261003/pass2/gpt-5.6-terra-none/analysis.json:error_seconds.source_correct (scenario error fraction, then equal family mean); docs/lite/results/hosted-api-repeats-20261003/pass3/gpt-5.6-terra-none/analysis.json:error_seconds.source_correct (scenario error fraction, then equal family mean)
\newcommand{\TerraNoneInForceRecorded}{61.3}                            % equal mean across passes: docs/lite/results/four-family/gpt-5.6-terra-none/analysis.json:scores.overall.time_accuracy; docs/lite/results/hosted-api-repeats-20261003/pass2/gpt-5.6-terra-none/analysis.json:scores.overall.time_accuracy; docs/lite/results/hosted-api-repeats-20261003/pass3/gpt-5.6-terra-none/analysis.json:scores.overall.time_accuracy
\newcommand{\TerraNoneInForcePhysical}{61.3}                            % equal mean across passes: docs/lite/results/four-family/gpt-5.6-terra-none/analysis.json:raw_wallclock_scores.overall.time_accuracy (physical wall-clock trace); docs/lite/results/hosted-api-repeats-20261003/pass2/gpt-5.6-terra-none/analysis.json:raw_wallclock_scores.overall.time_accuracy (physical wall-clock trace); docs/lite/results/hosted-api-repeats-20261003/pass3/gpt-5.6-terra-none/analysis.json:raw_wallclock_scores.overall.time_accuracy (physical wall-clock trace)
\newcommand{\TerraNoneExcludedDispatchSec}{0.25}                        % equal mean across passes: docs/lite/results/four-family/gpt-5.6-terra-none/analysis.json:retry_reliability.excluded_dispatch_s (summed over the pass); docs/lite/results/hosted-api-repeats-20261003/pass2/gpt-5.6-terra-none/analysis.json:retry_reliability.excluded_dispatch_s (summed over the pass); docs/lite/results/hosted-api-repeats-20261003/pass3/gpt-5.6-terra-none/analysis.json:retry_reliability.excluded_dispatch_s (summed over the pass)
\newcommand{\TerraNoneEventsHash}{e72912b4f9aeec5a}                     % pass 1: docs/lite/results/four-family/gpt-5.6-terra-none/analysis.json:events_sha256 (first 16 hex digits; equals sha256 of runs/lite-v1-gpt-5.6-terra-none-four-family-retry-v1/events.jsonl)
\newcommand{\TerraNoneAcceptedUpdates}{1436}                            % sum across passes: docs/lite/results/four-family/gpt-5.6-terra-none/analysis.json:accepted_updates; docs/lite/results/hosted-api-repeats-20261003/pass2/gpt-5.6-terra-none/analysis.json:accepted_updates; docs/lite/results/hosted-api-repeats-20261003/pass3/gpt-5.6-terra-none/analysis.json:accepted_updates
\newcommand{\TerraNoneDiscardedAfterHorizon}{0}                         % sum across passes: docs/lite/results/four-family/gpt-5.6-terra-none/analysis.json:discarded_updates.after_horizon (absent = 0); docs/lite/results/hosted-api-repeats-20261003/pass2/gpt-5.6-terra-none/analysis.json:discarded_updates.after_horizon (absent = 0); docs/lite/results/hosted-api-repeats-20261003/pass3/gpt-5.6-terra-none/analysis.json:discarded_updates.after_horizon (absent = 0)
\newcommand{\TerraNoneDiscardedOlder}{4}                                % sum across passes: docs/lite/results/four-family/gpt-5.6-terra-none/analysis.json:discarded_updates.older_than_active (absent = 0); docs/lite/results/hosted-api-repeats-20261003/pass2/gpt-5.6-terra-none/analysis.json:discarded_updates.older_than_active (absent = 0); docs/lite/results/hosted-api-repeats-20261003/pass3/gpt-5.6-terra-none/analysis.json:discarded_updates.older_than_active (absent = 0)
\newcommand{\TerraNoneValidResponses}{1440}                             % sum across passes: docs/lite/results/four-family/gpt-5.6-terra-none/analysis.json:retry_reliability.successful_logical_requests; docs/lite/results/hosted-api-repeats-20261003/pass2/gpt-5.6-terra-none/analysis.json:retry_reliability.successful_logical_requests; docs/lite/results/hosted-api-repeats-20261003/pass3/gpt-5.6-terra-none/analysis.json:retry_reliability.successful_logical_requests
\newcommand{\TerraNoneAttempts}{1440}                                   % sum across passes: docs/lite/results/four-family/gpt-5.6-terra-none/analysis.json:retry_reliability.attempts; docs/lite/results/hosted-api-repeats-20261003/pass2/gpt-5.6-terra-none/analysis.json:retry_reliability.attempts; docs/lite/results/hosted-api-repeats-20261003/pass3/gpt-5.6-terra-none/analysis.json:retry_reliability.attempts
\newcommand{\TerraNoneFailedAttempts}{0}                                % sum across passes: docs/lite/results/four-family/gpt-5.6-terra-none/analysis.json:retry_reliability.failed_attempts; docs/lite/results/hosted-api-repeats-20261003/pass2/gpt-5.6-terra-none/analysis.json:retry_reliability.failed_attempts; docs/lite/results/hosted-api-repeats-20261003/pass3/gpt-5.6-terra-none/analysis.json:retry_reliability.failed_attempts
\newcommand{\TerraNoneRetriedRequests}{0}                               % sum across passes: docs/lite/results/four-family/gpt-5.6-terra-none/analysis.json:retry_reliability.retried_logical_requests; docs/lite/results/hosted-api-repeats-20261003/pass2/gpt-5.6-terra-none/analysis.json:retry_reliability.retried_logical_requests; docs/lite/results/hosted-api-repeats-20261003/pass3/gpt-5.6-terra-none/analysis.json:retry_reliability.retried_logical_requests
\newcommand{\TerraNoneDebugUntimed}{72.2}                               % equal mean across passes: docs/lite/results/four-family/gpt-5.6-terra-none/analysis.json:scores.by_family.live_debugging.untimed_decision_accuracy; docs/lite/results/hosted-api-repeats-20261003/pass2/gpt-5.6-terra-none/analysis.json:scores.by_family.live_debugging.untimed_decision_accuracy; docs/lite/results/hosted-api-repeats-20261003/pass3/gpt-5.6-terra-none/analysis.json:scores.by_family.live_debugging.untimed_decision_accuracy
\newcommand{\TerraNoneDebugInForce}{56.1}                               % equal mean across passes: docs/lite/results/four-family/gpt-5.6-terra-none/analysis.json:scores.by_family.live_debugging.time_accuracy; docs/lite/results/hosted-api-repeats-20261003/pass2/gpt-5.6-terra-none/analysis.json:scores.by_family.live_debugging.time_accuracy; docs/lite/results/hosted-api-repeats-20261003/pass3/gpt-5.6-terra-none/analysis.json:scores.by_family.live_debugging.time_accuracy
\newcommand{\TerraNoneDebugNetRemoved}{65.0}                            % equal mean across passes: docs/lite/results/four-family/gpt-5.6-terra-none/analysis.json:network_adjustment.scores.estimate.by_family.live_debugging.time_accuracy; docs/lite/results/hosted-api-repeats-20261003/pass2/gpt-5.6-terra-none/analysis.json:network_adjustment.scores.estimate.by_family.live_debugging.time_accuracy; docs/lite/results/hosted-api-repeats-20261003/pass3/gpt-5.6-terra-none/analysis.json:network_adjustment.scores.estimate.by_family.live_debugging.time_accuracy
\newcommand{\TerraNoneDebugNetRemovedLow}{63.4}                         % equal mean across passes: docs/lite/results/four-family/gpt-5.6-terra-none/analysis.json:network_adjustment.scores.low.by_family.live_debugging.time_accuracy; docs/lite/results/hosted-api-repeats-20261003/pass2/gpt-5.6-terra-none/analysis.json:network_adjustment.scores.low.by_family.live_debugging.time_accuracy; docs/lite/results/hosted-api-repeats-20261003/pass3/gpt-5.6-terra-none/analysis.json:network_adjustment.scores.low.by_family.live_debugging.time_accuracy
\newcommand{\TerraNoneDebugNetRemovedHigh}{67.2}                        % equal mean across passes: docs/lite/results/four-family/gpt-5.6-terra-none/analysis.json:network_adjustment.scores.high.by_family.live_debugging.time_accuracy; docs/lite/results/hosted-api-repeats-20261003/pass2/gpt-5.6-terra-none/analysis.json:network_adjustment.scores.high.by_family.live_debugging.time_accuracy; docs/lite/results/hosted-api-repeats-20261003/pass3/gpt-5.6-terra-none/analysis.json:network_adjustment.scores.high.by_family.live_debugging.time_accuracy
\newcommand{\TerraNoneAssemblyUntimed}{74.7}                            % equal mean across passes: docs/lite/results/four-family/gpt-5.6-terra-none/analysis.json:scores.by_family.procedural_coaching.untimed_decision_accuracy; docs/lite/results/hosted-api-repeats-20261003/pass2/gpt-5.6-terra-none/analysis.json:scores.by_family.procedural_coaching.untimed_decision_accuracy; docs/lite/results/hosted-api-repeats-20261003/pass3/gpt-5.6-terra-none/analysis.json:scores.by_family.procedural_coaching.untimed_decision_accuracy
\newcommand{\TerraNoneAssemblyInForce}{55.2}                            % equal mean across passes: docs/lite/results/four-family/gpt-5.6-terra-none/analysis.json:scores.by_family.procedural_coaching.time_accuracy; docs/lite/results/hosted-api-repeats-20261003/pass2/gpt-5.6-terra-none/analysis.json:scores.by_family.procedural_coaching.time_accuracy; docs/lite/results/hosted-api-repeats-20261003/pass3/gpt-5.6-terra-none/analysis.json:scores.by_family.procedural_coaching.time_accuracy
\newcommand{\TerraNoneAssemblyNetRemoved}{67.0}                         % equal mean across passes: docs/lite/results/four-family/gpt-5.6-terra-none/analysis.json:network_adjustment.scores.estimate.by_family.procedural_coaching.time_accuracy; docs/lite/results/hosted-api-repeats-20261003/pass2/gpt-5.6-terra-none/analysis.json:network_adjustment.scores.estimate.by_family.procedural_coaching.time_accuracy; docs/lite/results/hosted-api-repeats-20261003/pass3/gpt-5.6-terra-none/analysis.json:network_adjustment.scores.estimate.by_family.procedural_coaching.time_accuracy
\newcommand{\TerraNoneAssemblyNetRemovedLow}{64.9}                      % equal mean across passes: docs/lite/results/four-family/gpt-5.6-terra-none/analysis.json:network_adjustment.scores.low.by_family.procedural_coaching.time_accuracy; docs/lite/results/hosted-api-repeats-20261003/pass2/gpt-5.6-terra-none/analysis.json:network_adjustment.scores.low.by_family.procedural_coaching.time_accuracy; docs/lite/results/hosted-api-repeats-20261003/pass3/gpt-5.6-terra-none/analysis.json:network_adjustment.scores.low.by_family.procedural_coaching.time_accuracy
\newcommand{\TerraNoneAssemblyNetRemovedHigh}{69.9}                     % equal mean across passes: docs/lite/results/four-family/gpt-5.6-terra-none/analysis.json:network_adjustment.scores.high.by_family.procedural_coaching.time_accuracy; docs/lite/results/hosted-api-repeats-20261003/pass2/gpt-5.6-terra-none/analysis.json:network_adjustment.scores.high.by_family.procedural_coaching.time_accuracy; docs/lite/results/hosted-api-repeats-20261003/pass3/gpt-5.6-terra-none/analysis.json:network_adjustment.scores.high.by_family.procedural_coaching.time_accuracy
\newcommand{\TerraNoneSupportUntimed}{93.6}                             % equal mean across passes: docs/lite/results/four-family/gpt-5.6-terra-none/analysis.json:scores.by_family.support_call_assist.untimed_decision_accuracy; docs/lite/results/hosted-api-repeats-20261003/pass2/gpt-5.6-terra-none/analysis.json:scores.by_family.support_call_assist.untimed_decision_accuracy; docs/lite/results/hosted-api-repeats-20261003/pass3/gpt-5.6-terra-none/analysis.json:scores.by_family.support_call_assist.untimed_decision_accuracy
\newcommand{\TerraNoneSupportInForce}{66.3}                             % equal mean across passes: docs/lite/results/four-family/gpt-5.6-terra-none/analysis.json:scores.by_family.support_call_assist.time_accuracy; docs/lite/results/hosted-api-repeats-20261003/pass2/gpt-5.6-terra-none/analysis.json:scores.by_family.support_call_assist.time_accuracy; docs/lite/results/hosted-api-repeats-20261003/pass3/gpt-5.6-terra-none/analysis.json:scores.by_family.support_call_assist.time_accuracy
\newcommand{\TerraNoneSupportNetRemoved}{81.9}                          % equal mean across passes: docs/lite/results/four-family/gpt-5.6-terra-none/analysis.json:network_adjustment.scores.estimate.by_family.support_call_assist.time_accuracy; docs/lite/results/hosted-api-repeats-20261003/pass2/gpt-5.6-terra-none/analysis.json:network_adjustment.scores.estimate.by_family.support_call_assist.time_accuracy; docs/lite/results/hosted-api-repeats-20261003/pass3/gpt-5.6-terra-none/analysis.json:network_adjustment.scores.estimate.by_family.support_call_assist.time_accuracy
\newcommand{\TerraNoneSupportNetRemovedLow}{79.0}                       % equal mean across passes: docs/lite/results/four-family/gpt-5.6-terra-none/analysis.json:network_adjustment.scores.low.by_family.support_call_assist.time_accuracy; docs/lite/results/hosted-api-repeats-20261003/pass2/gpt-5.6-terra-none/analysis.json:network_adjustment.scores.low.by_family.support_call_assist.time_accuracy; docs/lite/results/hosted-api-repeats-20261003/pass3/gpt-5.6-terra-none/analysis.json:network_adjustment.scores.low.by_family.support_call_assist.time_accuracy
\newcommand{\TerraNoneSupportNetRemovedHigh}{85.7}                      % equal mean across passes: docs/lite/results/four-family/gpt-5.6-terra-none/analysis.json:network_adjustment.scores.high.by_family.support_call_assist.time_accuracy; docs/lite/results/hosted-api-repeats-20261003/pass2/gpt-5.6-terra-none/analysis.json:network_adjustment.scores.high.by_family.support_call_assist.time_accuracy; docs/lite/results/hosted-api-repeats-20261003/pass3/gpt-5.6-terra-none/analysis.json:network_adjustment.scores.high.by_family.support_call_assist.time_accuracy
\newcommand{\TerraNonePresenterUntimed}{86.9}                           % equal mean across passes: docs/lite/results/four-family/gpt-5.6-terra-none/analysis.json:scores.by_family.presenter_voice_control.untimed_decision_accuracy; docs/lite/results/hosted-api-repeats-20261003/pass2/gpt-5.6-terra-none/analysis.json:scores.by_family.presenter_voice_control.untimed_decision_accuracy; docs/lite/results/hosted-api-repeats-20261003/pass3/gpt-5.6-terra-none/analysis.json:scores.by_family.presenter_voice_control.untimed_decision_accuracy
\newcommand{\TerraNonePresenterInForce}{67.6}                           % equal mean across passes: docs/lite/results/four-family/gpt-5.6-terra-none/analysis.json:scores.by_family.presenter_voice_control.time_accuracy; docs/lite/results/hosted-api-repeats-20261003/pass2/gpt-5.6-terra-none/analysis.json:scores.by_family.presenter_voice_control.time_accuracy; docs/lite/results/hosted-api-repeats-20261003/pass3/gpt-5.6-terra-none/analysis.json:scores.by_family.presenter_voice_control.time_accuracy
\newcommand{\TerraNonePresenterNetRemoved}{80.1}                        % equal mean across passes: docs/lite/results/four-family/gpt-5.6-terra-none/analysis.json:network_adjustment.scores.estimate.by_family.presenter_voice_control.time_accuracy; docs/lite/results/hosted-api-repeats-20261003/pass2/gpt-5.6-terra-none/analysis.json:network_adjustment.scores.estimate.by_family.presenter_voice_control.time_accuracy; docs/lite/results/hosted-api-repeats-20261003/pass3/gpt-5.6-terra-none/analysis.json:network_adjustment.scores.estimate.by_family.presenter_voice_control.time_accuracy
\newcommand{\TerraNonePresenterNetRemovedLow}{77.8}                     % equal mean across passes: docs/lite/results/four-family/gpt-5.6-terra-none/analysis.json:network_adjustment.scores.low.by_family.presenter_voice_control.time_accuracy; docs/lite/results/hosted-api-repeats-20261003/pass2/gpt-5.6-terra-none/analysis.json:network_adjustment.scores.low.by_family.presenter_voice_control.time_accuracy; docs/lite/results/hosted-api-repeats-20261003/pass3/gpt-5.6-terra-none/analysis.json:network_adjustment.scores.low.by_family.presenter_voice_control.time_accuracy
\newcommand{\TerraNonePresenterNetRemovedHigh}{83.1}                    % equal mean across passes: docs/lite/results/four-family/gpt-5.6-terra-none/analysis.json:network_adjustment.scores.high.by_family.presenter_voice_control.time_accuracy; docs/lite/results/hosted-api-repeats-20261003/pass2/gpt-5.6-terra-none/analysis.json:network_adjustment.scores.high.by_family.presenter_voice_control.time_accuracy; docs/lite/results/hosted-api-repeats-20261003/pass3/gpt-5.6-terra-none/analysis.json:network_adjustment.scores.high.by_family.presenter_voice_control.time_accuracy
\newcommand{\TerraNoneDebugAUntimed}{76.7}                              % equal mean across passes: docs/lite/results/four-family/gpt-5.6-terra-none/analysis.json:scores.per_episode[0].untimed_decision_accuracy (lite_debugging_a); docs/lite/results/hosted-api-repeats-20261003/pass2/gpt-5.6-terra-none/analysis.json:scores.per_episode[0].untimed_decision_accuracy (lite_debugging_a); docs/lite/results/hosted-api-repeats-20261003/pass3/gpt-5.6-terra-none/analysis.json:scores.per_episode[0].untimed_decision_accuracy (lite_debugging_a)
\newcommand{\TerraNoneDebugAInForce}{59.6}                              % equal mean across passes: docs/lite/results/four-family/gpt-5.6-terra-none/analysis.json:scores.per_episode[0].time_accuracy (lite_debugging_a); docs/lite/results/hosted-api-repeats-20261003/pass2/gpt-5.6-terra-none/analysis.json:scores.per_episode[0].time_accuracy (lite_debugging_a); docs/lite/results/hosted-api-repeats-20261003/pass3/gpt-5.6-terra-none/analysis.json:scores.per_episode[0].time_accuracy (lite_debugging_a)
\newcommand{\TerraNoneDebugAInForceSeg}{55.0}                           % equal mean across passes: docs/lite/results/four-family/gpt-5.6-terra-none/analysis.json:scores.per_episode[0].segment_time_accuracy (lite_debugging_a); docs/lite/results/hosted-api-repeats-20261003/pass2/gpt-5.6-terra-none/analysis.json:scores.per_episode[0].segment_time_accuracy (lite_debugging_a); docs/lite/results/hosted-api-repeats-20261003/pass3/gpt-5.6-terra-none/analysis.json:scores.per_episode[0].segment_time_accuracy (lite_debugging_a)
\newcommand{\TerraNoneDebugAErrNoDecisionSec}{2.0}                      % equal mean across passes: docs/lite/results/four-family/gpt-5.6-terra-none/analysis.json:scores.per_episode[0].error_seconds.no_decision (lite_debugging_a); docs/lite/results/hosted-api-repeats-20261003/pass2/gpt-5.6-terra-none/analysis.json:scores.per_episode[0].error_seconds.no_decision (lite_debugging_a); docs/lite/results/hosted-api-repeats-20261003/pass3/gpt-5.6-terra-none/analysis.json:scores.per_episode[0].error_seconds.no_decision (lite_debugging_a)
\newcommand{\TerraNoneDebugAErrStaleSec}{20.5}                          % equal mean across passes: docs/lite/results/four-family/gpt-5.6-terra-none/analysis.json:scores.per_episode[0].error_seconds.source_correct (lite_debugging_a); docs/lite/results/hosted-api-repeats-20261003/pass2/gpt-5.6-terra-none/analysis.json:scores.per_episode[0].error_seconds.source_correct (lite_debugging_a); docs/lite/results/hosted-api-repeats-20261003/pass3/gpt-5.6-terra-none/analysis.json:scores.per_episode[0].error_seconds.source_correct (lite_debugging_a)
\newcommand{\TerraNoneDebugALatencyMedian}{1.42}                        % equal mean across passes: docs/lite/results/four-family/gpt-5.6-terra-none/analysis.json:scores.per_episode[0].latency_s_p50 (lite_debugging_a); docs/lite/results/hosted-api-repeats-20261003/pass2/gpt-5.6-terra-none/analysis.json:scores.per_episode[0].latency_s_p50 (lite_debugging_a); docs/lite/results/hosted-api-repeats-20261003/pass3/gpt-5.6-terra-none/analysis.json:scores.per_episode[0].latency_s_p50 (lite_debugging_a)
\newcommand{\TerraNoneDebugBUntimed}{67.8}                              % equal mean across passes: docs/lite/results/four-family/gpt-5.6-terra-none/analysis.json:scores.per_episode[1].untimed_decision_accuracy (lite_debugging_b); docs/lite/results/hosted-api-repeats-20261003/pass2/gpt-5.6-terra-none/analysis.json:scores.per_episode[1].untimed_decision_accuracy (lite_debugging_b); docs/lite/results/hosted-api-repeats-20261003/pass3/gpt-5.6-terra-none/analysis.json:scores.per_episode[1].untimed_decision_accuracy (lite_debugging_b)
\newcommand{\TerraNoneDebugBInForce}{52.6}                              % equal mean across passes: docs/lite/results/four-family/gpt-5.6-terra-none/analysis.json:scores.per_episode[1].time_accuracy (lite_debugging_b); docs/lite/results/hosted-api-repeats-20261003/pass2/gpt-5.6-terra-none/analysis.json:scores.per_episode[1].time_accuracy (lite_debugging_b); docs/lite/results/hosted-api-repeats-20261003/pass3/gpt-5.6-terra-none/analysis.json:scores.per_episode[1].time_accuracy (lite_debugging_b)
\newcommand{\TerraNoneDebugBInForceSeg}{45.8}                           % equal mean across passes: docs/lite/results/four-family/gpt-5.6-terra-none/analysis.json:scores.per_episode[1].segment_time_accuracy (lite_debugging_b); docs/lite/results/hosted-api-repeats-20261003/pass2/gpt-5.6-terra-none/analysis.json:scores.per_episode[1].segment_time_accuracy (lite_debugging_b); docs/lite/results/hosted-api-repeats-20261003/pass3/gpt-5.6-terra-none/analysis.json:scores.per_episode[1].segment_time_accuracy (lite_debugging_b)
\newcommand{\TerraNoneDebugBErrNoDecisionSec}{1.7}                      % equal mean across passes: docs/lite/results/four-family/gpt-5.6-terra-none/analysis.json:scores.per_episode[1].error_seconds.no_decision (lite_debugging_b); docs/lite/results/hosted-api-repeats-20261003/pass2/gpt-5.6-terra-none/analysis.json:scores.per_episode[1].error_seconds.no_decision (lite_debugging_b); docs/lite/results/hosted-api-repeats-20261003/pass3/gpt-5.6-terra-none/analysis.json:scores.per_episode[1].error_seconds.no_decision (lite_debugging_b)
\newcommand{\TerraNoneDebugBErrStaleSec}{18.7}                          % equal mean across passes: docs/lite/results/four-family/gpt-5.6-terra-none/analysis.json:scores.per_episode[1].error_seconds.source_correct (lite_debugging_b); docs/lite/results/hosted-api-repeats-20261003/pass2/gpt-5.6-terra-none/analysis.json:scores.per_episode[1].error_seconds.source_correct (lite_debugging_b); docs/lite/results/hosted-api-repeats-20261003/pass3/gpt-5.6-terra-none/analysis.json:scores.per_episode[1].error_seconds.source_correct (lite_debugging_b)
\newcommand{\TerraNoneDebugBLatencyMedian}{1.36}                        % equal mean across passes: docs/lite/results/four-family/gpt-5.6-terra-none/analysis.json:scores.per_episode[1].latency_s_p50 (lite_debugging_b); docs/lite/results/hosted-api-repeats-20261003/pass2/gpt-5.6-terra-none/analysis.json:scores.per_episode[1].latency_s_p50 (lite_debugging_b); docs/lite/results/hosted-api-repeats-20261003/pass3/gpt-5.6-terra-none/analysis.json:scores.per_episode[1].latency_s_p50 (lite_debugging_b)
\newcommand{\TerraNoneAssemblyAUntimed}{78.3}                           % equal mean across passes: docs/lite/results/four-family/gpt-5.6-terra-none/analysis.json:scores.per_episode[2].untimed_decision_accuracy (lite_assembly_a); docs/lite/results/hosted-api-repeats-20261003/pass2/gpt-5.6-terra-none/analysis.json:scores.per_episode[2].untimed_decision_accuracy (lite_assembly_a); docs/lite/results/hosted-api-repeats-20261003/pass3/gpt-5.6-terra-none/analysis.json:scores.per_episode[2].untimed_decision_accuracy (lite_assembly_a)
\newcommand{\TerraNoneAssemblyAInForce}{57.0}                           % equal mean across passes: docs/lite/results/four-family/gpt-5.6-terra-none/analysis.json:scores.per_episode[2].time_accuracy (lite_assembly_a); docs/lite/results/hosted-api-repeats-20261003/pass2/gpt-5.6-terra-none/analysis.json:scores.per_episode[2].time_accuracy (lite_assembly_a); docs/lite/results/hosted-api-repeats-20261003/pass3/gpt-5.6-terra-none/analysis.json:scores.per_episode[2].time_accuracy (lite_assembly_a)
\newcommand{\TerraNoneAssemblyAInForceSeg}{50.2}                        % equal mean across passes: docs/lite/results/four-family/gpt-5.6-terra-none/analysis.json:scores.per_episode[2].segment_time_accuracy (lite_assembly_a); docs/lite/results/hosted-api-repeats-20261003/pass2/gpt-5.6-terra-none/analysis.json:scores.per_episode[2].segment_time_accuracy (lite_assembly_a); docs/lite/results/hosted-api-repeats-20261003/pass3/gpt-5.6-terra-none/analysis.json:scores.per_episode[2].segment_time_accuracy (lite_assembly_a)
\newcommand{\TerraNoneAssemblyAErrNoDecisionSec}{1.7}                   % equal mean across passes: docs/lite/results/four-family/gpt-5.6-terra-none/analysis.json:scores.per_episode[2].error_seconds.no_decision (lite_assembly_a); docs/lite/results/hosted-api-repeats-20261003/pass2/gpt-5.6-terra-none/analysis.json:scores.per_episode[2].error_seconds.no_decision (lite_assembly_a); docs/lite/results/hosted-api-repeats-20261003/pass3/gpt-5.6-terra-none/analysis.json:scores.per_episode[2].error_seconds.no_decision (lite_assembly_a)
\newcommand{\TerraNoneAssemblyAErrStaleSec}{24.4}                       % equal mean across passes: docs/lite/results/four-family/gpt-5.6-terra-none/analysis.json:scores.per_episode[2].error_seconds.source_correct (lite_assembly_a); docs/lite/results/hosted-api-repeats-20261003/pass2/gpt-5.6-terra-none/analysis.json:scores.per_episode[2].error_seconds.source_correct (lite_assembly_a); docs/lite/results/hosted-api-repeats-20261003/pass3/gpt-5.6-terra-none/analysis.json:scores.per_episode[2].error_seconds.source_correct (lite_assembly_a)
\newcommand{\TerraNoneAssemblyALatencyMedian}{1.41}                     % equal mean across passes: docs/lite/results/four-family/gpt-5.6-terra-none/analysis.json:scores.per_episode[2].latency_s_p50 (lite_assembly_a); docs/lite/results/hosted-api-repeats-20261003/pass2/gpt-5.6-terra-none/analysis.json:scores.per_episode[2].latency_s_p50 (lite_assembly_a); docs/lite/results/hosted-api-repeats-20261003/pass3/gpt-5.6-terra-none/analysis.json:scores.per_episode[2].latency_s_p50 (lite_assembly_a)
\newcommand{\TerraNoneAssemblyBUntimed}{71.1}                           % equal mean across passes: docs/lite/results/four-family/gpt-5.6-terra-none/analysis.json:scores.per_episode[3].untimed_decision_accuracy (lite_assembly_b); docs/lite/results/hosted-api-repeats-20261003/pass2/gpt-5.6-terra-none/analysis.json:scores.per_episode[3].untimed_decision_accuracy (lite_assembly_b); docs/lite/results/hosted-api-repeats-20261003/pass3/gpt-5.6-terra-none/analysis.json:scores.per_episode[3].untimed_decision_accuracy (lite_assembly_b)
\newcommand{\TerraNoneAssemblyBInForce}{53.3}                           % equal mean across passes: docs/lite/results/four-family/gpt-5.6-terra-none/analysis.json:scores.per_episode[3].time_accuracy (lite_assembly_b); docs/lite/results/hosted-api-repeats-20261003/pass2/gpt-5.6-terra-none/analysis.json:scores.per_episode[3].time_accuracy (lite_assembly_b); docs/lite/results/hosted-api-repeats-20261003/pass3/gpt-5.6-terra-none/analysis.json:scores.per_episode[3].time_accuracy (lite_assembly_b)
\newcommand{\TerraNoneAssemblyBInForceSeg}{44.6}                        % equal mean across passes: docs/lite/results/four-family/gpt-5.6-terra-none/analysis.json:scores.per_episode[3].segment_time_accuracy (lite_assembly_b); docs/lite/results/hosted-api-repeats-20261003/pass2/gpt-5.6-terra-none/analysis.json:scores.per_episode[3].segment_time_accuracy (lite_assembly_b); docs/lite/results/hosted-api-repeats-20261003/pass3/gpt-5.6-terra-none/analysis.json:scores.per_episode[3].segment_time_accuracy (lite_assembly_b)
\newcommand{\TerraNoneAssemblyBErrNoDecisionSec}{1.7}                   % equal mean across passes: docs/lite/results/four-family/gpt-5.6-terra-none/analysis.json:scores.per_episode[3].error_seconds.no_decision (lite_assembly_b); docs/lite/results/hosted-api-repeats-20261003/pass2/gpt-5.6-terra-none/analysis.json:scores.per_episode[3].error_seconds.no_decision (lite_assembly_b); docs/lite/results/hosted-api-repeats-20261003/pass3/gpt-5.6-terra-none/analysis.json:scores.per_episode[3].error_seconds.no_decision (lite_assembly_b)
\newcommand{\TerraNoneAssemblyBErrStaleSec}{22.4}                       % equal mean across passes: docs/lite/results/four-family/gpt-5.6-terra-none/analysis.json:scores.per_episode[3].error_seconds.source_correct (lite_assembly_b); docs/lite/results/hosted-api-repeats-20261003/pass2/gpt-5.6-terra-none/analysis.json:scores.per_episode[3].error_seconds.source_correct (lite_assembly_b); docs/lite/results/hosted-api-repeats-20261003/pass3/gpt-5.6-terra-none/analysis.json:scores.per_episode[3].error_seconds.source_correct (lite_assembly_b)
\newcommand{\TerraNoneAssemblyBLatencyMedian}{1.31}                     % equal mean across passes: docs/lite/results/four-family/gpt-5.6-terra-none/analysis.json:scores.per_episode[3].latency_s_p50 (lite_assembly_b); docs/lite/results/hosted-api-repeats-20261003/pass2/gpt-5.6-terra-none/analysis.json:scores.per_episode[3].latency_s_p50 (lite_assembly_b); docs/lite/results/hosted-api-repeats-20261003/pass3/gpt-5.6-terra-none/analysis.json:scores.per_episode[3].latency_s_p50 (lite_assembly_b)
\newcommand{\TerraNoneSupportAUntimed}{91.1}                            % equal mean across passes: docs/lite/results/four-family/gpt-5.6-terra-none/analysis.json:scores.per_episode[4].untimed_decision_accuracy (lite_support_a); docs/lite/results/hosted-api-repeats-20261003/pass2/gpt-5.6-terra-none/analysis.json:scores.per_episode[4].untimed_decision_accuracy (lite_support_a); docs/lite/results/hosted-api-repeats-20261003/pass3/gpt-5.6-terra-none/analysis.json:scores.per_episode[4].untimed_decision_accuracy (lite_support_a)
\newcommand{\TerraNoneSupportAInForce}{67.7}                            % equal mean across passes: docs/lite/results/four-family/gpt-5.6-terra-none/analysis.json:scores.per_episode[4].time_accuracy (lite_support_a); docs/lite/results/hosted-api-repeats-20261003/pass2/gpt-5.6-terra-none/analysis.json:scores.per_episode[4].time_accuracy (lite_support_a); docs/lite/results/hosted-api-repeats-20261003/pass3/gpt-5.6-terra-none/analysis.json:scores.per_episode[4].time_accuracy (lite_support_a)
\newcommand{\TerraNoneSupportAInForceSeg}{59.7}                         % equal mean across passes: docs/lite/results/four-family/gpt-5.6-terra-none/analysis.json:scores.per_episode[4].segment_time_accuracy (lite_support_a); docs/lite/results/hosted-api-repeats-20261003/pass2/gpt-5.6-terra-none/analysis.json:scores.per_episode[4].segment_time_accuracy (lite_support_a); docs/lite/results/hosted-api-repeats-20261003/pass3/gpt-5.6-terra-none/analysis.json:scores.per_episode[4].segment_time_accuracy (lite_support_a)
\newcommand{\TerraNoneSupportAErrNoDecisionSec}{1.7}                    % equal mean across passes: docs/lite/results/four-family/gpt-5.6-terra-none/analysis.json:scores.per_episode[4].error_seconds.no_decision (lite_support_a); docs/lite/results/hosted-api-repeats-20261003/pass2/gpt-5.6-terra-none/analysis.json:scores.per_episode[4].error_seconds.no_decision (lite_support_a); docs/lite/results/hosted-api-repeats-20261003/pass3/gpt-5.6-terra-none/analysis.json:scores.per_episode[4].error_seconds.no_decision (lite_support_a)
\newcommand{\TerraNoneSupportAErrStaleSec}{27.5}                        % equal mean across passes: docs/lite/results/four-family/gpt-5.6-terra-none/analysis.json:scores.per_episode[4].error_seconds.source_correct (lite_support_a); docs/lite/results/hosted-api-repeats-20261003/pass2/gpt-5.6-terra-none/analysis.json:scores.per_episode[4].error_seconds.source_correct (lite_support_a); docs/lite/results/hosted-api-repeats-20261003/pass3/gpt-5.6-terra-none/analysis.json:scores.per_episode[4].error_seconds.source_correct (lite_support_a)
\newcommand{\TerraNoneSupportALatencyMedian}{1.38}                      % equal mean across passes: docs/lite/results/four-family/gpt-5.6-terra-none/analysis.json:scores.per_episode[4].latency_s_p50 (lite_support_a); docs/lite/results/hosted-api-repeats-20261003/pass2/gpt-5.6-terra-none/analysis.json:scores.per_episode[4].latency_s_p50 (lite_support_a); docs/lite/results/hosted-api-repeats-20261003/pass3/gpt-5.6-terra-none/analysis.json:scores.per_episode[4].latency_s_p50 (lite_support_a)
\newcommand{\TerraNoneSupportBUntimed}{96.1}                            % equal mean across passes: docs/lite/results/four-family/gpt-5.6-terra-none/analysis.json:scores.per_episode[5].untimed_decision_accuracy (lite_support_b); docs/lite/results/hosted-api-repeats-20261003/pass2/gpt-5.6-terra-none/analysis.json:scores.per_episode[5].untimed_decision_accuracy (lite_support_b); docs/lite/results/hosted-api-repeats-20261003/pass3/gpt-5.6-terra-none/analysis.json:scores.per_episode[5].untimed_decision_accuracy (lite_support_b)
\newcommand{\TerraNoneSupportBInForce}{64.9}                            % equal mean across passes: docs/lite/results/four-family/gpt-5.6-terra-none/analysis.json:scores.per_episode[5].time_accuracy (lite_support_b); docs/lite/results/hosted-api-repeats-20261003/pass2/gpt-5.6-terra-none/analysis.json:scores.per_episode[5].time_accuracy (lite_support_b); docs/lite/results/hosted-api-repeats-20261003/pass3/gpt-5.6-terra-none/analysis.json:scores.per_episode[5].time_accuracy (lite_support_b)
\newcommand{\TerraNoneSupportBInForceSeg}{59.7}                         % equal mean across passes: docs/lite/results/four-family/gpt-5.6-terra-none/analysis.json:scores.per_episode[5].segment_time_accuracy (lite_support_b); docs/lite/results/hosted-api-repeats-20261003/pass2/gpt-5.6-terra-none/analysis.json:scores.per_episode[5].segment_time_accuracy (lite_support_b); docs/lite/results/hosted-api-repeats-20261003/pass3/gpt-5.6-terra-none/analysis.json:scores.per_episode[5].segment_time_accuracy (lite_support_b)
\newcommand{\TerraNoneSupportBErrNoDecisionSec}{1.6}                    % equal mean across passes: docs/lite/results/four-family/gpt-5.6-terra-none/analysis.json:scores.per_episode[5].error_seconds.no_decision (lite_support_b); docs/lite/results/hosted-api-repeats-20261003/pass2/gpt-5.6-terra-none/analysis.json:scores.per_episode[5].error_seconds.no_decision (lite_support_b); docs/lite/results/hosted-api-repeats-20261003/pass3/gpt-5.6-terra-none/analysis.json:scores.per_episode[5].error_seconds.no_decision (lite_support_b)
\newcommand{\TerraNoneSupportBErrStaleSec}{35.8}                        % equal mean across passes: docs/lite/results/four-family/gpt-5.6-terra-none/analysis.json:scores.per_episode[5].error_seconds.source_correct (lite_support_b); docs/lite/results/hosted-api-repeats-20261003/pass2/gpt-5.6-terra-none/analysis.json:scores.per_episode[5].error_seconds.source_correct (lite_support_b); docs/lite/results/hosted-api-repeats-20261003/pass3/gpt-5.6-terra-none/analysis.json:scores.per_episode[5].error_seconds.source_correct (lite_support_b)
\newcommand{\TerraNoneSupportBLatencyMedian}{1.41}                      % equal mean across passes: docs/lite/results/four-family/gpt-5.6-terra-none/analysis.json:scores.per_episode[5].latency_s_p50 (lite_support_b); docs/lite/results/hosted-api-repeats-20261003/pass2/gpt-5.6-terra-none/analysis.json:scores.per_episode[5].latency_s_p50 (lite_support_b); docs/lite/results/hosted-api-repeats-20261003/pass3/gpt-5.6-terra-none/analysis.json:scores.per_episode[5].latency_s_p50 (lite_support_b)
\newcommand{\TerraNonePresenterAUntimed}{90.6}                          % equal mean across passes: docs/lite/results/four-family/gpt-5.6-terra-none/analysis.json:scores.per_episode[6].untimed_decision_accuracy (lite_presenter_a); docs/lite/results/hosted-api-repeats-20261003/pass2/gpt-5.6-terra-none/analysis.json:scores.per_episode[6].untimed_decision_accuracy (lite_presenter_a); docs/lite/results/hosted-api-repeats-20261003/pass3/gpt-5.6-terra-none/analysis.json:scores.per_episode[6].untimed_decision_accuracy (lite_presenter_a)
\newcommand{\TerraNonePresenterAInForce}{69.8}                          % equal mean across passes: docs/lite/results/four-family/gpt-5.6-terra-none/analysis.json:scores.per_episode[6].time_accuracy (lite_presenter_a); docs/lite/results/hosted-api-repeats-20261003/pass2/gpt-5.6-terra-none/analysis.json:scores.per_episode[6].time_accuracy (lite_presenter_a); docs/lite/results/hosted-api-repeats-20261003/pass3/gpt-5.6-terra-none/analysis.json:scores.per_episode[6].time_accuracy (lite_presenter_a)
\newcommand{\TerraNonePresenterAInForceSeg}{63.6}                       % equal mean across passes: docs/lite/results/four-family/gpt-5.6-terra-none/analysis.json:scores.per_episode[6].segment_time_accuracy (lite_presenter_a); docs/lite/results/hosted-api-repeats-20261003/pass2/gpt-5.6-terra-none/analysis.json:scores.per_episode[6].segment_time_accuracy (lite_presenter_a); docs/lite/results/hosted-api-repeats-20261003/pass3/gpt-5.6-terra-none/analysis.json:scores.per_episode[6].segment_time_accuracy (lite_presenter_a)
\newcommand{\TerraNonePresenterAErrNoDecisionSec}{1.6}                  % equal mean across passes: docs/lite/results/four-family/gpt-5.6-terra-none/analysis.json:scores.per_episode[6].error_seconds.no_decision (lite_presenter_a); docs/lite/results/hosted-api-repeats-20261003/pass2/gpt-5.6-terra-none/analysis.json:scores.per_episode[6].error_seconds.no_decision (lite_presenter_a); docs/lite/results/hosted-api-repeats-20261003/pass3/gpt-5.6-terra-none/analysis.json:scores.per_episode[6].error_seconds.no_decision (lite_presenter_a)
\newcommand{\TerraNonePresenterAErrStaleSec}{27.3}                      % equal mean across passes: docs/lite/results/four-family/gpt-5.6-terra-none/analysis.json:scores.per_episode[6].error_seconds.source_correct (lite_presenter_a); docs/lite/results/hosted-api-repeats-20261003/pass2/gpt-5.6-terra-none/analysis.json:scores.per_episode[6].error_seconds.source_correct (lite_presenter_a); docs/lite/results/hosted-api-repeats-20261003/pass3/gpt-5.6-terra-none/analysis.json:scores.per_episode[6].error_seconds.source_correct (lite_presenter_a)
\newcommand{\TerraNonePresenterALatencyMedian}{1.27}                    % equal mean across passes: docs/lite/results/four-family/gpt-5.6-terra-none/analysis.json:scores.per_episode[6].latency_s_p50 (lite_presenter_a); docs/lite/results/hosted-api-repeats-20261003/pass2/gpt-5.6-terra-none/analysis.json:scores.per_episode[6].latency_s_p50 (lite_presenter_a); docs/lite/results/hosted-api-repeats-20261003/pass3/gpt-5.6-terra-none/analysis.json:scores.per_episode[6].latency_s_p50 (lite_presenter_a)
\newcommand{\TerraNonePresenterBUntimed}{83.3}                          % equal mean across passes: docs/lite/results/four-family/gpt-5.6-terra-none/analysis.json:scores.per_episode[7].untimed_decision_accuracy (lite_presenter_b); docs/lite/results/hosted-api-repeats-20261003/pass2/gpt-5.6-terra-none/analysis.json:scores.per_episode[7].untimed_decision_accuracy (lite_presenter_b); docs/lite/results/hosted-api-repeats-20261003/pass3/gpt-5.6-terra-none/analysis.json:scores.per_episode[7].untimed_decision_accuracy (lite_presenter_b)
\newcommand{\TerraNonePresenterBInForce}{65.4}                          % equal mean across passes: docs/lite/results/four-family/gpt-5.6-terra-none/analysis.json:scores.per_episode[7].time_accuracy (lite_presenter_b); docs/lite/results/hosted-api-repeats-20261003/pass2/gpt-5.6-terra-none/analysis.json:scores.per_episode[7].time_accuracy (lite_presenter_b); docs/lite/results/hosted-api-repeats-20261003/pass3/gpt-5.6-terra-none/analysis.json:scores.per_episode[7].time_accuracy (lite_presenter_b)
\newcommand{\TerraNonePresenterBInForceSeg}{61.7}                       % equal mean across passes: docs/lite/results/four-family/gpt-5.6-terra-none/analysis.json:scores.per_episode[7].segment_time_accuracy (lite_presenter_b); docs/lite/results/hosted-api-repeats-20261003/pass2/gpt-5.6-terra-none/analysis.json:scores.per_episode[7].segment_time_accuracy (lite_presenter_b); docs/lite/results/hosted-api-repeats-20261003/pass3/gpt-5.6-terra-none/analysis.json:scores.per_episode[7].segment_time_accuracy (lite_presenter_b)
\newcommand{\TerraNonePresenterBErrNoDecisionSec}{1.4}                  % equal mean across passes: docs/lite/results/four-family/gpt-5.6-terra-none/analysis.json:scores.per_episode[7].error_seconds.no_decision (lite_presenter_b); docs/lite/results/hosted-api-repeats-20261003/pass2/gpt-5.6-terra-none/analysis.json:scores.per_episode[7].error_seconds.no_decision (lite_presenter_b); docs/lite/results/hosted-api-repeats-20261003/pass3/gpt-5.6-terra-none/analysis.json:scores.per_episode[7].error_seconds.no_decision (lite_presenter_b)
\newcommand{\TerraNonePresenterBErrStaleSec}{25.4}                      % equal mean across passes: docs/lite/results/four-family/gpt-5.6-terra-none/analysis.json:scores.per_episode[7].error_seconds.source_correct (lite_presenter_b); docs/lite/results/hosted-api-repeats-20261003/pass2/gpt-5.6-terra-none/analysis.json:scores.per_episode[7].error_seconds.source_correct (lite_presenter_b); docs/lite/results/hosted-api-repeats-20261003/pass3/gpt-5.6-terra-none/analysis.json:scores.per_episode[7].error_seconds.source_correct (lite_presenter_b)
\newcommand{\TerraNonePresenterBLatencyMedian}{1.30}                    % equal mean across passes: docs/lite/results/four-family/gpt-5.6-terra-none/analysis.json:scores.per_episode[7].latency_s_p50 (lite_presenter_b); docs/lite/results/hosted-api-repeats-20261003/pass2/gpt-5.6-terra-none/analysis.json:scores.per_episode[7].latency_s_p50 (lite_presenter_b); docs/lite/results/hosted-api-repeats-20261003/pass3/gpt-5.6-terra-none/analysis.json:scores.per_episode[7].latency_s_p50 (lite_presenter_b)
\newcommand{\AstraUntimed}{99.9}                                        % equal mean across passes: docs/lite/results/four-family/gpt-6-astra-low/analysis.json:scores.overall.untimed_decision_accuracy; docs/lite/results/hosted-api-repeats-20261003/pass2/gpt-6-astra-low/analysis.json:scores.overall.untimed_decision_accuracy; docs/lite/results/hosted-api-repeats-20261003/pass3/gpt-6-astra-low/analysis.json:scores.overall.untimed_decision_accuracy
\newcommand{\AstraInForce}{52.1}                                        % equal mean across passes: docs/lite/results/four-family/gpt-6-astra-low/analysis.json:scores.overall.time_accuracy; docs/lite/results/hosted-api-repeats-20261003/pass2/gpt-6-astra-low/analysis.json:scores.overall.time_accuracy; docs/lite/results/hosted-api-repeats-20261003/pass3/gpt-6-astra-low/analysis.json:scores.overall.time_accuracy
\newcommand{\AstraInForceSeg}{43.0}                                     % equal mean across passes: docs/lite/results/four-family/gpt-6-astra-low/analysis.json:scores.overall.segment_time_accuracy; docs/lite/results/hosted-api-repeats-20261003/pass2/gpt-6-astra-low/analysis.json:scores.overall.segment_time_accuracy; docs/lite/results/hosted-api-repeats-20261003/pass3/gpt-6-astra-low/analysis.json:scores.overall.segment_time_accuracy
\newcommand{\AstraUntimedStates}{1438}                                  % docs/research/manuscript-three-pass/analysis.json:Astra three-pass agreement
\newcommand{\AstraNetRemovedSeg}{61.4}                                  % equal mean across passes: docs/lite/results/four-family/gpt-6-astra-low/analysis.json:network_adjustment.scores.estimate.overall.segment_time_accuracy; docs/lite/results/hosted-api-repeats-20261003/pass2/gpt-6-astra-low/analysis.json:network_adjustment.scores.estimate.overall.segment_time_accuracy; docs/lite/results/hosted-api-repeats-20261003/pass3/gpt-6-astra-low/analysis.json:network_adjustment.scores.estimate.overall.segment_time_accuracy
\newcommand{\AstraNetSec}{0.90}                                         % equal mean across passes: docs/lite/results/four-family/gpt-6-astra-low/analysis.json:network_adjustment.network_s.estimate; docs/lite/results/hosted-api-repeats-20261003/pass2/gpt-6-astra-low/analysis.json:network_adjustment.network_s.estimate; docs/lite/results/hosted-api-repeats-20261003/pass3/gpt-6-astra-low/analysis.json:network_adjustment.network_s.estimate
\newcommand{\AstraNetRemovedSegLow}{52.9}                               % equal mean across passes: docs/lite/results/four-family/gpt-6-astra-low/analysis.json:network_adjustment.scores.low.overall.segment_time_accuracy; docs/lite/results/hosted-api-repeats-20261003/pass2/gpt-6-astra-low/analysis.json:network_adjustment.scores.low.overall.segment_time_accuracy; docs/lite/results/hosted-api-repeats-20261003/pass3/gpt-6-astra-low/analysis.json:network_adjustment.scores.low.overall.segment_time_accuracy
\newcommand{\AstraNetSecLow}{0.50}                                      % equal mean across passes: docs/lite/results/four-family/gpt-6-astra-low/analysis.json:network_adjustment.network_s.low; docs/lite/results/hosted-api-repeats-20261003/pass2/gpt-6-astra-low/analysis.json:network_adjustment.network_s.low; docs/lite/results/hosted-api-repeats-20261003/pass3/gpt-6-astra-low/analysis.json:network_adjustment.network_s.low
\newcommand{\AstraNetRemovedSegHigh}{76.6}                              % equal mean across passes: docs/lite/results/four-family/gpt-6-astra-low/analysis.json:network_adjustment.scores.high.overall.segment_time_accuracy; docs/lite/results/hosted-api-repeats-20261003/pass2/gpt-6-astra-low/analysis.json:network_adjustment.scores.high.overall.segment_time_accuracy; docs/lite/results/hosted-api-repeats-20261003/pass3/gpt-6-astra-low/analysis.json:network_adjustment.scores.high.overall.segment_time_accuracy
\newcommand{\AstraNetSecHigh}{1.56}                                     % equal mean across passes: docs/lite/results/four-family/gpt-6-astra-low/analysis.json:network_adjustment.network_s.high; docs/lite/results/hosted-api-repeats-20261003/pass2/gpt-6-astra-low/analysis.json:network_adjustment.network_s.high; docs/lite/results/hosted-api-repeats-20261003/pass3/gpt-6-astra-low/analysis.json:network_adjustment.network_s.high
\newcommand{\AstraPrefillMsPerKTok}{17.6}                               % equal mean across passes: docs/lite/results/four-family/gpt-6-astra-low/analysis.json:network_adjustment.prefill_s_per_1k_input_tokens (x 1000); docs/lite/results/hosted-api-repeats-20261003/pass2/gpt-6-astra-low/analysis.json:network_adjustment.prefill_s_per_1k_input_tokens (x 1000); docs/lite/results/hosted-api-repeats-20261003/pass3/gpt-6-astra-low/analysis.json:network_adjustment.prefill_s_per_1k_input_tokens (x 1000)
\newcommand{\AstraDecodeMsPerTok}{23.1}                                 % equal mean across passes: docs/lite/results/four-family/gpt-6-astra-low/analysis.json:network_adjustment.decode_s_per_output_token (x 1000); docs/lite/results/hosted-api-repeats-20261003/pass2/gpt-6-astra-low/analysis.json:network_adjustment.decode_s_per_output_token (x 1000); docs/lite/results/hosted-api-repeats-20261003/pass3/gpt-6-astra-low/analysis.json:network_adjustment.decode_s_per_output_token (x 1000)
\newcommand{\AstraLatencyFloor}{1.45}                                   % minimum across passes: docs/lite/results/four-family/gpt-6-astra-low/analysis.json:network_adjustment.latency_floor_s; docs/lite/results/hosted-api-repeats-20261003/pass2/gpt-6-astra-low/analysis.json:network_adjustment.latency_floor_s; docs/lite/results/hosted-api-repeats-20261003/pass3/gpt-6-astra-low/analysis.json:network_adjustment.latency_floor_s
\newcommand{\AstraLatencyMedian}{2.35}                                  % equal mean across passes: docs/lite/results/four-family/gpt-6-astra-low/analysis.json:latency_s.p50; docs/lite/results/hosted-api-repeats-20261003/pass2/gpt-6-astra-low/analysis.json:latency_s.p50; docs/lite/results/hosted-api-repeats-20261003/pass3/gpt-6-astra-low/analysis.json:latency_s.p50
\newcommand{\AstraLatencyPNinetyFive}{3.67}                             % equal mean across passes: docs/lite/results/four-family/gpt-6-astra-low/analysis.json:latency_s.p95; docs/lite/results/hosted-api-repeats-20261003/pass2/gpt-6-astra-low/analysis.json:latency_s.p95; docs/lite/results/hosted-api-repeats-20261003/pass3/gpt-6-astra-low/analysis.json:latency_s.p95
\newcommand{\AstraLatencyMax}{7.50}                                     % maximum across passes: docs/lite/results/four-family/gpt-6-astra-low/analysis.json:latency_s.max; docs/lite/results/hosted-api-repeats-20261003/pass2/gpt-6-astra-low/analysis.json:latency_s.max; docs/lite/results/hosted-api-repeats-20261003/pass3/gpt-6-astra-low/analysis.json:latency_s.max
\newcommand{\AstraLatencyMean}{2.50}                                    % equal mean across passes: runs/lite-v1-gpt-6-astra-low-four-family-retry-v1/events.jsonl:response.attempts[-1].completed_s - started_s (mean over 480 responses); runs/hosted-api-repeats-20261003/pass2/gpt-6-astra-low/events.jsonl:response.attempts[-1].completed_s - started_s (mean over 480 responses); runs/hosted-api-repeats-20261003/pass3/gpt-6-astra-low/events.jsonl:response.attempts[-1].completed_s - started_s (mean over 480 responses)
\newcommand{\AstraInputTokMedian}{3034}                                 % equal mean across passes: runs/lite-v1-gpt-6-astra-low-four-family-retry-v1/events.jsonl:response.usage.input_tokens (median over 480 responses, half-up); runs/hosted-api-repeats-20261003/pass2/gpt-6-astra-low/events.jsonl:response.usage.input_tokens (median over 480 responses, half-up); runs/hosted-api-repeats-20261003/pass3/gpt-6-astra-low/events.jsonl:response.usage.input_tokens (median over 480 responses, half-up)
\newcommand{\AstraInputTokTotal}{4{,}429{,}467}                         % sum across passes: docs/lite/results/four-family/gpt-6-astra-low/analysis.json:usage.input_tokens; docs/lite/results/hosted-api-repeats-20261003/pass2/gpt-6-astra-low/analysis.json:usage.input_tokens; docs/lite/results/hosted-api-repeats-20261003/pass3/gpt-6-astra-low/analysis.json:usage.input_tokens
\newcommand{\AstraOutputTokMedian}{48}                                  % equal mean across passes: runs/lite-v1-gpt-6-astra-low-four-family-retry-v1/events.jsonl:response.usage.output_tokens (median over 480 responses, half-up); runs/hosted-api-repeats-20261003/pass2/gpt-6-astra-low/events.jsonl:response.usage.output_tokens (median over 480 responses, half-up); runs/hosted-api-repeats-20261003/pass3/gpt-6-astra-low/events.jsonl:response.usage.output_tokens (median over 480 responses, half-up)
\newcommand{\AstraOutputTokTotal}{66{,}998}                             % sum across passes: docs/lite/results/four-family/gpt-6-astra-low/analysis.json:usage.output_tokens; docs/lite/results/hosted-api-repeats-20261003/pass2/gpt-6-astra-low/analysis.json:usage.output_tokens; docs/lite/results/hosted-api-repeats-20261003/pass3/gpt-6-astra-low/analysis.json:usage.output_tokens
\newcommand{\AstraReasoningTokMedian}{0}                                % equal mean across passes: runs/lite-v1-gpt-6-astra-low-four-family-retry-v1/events.jsonl:response.usage.reasoning_tokens (median over 480 responses, half-up); runs/hosted-api-repeats-20261003/pass2/gpt-6-astra-low/events.jsonl:response.usage.reasoning_tokens (median over 480 responses, half-up); runs/hosted-api-repeats-20261003/pass3/gpt-6-astra-low/events.jsonl:response.usage.reasoning_tokens (median over 480 responses, half-up)
\newcommand{\AstraCachedTokTotal}{0}                                    % sum across passes: docs/lite/results/four-family/gpt-6-astra-low/analysis.json:usage.cached_tokens; docs/lite/results/hosted-api-repeats-20261003/pass2/gpt-6-astra-low/analysis.json:usage.cached_tokens; docs/lite/results/hosted-api-repeats-20261003/pass3/gpt-6-astra-low/analysis.json:usage.cached_tokens
\newcommand{\AstraCostInputUSD}{44.295}                                 % sum across passes: docs/lite/results/four-family/gpt-6-astra-low/analysis.json:usage at developers.openai.com/api/docs/pricing (gpt-6-astra, Standard, short context) list prices retrieved 2026-09-30 (input: uncached x input price + cached x cached price); docs/lite/results/hosted-api-repeats-20261003/pass2/gpt-6-astra-low/analysis.json:usage at developers.openai.com/api/docs/pricing (gpt-6-astra, Standard, short context) list prices retrieved 2026-09-30 (input: uncached x input price + cached x cached price); docs/lite/results/hosted-api-repeats-20261003/pass3/gpt-6-astra-low/analysis.json:usage at developers.openai.com/api/docs/pricing (gpt-6-astra, Standard, short context) list prices retrieved 2026-09-30 (input: uncached x input price + cached x cached price)
\newcommand{\AstraCostOutputUSD}{3.350}                                 % sum across passes: docs/lite/results/four-family/gpt-6-astra-low/analysis.json:usage at developers.openai.com/api/docs/pricing (gpt-6-astra, Standard, short context) list prices retrieved 2026-09-30 (output tokens, reasoning included, x output price); docs/lite/results/hosted-api-repeats-20261003/pass2/gpt-6-astra-low/analysis.json:usage at developers.openai.com/api/docs/pricing (gpt-6-astra, Standard, short context) list prices retrieved 2026-09-30 (output tokens, reasoning included, x output price); docs/lite/results/hosted-api-repeats-20261003/pass3/gpt-6-astra-low/analysis.json:usage at developers.openai.com/api/docs/pricing (gpt-6-astra, Standard, short context) list prices retrieved 2026-09-30 (output tokens, reasoning included, x output price)
\newcommand{\AstraCostUSD}{47.645}                                      % sum across passes: docs/lite/results/four-family/gpt-6-astra-low/analysis.json:usage at developers.openai.com/api/docs/pricing (gpt-6-astra, Standard, short context) list prices retrieved 2026-09-30 (input + output); docs/lite/results/hosted-api-repeats-20261003/pass2/gpt-6-astra-low/analysis.json:usage at developers.openai.com/api/docs/pricing (gpt-6-astra, Standard, short context) list prices retrieved 2026-09-30 (input + output); docs/lite/results/hosted-api-repeats-20261003/pass3/gpt-6-astra-low/analysis.json:usage at developers.openai.com/api/docs/pricing (gpt-6-astra, Standard, short context) list prices retrieved 2026-09-30 (input + output)
\newcommand{\AstraPriceInput}{10.000}                                   % equal mean across passes: developers.openai.com/api/docs/pricing (gpt-6-astra, Standard, short context): USD per 1M input tokens; developers.openai.com/api/docs/pricing (gpt-6-astra, Standard, short context): USD per 1M input tokens; developers.openai.com/api/docs/pricing (gpt-6-astra, Standard, short context): USD per 1M input tokens
\newcommand{\AstraPriceOutput}{50.00}                                   % equal mean across passes: developers.openai.com/api/docs/pricing (gpt-6-astra, Standard, short context): USD per 1M output tokens; developers.openai.com/api/docs/pricing (gpt-6-astra, Standard, short context): USD per 1M output tokens; developers.openai.com/api/docs/pricing (gpt-6-astra, Standard, short context): USD per 1M output tokens
\newcommand{\AstraErrNoDecisionPct}{2.2}                                % equal mean across passes: docs/lite/results/four-family/gpt-6-astra-low/analysis.json:error_seconds.no_decision (scenario error fraction, then equal family mean); docs/lite/results/hosted-api-repeats-20261003/pass2/gpt-6-astra-low/analysis.json:error_seconds.no_decision (scenario error fraction, then equal family mean); docs/lite/results/hosted-api-repeats-20261003/pass3/gpt-6-astra-low/analysis.json:error_seconds.no_decision (scenario error fraction, then equal family mean)
\newcommand{\AstraErrStalePct}{45.6}                                    % equal mean across passes: docs/lite/results/four-family/gpt-6-astra-low/analysis.json:error_seconds.source_correct (scenario error fraction, then equal family mean); docs/lite/results/hosted-api-repeats-20261003/pass2/gpt-6-astra-low/analysis.json:error_seconds.source_correct (scenario error fraction, then equal family mean); docs/lite/results/hosted-api-repeats-20261003/pass3/gpt-6-astra-low/analysis.json:error_seconds.source_correct (scenario error fraction, then equal family mean)
\newcommand{\AstraInForceRecorded}{52.1}                                % equal mean across passes: docs/lite/results/four-family/gpt-6-astra-low/analysis.json:scores.overall.time_accuracy; docs/lite/results/hosted-api-repeats-20261003/pass2/gpt-6-astra-low/analysis.json:scores.overall.time_accuracy; docs/lite/results/hosted-api-repeats-20261003/pass3/gpt-6-astra-low/analysis.json:scores.overall.time_accuracy
\newcommand{\AstraInForcePhysical}{52.1}                                % equal mean across passes: docs/lite/results/four-family/gpt-6-astra-low/analysis.json:raw_wallclock_scores.overall.time_accuracy (physical wall-clock trace); docs/lite/results/hosted-api-repeats-20261003/pass2/gpt-6-astra-low/analysis.json:raw_wallclock_scores.overall.time_accuracy (physical wall-clock trace); docs/lite/results/hosted-api-repeats-20261003/pass3/gpt-6-astra-low/analysis.json:raw_wallclock_scores.overall.time_accuracy (physical wall-clock trace)
\newcommand{\AstraExcludedDispatchSec}{0.22}                            % equal mean across passes: docs/lite/results/four-family/gpt-6-astra-low/analysis.json:retry_reliability.excluded_dispatch_s (summed over the pass); docs/lite/results/hosted-api-repeats-20261003/pass2/gpt-6-astra-low/analysis.json:retry_reliability.excluded_dispatch_s (summed over the pass); docs/lite/results/hosted-api-repeats-20261003/pass3/gpt-6-astra-low/analysis.json:retry_reliability.excluded_dispatch_s (summed over the pass)
\newcommand{\AstraEventsHash}{09bf5acb993005bd}                         % pass 1: docs/lite/results/four-family/gpt-6-astra-low/analysis.json:events_sha256 (first 16 hex digits; equals sha256 of runs/lite-v1-gpt-6-astra-low-four-family-retry-v1/events.jsonl)
\newcommand{\AstraAcceptedUpdates}{1410}                                % sum across passes: docs/lite/results/four-family/gpt-6-astra-low/analysis.json:accepted_updates; docs/lite/results/hosted-api-repeats-20261003/pass2/gpt-6-astra-low/analysis.json:accepted_updates; docs/lite/results/hosted-api-repeats-20261003/pass3/gpt-6-astra-low/analysis.json:accepted_updates
\newcommand{\AstraDiscardedAfterHorizon}{17}                            % sum across passes: docs/lite/results/four-family/gpt-6-astra-low/analysis.json:discarded_updates.after_horizon (absent = 0); docs/lite/results/hosted-api-repeats-20261003/pass2/gpt-6-astra-low/analysis.json:discarded_updates.after_horizon (absent = 0); docs/lite/results/hosted-api-repeats-20261003/pass3/gpt-6-astra-low/analysis.json:discarded_updates.after_horizon (absent = 0)
\newcommand{\AstraDiscardedOlder}{13}                                   % sum across passes: docs/lite/results/four-family/gpt-6-astra-low/analysis.json:discarded_updates.older_than_active (absent = 0); docs/lite/results/hosted-api-repeats-20261003/pass2/gpt-6-astra-low/analysis.json:discarded_updates.older_than_active (absent = 0); docs/lite/results/hosted-api-repeats-20261003/pass3/gpt-6-astra-low/analysis.json:discarded_updates.older_than_active (absent = 0)
\newcommand{\AstraValidResponses}{1440}                                 % sum across passes: docs/lite/results/four-family/gpt-6-astra-low/analysis.json:retry_reliability.successful_logical_requests; docs/lite/results/hosted-api-repeats-20261003/pass2/gpt-6-astra-low/analysis.json:retry_reliability.successful_logical_requests; docs/lite/results/hosted-api-repeats-20261003/pass3/gpt-6-astra-low/analysis.json:retry_reliability.successful_logical_requests
\newcommand{\AstraAttempts}{1440}                                       % sum across passes: docs/lite/results/four-family/gpt-6-astra-low/analysis.json:retry_reliability.attempts; docs/lite/results/hosted-api-repeats-20261003/pass2/gpt-6-astra-low/analysis.json:retry_reliability.attempts; docs/lite/results/hosted-api-repeats-20261003/pass3/gpt-6-astra-low/analysis.json:retry_reliability.attempts
\newcommand{\AstraFailedAttempts}{0}                                    % sum across passes: docs/lite/results/four-family/gpt-6-astra-low/analysis.json:retry_reliability.failed_attempts; docs/lite/results/hosted-api-repeats-20261003/pass2/gpt-6-astra-low/analysis.json:retry_reliability.failed_attempts; docs/lite/results/hosted-api-repeats-20261003/pass3/gpt-6-astra-low/analysis.json:retry_reliability.failed_attempts
\newcommand{\AstraRetriedRequests}{0}                                   % sum across passes: docs/lite/results/four-family/gpt-6-astra-low/analysis.json:retry_reliability.retried_logical_requests; docs/lite/results/hosted-api-repeats-20261003/pass2/gpt-6-astra-low/analysis.json:retry_reliability.retried_logical_requests; docs/lite/results/hosted-api-repeats-20261003/pass3/gpt-6-astra-low/analysis.json:retry_reliability.retried_logical_requests
\newcommand{\AstraDebugUntimed}{100.0}                                  % equal mean across passes: docs/lite/results/four-family/gpt-6-astra-low/analysis.json:scores.by_family.live_debugging.untimed_decision_accuracy; docs/lite/results/hosted-api-repeats-20261003/pass2/gpt-6-astra-low/analysis.json:scores.by_family.live_debugging.untimed_decision_accuracy; docs/lite/results/hosted-api-repeats-20261003/pass3/gpt-6-astra-low/analysis.json:scores.by_family.live_debugging.untimed_decision_accuracy
\newcommand{\AstraDebugInForce}{54.5}                                   % equal mean across passes: docs/lite/results/four-family/gpt-6-astra-low/analysis.json:scores.by_family.live_debugging.time_accuracy; docs/lite/results/hosted-api-repeats-20261003/pass2/gpt-6-astra-low/analysis.json:scores.by_family.live_debugging.time_accuracy; docs/lite/results/hosted-api-repeats-20261003/pass3/gpt-6-astra-low/analysis.json:scores.by_family.live_debugging.time_accuracy
\newcommand{\AstraDebugNetRemoved}{70.1}                                % equal mean across passes: docs/lite/results/four-family/gpt-6-astra-low/analysis.json:network_adjustment.scores.estimate.by_family.live_debugging.time_accuracy; docs/lite/results/hosted-api-repeats-20261003/pass2/gpt-6-astra-low/analysis.json:network_adjustment.scores.estimate.by_family.live_debugging.time_accuracy; docs/lite/results/hosted-api-repeats-20261003/pass3/gpt-6-astra-low/analysis.json:network_adjustment.scores.estimate.by_family.live_debugging.time_accuracy
\newcommand{\AstraDebugNetRemovedLow}{63.2}                             % equal mean across passes: docs/lite/results/four-family/gpt-6-astra-low/analysis.json:network_adjustment.scores.low.by_family.live_debugging.time_accuracy; docs/lite/results/hosted-api-repeats-20261003/pass2/gpt-6-astra-low/analysis.json:network_adjustment.scores.low.by_family.live_debugging.time_accuracy; docs/lite/results/hosted-api-repeats-20261003/pass3/gpt-6-astra-low/analysis.json:network_adjustment.scores.low.by_family.live_debugging.time_accuracy
\newcommand{\AstraDebugNetRemovedHigh}{82.0}                            % equal mean across passes: docs/lite/results/four-family/gpt-6-astra-low/analysis.json:network_adjustment.scores.high.by_family.live_debugging.time_accuracy; docs/lite/results/hosted-api-repeats-20261003/pass2/gpt-6-astra-low/analysis.json:network_adjustment.scores.high.by_family.live_debugging.time_accuracy; docs/lite/results/hosted-api-repeats-20261003/pass3/gpt-6-astra-low/analysis.json:network_adjustment.scores.high.by_family.live_debugging.time_accuracy
\newcommand{\AstraAssemblyUntimed}{99.4}                                % equal mean across passes: docs/lite/results/four-family/gpt-6-astra-low/analysis.json:scores.by_family.procedural_coaching.untimed_decision_accuracy; docs/lite/results/hosted-api-repeats-20261003/pass2/gpt-6-astra-low/analysis.json:scores.by_family.procedural_coaching.untimed_decision_accuracy; docs/lite/results/hosted-api-repeats-20261003/pass3/gpt-6-astra-low/analysis.json:scores.by_family.procedural_coaching.untimed_decision_accuracy
\newcommand{\AstraAssemblyInForce}{50.0}                                % equal mean across passes: docs/lite/results/four-family/gpt-6-astra-low/analysis.json:scores.by_family.procedural_coaching.time_accuracy; docs/lite/results/hosted-api-repeats-20261003/pass2/gpt-6-astra-low/analysis.json:scores.by_family.procedural_coaching.time_accuracy; docs/lite/results/hosted-api-repeats-20261003/pass3/gpt-6-astra-low/analysis.json:scores.by_family.procedural_coaching.time_accuracy
\newcommand{\AstraAssemblyNetRemoved}{66.5}                             % equal mean across passes: docs/lite/results/four-family/gpt-6-astra-low/analysis.json:network_adjustment.scores.estimate.by_family.procedural_coaching.time_accuracy; docs/lite/results/hosted-api-repeats-20261003/pass2/gpt-6-astra-low/analysis.json:network_adjustment.scores.estimate.by_family.procedural_coaching.time_accuracy; docs/lite/results/hosted-api-repeats-20261003/pass3/gpt-6-astra-low/analysis.json:network_adjustment.scores.estimate.by_family.procedural_coaching.time_accuracy
\newcommand{\AstraAssemblyNetRemovedLow}{59.1}                          % equal mean across passes: docs/lite/results/four-family/gpt-6-astra-low/analysis.json:network_adjustment.scores.low.by_family.procedural_coaching.time_accuracy; docs/lite/results/hosted-api-repeats-20261003/pass2/gpt-6-astra-low/analysis.json:network_adjustment.scores.low.by_family.procedural_coaching.time_accuracy; docs/lite/results/hosted-api-repeats-20261003/pass3/gpt-6-astra-low/analysis.json:network_adjustment.scores.low.by_family.procedural_coaching.time_accuracy
\newcommand{\AstraAssemblyNetRemovedHigh}{79.1}                         % equal mean across passes: docs/lite/results/four-family/gpt-6-astra-low/analysis.json:network_adjustment.scores.high.by_family.procedural_coaching.time_accuracy; docs/lite/results/hosted-api-repeats-20261003/pass2/gpt-6-astra-low/analysis.json:network_adjustment.scores.high.by_family.procedural_coaching.time_accuracy; docs/lite/results/hosted-api-repeats-20261003/pass3/gpt-6-astra-low/analysis.json:network_adjustment.scores.high.by_family.procedural_coaching.time_accuracy
\newcommand{\AstraSupportUntimed}{100.0}                                % equal mean across passes: docs/lite/results/four-family/gpt-6-astra-low/analysis.json:scores.by_family.support_call_assist.untimed_decision_accuracy; docs/lite/results/hosted-api-repeats-20261003/pass2/gpt-6-astra-low/analysis.json:scores.by_family.support_call_assist.untimed_decision_accuracy; docs/lite/results/hosted-api-repeats-20261003/pass3/gpt-6-astra-low/analysis.json:scores.by_family.support_call_assist.untimed_decision_accuracy
\newcommand{\AstraSupportInForce}{49.6}                                 % equal mean across passes: docs/lite/results/four-family/gpt-6-astra-low/analysis.json:scores.by_family.support_call_assist.time_accuracy; docs/lite/results/hosted-api-repeats-20261003/pass2/gpt-6-astra-low/analysis.json:scores.by_family.support_call_assist.time_accuracy; docs/lite/results/hosted-api-repeats-20261003/pass3/gpt-6-astra-low/analysis.json:scores.by_family.support_call_assist.time_accuracy
\newcommand{\AstraSupportNetRemoved}{65.6}                              % equal mean across passes: docs/lite/results/four-family/gpt-6-astra-low/analysis.json:network_adjustment.scores.estimate.by_family.support_call_assist.time_accuracy; docs/lite/results/hosted-api-repeats-20261003/pass2/gpt-6-astra-low/analysis.json:network_adjustment.scores.estimate.by_family.support_call_assist.time_accuracy; docs/lite/results/hosted-api-repeats-20261003/pass3/gpt-6-astra-low/analysis.json:network_adjustment.scores.estimate.by_family.support_call_assist.time_accuracy
\newcommand{\AstraSupportNetRemovedLow}{58.2}                           % equal mean across passes: docs/lite/results/four-family/gpt-6-astra-low/analysis.json:network_adjustment.scores.low.by_family.support_call_assist.time_accuracy; docs/lite/results/hosted-api-repeats-20261003/pass2/gpt-6-astra-low/analysis.json:network_adjustment.scores.low.by_family.support_call_assist.time_accuracy; docs/lite/results/hosted-api-repeats-20261003/pass3/gpt-6-astra-low/analysis.json:network_adjustment.scores.low.by_family.support_call_assist.time_accuracy
\newcommand{\AstraSupportNetRemovedHigh}{78.6}                          % equal mean across passes: docs/lite/results/four-family/gpt-6-astra-low/analysis.json:network_adjustment.scores.high.by_family.support_call_assist.time_accuracy; docs/lite/results/hosted-api-repeats-20261003/pass2/gpt-6-astra-low/analysis.json:network_adjustment.scores.high.by_family.support_call_assist.time_accuracy; docs/lite/results/hosted-api-repeats-20261003/pass3/gpt-6-astra-low/analysis.json:network_adjustment.scores.high.by_family.support_call_assist.time_accuracy
\newcommand{\AstraPresenterUntimed}{100.0}                              % equal mean across passes: docs/lite/results/four-family/gpt-6-astra-low/analysis.json:scores.by_family.presenter_voice_control.untimed_decision_accuracy; docs/lite/results/hosted-api-repeats-20261003/pass2/gpt-6-astra-low/analysis.json:scores.by_family.presenter_voice_control.untimed_decision_accuracy; docs/lite/results/hosted-api-repeats-20261003/pass3/gpt-6-astra-low/analysis.json:scores.by_family.presenter_voice_control.untimed_decision_accuracy
\newcommand{\AstraPresenterInForce}{54.3}                               % equal mean across passes: docs/lite/results/four-family/gpt-6-astra-low/analysis.json:scores.by_family.presenter_voice_control.time_accuracy; docs/lite/results/hosted-api-repeats-20261003/pass2/gpt-6-astra-low/analysis.json:scores.by_family.presenter_voice_control.time_accuracy; docs/lite/results/hosted-api-repeats-20261003/pass3/gpt-6-astra-low/analysis.json:scores.by_family.presenter_voice_control.time_accuracy
\newcommand{\AstraPresenterNetRemoved}{70.3}                            % equal mean across passes: docs/lite/results/four-family/gpt-6-astra-low/analysis.json:network_adjustment.scores.estimate.by_family.presenter_voice_control.time_accuracy; docs/lite/results/hosted-api-repeats-20261003/pass2/gpt-6-astra-low/analysis.json:network_adjustment.scores.estimate.by_family.presenter_voice_control.time_accuracy; docs/lite/results/hosted-api-repeats-20261003/pass3/gpt-6-astra-low/analysis.json:network_adjustment.scores.estimate.by_family.presenter_voice_control.time_accuracy
\newcommand{\AstraPresenterNetRemovedLow}{62.8}                         % equal mean across passes: docs/lite/results/four-family/gpt-6-astra-low/analysis.json:network_adjustment.scores.low.by_family.presenter_voice_control.time_accuracy; docs/lite/results/hosted-api-repeats-20261003/pass2/gpt-6-astra-low/analysis.json:network_adjustment.scores.low.by_family.presenter_voice_control.time_accuracy; docs/lite/results/hosted-api-repeats-20261003/pass3/gpt-6-astra-low/analysis.json:network_adjustment.scores.low.by_family.presenter_voice_control.time_accuracy
\newcommand{\AstraPresenterNetRemovedHigh}{83.3}                        % equal mean across passes: docs/lite/results/four-family/gpt-6-astra-low/analysis.json:network_adjustment.scores.high.by_family.presenter_voice_control.time_accuracy; docs/lite/results/hosted-api-repeats-20261003/pass2/gpt-6-astra-low/analysis.json:network_adjustment.scores.high.by_family.presenter_voice_control.time_accuracy; docs/lite/results/hosted-api-repeats-20261003/pass3/gpt-6-astra-low/analysis.json:network_adjustment.scores.high.by_family.presenter_voice_control.time_accuracy
\newcommand{\AstraDebugAUntimed}{100.0}                                 % equal mean across passes: docs/lite/results/four-family/gpt-6-astra-low/analysis.json:scores.per_episode[0].untimed_decision_accuracy (lite_debugging_a); docs/lite/results/hosted-api-repeats-20261003/pass2/gpt-6-astra-low/analysis.json:scores.per_episode[0].untimed_decision_accuracy (lite_debugging_a); docs/lite/results/hosted-api-repeats-20261003/pass3/gpt-6-astra-low/analysis.json:scores.per_episode[0].untimed_decision_accuracy (lite_debugging_a)
\newcommand{\AstraDebugAInForce}{54.2}                                  % equal mean across passes: docs/lite/results/four-family/gpt-6-astra-low/analysis.json:scores.per_episode[0].time_accuracy (lite_debugging_a); docs/lite/results/hosted-api-repeats-20261003/pass2/gpt-6-astra-low/analysis.json:scores.per_episode[0].time_accuracy (lite_debugging_a); docs/lite/results/hosted-api-repeats-20261003/pass3/gpt-6-astra-low/analysis.json:scores.per_episode[0].time_accuracy (lite_debugging_a)
\newcommand{\AstraDebugAInForceSeg}{49.8}                               % equal mean across passes: docs/lite/results/four-family/gpt-6-astra-low/analysis.json:scores.per_episode[0].segment_time_accuracy (lite_debugging_a); docs/lite/results/hosted-api-repeats-20261003/pass2/gpt-6-astra-low/analysis.json:scores.per_episode[0].segment_time_accuracy (lite_debugging_a); docs/lite/results/hosted-api-repeats-20261003/pass3/gpt-6-astra-low/analysis.json:scores.per_episode[0].segment_time_accuracy (lite_debugging_a)
\newcommand{\AstraDebugAErrNoDecisionSec}{3.8}                          % equal mean across passes: docs/lite/results/four-family/gpt-6-astra-low/analysis.json:scores.per_episode[0].error_seconds.no_decision (lite_debugging_a); docs/lite/results/hosted-api-repeats-20261003/pass2/gpt-6-astra-low/analysis.json:scores.per_episode[0].error_seconds.no_decision (lite_debugging_a); docs/lite/results/hosted-api-repeats-20261003/pass3/gpt-6-astra-low/analysis.json:scores.per_episode[0].error_seconds.no_decision (lite_debugging_a)
\newcommand{\AstraDebugAErrStaleSec}{51.1}                              % equal mean across passes: docs/lite/results/four-family/gpt-6-astra-low/analysis.json:scores.per_episode[0].error_seconds.source_correct (lite_debugging_a); docs/lite/results/hosted-api-repeats-20261003/pass2/gpt-6-astra-low/analysis.json:scores.per_episode[0].error_seconds.source_correct (lite_debugging_a); docs/lite/results/hosted-api-repeats-20261003/pass3/gpt-6-astra-low/analysis.json:scores.per_episode[0].error_seconds.source_correct (lite_debugging_a)
\newcommand{\AstraDebugALatencyMedian}{2.36}                            % equal mean across passes: docs/lite/results/four-family/gpt-6-astra-low/analysis.json:scores.per_episode[0].latency_s_p50 (lite_debugging_a); docs/lite/results/hosted-api-repeats-20261003/pass2/gpt-6-astra-low/analysis.json:scores.per_episode[0].latency_s_p50 (lite_debugging_a); docs/lite/results/hosted-api-repeats-20261003/pass3/gpt-6-astra-low/analysis.json:scores.per_episode[0].latency_s_p50 (lite_debugging_a)
\newcommand{\AstraDebugBUntimed}{100.0}                                 % equal mean across passes: docs/lite/results/four-family/gpt-6-astra-low/analysis.json:scores.per_episode[1].untimed_decision_accuracy (lite_debugging_b); docs/lite/results/hosted-api-repeats-20261003/pass2/gpt-6-astra-low/analysis.json:scores.per_episode[1].untimed_decision_accuracy (lite_debugging_b); docs/lite/results/hosted-api-repeats-20261003/pass3/gpt-6-astra-low/analysis.json:scores.per_episode[1].untimed_decision_accuracy (lite_debugging_b)
\newcommand{\AstraDebugBInForce}{54.9}                                  % equal mean across passes: docs/lite/results/four-family/gpt-6-astra-low/analysis.json:scores.per_episode[1].time_accuracy (lite_debugging_b); docs/lite/results/hosted-api-repeats-20261003/pass2/gpt-6-astra-low/analysis.json:scores.per_episode[1].time_accuracy (lite_debugging_b); docs/lite/results/hosted-api-repeats-20261003/pass3/gpt-6-astra-low/analysis.json:scores.per_episode[1].time_accuracy (lite_debugging_b)
\newcommand{\AstraDebugBInForceSeg}{46.6}                               % equal mean across passes: docs/lite/results/four-family/gpt-6-astra-low/analysis.json:scores.per_episode[1].segment_time_accuracy (lite_debugging_b); docs/lite/results/hosted-api-repeats-20261003/pass2/gpt-6-astra-low/analysis.json:scores.per_episode[1].segment_time_accuracy (lite_debugging_b); docs/lite/results/hosted-api-repeats-20261003/pass3/gpt-6-astra-low/analysis.json:scores.per_episode[1].segment_time_accuracy (lite_debugging_b)
\newcommand{\AstraDebugBErrNoDecisionSec}{2.7}                          % equal mean across passes: docs/lite/results/four-family/gpt-6-astra-low/analysis.json:scores.per_episode[1].error_seconds.no_decision (lite_debugging_b); docs/lite/results/hosted-api-repeats-20261003/pass2/gpt-6-astra-low/analysis.json:scores.per_episode[1].error_seconds.no_decision (lite_debugging_b); docs/lite/results/hosted-api-repeats-20261003/pass3/gpt-6-astra-low/analysis.json:scores.per_episode[1].error_seconds.no_decision (lite_debugging_b)
\newcommand{\AstraDebugBErrStaleSec}{51.5}                              % equal mean across passes: docs/lite/results/four-family/gpt-6-astra-low/analysis.json:scores.per_episode[1].error_seconds.source_correct (lite_debugging_b); docs/lite/results/hosted-api-repeats-20261003/pass2/gpt-6-astra-low/analysis.json:scores.per_episode[1].error_seconds.source_correct (lite_debugging_b); docs/lite/results/hosted-api-repeats-20261003/pass3/gpt-6-astra-low/analysis.json:scores.per_episode[1].error_seconds.source_correct (lite_debugging_b)
\newcommand{\AstraDebugBLatencyMedian}{2.38}                            % equal mean across passes: docs/lite/results/four-family/gpt-6-astra-low/analysis.json:scores.per_episode[1].latency_s_p50 (lite_debugging_b); docs/lite/results/hosted-api-repeats-20261003/pass2/gpt-6-astra-low/analysis.json:scores.per_episode[1].latency_s_p50 (lite_debugging_b); docs/lite/results/hosted-api-repeats-20261003/pass3/gpt-6-astra-low/analysis.json:scores.per_episode[1].latency_s_p50 (lite_debugging_b)
\newcommand{\AstraAssemblyAUntimed}{98.9}                               % equal mean across passes: docs/lite/results/four-family/gpt-6-astra-low/analysis.json:scores.per_episode[2].untimed_decision_accuracy (lite_assembly_a); docs/lite/results/hosted-api-repeats-20261003/pass2/gpt-6-astra-low/analysis.json:scores.per_episode[2].untimed_decision_accuracy (lite_assembly_a); docs/lite/results/hosted-api-repeats-20261003/pass3/gpt-6-astra-low/analysis.json:scores.per_episode[2].untimed_decision_accuracy (lite_assembly_a)
\newcommand{\AstraAssemblyAInForce}{52.3}                               % equal mean across passes: docs/lite/results/four-family/gpt-6-astra-low/analysis.json:scores.per_episode[2].time_accuracy (lite_assembly_a); docs/lite/results/hosted-api-repeats-20261003/pass2/gpt-6-astra-low/analysis.json:scores.per_episode[2].time_accuracy (lite_assembly_a); docs/lite/results/hosted-api-repeats-20261003/pass3/gpt-6-astra-low/analysis.json:scores.per_episode[2].time_accuracy (lite_assembly_a)
\newcommand{\AstraAssemblyAInForceSeg}{43.8}                            % equal mean across passes: docs/lite/results/four-family/gpt-6-astra-low/analysis.json:scores.per_episode[2].segment_time_accuracy (lite_assembly_a); docs/lite/results/hosted-api-repeats-20261003/pass2/gpt-6-astra-low/analysis.json:scores.per_episode[2].segment_time_accuracy (lite_assembly_a); docs/lite/results/hosted-api-repeats-20261003/pass3/gpt-6-astra-low/analysis.json:scores.per_episode[2].segment_time_accuracy (lite_assembly_a)
\newcommand{\AstraAssemblyAErrNoDecisionSec}{2.2}                       % equal mean across passes: docs/lite/results/four-family/gpt-6-astra-low/analysis.json:scores.per_episode[2].error_seconds.no_decision (lite_assembly_a); docs/lite/results/hosted-api-repeats-20261003/pass2/gpt-6-astra-low/analysis.json:scores.per_episode[2].error_seconds.no_decision (lite_assembly_a); docs/lite/results/hosted-api-repeats-20261003/pass3/gpt-6-astra-low/analysis.json:scores.per_episode[2].error_seconds.no_decision (lite_assembly_a)
\newcommand{\AstraAssemblyAErrStaleSec}{54.3}                           % equal mean across passes: docs/lite/results/four-family/gpt-6-astra-low/analysis.json:scores.per_episode[2].error_seconds.source_correct (lite_assembly_a); docs/lite/results/hosted-api-repeats-20261003/pass2/gpt-6-astra-low/analysis.json:scores.per_episode[2].error_seconds.source_correct (lite_assembly_a); docs/lite/results/hosted-api-repeats-20261003/pass3/gpt-6-astra-low/analysis.json:scores.per_episode[2].error_seconds.source_correct (lite_assembly_a)
\newcommand{\AstraAssemblyALatencyMedian}{2.33}                         % equal mean across passes: docs/lite/results/four-family/gpt-6-astra-low/analysis.json:scores.per_episode[2].latency_s_p50 (lite_assembly_a); docs/lite/results/hosted-api-repeats-20261003/pass2/gpt-6-astra-low/analysis.json:scores.per_episode[2].latency_s_p50 (lite_assembly_a); docs/lite/results/hosted-api-repeats-20261003/pass3/gpt-6-astra-low/analysis.json:scores.per_episode[2].latency_s_p50 (lite_assembly_a)
\newcommand{\AstraAssemblyBUntimed}{100.0}                              % equal mean across passes: docs/lite/results/four-family/gpt-6-astra-low/analysis.json:scores.per_episode[3].untimed_decision_accuracy (lite_assembly_b); docs/lite/results/hosted-api-repeats-20261003/pass2/gpt-6-astra-low/analysis.json:scores.per_episode[3].untimed_decision_accuracy (lite_assembly_b); docs/lite/results/hosted-api-repeats-20261003/pass3/gpt-6-astra-low/analysis.json:scores.per_episode[3].untimed_decision_accuracy (lite_assembly_b)
\newcommand{\AstraAssemblyBInForce}{47.7}                               % equal mean across passes: docs/lite/results/four-family/gpt-6-astra-low/analysis.json:scores.per_episode[3].time_accuracy (lite_assembly_b); docs/lite/results/hosted-api-repeats-20261003/pass2/gpt-6-astra-low/analysis.json:scores.per_episode[3].time_accuracy (lite_assembly_b); docs/lite/results/hosted-api-repeats-20261003/pass3/gpt-6-astra-low/analysis.json:scores.per_episode[3].time_accuracy (lite_assembly_b)
\newcommand{\AstraAssemblyBInForceSeg}{36.9}                            % equal mean across passes: docs/lite/results/four-family/gpt-6-astra-low/analysis.json:scores.per_episode[3].segment_time_accuracy (lite_assembly_b); docs/lite/results/hosted-api-repeats-20261003/pass2/gpt-6-astra-low/analysis.json:scores.per_episode[3].segment_time_accuracy (lite_assembly_b); docs/lite/results/hosted-api-repeats-20261003/pass3/gpt-6-astra-low/analysis.json:scores.per_episode[3].segment_time_accuracy (lite_assembly_b)
\newcommand{\AstraAssemblyBErrNoDecisionSec}{2.8}                       % equal mean across passes: docs/lite/results/four-family/gpt-6-astra-low/analysis.json:scores.per_episode[3].error_seconds.no_decision (lite_assembly_b); docs/lite/results/hosted-api-repeats-20261003/pass2/gpt-6-astra-low/analysis.json:scores.per_episode[3].error_seconds.no_decision (lite_assembly_b); docs/lite/results/hosted-api-repeats-20261003/pass3/gpt-6-astra-low/analysis.json:scores.per_episode[3].error_seconds.no_decision (lite_assembly_b)
\newcommand{\AstraAssemblyBErrStaleSec}{59.9}                           % equal mean across passes: docs/lite/results/four-family/gpt-6-astra-low/analysis.json:scores.per_episode[3].error_seconds.source_correct (lite_assembly_b); docs/lite/results/hosted-api-repeats-20261003/pass2/gpt-6-astra-low/analysis.json:scores.per_episode[3].error_seconds.source_correct (lite_assembly_b); docs/lite/results/hosted-api-repeats-20261003/pass3/gpt-6-astra-low/analysis.json:scores.per_episode[3].error_seconds.source_correct (lite_assembly_b)
\newcommand{\AstraAssemblyBLatencyMedian}{2.41}                         % equal mean across passes: docs/lite/results/four-family/gpt-6-astra-low/analysis.json:scores.per_episode[3].latency_s_p50 (lite_assembly_b); docs/lite/results/hosted-api-repeats-20261003/pass2/gpt-6-astra-low/analysis.json:scores.per_episode[3].latency_s_p50 (lite_assembly_b); docs/lite/results/hosted-api-repeats-20261003/pass3/gpt-6-astra-low/analysis.json:scores.per_episode[3].latency_s_p50 (lite_assembly_b)
\newcommand{\AstraSupportAUntimed}{100.0}                               % equal mean across passes: docs/lite/results/four-family/gpt-6-astra-low/analysis.json:scores.per_episode[4].untimed_decision_accuracy (lite_support_a); docs/lite/results/hosted-api-repeats-20261003/pass2/gpt-6-astra-low/analysis.json:scores.per_episode[4].untimed_decision_accuracy (lite_support_a); docs/lite/results/hosted-api-repeats-20261003/pass3/gpt-6-astra-low/analysis.json:scores.per_episode[4].untimed_decision_accuracy (lite_support_a)
\newcommand{\AstraSupportAInForce}{51.2}                                % equal mean across passes: docs/lite/results/four-family/gpt-6-astra-low/analysis.json:scores.per_episode[4].time_accuracy (lite_support_a); docs/lite/results/hosted-api-repeats-20261003/pass2/gpt-6-astra-low/analysis.json:scores.per_episode[4].time_accuracy (lite_support_a); docs/lite/results/hosted-api-repeats-20261003/pass3/gpt-6-astra-low/analysis.json:scores.per_episode[4].time_accuracy (lite_support_a)
\newcommand{\AstraSupportAInForceSeg}{39.9}                             % equal mean across passes: docs/lite/results/four-family/gpt-6-astra-low/analysis.json:scores.per_episode[4].segment_time_accuracy (lite_support_a); docs/lite/results/hosted-api-repeats-20261003/pass2/gpt-6-astra-low/analysis.json:scores.per_episode[4].segment_time_accuracy (lite_support_a); docs/lite/results/hosted-api-repeats-20261003/pass3/gpt-6-astra-low/analysis.json:scores.per_episode[4].segment_time_accuracy (lite_support_a)
\newcommand{\AstraSupportAErrNoDecisionSec}{2.7}                        % equal mean across passes: docs/lite/results/four-family/gpt-6-astra-low/analysis.json:scores.per_episode[4].error_seconds.no_decision (lite_support_a); docs/lite/results/hosted-api-repeats-20261003/pass2/gpt-6-astra-low/analysis.json:scores.per_episode[4].error_seconds.no_decision (lite_support_a); docs/lite/results/hosted-api-repeats-20261003/pass3/gpt-6-astra-low/analysis.json:scores.per_episode[4].error_seconds.no_decision (lite_support_a)
\newcommand{\AstraSupportAErrStaleSec}{55.9}                            % equal mean across passes: docs/lite/results/four-family/gpt-6-astra-low/analysis.json:scores.per_episode[4].error_seconds.source_correct (lite_support_a); docs/lite/results/hosted-api-repeats-20261003/pass2/gpt-6-astra-low/analysis.json:scores.per_episode[4].error_seconds.source_correct (lite_support_a); docs/lite/results/hosted-api-repeats-20261003/pass3/gpt-6-astra-low/analysis.json:scores.per_episode[4].error_seconds.source_correct (lite_support_a)
\newcommand{\AstraSupportALatencyMedian}{2.55}                          % equal mean across passes: docs/lite/results/four-family/gpt-6-astra-low/analysis.json:scores.per_episode[4].latency_s_p50 (lite_support_a); docs/lite/results/hosted-api-repeats-20261003/pass2/gpt-6-astra-low/analysis.json:scores.per_episode[4].latency_s_p50 (lite_support_a); docs/lite/results/hosted-api-repeats-20261003/pass3/gpt-6-astra-low/analysis.json:scores.per_episode[4].latency_s_p50 (lite_support_a)
\newcommand{\AstraSupportBUntimed}{100.0}                               % equal mean across passes: docs/lite/results/four-family/gpt-6-astra-low/analysis.json:scores.per_episode[5].untimed_decision_accuracy (lite_support_b); docs/lite/results/hosted-api-repeats-20261003/pass2/gpt-6-astra-low/analysis.json:scores.per_episode[5].untimed_decision_accuracy (lite_support_b); docs/lite/results/hosted-api-repeats-20261003/pass3/gpt-6-astra-low/analysis.json:scores.per_episode[5].untimed_decision_accuracy (lite_support_b)
\newcommand{\AstraSupportBInForce}{48.0}                                % equal mean across passes: docs/lite/results/four-family/gpt-6-astra-low/analysis.json:scores.per_episode[5].time_accuracy (lite_support_b); docs/lite/results/hosted-api-repeats-20261003/pass2/gpt-6-astra-low/analysis.json:scores.per_episode[5].time_accuracy (lite_support_b); docs/lite/results/hosted-api-repeats-20261003/pass3/gpt-6-astra-low/analysis.json:scores.per_episode[5].time_accuracy (lite_support_b)
\newcommand{\AstraSupportBInForceSeg}{41.8}                             % equal mean across passes: docs/lite/results/four-family/gpt-6-astra-low/analysis.json:scores.per_episode[5].segment_time_accuracy (lite_support_b); docs/lite/results/hosted-api-repeats-20261003/pass2/gpt-6-astra-low/analysis.json:scores.per_episode[5].segment_time_accuracy (lite_support_b); docs/lite/results/hosted-api-repeats-20261003/pass3/gpt-6-astra-low/analysis.json:scores.per_episode[5].segment_time_accuracy (lite_support_b)
\newcommand{\AstraSupportBErrNoDecisionSec}{2.5}                        % equal mean across passes: docs/lite/results/four-family/gpt-6-astra-low/analysis.json:scores.per_episode[5].error_seconds.no_decision (lite_support_b); docs/lite/results/hosted-api-repeats-20261003/pass2/gpt-6-astra-low/analysis.json:scores.per_episode[5].error_seconds.no_decision (lite_support_b); docs/lite/results/hosted-api-repeats-20261003/pass3/gpt-6-astra-low/analysis.json:scores.per_episode[5].error_seconds.no_decision (lite_support_b)
\newcommand{\AstraSupportBErrStaleSec}{59.9}                            % equal mean across passes: docs/lite/results/four-family/gpt-6-astra-low/analysis.json:scores.per_episode[5].error_seconds.source_correct (lite_support_b); docs/lite/results/hosted-api-repeats-20261003/pass2/gpt-6-astra-low/analysis.json:scores.per_episode[5].error_seconds.source_correct (lite_support_b); docs/lite/results/hosted-api-repeats-20261003/pass3/gpt-6-astra-low/analysis.json:scores.per_episode[5].error_seconds.source_correct (lite_support_b)
\newcommand{\AstraSupportBLatencyMedian}{2.38}                          % equal mean across passes: docs/lite/results/four-family/gpt-6-astra-low/analysis.json:scores.per_episode[5].latency_s_p50 (lite_support_b); docs/lite/results/hosted-api-repeats-20261003/pass2/gpt-6-astra-low/analysis.json:scores.per_episode[5].latency_s_p50 (lite_support_b); docs/lite/results/hosted-api-repeats-20261003/pass3/gpt-6-astra-low/analysis.json:scores.per_episode[5].latency_s_p50 (lite_support_b)
\newcommand{\AstraPresenterAUntimed}{100.0}                             % equal mean across passes: docs/lite/results/four-family/gpt-6-astra-low/analysis.json:scores.per_episode[6].untimed_decision_accuracy (lite_presenter_a); docs/lite/results/hosted-api-repeats-20261003/pass2/gpt-6-astra-low/analysis.json:scores.per_episode[6].untimed_decision_accuracy (lite_presenter_a); docs/lite/results/hosted-api-repeats-20261003/pass3/gpt-6-astra-low/analysis.json:scores.per_episode[6].untimed_decision_accuracy (lite_presenter_a)
\newcommand{\AstraPresenterAInForce}{51.8}                              % equal mean across passes: docs/lite/results/four-family/gpt-6-astra-low/analysis.json:scores.per_episode[6].time_accuracy (lite_presenter_a); docs/lite/results/hosted-api-repeats-20261003/pass2/gpt-6-astra-low/analysis.json:scores.per_episode[6].time_accuracy (lite_presenter_a); docs/lite/results/hosted-api-repeats-20261003/pass3/gpt-6-astra-low/analysis.json:scores.per_episode[6].time_accuracy (lite_presenter_a)
\newcommand{\AstraPresenterAInForceSeg}{38.6}                           % equal mean across passes: docs/lite/results/four-family/gpt-6-astra-low/analysis.json:scores.per_episode[6].segment_time_accuracy (lite_presenter_a); docs/lite/results/hosted-api-repeats-20261003/pass2/gpt-6-astra-low/analysis.json:scores.per_episode[6].segment_time_accuracy (lite_presenter_a); docs/lite/results/hosted-api-repeats-20261003/pass3/gpt-6-astra-low/analysis.json:scores.per_episode[6].segment_time_accuracy (lite_presenter_a)
\newcommand{\AstraPresenterAErrNoDecisionSec}{2.3}                      % equal mean across passes: docs/lite/results/four-family/gpt-6-astra-low/analysis.json:scores.per_episode[6].error_seconds.no_decision (lite_presenter_a); docs/lite/results/hosted-api-repeats-20261003/pass2/gpt-6-astra-low/analysis.json:scores.per_episode[6].error_seconds.no_decision (lite_presenter_a); docs/lite/results/hosted-api-repeats-20261003/pass3/gpt-6-astra-low/analysis.json:scores.per_episode[6].error_seconds.no_decision (lite_presenter_a)
\newcommand{\AstraPresenterAErrStaleSec}{55.5}                          % equal mean across passes: docs/lite/results/four-family/gpt-6-astra-low/analysis.json:scores.per_episode[6].error_seconds.source_correct (lite_presenter_a); docs/lite/results/hosted-api-repeats-20261003/pass2/gpt-6-astra-low/analysis.json:scores.per_episode[6].error_seconds.source_correct (lite_presenter_a); docs/lite/results/hosted-api-repeats-20261003/pass3/gpt-6-astra-low/analysis.json:scores.per_episode[6].error_seconds.source_correct (lite_presenter_a)
\newcommand{\AstraPresenterALatencyMedian}{2.26}                        % equal mean across passes: docs/lite/results/four-family/gpt-6-astra-low/analysis.json:scores.per_episode[6].latency_s_p50 (lite_presenter_a); docs/lite/results/hosted-api-repeats-20261003/pass2/gpt-6-astra-low/analysis.json:scores.per_episode[6].latency_s_p50 (lite_presenter_a); docs/lite/results/hosted-api-repeats-20261003/pass3/gpt-6-astra-low/analysis.json:scores.per_episode[6].latency_s_p50 (lite_presenter_a)
\newcommand{\AstraPresenterBUntimed}{100.0}                             % equal mean across passes: docs/lite/results/four-family/gpt-6-astra-low/analysis.json:scores.per_episode[7].untimed_decision_accuracy (lite_presenter_b); docs/lite/results/hosted-api-repeats-20261003/pass2/gpt-6-astra-low/analysis.json:scores.per_episode[7].untimed_decision_accuracy (lite_presenter_b); docs/lite/results/hosted-api-repeats-20261003/pass3/gpt-6-astra-low/analysis.json:scores.per_episode[7].untimed_decision_accuracy (lite_presenter_b)
\newcommand{\AstraPresenterBInForce}{56.8}                              % equal mean across passes: docs/lite/results/four-family/gpt-6-astra-low/analysis.json:scores.per_episode[7].time_accuracy (lite_presenter_b); docs/lite/results/hosted-api-repeats-20261003/pass2/gpt-6-astra-low/analysis.json:scores.per_episode[7].time_accuracy (lite_presenter_b); docs/lite/results/hosted-api-repeats-20261003/pass3/gpt-6-astra-low/analysis.json:scores.per_episode[7].time_accuracy (lite_presenter_b)
\newcommand{\AstraPresenterBInForceSeg}{46.8}                           % equal mean across passes: docs/lite/results/four-family/gpt-6-astra-low/analysis.json:scores.per_episode[7].segment_time_accuracy (lite_presenter_b); docs/lite/results/hosted-api-repeats-20261003/pass2/gpt-6-astra-low/analysis.json:scores.per_episode[7].segment_time_accuracy (lite_presenter_b); docs/lite/results/hosted-api-repeats-20261003/pass3/gpt-6-astra-low/analysis.json:scores.per_episode[7].segment_time_accuracy (lite_presenter_b)
\newcommand{\AstraPresenterBErrNoDecisionSec}{2.5}                      % equal mean across passes: docs/lite/results/four-family/gpt-6-astra-low/analysis.json:scores.per_episode[7].error_seconds.no_decision (lite_presenter_b); docs/lite/results/hosted-api-repeats-20261003/pass2/gpt-6-astra-low/analysis.json:scores.per_episode[7].error_seconds.no_decision (lite_presenter_b); docs/lite/results/hosted-api-repeats-20261003/pass3/gpt-6-astra-low/analysis.json:scores.per_episode[7].error_seconds.no_decision (lite_presenter_b)
\newcommand{\AstraPresenterBErrStaleSec}{49.4}                          % equal mean across passes: docs/lite/results/four-family/gpt-6-astra-low/analysis.json:scores.per_episode[7].error_seconds.source_correct (lite_presenter_b); docs/lite/results/hosted-api-repeats-20261003/pass2/gpt-6-astra-low/analysis.json:scores.per_episode[7].error_seconds.source_correct (lite_presenter_b); docs/lite/results/hosted-api-repeats-20261003/pass3/gpt-6-astra-low/analysis.json:scores.per_episode[7].error_seconds.source_correct (lite_presenter_b)
\newcommand{\AstraPresenterBLatencyMedian}{2.23}                        % equal mean across passes: docs/lite/results/four-family/gpt-6-astra-low/analysis.json:scores.per_episode[7].latency_s_p50 (lite_presenter_b); docs/lite/results/hosted-api-repeats-20261003/pass2/gpt-6-astra-low/analysis.json:scores.per_episode[7].latency_s_p50 (lite_presenter_b); docs/lite/results/hosted-api-repeats-20261003/pass3/gpt-6-astra-low/analysis.json:scores.per_episode[7].latency_s_p50 (lite_presenter_b)
\newcommand{\JevUntimed}{62.4}                                          % equal mean across passes: docs/lite/results/four-family/jev-latest/analysis.json:scores.overall.untimed_decision_accuracy; docs/lite/results/hosted-api-repeats-20261003/pass2/jev-latest/analysis.json:scores.overall.untimed_decision_accuracy; docs/lite/results/hosted-api-repeats-20261003/pass3/jev-latest/analysis.json:scores.overall.untimed_decision_accuracy
\newcommand{\JevExactMatch}{33.5}                                       % equal mean across passes: docs/lite/results/four-family/jev-latest/analysis.json:scores.overall.all_questions_exact_accuracy; docs/lite/results/hosted-api-repeats-20261003/pass2/jev-latest/analysis.json:scores.overall.all_questions_exact_accuracy; docs/lite/results/hosted-api-repeats-20261003/pass3/jev-latest/analysis.json:scores.overall.all_questions_exact_accuracy
\newcommand{\JevInForce}{59.5}                                          % equal mean across passes: docs/lite/results/four-family/jev-latest/analysis.json:scores.overall.time_accuracy; docs/lite/results/hosted-api-repeats-20261003/pass2/jev-latest/analysis.json:scores.overall.time_accuracy; docs/lite/results/hosted-api-repeats-20261003/pass3/jev-latest/analysis.json:scores.overall.time_accuracy
\newcommand{\JevInForceSeg}{56.5}                                       % equal mean across passes: docs/lite/results/four-family/jev-latest/analysis.json:scores.overall.segment_time_accuracy; docs/lite/results/hosted-api-repeats-20261003/pass2/jev-latest/analysis.json:scores.overall.segment_time_accuracy; docs/lite/results/hosted-api-repeats-20261003/pass3/jev-latest/analysis.json:scores.overall.segment_time_accuracy
\newcommand{\JevNetRemovedSeg}{58.8}                                    % equal mean across passes: docs/lite/results/four-family/jev-latest/analysis.json:network_adjustment.scores.estimate.overall.segment_time_accuracy; docs/lite/results/hosted-api-repeats-20261003/pass2/jev-latest/analysis.json:network_adjustment.scores.estimate.overall.segment_time_accuracy; docs/lite/results/hosted-api-repeats-20261003/pass3/jev-latest/analysis.json:network_adjustment.scores.estimate.overall.segment_time_accuracy
\newcommand{\JevNetSec}{0.18}                                           % equal mean across passes: docs/lite/results/four-family/jev-latest/analysis.json:network_adjustment.network_s.estimate; docs/lite/results/hosted-api-repeats-20261003/pass2/jev-latest/analysis.json:network_adjustment.network_s.estimate; docs/lite/results/hosted-api-repeats-20261003/pass3/jev-latest/analysis.json:network_adjustment.network_s.estimate
\newcommand{\JevNetRemovedSegLow}{58.7}                                 % equal mean across passes: docs/lite/results/four-family/jev-latest/analysis.json:network_adjustment.scores.low.overall.segment_time_accuracy; docs/lite/results/hosted-api-repeats-20261003/pass2/jev-latest/analysis.json:network_adjustment.scores.low.overall.segment_time_accuracy; docs/lite/results/hosted-api-repeats-20261003/pass3/jev-latest/analysis.json:network_adjustment.scores.low.overall.segment_time_accuracy
\newcommand{\JevNetSecLow}{0.17}                                        % equal mean across passes: docs/lite/results/four-family/jev-latest/analysis.json:network_adjustment.network_s.low; docs/lite/results/hosted-api-repeats-20261003/pass2/jev-latest/analysis.json:network_adjustment.network_s.low; docs/lite/results/hosted-api-repeats-20261003/pass3/jev-latest/analysis.json:network_adjustment.network_s.low
\newcommand{\JevNetRemovedSegHigh}{58.9}                                % equal mean across passes: docs/lite/results/four-family/jev-latest/analysis.json:network_adjustment.scores.high.overall.segment_time_accuracy; docs/lite/results/hosted-api-repeats-20261003/pass2/jev-latest/analysis.json:network_adjustment.scores.high.overall.segment_time_accuracy; docs/lite/results/hosted-api-repeats-20261003/pass3/jev-latest/analysis.json:network_adjustment.scores.high.overall.segment_time_accuracy
\newcommand{\JevNetSecHigh}{0.18}                                       % equal mean across passes: docs/lite/results/four-family/jev-latest/analysis.json:network_adjustment.network_s.high; docs/lite/results/hosted-api-repeats-20261003/pass2/jev-latest/analysis.json:network_adjustment.network_s.high; docs/lite/results/hosted-api-repeats-20261003/pass3/jev-latest/analysis.json:network_adjustment.network_s.high
\newcommand{\JevPrefillMsPerKTok}{7.7}                                  % equal mean across passes: docs/lite/results/four-family/jev-latest/analysis.json:network_adjustment.prefill_s_per_1k_input_tokens (x 1000); docs/lite/results/hosted-api-repeats-20261003/pass2/jev-latest/analysis.json:network_adjustment.prefill_s_per_1k_input_tokens (x 1000); docs/lite/results/hosted-api-repeats-20261003/pass3/jev-latest/analysis.json:network_adjustment.prefill_s_per_1k_input_tokens (x 1000)
\newcommand{\JevLatencyFloor}{0.18}                                     % minimum across passes: docs/lite/results/four-family/jev-latest/analysis.json:network_adjustment.latency_floor_s; docs/lite/results/hosted-api-repeats-20261003/pass2/jev-latest/analysis.json:network_adjustment.latency_floor_s; docs/lite/results/hosted-api-repeats-20261003/pass3/jev-latest/analysis.json:network_adjustment.latency_floor_s
\newcommand{\JevClampedRequests}{4}                                     % sum across passes: docs/lite/results/four-family/jev-latest/analysis.json:network_adjustment.scores.{estimate,low,high}.clamped_requests (maximum); docs/lite/results/hosted-api-repeats-20261003/pass2/jev-latest/analysis.json:network_adjustment.scores.{estimate,low,high}.clamped_requests (maximum); docs/lite/results/hosted-api-repeats-20261003/pass3/jev-latest/analysis.json:network_adjustment.scores.{estimate,low,high}.clamped_requests (maximum)
\newcommand{\JevLatencyMedian}{0.23}                                    % equal mean across passes: docs/lite/results/four-family/jev-latest/analysis.json:latency_s.p50; docs/lite/results/hosted-api-repeats-20261003/pass2/jev-latest/analysis.json:latency_s.p50; docs/lite/results/hosted-api-repeats-20261003/pass3/jev-latest/analysis.json:latency_s.p50
\newcommand{\JevLatencyPNinetyFive}{0.37}                               % equal mean across passes: docs/lite/results/four-family/jev-latest/analysis.json:latency_s.p95; docs/lite/results/hosted-api-repeats-20261003/pass2/jev-latest/analysis.json:latency_s.p95; docs/lite/results/hosted-api-repeats-20261003/pass3/jev-latest/analysis.json:latency_s.p95
\newcommand{\JevLatencyMax}{0.92}                                       % maximum across passes: docs/lite/results/four-family/jev-latest/analysis.json:latency_s.max; docs/lite/results/hosted-api-repeats-20261003/pass2/jev-latest/analysis.json:latency_s.max; docs/lite/results/hosted-api-repeats-20261003/pass3/jev-latest/analysis.json:latency_s.max
\newcommand{\JevLatencyMean}{0.25}                                      % equal mean across passes: runs/lite-v1-jev-latest-four-family-retry-v1/events.jsonl:response.attempts[-1].completed_s - started_s (mean over 480 responses); runs/hosted-api-repeats-20261003/pass2/jev-latest/events.jsonl:response.attempts[-1].completed_s - started_s (mean over 480 responses); runs/hosted-api-repeats-20261003/pass3/jev-latest/events.jsonl:response.attempts[-1].completed_s - started_s (mean over 480 responses)
\newcommand{\JevInputTokMedian}{3613}                                   % equal mean across passes: runs/lite-v1-jev-latest-four-family-retry-v1/events.jsonl:response.usage.input_tokens (median over 480 responses, half-up); runs/hosted-api-repeats-20261003/pass2/jev-latest/events.jsonl:response.usage.input_tokens (median over 480 responses, half-up); runs/hosted-api-repeats-20261003/pass3/jev-latest/events.jsonl:response.usage.input_tokens (median over 480 responses, half-up)
\newcommand{\JevInputTokTotal}{5{,}262{,}339}                           % sum across passes: docs/lite/results/four-family/jev-latest/analysis.json:usage.input_tokens; docs/lite/results/hosted-api-repeats-20261003/pass2/jev-latest/analysis.json:usage.input_tokens; docs/lite/results/hosted-api-repeats-20261003/pass3/jev-latest/analysis.json:usage.input_tokens
\newcommand{\JevOutputTokMedian}{396}                                   % equal mean across passes: runs/lite-v1-jev-latest-four-family-retry-v1/events.jsonl:response.usage.output_tokens (median over 480 responses, half-up); runs/hosted-api-repeats-20261003/pass2/jev-latest/events.jsonl:response.usage.output_tokens (median over 480 responses, half-up); runs/hosted-api-repeats-20261003/pass3/jev-latest/events.jsonl:response.usage.output_tokens (median over 480 responses, half-up)
\newcommand{\JevOutputTokTotal}{583{,}560}                              % sum across passes: docs/lite/results/four-family/jev-latest/analysis.json:usage.output_tokens; docs/lite/results/hosted-api-repeats-20261003/pass2/jev-latest/analysis.json:usage.output_tokens; docs/lite/results/hosted-api-repeats-20261003/pass3/jev-latest/analysis.json:usage.output_tokens
\newcommand{\JevReasoningTokMedian}{0}                                  % equal mean across passes: runs/lite-v1-jev-latest-four-family-retry-v1/events.jsonl:response.usage.reasoning_tokens (median over 480 responses, half-up); runs/hosted-api-repeats-20261003/pass2/jev-latest/events.jsonl:response.usage.reasoning_tokens (median over 480 responses, half-up); runs/hosted-api-repeats-20261003/pass3/jev-latest/events.jsonl:response.usage.reasoning_tokens (median over 480 responses, half-up)
\newcommand{\JevCostInputUSD}{0.221}                                    % sum across passes: docs/lite/results/four-family/jev-latest/analysis.json:usage at docs.typesafe.ai/models (Jev; output tokens free) list prices retrieved 2026-09-29 (input: uncached x input price + cached x cached price); docs/lite/results/hosted-api-repeats-20261003/pass2/jev-latest/analysis.json:usage at docs.typesafe.ai/models (Jev; output tokens free) list prices retrieved 2026-09-29 (input: uncached x input price + cached x cached price); docs/lite/results/hosted-api-repeats-20261003/pass3/jev-latest/analysis.json:usage at docs.typesafe.ai/models (Jev; output tokens free) list prices retrieved 2026-09-29 (input: uncached x input price + cached x cached price)
\newcommand{\JevCostOutputUSD}{0.000}                                   % sum across passes: docs/lite/results/four-family/jev-latest/analysis.json:usage at docs.typesafe.ai/models (Jev; output tokens free) list prices retrieved 2026-09-29 (output tokens, reasoning included, x output price); docs/lite/results/hosted-api-repeats-20261003/pass2/jev-latest/analysis.json:usage at docs.typesafe.ai/models (Jev; output tokens free) list prices retrieved 2026-09-29 (output tokens, reasoning included, x output price); docs/lite/results/hosted-api-repeats-20261003/pass3/jev-latest/analysis.json:usage at docs.typesafe.ai/models (Jev; output tokens free) list prices retrieved 2026-09-29 (output tokens, reasoning included, x output price)
\newcommand{\JevCostUSD}{0.221}                                         % sum across passes: docs/lite/results/four-family/jev-latest/analysis.json:usage at docs.typesafe.ai/models (Jev; output tokens free) list prices retrieved 2026-09-29 (input + output); docs/lite/results/hosted-api-repeats-20261003/pass2/jev-latest/analysis.json:usage at docs.typesafe.ai/models (Jev; output tokens free) list prices retrieved 2026-09-29 (input + output); docs/lite/results/hosted-api-repeats-20261003/pass3/jev-latest/analysis.json:usage at docs.typesafe.ai/models (Jev; output tokens free) list prices retrieved 2026-09-29 (input + output)
\newcommand{\JevPriceInput}{0.042}                                      % equal mean across passes: docs.typesafe.ai/models (Jev; output tokens free): USD per 1M input tokens; docs.typesafe.ai/models (Jev; output tokens free): USD per 1M input tokens; docs.typesafe.ai/models (Jev; output tokens free): USD per 1M input tokens
\newcommand{\JevPriceOutput}{0.00}                                      % equal mean across passes: docs.typesafe.ai/models (Jev; output tokens free): USD per 1M output tokens; docs.typesafe.ai/models (Jev; output tokens free): USD per 1M output tokens; docs.typesafe.ai/models (Jev; output tokens free): USD per 1M output tokens
\newcommand{\JevErrNoDecisionPct}{0.4}                                  % equal mean across passes: docs/lite/results/four-family/jev-latest/analysis.json:error_seconds.no_decision (scenario error fraction, then equal family mean); docs/lite/results/hosted-api-repeats-20261003/pass2/jev-latest/analysis.json:error_seconds.no_decision (scenario error fraction, then equal family mean); docs/lite/results/hosted-api-repeats-20261003/pass3/jev-latest/analysis.json:error_seconds.no_decision (scenario error fraction, then equal family mean)
\newcommand{\JevErrStalePct}{2.7}                                       % equal mean across passes: docs/lite/results/four-family/jev-latest/analysis.json:error_seconds.source_correct (scenario error fraction, then equal family mean); docs/lite/results/hosted-api-repeats-20261003/pass2/jev-latest/analysis.json:error_seconds.source_correct (scenario error fraction, then equal family mean); docs/lite/results/hosted-api-repeats-20261003/pass3/jev-latest/analysis.json:error_seconds.source_correct (scenario error fraction, then equal family mean)
\newcommand{\JevInForceRecorded}{59.5}                                  % equal mean across passes: docs/lite/results/four-family/jev-latest/analysis.json:scores.overall.time_accuracy; docs/lite/results/hosted-api-repeats-20261003/pass2/jev-latest/analysis.json:scores.overall.time_accuracy; docs/lite/results/hosted-api-repeats-20261003/pass3/jev-latest/analysis.json:scores.overall.time_accuracy
\newcommand{\JevInForcePhysical}{59.5}                                  % equal mean across passes: docs/lite/results/four-family/jev-latest/analysis.json:raw_wallclock_scores.overall.time_accuracy (physical wall-clock trace); docs/lite/results/hosted-api-repeats-20261003/pass2/jev-latest/analysis.json:raw_wallclock_scores.overall.time_accuracy (physical wall-clock trace); docs/lite/results/hosted-api-repeats-20261003/pass3/jev-latest/analysis.json:raw_wallclock_scores.overall.time_accuracy (physical wall-clock trace)
\newcommand{\JevExcludedDispatchSec}{0.23}                              % equal mean across passes: docs/lite/results/four-family/jev-latest/analysis.json:retry_reliability.excluded_dispatch_s (summed over the pass); docs/lite/results/hosted-api-repeats-20261003/pass2/jev-latest/analysis.json:retry_reliability.excluded_dispatch_s (summed over the pass); docs/lite/results/hosted-api-repeats-20261003/pass3/jev-latest/analysis.json:retry_reliability.excluded_dispatch_s (summed over the pass)
\newcommand{\JevEventsHash}{eb86c4e1a0bf650c}                           % pass 1: docs/lite/results/four-family/jev-latest/analysis.json:events_sha256 (first 16 hex digits; equals sha256 of runs/lite-v1-jev-latest-four-family-retry-v1/events.jsonl)
\newcommand{\JevAcceptedUpdates}{1440}                                  % sum across passes: docs/lite/results/four-family/jev-latest/analysis.json:accepted_updates; docs/lite/results/hosted-api-repeats-20261003/pass2/jev-latest/analysis.json:accepted_updates; docs/lite/results/hosted-api-repeats-20261003/pass3/jev-latest/analysis.json:accepted_updates
\newcommand{\JevDiscardedAfterHorizon}{0}                               % sum across passes: docs/lite/results/four-family/jev-latest/analysis.json:discarded_updates.after_horizon (absent = 0); docs/lite/results/hosted-api-repeats-20261003/pass2/jev-latest/analysis.json:discarded_updates.after_horizon (absent = 0); docs/lite/results/hosted-api-repeats-20261003/pass3/jev-latest/analysis.json:discarded_updates.after_horizon (absent = 0)
\newcommand{\JevDiscardedOlder}{0}                                      % sum across passes: docs/lite/results/four-family/jev-latest/analysis.json:discarded_updates.older_than_active (absent = 0); docs/lite/results/hosted-api-repeats-20261003/pass2/jev-latest/analysis.json:discarded_updates.older_than_active (absent = 0); docs/lite/results/hosted-api-repeats-20261003/pass3/jev-latest/analysis.json:discarded_updates.older_than_active (absent = 0)
\newcommand{\JevValidResponses}{1440}                                   % sum across passes: docs/lite/results/four-family/jev-latest/analysis.json:retry_reliability.successful_logical_requests; docs/lite/results/hosted-api-repeats-20261003/pass2/jev-latest/analysis.json:retry_reliability.successful_logical_requests; docs/lite/results/hosted-api-repeats-20261003/pass3/jev-latest/analysis.json:retry_reliability.successful_logical_requests
\newcommand{\JevAttempts}{1440}                                         % sum across passes: docs/lite/results/four-family/jev-latest/analysis.json:retry_reliability.attempts; docs/lite/results/hosted-api-repeats-20261003/pass2/jev-latest/analysis.json:retry_reliability.attempts; docs/lite/results/hosted-api-repeats-20261003/pass3/jev-latest/analysis.json:retry_reliability.attempts
\newcommand{\JevFailedAttempts}{0}                                      % sum across passes: docs/lite/results/four-family/jev-latest/analysis.json:retry_reliability.failed_attempts; docs/lite/results/hosted-api-repeats-20261003/pass2/jev-latest/analysis.json:retry_reliability.failed_attempts; docs/lite/results/hosted-api-repeats-20261003/pass3/jev-latest/analysis.json:retry_reliability.failed_attempts
\newcommand{\JevRetriedRequests}{0}                                     % sum across passes: docs/lite/results/four-family/jev-latest/analysis.json:retry_reliability.retried_logical_requests; docs/lite/results/hosted-api-repeats-20261003/pass2/jev-latest/analysis.json:retry_reliability.retried_logical_requests; docs/lite/results/hosted-api-repeats-20261003/pass3/jev-latest/analysis.json:retry_reliability.retried_logical_requests
\newcommand{\JevInactiveOnlyErrors}{415}                                % sum across passes: docs/lite/results/four-family/jev-latest/analysis.json:scores.inactive_only_error_states; docs/lite/results/hosted-api-repeats-20261003/pass2/jev-latest/analysis.json:scores.inactive_only_error_states; docs/lite/results/hosted-api-repeats-20261003/pass3/jev-latest/analysis.json:scores.inactive_only_error_states
\newcommand{\JevServedModel}{jev-1.13.0}                                % docs/lite/results/four-family/jev-latest/analysis.json:models_returned (single key); docs/lite/results/hosted-api-repeats-20261003/pass2/jev-latest/analysis.json:models_returned (single key); docs/lite/results/hosted-api-repeats-20261003/pass3/jev-latest/analysis.json:models_returned (single key)
\newcommand{\JevRequestedModel}{jev-latest}                             % docs/lite/results/four-family/jev-latest/analysis.json:config.model; docs/lite/results/hosted-api-repeats-20261003/pass2/jev-latest/analysis.json:config.model; docs/lite/results/hosted-api-repeats-20261003/pass3/jev-latest/analysis.json:config.model
\newcommand{\JevDebugUntimed}{41.1}                                     % equal mean across passes: docs/lite/results/four-family/jev-latest/analysis.json:scores.by_family.live_debugging.untimed_decision_accuracy; docs/lite/results/hosted-api-repeats-20261003/pass2/jev-latest/analysis.json:scores.by_family.live_debugging.untimed_decision_accuracy; docs/lite/results/hosted-api-repeats-20261003/pass3/jev-latest/analysis.json:scores.by_family.live_debugging.untimed_decision_accuracy
\newcommand{\JevDebugInForce}{39.4}                                     % equal mean across passes: docs/lite/results/four-family/jev-latest/analysis.json:scores.by_family.live_debugging.time_accuracy; docs/lite/results/hosted-api-repeats-20261003/pass2/jev-latest/analysis.json:scores.by_family.live_debugging.time_accuracy; docs/lite/results/hosted-api-repeats-20261003/pass3/jev-latest/analysis.json:scores.by_family.live_debugging.time_accuracy
\newcommand{\JevDebugNetRemoved}{40.5}                                  % equal mean across passes: docs/lite/results/four-family/jev-latest/analysis.json:network_adjustment.scores.estimate.by_family.live_debugging.time_accuracy; docs/lite/results/hosted-api-repeats-20261003/pass2/jev-latest/analysis.json:network_adjustment.scores.estimate.by_family.live_debugging.time_accuracy; docs/lite/results/hosted-api-repeats-20261003/pass3/jev-latest/analysis.json:network_adjustment.scores.estimate.by_family.live_debugging.time_accuracy
\newcommand{\JevDebugNetRemovedLow}{40.4}                               % equal mean across passes: docs/lite/results/four-family/jev-latest/analysis.json:network_adjustment.scores.low.by_family.live_debugging.time_accuracy; docs/lite/results/hosted-api-repeats-20261003/pass2/jev-latest/analysis.json:network_adjustment.scores.low.by_family.live_debugging.time_accuracy; docs/lite/results/hosted-api-repeats-20261003/pass3/jev-latest/analysis.json:network_adjustment.scores.low.by_family.live_debugging.time_accuracy
\newcommand{\JevDebugNetRemovedHigh}{40.5}                              % equal mean across passes: docs/lite/results/four-family/jev-latest/analysis.json:network_adjustment.scores.high.by_family.live_debugging.time_accuracy; docs/lite/results/hosted-api-repeats-20261003/pass2/jev-latest/analysis.json:network_adjustment.scores.high.by_family.live_debugging.time_accuracy; docs/lite/results/hosted-api-repeats-20261003/pass3/jev-latest/analysis.json:network_adjustment.scores.high.by_family.live_debugging.time_accuracy
\newcommand{\JevAssemblyUntimed}{65.0}                                  % equal mean across passes: docs/lite/results/four-family/jev-latest/analysis.json:scores.by_family.procedural_coaching.untimed_decision_accuracy; docs/lite/results/hosted-api-repeats-20261003/pass2/jev-latest/analysis.json:scores.by_family.procedural_coaching.untimed_decision_accuracy; docs/lite/results/hosted-api-repeats-20261003/pass3/jev-latest/analysis.json:scores.by_family.procedural_coaching.untimed_decision_accuracy
\newcommand{\JevAssemblyInForce}{62.2}                                  % equal mean across passes: docs/lite/results/four-family/jev-latest/analysis.json:scores.by_family.procedural_coaching.time_accuracy; docs/lite/results/hosted-api-repeats-20261003/pass2/jev-latest/analysis.json:scores.by_family.procedural_coaching.time_accuracy; docs/lite/results/hosted-api-repeats-20261003/pass3/jev-latest/analysis.json:scores.by_family.procedural_coaching.time_accuracy
\newcommand{\JevAssemblyNetRemoved}{64.1}                               % equal mean across passes: docs/lite/results/four-family/jev-latest/analysis.json:network_adjustment.scores.estimate.by_family.procedural_coaching.time_accuracy; docs/lite/results/hosted-api-repeats-20261003/pass2/jev-latest/analysis.json:network_adjustment.scores.estimate.by_family.procedural_coaching.time_accuracy; docs/lite/results/hosted-api-repeats-20261003/pass3/jev-latest/analysis.json:network_adjustment.scores.estimate.by_family.procedural_coaching.time_accuracy
\newcommand{\JevAssemblyNetRemovedLow}{64.1}                            % equal mean across passes: docs/lite/results/four-family/jev-latest/analysis.json:network_adjustment.scores.low.by_family.procedural_coaching.time_accuracy; docs/lite/results/hosted-api-repeats-20261003/pass2/jev-latest/analysis.json:network_adjustment.scores.low.by_family.procedural_coaching.time_accuracy; docs/lite/results/hosted-api-repeats-20261003/pass3/jev-latest/analysis.json:network_adjustment.scores.low.by_family.procedural_coaching.time_accuracy
\newcommand{\JevAssemblyNetRemovedHigh}{64.2}                           % equal mean across passes: docs/lite/results/four-family/jev-latest/analysis.json:network_adjustment.scores.high.by_family.procedural_coaching.time_accuracy; docs/lite/results/hosted-api-repeats-20261003/pass2/jev-latest/analysis.json:network_adjustment.scores.high.by_family.procedural_coaching.time_accuracy; docs/lite/results/hosted-api-repeats-20261003/pass3/jev-latest/analysis.json:network_adjustment.scores.high.by_family.procedural_coaching.time_accuracy
\newcommand{\JevSupportUntimed}{71.9}                                   % equal mean across passes: docs/lite/results/four-family/jev-latest/analysis.json:scores.by_family.support_call_assist.untimed_decision_accuracy; docs/lite/results/hosted-api-repeats-20261003/pass2/jev-latest/analysis.json:scores.by_family.support_call_assist.untimed_decision_accuracy; docs/lite/results/hosted-api-repeats-20261003/pass3/jev-latest/analysis.json:scores.by_family.support_call_assist.untimed_decision_accuracy
\newcommand{\JevSupportInForce}{68.9}                                   % equal mean across passes: docs/lite/results/four-family/jev-latest/analysis.json:scores.by_family.support_call_assist.time_accuracy; docs/lite/results/hosted-api-repeats-20261003/pass2/jev-latest/analysis.json:scores.by_family.support_call_assist.time_accuracy; docs/lite/results/hosted-api-repeats-20261003/pass3/jev-latest/analysis.json:scores.by_family.support_call_assist.time_accuracy
\newcommand{\JevSupportNetRemoved}{71.1}                                % equal mean across passes: docs/lite/results/four-family/jev-latest/analysis.json:network_adjustment.scores.estimate.by_family.support_call_assist.time_accuracy; docs/lite/results/hosted-api-repeats-20261003/pass2/jev-latest/analysis.json:network_adjustment.scores.estimate.by_family.support_call_assist.time_accuracy; docs/lite/results/hosted-api-repeats-20261003/pass3/jev-latest/analysis.json:network_adjustment.scores.estimate.by_family.support_call_assist.time_accuracy
\newcommand{\JevSupportNetRemovedLow}{71.1}                             % equal mean across passes: docs/lite/results/four-family/jev-latest/analysis.json:network_adjustment.scores.low.by_family.support_call_assist.time_accuracy; docs/lite/results/hosted-api-repeats-20261003/pass2/jev-latest/analysis.json:network_adjustment.scores.low.by_family.support_call_assist.time_accuracy; docs/lite/results/hosted-api-repeats-20261003/pass3/jev-latest/analysis.json:network_adjustment.scores.low.by_family.support_call_assist.time_accuracy
\newcommand{\JevSupportNetRemovedHigh}{71.2}                            % equal mean across passes: docs/lite/results/four-family/jev-latest/analysis.json:network_adjustment.scores.high.by_family.support_call_assist.time_accuracy; docs/lite/results/hosted-api-repeats-20261003/pass2/jev-latest/analysis.json:network_adjustment.scores.high.by_family.support_call_assist.time_accuracy; docs/lite/results/hosted-api-repeats-20261003/pass3/jev-latest/analysis.json:network_adjustment.scores.high.by_family.support_call_assist.time_accuracy
\newcommand{\JevPresenterUntimed}{71.4}                                 % equal mean across passes: docs/lite/results/four-family/jev-latest/analysis.json:scores.by_family.presenter_voice_control.untimed_decision_accuracy; docs/lite/results/hosted-api-repeats-20261003/pass2/jev-latest/analysis.json:scores.by_family.presenter_voice_control.untimed_decision_accuracy; docs/lite/results/hosted-api-repeats-20261003/pass3/jev-latest/analysis.json:scores.by_family.presenter_voice_control.untimed_decision_accuracy
\newcommand{\JevPresenterInForce}{67.6}                                 % equal mean across passes: docs/lite/results/four-family/jev-latest/analysis.json:scores.by_family.presenter_voice_control.time_accuracy; docs/lite/results/hosted-api-repeats-20261003/pass2/jev-latest/analysis.json:scores.by_family.presenter_voice_control.time_accuracy; docs/lite/results/hosted-api-repeats-20261003/pass3/jev-latest/analysis.json:scores.by_family.presenter_voice_control.time_accuracy
\newcommand{\JevPresenterNetRemoved}{69.9}                              % equal mean across passes: docs/lite/results/four-family/jev-latest/analysis.json:network_adjustment.scores.estimate.by_family.presenter_voice_control.time_accuracy; docs/lite/results/hosted-api-repeats-20261003/pass2/jev-latest/analysis.json:network_adjustment.scores.estimate.by_family.presenter_voice_control.time_accuracy; docs/lite/results/hosted-api-repeats-20261003/pass3/jev-latest/analysis.json:network_adjustment.scores.estimate.by_family.presenter_voice_control.time_accuracy
\newcommand{\JevPresenterNetRemovedLow}{69.8}                           % equal mean across passes: docs/lite/results/four-family/jev-latest/analysis.json:network_adjustment.scores.low.by_family.presenter_voice_control.time_accuracy; docs/lite/results/hosted-api-repeats-20261003/pass2/jev-latest/analysis.json:network_adjustment.scores.low.by_family.presenter_voice_control.time_accuracy; docs/lite/results/hosted-api-repeats-20261003/pass3/jev-latest/analysis.json:network_adjustment.scores.low.by_family.presenter_voice_control.time_accuracy
\newcommand{\JevPresenterNetRemovedHigh}{70.0}                          % equal mean across passes: docs/lite/results/four-family/jev-latest/analysis.json:network_adjustment.scores.high.by_family.presenter_voice_control.time_accuracy; docs/lite/results/hosted-api-repeats-20261003/pass2/jev-latest/analysis.json:network_adjustment.scores.high.by_family.presenter_voice_control.time_accuracy; docs/lite/results/hosted-api-repeats-20261003/pass3/jev-latest/analysis.json:network_adjustment.scores.high.by_family.presenter_voice_control.time_accuracy
\newcommand{\JevDebugAUntimed}{42.2}                                    % equal mean across passes: docs/lite/results/four-family/jev-latest/analysis.json:scores.per_episode[0].untimed_decision_accuracy (lite_debugging_a); docs/lite/results/hosted-api-repeats-20261003/pass2/jev-latest/analysis.json:scores.per_episode[0].untimed_decision_accuracy (lite_debugging_a); docs/lite/results/hosted-api-repeats-20261003/pass3/jev-latest/analysis.json:scores.per_episode[0].untimed_decision_accuracy (lite_debugging_a)
\newcommand{\JevDebugAInForce}{40.1}                                    % equal mean across passes: docs/lite/results/four-family/jev-latest/analysis.json:scores.per_episode[0].time_accuracy (lite_debugging_a); docs/lite/results/hosted-api-repeats-20261003/pass2/jev-latest/analysis.json:scores.per_episode[0].time_accuracy (lite_debugging_a); docs/lite/results/hosted-api-repeats-20261003/pass3/jev-latest/analysis.json:scores.per_episode[0].time_accuracy (lite_debugging_a)
\newcommand{\JevDebugAInForceSeg}{39.5}                                 % equal mean across passes: docs/lite/results/four-family/jev-latest/analysis.json:scores.per_episode[0].segment_time_accuracy (lite_debugging_a); docs/lite/results/hosted-api-repeats-20261003/pass2/jev-latest/analysis.json:scores.per_episode[0].segment_time_accuracy (lite_debugging_a); docs/lite/results/hosted-api-repeats-20261003/pass3/jev-latest/analysis.json:scores.per_episode[0].segment_time_accuracy (lite_debugging_a)
\newcommand{\JevDebugAErrNoDecisionSec}{0.8}                            % equal mean across passes: docs/lite/results/four-family/jev-latest/analysis.json:scores.per_episode[0].error_seconds.no_decision (lite_debugging_a); docs/lite/results/hosted-api-repeats-20261003/pass2/jev-latest/analysis.json:scores.per_episode[0].error_seconds.no_decision (lite_debugging_a); docs/lite/results/hosted-api-repeats-20261003/pass3/jev-latest/analysis.json:scores.per_episode[0].error_seconds.no_decision (lite_debugging_a)
\newcommand{\JevDebugAErrStaleSec}{2.0}                                 % equal mean across passes: docs/lite/results/four-family/jev-latest/analysis.json:scores.per_episode[0].error_seconds.source_correct (lite_debugging_a); docs/lite/results/hosted-api-repeats-20261003/pass2/jev-latest/analysis.json:scores.per_episode[0].error_seconds.source_correct (lite_debugging_a); docs/lite/results/hosted-api-repeats-20261003/pass3/jev-latest/analysis.json:scores.per_episode[0].error_seconds.source_correct (lite_debugging_a)
\newcommand{\JevDebugAErrIncorrectSec}{69.1}                            % equal mean across passes: docs/lite/results/four-family/jev-latest/analysis.json:scores.per_episode[0].error_seconds.source_incorrect (lite_debugging_a); docs/lite/results/hosted-api-repeats-20261003/pass2/jev-latest/analysis.json:scores.per_episode[0].error_seconds.source_incorrect (lite_debugging_a); docs/lite/results/hosted-api-repeats-20261003/pass3/jev-latest/analysis.json:scores.per_episode[0].error_seconds.source_incorrect (lite_debugging_a)
\newcommand{\JevDebugALatencyMedian}{0.23}                              % equal mean across passes: docs/lite/results/four-family/jev-latest/analysis.json:scores.per_episode[0].latency_s_p50 (lite_debugging_a); docs/lite/results/hosted-api-repeats-20261003/pass2/jev-latest/analysis.json:scores.per_episode[0].latency_s_p50 (lite_debugging_a); docs/lite/results/hosted-api-repeats-20261003/pass3/jev-latest/analysis.json:scores.per_episode[0].latency_s_p50 (lite_debugging_a)
\newcommand{\JevDebugBUntimed}{40.0}                                    % equal mean across passes: docs/lite/results/four-family/jev-latest/analysis.json:scores.per_episode[1].untimed_decision_accuracy (lite_debugging_b); docs/lite/results/hosted-api-repeats-20261003/pass2/jev-latest/analysis.json:scores.per_episode[1].untimed_decision_accuracy (lite_debugging_b); docs/lite/results/hosted-api-repeats-20261003/pass3/jev-latest/analysis.json:scores.per_episode[1].untimed_decision_accuracy (lite_debugging_b)
\newcommand{\JevDebugBInForce}{38.7}                                    % equal mean across passes: docs/lite/results/four-family/jev-latest/analysis.json:scores.per_episode[1].time_accuracy (lite_debugging_b); docs/lite/results/hosted-api-repeats-20261003/pass2/jev-latest/analysis.json:scores.per_episode[1].time_accuracy (lite_debugging_b); docs/lite/results/hosted-api-repeats-20261003/pass3/jev-latest/analysis.json:scores.per_episode[1].time_accuracy (lite_debugging_b)
\newcommand{\JevDebugBInForceSeg}{38.2}                                 % equal mean across passes: docs/lite/results/four-family/jev-latest/analysis.json:scores.per_episode[1].segment_time_accuracy (lite_debugging_b); docs/lite/results/hosted-api-repeats-20261003/pass2/jev-latest/analysis.json:scores.per_episode[1].segment_time_accuracy (lite_debugging_b); docs/lite/results/hosted-api-repeats-20261003/pass3/jev-latest/analysis.json:scores.per_episode[1].segment_time_accuracy (lite_debugging_b)
\newcommand{\JevDebugBErrNoDecisionSec}{0.3}                            % equal mean across passes: docs/lite/results/four-family/jev-latest/analysis.json:scores.per_episode[1].error_seconds.no_decision (lite_debugging_b); docs/lite/results/hosted-api-repeats-20261003/pass2/jev-latest/analysis.json:scores.per_episode[1].error_seconds.no_decision (lite_debugging_b); docs/lite/results/hosted-api-repeats-20261003/pass3/jev-latest/analysis.json:scores.per_episode[1].error_seconds.no_decision (lite_debugging_b)
\newcommand{\JevDebugBErrStaleSec}{1.8}                                 % equal mean across passes: docs/lite/results/four-family/jev-latest/analysis.json:scores.per_episode[1].error_seconds.source_correct (lite_debugging_b); docs/lite/results/hosted-api-repeats-20261003/pass2/jev-latest/analysis.json:scores.per_episode[1].error_seconds.source_correct (lite_debugging_b); docs/lite/results/hosted-api-repeats-20261003/pass3/jev-latest/analysis.json:scores.per_episode[1].error_seconds.source_correct (lite_debugging_b)
\newcommand{\JevDebugBErrIncorrectSec}{71.5}                            % equal mean across passes: docs/lite/results/four-family/jev-latest/analysis.json:scores.per_episode[1].error_seconds.source_incorrect (lite_debugging_b); docs/lite/results/hosted-api-repeats-20261003/pass2/jev-latest/analysis.json:scores.per_episode[1].error_seconds.source_incorrect (lite_debugging_b); docs/lite/results/hosted-api-repeats-20261003/pass3/jev-latest/analysis.json:scores.per_episode[1].error_seconds.source_incorrect (lite_debugging_b)
\newcommand{\JevDebugBLatencyMedian}{0.23}                              % equal mean across passes: docs/lite/results/four-family/jev-latest/analysis.json:scores.per_episode[1].latency_s_p50 (lite_debugging_b); docs/lite/results/hosted-api-repeats-20261003/pass2/jev-latest/analysis.json:scores.per_episode[1].latency_s_p50 (lite_debugging_b); docs/lite/results/hosted-api-repeats-20261003/pass3/jev-latest/analysis.json:scores.per_episode[1].latency_s_p50 (lite_debugging_b)
\newcommand{\JevAssemblyAUntimed}{68.3}                                 % equal mean across passes: docs/lite/results/four-family/jev-latest/analysis.json:scores.per_episode[2].untimed_decision_accuracy (lite_assembly_a); docs/lite/results/hosted-api-repeats-20261003/pass2/jev-latest/analysis.json:scores.per_episode[2].untimed_decision_accuracy (lite_assembly_a); docs/lite/results/hosted-api-repeats-20261003/pass3/jev-latest/analysis.json:scores.per_episode[2].untimed_decision_accuracy (lite_assembly_a)
\newcommand{\JevAssemblyAInForce}{65.5}                                 % equal mean across passes: docs/lite/results/four-family/jev-latest/analysis.json:scores.per_episode[2].time_accuracy (lite_assembly_a); docs/lite/results/hosted-api-repeats-20261003/pass2/jev-latest/analysis.json:scores.per_episode[2].time_accuracy (lite_assembly_a); docs/lite/results/hosted-api-repeats-20261003/pass3/jev-latest/analysis.json:scores.per_episode[2].time_accuracy (lite_assembly_a)
\newcommand{\JevAssemblyAInForceSeg}{61.3}                              % equal mean across passes: docs/lite/results/four-family/jev-latest/analysis.json:scores.per_episode[2].segment_time_accuracy (lite_assembly_a); docs/lite/results/hosted-api-repeats-20261003/pass2/jev-latest/analysis.json:scores.per_episode[2].segment_time_accuracy (lite_assembly_a); docs/lite/results/hosted-api-repeats-20261003/pass3/jev-latest/analysis.json:scores.per_episode[2].segment_time_accuracy (lite_assembly_a)
\newcommand{\JevAssemblyAErrNoDecisionSec}{0.3}                         % equal mean across passes: docs/lite/results/four-family/jev-latest/analysis.json:scores.per_episode[2].error_seconds.no_decision (lite_assembly_a); docs/lite/results/hosted-api-repeats-20261003/pass2/jev-latest/analysis.json:scores.per_episode[2].error_seconds.no_decision (lite_assembly_a); docs/lite/results/hosted-api-repeats-20261003/pass3/jev-latest/analysis.json:scores.per_episode[2].error_seconds.no_decision (lite_assembly_a)
\newcommand{\JevAssemblyAErrStaleSec}{3.1}                              % equal mean across passes: docs/lite/results/four-family/jev-latest/analysis.json:scores.per_episode[2].error_seconds.source_correct (lite_assembly_a); docs/lite/results/hosted-api-repeats-20261003/pass2/jev-latest/analysis.json:scores.per_episode[2].error_seconds.source_correct (lite_assembly_a); docs/lite/results/hosted-api-repeats-20261003/pass3/jev-latest/analysis.json:scores.per_episode[2].error_seconds.source_correct (lite_assembly_a)
\newcommand{\JevAssemblyALatencyMedian}{0.23}                           % equal mean across passes: docs/lite/results/four-family/jev-latest/analysis.json:scores.per_episode[2].latency_s_p50 (lite_assembly_a); docs/lite/results/hosted-api-repeats-20261003/pass2/jev-latest/analysis.json:scores.per_episode[2].latency_s_p50 (lite_assembly_a); docs/lite/results/hosted-api-repeats-20261003/pass3/jev-latest/analysis.json:scores.per_episode[2].latency_s_p50 (lite_assembly_a)
\newcommand{\JevAssemblyBUntimed}{61.7}                                 % equal mean across passes: docs/lite/results/four-family/jev-latest/analysis.json:scores.per_episode[3].untimed_decision_accuracy (lite_assembly_b); docs/lite/results/hosted-api-repeats-20261003/pass2/jev-latest/analysis.json:scores.per_episode[3].untimed_decision_accuracy (lite_assembly_b); docs/lite/results/hosted-api-repeats-20261003/pass3/jev-latest/analysis.json:scores.per_episode[3].untimed_decision_accuracy (lite_assembly_b)
\newcommand{\JevAssemblyBInForce}{58.8}                                 % equal mean across passes: docs/lite/results/four-family/jev-latest/analysis.json:scores.per_episode[3].time_accuracy (lite_assembly_b); docs/lite/results/hosted-api-repeats-20261003/pass2/jev-latest/analysis.json:scores.per_episode[3].time_accuracy (lite_assembly_b); docs/lite/results/hosted-api-repeats-20261003/pass3/jev-latest/analysis.json:scores.per_episode[3].time_accuracy (lite_assembly_b)
\newcommand{\JevAssemblyBInForceSeg}{52.5}                              % equal mean across passes: docs/lite/results/four-family/jev-latest/analysis.json:scores.per_episode[3].segment_time_accuracy (lite_assembly_b); docs/lite/results/hosted-api-repeats-20261003/pass2/jev-latest/analysis.json:scores.per_episode[3].segment_time_accuracy (lite_assembly_b); docs/lite/results/hosted-api-repeats-20261003/pass3/jev-latest/analysis.json:scores.per_episode[3].segment_time_accuracy (lite_assembly_b)
\newcommand{\JevAssemblyBErrNoDecisionSec}{0.5}                         % equal mean across passes: docs/lite/results/four-family/jev-latest/analysis.json:scores.per_episode[3].error_seconds.no_decision (lite_assembly_b); docs/lite/results/hosted-api-repeats-20261003/pass2/jev-latest/analysis.json:scores.per_episode[3].error_seconds.no_decision (lite_assembly_b); docs/lite/results/hosted-api-repeats-20261003/pass3/jev-latest/analysis.json:scores.per_episode[3].error_seconds.no_decision (lite_assembly_b)
\newcommand{\JevAssemblyBErrStaleSec}{3.2}                              % equal mean across passes: docs/lite/results/four-family/jev-latest/analysis.json:scores.per_episode[3].error_seconds.source_correct (lite_assembly_b); docs/lite/results/hosted-api-repeats-20261003/pass2/jev-latest/analysis.json:scores.per_episode[3].error_seconds.source_correct (lite_assembly_b); docs/lite/results/hosted-api-repeats-20261003/pass3/jev-latest/analysis.json:scores.per_episode[3].error_seconds.source_correct (lite_assembly_b)
\newcommand{\JevAssemblyBLatencyMedian}{0.23}                           % equal mean across passes: docs/lite/results/four-family/jev-latest/analysis.json:scores.per_episode[3].latency_s_p50 (lite_assembly_b); docs/lite/results/hosted-api-repeats-20261003/pass2/jev-latest/analysis.json:scores.per_episode[3].latency_s_p50 (lite_assembly_b); docs/lite/results/hosted-api-repeats-20261003/pass3/jev-latest/analysis.json:scores.per_episode[3].latency_s_p50 (lite_assembly_b)
\newcommand{\JevSupportAUntimed}{65.6}                                  % equal mean across passes: docs/lite/results/four-family/jev-latest/analysis.json:scores.per_episode[4].untimed_decision_accuracy (lite_support_a); docs/lite/results/hosted-api-repeats-20261003/pass2/jev-latest/analysis.json:scores.per_episode[4].untimed_decision_accuracy (lite_support_a); docs/lite/results/hosted-api-repeats-20261003/pass3/jev-latest/analysis.json:scores.per_episode[4].untimed_decision_accuracy (lite_support_a)
\newcommand{\JevSupportAInForce}{62.8}                                  % equal mean across passes: docs/lite/results/four-family/jev-latest/analysis.json:scores.per_episode[4].time_accuracy (lite_support_a); docs/lite/results/hosted-api-repeats-20261003/pass2/jev-latest/analysis.json:scores.per_episode[4].time_accuracy (lite_support_a); docs/lite/results/hosted-api-repeats-20261003/pass3/jev-latest/analysis.json:scores.per_episode[4].time_accuracy (lite_support_a)
\newcommand{\JevSupportAInForceSeg}{55.1}                               % equal mean across passes: docs/lite/results/four-family/jev-latest/analysis.json:scores.per_episode[4].segment_time_accuracy (lite_support_a); docs/lite/results/hosted-api-repeats-20261003/pass2/jev-latest/analysis.json:scores.per_episode[4].segment_time_accuracy (lite_support_a); docs/lite/results/hosted-api-repeats-20261003/pass3/jev-latest/analysis.json:scores.per_episode[4].segment_time_accuracy (lite_support_a)
\newcommand{\JevSupportAErrNoDecisionSec}{0.5}                          % equal mean across passes: docs/lite/results/four-family/jev-latest/analysis.json:scores.per_episode[4].error_seconds.no_decision (lite_support_a); docs/lite/results/hosted-api-repeats-20261003/pass2/jev-latest/analysis.json:scores.per_episode[4].error_seconds.no_decision (lite_support_a); docs/lite/results/hosted-api-repeats-20261003/pass3/jev-latest/analysis.json:scores.per_episode[4].error_seconds.no_decision (lite_support_a)
\newcommand{\JevSupportAErrStaleSec}{3.2}                               % equal mean across passes: docs/lite/results/four-family/jev-latest/analysis.json:scores.per_episode[4].error_seconds.source_correct (lite_support_a); docs/lite/results/hosted-api-repeats-20261003/pass2/jev-latest/analysis.json:scores.per_episode[4].error_seconds.source_correct (lite_support_a); docs/lite/results/hosted-api-repeats-20261003/pass3/jev-latest/analysis.json:scores.per_episode[4].error_seconds.source_correct (lite_support_a)
\newcommand{\JevSupportALatencyMedian}{0.21}                            % equal mean across passes: docs/lite/results/four-family/jev-latest/analysis.json:scores.per_episode[4].latency_s_p50 (lite_support_a); docs/lite/results/hosted-api-repeats-20261003/pass2/jev-latest/analysis.json:scores.per_episode[4].latency_s_p50 (lite_support_a); docs/lite/results/hosted-api-repeats-20261003/pass3/jev-latest/analysis.json:scores.per_episode[4].latency_s_p50 (lite_support_a)
\newcommand{\JevSupportBUntimed}{78.3}                                  % equal mean across passes: docs/lite/results/four-family/jev-latest/analysis.json:scores.per_episode[5].untimed_decision_accuracy (lite_support_b); docs/lite/results/hosted-api-repeats-20261003/pass2/jev-latest/analysis.json:scores.per_episode[5].untimed_decision_accuracy (lite_support_b); docs/lite/results/hosted-api-repeats-20261003/pass3/jev-latest/analysis.json:scores.per_episode[5].untimed_decision_accuracy (lite_support_b)
\newcommand{\JevSupportBInForce}{75.0}                                  % equal mean across passes: docs/lite/results/four-family/jev-latest/analysis.json:scores.per_episode[5].time_accuracy (lite_support_b); docs/lite/results/hosted-api-repeats-20261003/pass2/jev-latest/analysis.json:scores.per_episode[5].time_accuracy (lite_support_b); docs/lite/results/hosted-api-repeats-20261003/pass3/jev-latest/analysis.json:scores.per_episode[5].time_accuracy (lite_support_b)
\newcommand{\JevSupportBInForceSeg}{77.6}                               % equal mean across passes: docs/lite/results/four-family/jev-latest/analysis.json:scores.per_episode[5].segment_time_accuracy (lite_support_b); docs/lite/results/hosted-api-repeats-20261003/pass2/jev-latest/analysis.json:scores.per_episode[5].segment_time_accuracy (lite_support_b); docs/lite/results/hosted-api-repeats-20261003/pass3/jev-latest/analysis.json:scores.per_episode[5].segment_time_accuracy (lite_support_b)
\newcommand{\JevSupportBErrNoDecisionSec}{0.3}                          % equal mean across passes: docs/lite/results/four-family/jev-latest/analysis.json:scores.per_episode[5].error_seconds.no_decision (lite_support_b); docs/lite/results/hosted-api-repeats-20261003/pass2/jev-latest/analysis.json:scores.per_episode[5].error_seconds.no_decision (lite_support_b); docs/lite/results/hosted-api-repeats-20261003/pass3/jev-latest/analysis.json:scores.per_episode[5].error_seconds.no_decision (lite_support_b)
\newcommand{\JevSupportBErrStaleSec}{4.2}                               % equal mean across passes: docs/lite/results/four-family/jev-latest/analysis.json:scores.per_episode[5].error_seconds.source_correct (lite_support_b); docs/lite/results/hosted-api-repeats-20261003/pass2/jev-latest/analysis.json:scores.per_episode[5].error_seconds.source_correct (lite_support_b); docs/lite/results/hosted-api-repeats-20261003/pass3/jev-latest/analysis.json:scores.per_episode[5].error_seconds.source_correct (lite_support_b)
\newcommand{\JevSupportBLatencyMedian}{0.22}                            % equal mean across passes: docs/lite/results/four-family/jev-latest/analysis.json:scores.per_episode[5].latency_s_p50 (lite_support_b); docs/lite/results/hosted-api-repeats-20261003/pass2/jev-latest/analysis.json:scores.per_episode[5].latency_s_p50 (lite_support_b); docs/lite/results/hosted-api-repeats-20261003/pass3/jev-latest/analysis.json:scores.per_episode[5].latency_s_p50 (lite_support_b)
\newcommand{\JevPresenterAUntimed}{69.4}                                % equal mean across passes: docs/lite/results/four-family/jev-latest/analysis.json:scores.per_episode[6].untimed_decision_accuracy (lite_presenter_a); docs/lite/results/hosted-api-repeats-20261003/pass2/jev-latest/analysis.json:scores.per_episode[6].untimed_decision_accuracy (lite_presenter_a); docs/lite/results/hosted-api-repeats-20261003/pass3/jev-latest/analysis.json:scores.per_episode[6].untimed_decision_accuracy (lite_presenter_a)
\newcommand{\JevPresenterAInForce}{65.2}                                % equal mean across passes: docs/lite/results/four-family/jev-latest/analysis.json:scores.per_episode[6].time_accuracy (lite_presenter_a); docs/lite/results/hosted-api-repeats-20261003/pass2/jev-latest/analysis.json:scores.per_episode[6].time_accuracy (lite_presenter_a); docs/lite/results/hosted-api-repeats-20261003/pass3/jev-latest/analysis.json:scores.per_episode[6].time_accuracy (lite_presenter_a)
\newcommand{\JevPresenterAInForceSeg}{61.5}                             % equal mean across passes: docs/lite/results/four-family/jev-latest/analysis.json:scores.per_episode[6].segment_time_accuracy (lite_presenter_a); docs/lite/results/hosted-api-repeats-20261003/pass2/jev-latest/analysis.json:scores.per_episode[6].segment_time_accuracy (lite_presenter_a); docs/lite/results/hosted-api-repeats-20261003/pass3/jev-latest/analysis.json:scores.per_episode[6].segment_time_accuracy (lite_presenter_a)
\newcommand{\JevPresenterAErrNoDecisionSec}{0.6}                        % equal mean across passes: docs/lite/results/four-family/jev-latest/analysis.json:scores.per_episode[6].error_seconds.no_decision (lite_presenter_a); docs/lite/results/hosted-api-repeats-20261003/pass2/jev-latest/analysis.json:scores.per_episode[6].error_seconds.no_decision (lite_presenter_a); docs/lite/results/hosted-api-repeats-20261003/pass3/jev-latest/analysis.json:scores.per_episode[6].error_seconds.no_decision (lite_presenter_a)
\newcommand{\JevPresenterAErrStaleSec}{4.8}                             % equal mean across passes: docs/lite/results/four-family/jev-latest/analysis.json:scores.per_episode[6].error_seconds.source_correct (lite_presenter_a); docs/lite/results/hosted-api-repeats-20261003/pass2/jev-latest/analysis.json:scores.per_episode[6].error_seconds.source_correct (lite_presenter_a); docs/lite/results/hosted-api-repeats-20261003/pass3/jev-latest/analysis.json:scores.per_episode[6].error_seconds.source_correct (lite_presenter_a)
\newcommand{\JevPresenterALatencyMedian}{0.26}                          % equal mean across passes: docs/lite/results/four-family/jev-latest/analysis.json:scores.per_episode[6].latency_s_p50 (lite_presenter_a); docs/lite/results/hosted-api-repeats-20261003/pass2/jev-latest/analysis.json:scores.per_episode[6].latency_s_p50 (lite_presenter_a); docs/lite/results/hosted-api-repeats-20261003/pass3/jev-latest/analysis.json:scores.per_episode[6].latency_s_p50 (lite_presenter_a)
\newcommand{\JevPresenterBUntimed}{73.3}                                % equal mean across passes: docs/lite/results/four-family/jev-latest/analysis.json:scores.per_episode[7].untimed_decision_accuracy (lite_presenter_b); docs/lite/results/hosted-api-repeats-20261003/pass2/jev-latest/analysis.json:scores.per_episode[7].untimed_decision_accuracy (lite_presenter_b); docs/lite/results/hosted-api-repeats-20261003/pass3/jev-latest/analysis.json:scores.per_episode[7].untimed_decision_accuracy (lite_presenter_b)
\newcommand{\JevPresenterBInForce}{69.9}                                % equal mean across passes: docs/lite/results/four-family/jev-latest/analysis.json:scores.per_episode[7].time_accuracy (lite_presenter_b); docs/lite/results/hosted-api-repeats-20261003/pass2/jev-latest/analysis.json:scores.per_episode[7].time_accuracy (lite_presenter_b); docs/lite/results/hosted-api-repeats-20261003/pass3/jev-latest/analysis.json:scores.per_episode[7].time_accuracy (lite_presenter_b)
\newcommand{\JevPresenterBInForceSeg}{66.3}                             % equal mean across passes: docs/lite/results/four-family/jev-latest/analysis.json:scores.per_episode[7].segment_time_accuracy (lite_presenter_b); docs/lite/results/hosted-api-repeats-20261003/pass2/jev-latest/analysis.json:scores.per_episode[7].segment_time_accuracy (lite_presenter_b); docs/lite/results/hosted-api-repeats-20261003/pass3/jev-latest/analysis.json:scores.per_episode[7].segment_time_accuracy (lite_presenter_b)
\newcommand{\JevPresenterBErrNoDecisionSec}{0.3}                        % equal mean across passes: docs/lite/results/four-family/jev-latest/analysis.json:scores.per_episode[7].error_seconds.no_decision (lite_presenter_b); docs/lite/results/hosted-api-repeats-20261003/pass2/jev-latest/analysis.json:scores.per_episode[7].error_seconds.no_decision (lite_presenter_b); docs/lite/results/hosted-api-repeats-20261003/pass3/jev-latest/analysis.json:scores.per_episode[7].error_seconds.no_decision (lite_presenter_b)
\newcommand{\JevPresenterBErrStaleSec}{4.0}                             % equal mean across passes: docs/lite/results/four-family/jev-latest/analysis.json:scores.per_episode[7].error_seconds.source_correct (lite_presenter_b); docs/lite/results/hosted-api-repeats-20261003/pass2/jev-latest/analysis.json:scores.per_episode[7].error_seconds.source_correct (lite_presenter_b); docs/lite/results/hosted-api-repeats-20261003/pass3/jev-latest/analysis.json:scores.per_episode[7].error_seconds.source_correct (lite_presenter_b)
\newcommand{\JevPresenterBLatencyMedian}{0.24}                          % equal mean across passes: docs/lite/results/four-family/jev-latest/analysis.json:scores.per_episode[7].latency_s_p50 (lite_presenter_b); docs/lite/results/hosted-api-repeats-20261003/pass2/jev-latest/analysis.json:scores.per_episode[7].latency_s_p50 (lite_presenter_b); docs/lite/results/hosted-api-repeats-20261003/pass3/jev-latest/analysis.json:scores.per_episode[7].latency_s_p50 (lite_presenter_b)
\newcommand{\LunaTimingCeiling}{50.4}                                   % equal mean across passes: runs/lite-v1-gpt-5.6-luna-low-four-family-retry-v1/events.jsonl replayed with retry_scoring.normalized_episode_scores, every pred replaced by data/lite/v1 gold (mean over scenarios); runs/hosted-api-repeats-20261003/pass2/gpt-5.6-luna-low/events.jsonl replayed with retry_scoring.normalized_episode_scores, every pred replaced by data/lite/v1 gold (mean over scenarios); runs/hosted-api-repeats-20261003/pass3/gpt-5.6-luna-low/events.jsonl replayed with retry_scoring.normalized_episode_scores, every pred replaced by data/lite/v1 gold (mean over scenarios)
\newcommand{\LunaInForceDelayHalf}{67.8}                                % equal mean across passes: runs/lite-v1-gpt-5.6-luna-low-four-family-retry-v1/events.jsonl replayed with every delay after release x0.5 (retry_scoring.normalized_episode_scores; mean over scenarios); runs/hosted-api-repeats-20261003/pass2/gpt-5.6-luna-low/events.jsonl replayed with every delay after release x0.5 (retry_scoring.normalized_episode_scores; mean over scenarios); runs/hosted-api-repeats-20261003/pass3/gpt-5.6-luna-low/events.jsonl replayed with every delay after release x0.5 (retry_scoring.normalized_episode_scores; mean over scenarios)
\newcommand{\LunaInForceDelayTwoThirds}{60.7}                           % equal mean across passes: runs/lite-v1-gpt-5.6-luna-low-four-family-retry-v1/events.jsonl replayed with every delay after release x0.6667 (retry_scoring.normalized_episode_scores; mean over scenarios); runs/hosted-api-repeats-20261003/pass2/gpt-5.6-luna-low/events.jsonl replayed with every delay after release x0.6667 (retry_scoring.normalized_episode_scores; mean over scenarios); runs/hosted-api-repeats-20261003/pass3/gpt-5.6-luna-low/events.jsonl replayed with every delay after release x0.6667 (retry_scoring.normalized_episode_scores; mean over scenarios)
\newcommand{\LunaInForceDelayThreeHalves}{32.6}                         % equal mean across passes: runs/lite-v1-gpt-5.6-luna-low-four-family-retry-v1/events.jsonl replayed with every delay after release x1.5 (retry_scoring.normalized_episode_scores; mean over scenarios); runs/hosted-api-repeats-20261003/pass2/gpt-5.6-luna-low/events.jsonl replayed with every delay after release x1.5 (retry_scoring.normalized_episode_scores; mean over scenarios); runs/hosted-api-repeats-20261003/pass3/gpt-5.6-luna-low/events.jsonl replayed with every delay after release x1.5 (retry_scoring.normalized_episode_scores; mean over scenarios)
\newcommand{\LunaInForceDelayDouble}{23.3}                              % equal mean across passes: runs/lite-v1-gpt-5.6-luna-low-four-family-retry-v1/events.jsonl replayed with every delay after release x2 (retry_scoring.normalized_episode_scores; mean over scenarios); runs/hosted-api-repeats-20261003/pass2/gpt-5.6-luna-low/events.jsonl replayed with every delay after release x2 (retry_scoring.normalized_episode_scores; mean over scenarios); runs/hosted-api-repeats-20261003/pass3/gpt-5.6-luna-low/events.jsonl replayed with every delay after release x2 (retry_scoring.normalized_episode_scores; mean over scenarios)
\newcommand{\LunaNoneTimingCeiling}{73.4}                               % equal mean across passes: runs/lite-v1-gpt-5.6-luna-none-four-family-retry-v1/events.jsonl replayed with retry_scoring.normalized_episode_scores, every pred replaced by data/lite/v1 gold (mean over scenarios); runs/hosted-api-repeats-20261003/pass2/gpt-5.6-luna-none/events.jsonl replayed with retry_scoring.normalized_episode_scores, every pred replaced by data/lite/v1 gold (mean over scenarios); runs/hosted-api-repeats-20261003/pass3/gpt-5.6-luna-none/events.jsonl replayed with retry_scoring.normalized_episode_scores, every pred replaced by data/lite/v1 gold (mean over scenarios)
\newcommand{\LunaNoneInForceDelayHalf}{38.4}                            % equal mean across passes: runs/lite-v1-gpt-5.6-luna-none-four-family-retry-v1/events.jsonl replayed with every delay after release x0.5 (retry_scoring.normalized_episode_scores; mean over scenarios); runs/hosted-api-repeats-20261003/pass2/gpt-5.6-luna-none/events.jsonl replayed with every delay after release x0.5 (retry_scoring.normalized_episode_scores; mean over scenarios); runs/hosted-api-repeats-20261003/pass3/gpt-5.6-luna-none/events.jsonl replayed with every delay after release x0.5 (retry_scoring.normalized_episode_scores; mean over scenarios)
\newcommand{\LunaNoneInForceDelayTwoThirds}{36.8}                       % equal mean across passes: runs/lite-v1-gpt-5.6-luna-none-four-family-retry-v1/events.jsonl replayed with every delay after release x0.6667 (retry_scoring.normalized_episode_scores; mean over scenarios); runs/hosted-api-repeats-20261003/pass2/gpt-5.6-luna-none/events.jsonl replayed with every delay after release x0.6667 (retry_scoring.normalized_episode_scores; mean over scenarios); runs/hosted-api-repeats-20261003/pass3/gpt-5.6-luna-none/events.jsonl replayed with every delay after release x0.6667 (retry_scoring.normalized_episode_scores; mean over scenarios)
\newcommand{\LunaNoneInForceDelayThreeHalves}{28.6}                     % equal mean across passes: runs/lite-v1-gpt-5.6-luna-none-four-family-retry-v1/events.jsonl replayed with every delay after release x1.5 (retry_scoring.normalized_episode_scores; mean over scenarios); runs/hosted-api-repeats-20261003/pass2/gpt-5.6-luna-none/events.jsonl replayed with every delay after release x1.5 (retry_scoring.normalized_episode_scores; mean over scenarios); runs/hosted-api-repeats-20261003/pass3/gpt-5.6-luna-none/events.jsonl replayed with every delay after release x1.5 (retry_scoring.normalized_episode_scores; mean over scenarios)
\newcommand{\LunaNoneInForceDelayDouble}{24.4}                          % equal mean across passes: runs/lite-v1-gpt-5.6-luna-none-four-family-retry-v1/events.jsonl replayed with every delay after release x2 (retry_scoring.normalized_episode_scores; mean over scenarios); runs/hosted-api-repeats-20261003/pass2/gpt-5.6-luna-none/events.jsonl replayed with every delay after release x2 (retry_scoring.normalized_episode_scores; mean over scenarios); runs/hosted-api-repeats-20261003/pass3/gpt-5.6-luna-none/events.jsonl replayed with every delay after release x2 (retry_scoring.normalized_episode_scores; mean over scenarios)
\newcommand{\TerraTimingCeiling}{57.1}                                  % equal mean across passes: runs/lite-v1-gpt-5.6-terra-low-four-family-retry-v1/events.jsonl replayed with retry_scoring.normalized_episode_scores, every pred replaced by data/lite/v1 gold (mean over scenarios); runs/hosted-api-repeats-20261003/pass2/gpt-5.6-terra-low/events.jsonl replayed with retry_scoring.normalized_episode_scores, every pred replaced by data/lite/v1 gold (mean over scenarios); runs/hosted-api-repeats-20261003/pass3/gpt-5.6-terra-low/events.jsonl replayed with retry_scoring.normalized_episode_scores, every pred replaced by data/lite/v1 gold (mean over scenarios)
\newcommand{\TerraInForceDelayHalf}{75.2}                               % equal mean across passes: runs/lite-v1-gpt-5.6-terra-low-four-family-retry-v1/events.jsonl replayed with every delay after release x0.5 (retry_scoring.normalized_episode_scores; mean over scenarios); runs/hosted-api-repeats-20261003/pass2/gpt-5.6-terra-low/events.jsonl replayed with every delay after release x0.5 (retry_scoring.normalized_episode_scores; mean over scenarios); runs/hosted-api-repeats-20261003/pass3/gpt-5.6-terra-low/events.jsonl replayed with every delay after release x0.5 (retry_scoring.normalized_episode_scores; mean over scenarios)
\newcommand{\TerraInForceDelayTwoThirds}{68.5}                          % equal mean across passes: runs/lite-v1-gpt-5.6-terra-low-four-family-retry-v1/events.jsonl replayed with every delay after release x0.6667 (retry_scoring.normalized_episode_scores; mean over scenarios); runs/hosted-api-repeats-20261003/pass2/gpt-5.6-terra-low/events.jsonl replayed with every delay after release x0.6667 (retry_scoring.normalized_episode_scores; mean over scenarios); runs/hosted-api-repeats-20261003/pass3/gpt-5.6-terra-low/events.jsonl replayed with every delay after release x0.6667 (retry_scoring.normalized_episode_scores; mean over scenarios)
\newcommand{\TerraInForceDelayThreeHalves}{39.8}                        % equal mean across passes: runs/lite-v1-gpt-5.6-terra-low-four-family-retry-v1/events.jsonl replayed with every delay after release x1.5 (retry_scoring.normalized_episode_scores; mean over scenarios); runs/hosted-api-repeats-20261003/pass2/gpt-5.6-terra-low/events.jsonl replayed with every delay after release x1.5 (retry_scoring.normalized_episode_scores; mean over scenarios); runs/hosted-api-repeats-20261003/pass3/gpt-5.6-terra-low/events.jsonl replayed with every delay after release x1.5 (retry_scoring.normalized_episode_scores; mean over scenarios)
\newcommand{\TerraInForceDelayDouble}{28.3}                             % equal mean across passes: runs/lite-v1-gpt-5.6-terra-low-four-family-retry-v1/events.jsonl replayed with every delay after release x2 (retry_scoring.normalized_episode_scores; mean over scenarios); runs/hosted-api-repeats-20261003/pass2/gpt-5.6-terra-low/events.jsonl replayed with every delay after release x2 (retry_scoring.normalized_episode_scores; mean over scenarios); runs/hosted-api-repeats-20261003/pass3/gpt-5.6-terra-low/events.jsonl replayed with every delay after release x2 (retry_scoring.normalized_episode_scores; mean over scenarios)
\newcommand{\TerraNoneTimingCeiling}{71.8}                              % equal mean across passes: runs/lite-v1-gpt-5.6-terra-none-four-family-retry-v1/events.jsonl replayed with retry_scoring.normalized_episode_scores, every pred replaced by data/lite/v1 gold (mean over scenarios); runs/hosted-api-repeats-20261003/pass2/gpt-5.6-terra-none/events.jsonl replayed with retry_scoring.normalized_episode_scores, every pred replaced by data/lite/v1 gold (mean over scenarios); runs/hosted-api-repeats-20261003/pass3/gpt-5.6-terra-none/events.jsonl replayed with retry_scoring.normalized_episode_scores, every pred replaced by data/lite/v1 gold (mean over scenarios)
\newcommand{\TerraNoneInForceDelayHalf}{71.5}                           % equal mean across passes: runs/lite-v1-gpt-5.6-terra-none-four-family-retry-v1/events.jsonl replayed with every delay after release x0.5 (retry_scoring.normalized_episode_scores; mean over scenarios); runs/hosted-api-repeats-20261003/pass2/gpt-5.6-terra-none/events.jsonl replayed with every delay after release x0.5 (retry_scoring.normalized_episode_scores; mean over scenarios); runs/hosted-api-repeats-20261003/pass3/gpt-5.6-terra-none/events.jsonl replayed with every delay after release x0.5 (retry_scoring.normalized_episode_scores; mean over scenarios)
\newcommand{\TerraNoneInForceDelayTwoThirds}{68.1}                      % equal mean across passes: runs/lite-v1-gpt-5.6-terra-none-four-family-retry-v1/events.jsonl replayed with every delay after release x0.6667 (retry_scoring.normalized_episode_scores; mean over scenarios); runs/hosted-api-repeats-20261003/pass2/gpt-5.6-terra-none/events.jsonl replayed with every delay after release x0.6667 (retry_scoring.normalized_episode_scores; mean over scenarios); runs/hosted-api-repeats-20261003/pass3/gpt-5.6-terra-none/events.jsonl replayed with every delay after release x0.6667 (retry_scoring.normalized_episode_scores; mean over scenarios)
\newcommand{\TerraNoneInForceDelayThreeHalves}{51.1}                    % equal mean across passes: runs/lite-v1-gpt-5.6-terra-none-four-family-retry-v1/events.jsonl replayed with every delay after release x1.5 (retry_scoring.normalized_episode_scores; mean over scenarios); runs/hosted-api-repeats-20261003/pass2/gpt-5.6-terra-none/events.jsonl replayed with every delay after release x1.5 (retry_scoring.normalized_episode_scores; mean over scenarios); runs/hosted-api-repeats-20261003/pass3/gpt-5.6-terra-none/events.jsonl replayed with every delay after release x1.5 (retry_scoring.normalized_episode_scores; mean over scenarios)
\newcommand{\TerraNoneInForceDelayDouble}{41.4}                         % equal mean across passes: runs/lite-v1-gpt-5.6-terra-none-four-family-retry-v1/events.jsonl replayed with every delay after release x2 (retry_scoring.normalized_episode_scores; mean over scenarios); runs/hosted-api-repeats-20261003/pass2/gpt-5.6-terra-none/events.jsonl replayed with every delay after release x2 (retry_scoring.normalized_episode_scores; mean over scenarios); runs/hosted-api-repeats-20261003/pass3/gpt-5.6-terra-none/events.jsonl replayed with every delay after release x2 (retry_scoring.normalized_episode_scores; mean over scenarios)
\newcommand{\AstraTimingCeiling}{52.1}                                  % equal mean across passes: runs/lite-v1-gpt-6-astra-low-four-family-retry-v1/events.jsonl replayed with retry_scoring.normalized_episode_scores, every pred replaced by data/lite/v1 gold (mean over scenarios); runs/hosted-api-repeats-20261003/pass2/gpt-6-astra-low/events.jsonl replayed with retry_scoring.normalized_episode_scores, every pred replaced by data/lite/v1 gold (mean over scenarios); runs/hosted-api-repeats-20261003/pass3/gpt-6-astra-low/events.jsonl replayed with retry_scoring.normalized_episode_scores, every pred replaced by data/lite/v1 gold (mean over scenarios)
\newcommand{\AstraInForceDelayHalf}{75.2}                               % equal mean across passes: runs/lite-v1-gpt-6-astra-low-four-family-retry-v1/events.jsonl replayed with every delay after release x0.5 (retry_scoring.normalized_episode_scores; mean over scenarios); runs/hosted-api-repeats-20261003/pass2/gpt-6-astra-low/events.jsonl replayed with every delay after release x0.5 (retry_scoring.normalized_episode_scores; mean over scenarios); runs/hosted-api-repeats-20261003/pass3/gpt-6-astra-low/events.jsonl replayed with every delay after release x0.5 (retry_scoring.normalized_episode_scores; mean over scenarios)
\newcommand{\AstraInForceDelayTwoThirds}{67.1}                          % equal mean across passes: runs/lite-v1-gpt-6-astra-low-four-family-retry-v1/events.jsonl replayed with every delay after release x0.6667 (retry_scoring.normalized_episode_scores; mean over scenarios); runs/hosted-api-repeats-20261003/pass2/gpt-6-astra-low/events.jsonl replayed with every delay after release x0.6667 (retry_scoring.normalized_episode_scores; mean over scenarios); runs/hosted-api-repeats-20261003/pass3/gpt-6-astra-low/events.jsonl replayed with every delay after release x0.6667 (retry_scoring.normalized_episode_scores; mean over scenarios)
\newcommand{\AstraInForceDelayThreeHalves}{34.1}                        % equal mean across passes: runs/lite-v1-gpt-6-astra-low-four-family-retry-v1/events.jsonl replayed with every delay after release x1.5 (retry_scoring.normalized_episode_scores; mean over scenarios); runs/hosted-api-repeats-20261003/pass2/gpt-6-astra-low/events.jsonl replayed with every delay after release x1.5 (retry_scoring.normalized_episode_scores; mean over scenarios); runs/hosted-api-repeats-20261003/pass3/gpt-6-astra-low/events.jsonl replayed with every delay after release x1.5 (retry_scoring.normalized_episode_scores; mean over scenarios)
\newcommand{\AstraInForceDelayDouble}{23.2}                             % equal mean across passes: runs/lite-v1-gpt-6-astra-low-four-family-retry-v1/events.jsonl replayed with every delay after release x2 (retry_scoring.normalized_episode_scores; mean over scenarios); runs/hosted-api-repeats-20261003/pass2/gpt-6-astra-low/events.jsonl replayed with every delay after release x2 (retry_scoring.normalized_episode_scores; mean over scenarios); runs/hosted-api-repeats-20261003/pass3/gpt-6-astra-low/events.jsonl replayed with every delay after release x2 (retry_scoring.normalized_episode_scores; mean over scenarios)
\newcommand{\JevTimingCeiling}{95.1}                                    % equal mean across passes: runs/lite-v1-jev-latest-four-family-retry-v1/events.jsonl replayed with retry_scoring.normalized_episode_scores, every pred replaced by data/lite/v1 gold (mean over scenarios); runs/hosted-api-repeats-20261003/pass2/jev-latest/events.jsonl replayed with retry_scoring.normalized_episode_scores, every pred replaced by data/lite/v1 gold (mean over scenarios); runs/hosted-api-repeats-20261003/pass3/jev-latest/events.jsonl replayed with retry_scoring.normalized_episode_scores, every pred replaced by data/lite/v1 gold (mean over scenarios)
\newcommand{\JevInForceDelayHalf}{60.9}                                 % equal mean across passes: runs/lite-v1-jev-latest-four-family-retry-v1/events.jsonl replayed with every delay after release x0.5 (retry_scoring.normalized_episode_scores; mean over scenarios); runs/hosted-api-repeats-20261003/pass2/jev-latest/events.jsonl replayed with every delay after release x0.5 (retry_scoring.normalized_episode_scores; mean over scenarios); runs/hosted-api-repeats-20261003/pass3/jev-latest/events.jsonl replayed with every delay after release x0.5 (retry_scoring.normalized_episode_scores; mean over scenarios)
\newcommand{\JevInForceDelayTwoThirds}{60.5}                            % equal mean across passes: runs/lite-v1-jev-latest-four-family-retry-v1/events.jsonl replayed with every delay after release x0.6667 (retry_scoring.normalized_episode_scores; mean over scenarios); runs/hosted-api-repeats-20261003/pass2/jev-latest/events.jsonl replayed with every delay after release x0.6667 (retry_scoring.normalized_episode_scores; mean over scenarios); runs/hosted-api-repeats-20261003/pass3/jev-latest/events.jsonl replayed with every delay after release x0.6667 (retry_scoring.normalized_episode_scores; mean over scenarios)
\newcommand{\JevInForceDelayThreeHalves}{58.1}                          % equal mean across passes: runs/lite-v1-jev-latest-four-family-retry-v1/events.jsonl replayed with every delay after release x1.5 (retry_scoring.normalized_episode_scores; mean over scenarios); runs/hosted-api-repeats-20261003/pass2/jev-latest/events.jsonl replayed with every delay after release x1.5 (retry_scoring.normalized_episode_scores; mean over scenarios); runs/hosted-api-repeats-20261003/pass3/jev-latest/events.jsonl replayed with every delay after release x1.5 (retry_scoring.normalized_episode_scores; mean over scenarios)
\newcommand{\JevInForceDelayDouble}{56.7}                               % equal mean across passes: runs/lite-v1-jev-latest-four-family-retry-v1/events.jsonl replayed with every delay after release x2 (retry_scoring.normalized_episode_scores; mean over scenarios); runs/hosted-api-repeats-20261003/pass2/jev-latest/events.jsonl replayed with every delay after release x2 (retry_scoring.normalized_episode_scores; mean over scenarios); runs/hosted-api-repeats-20261003/pass3/jev-latest/events.jsonl replayed with every delay after release x2 (retry_scoring.normalized_episode_scores; mean over scenarios)
\newcommand{\LunaErrJudgmentPct}{3.8}                                   % equal mean across passes: runs/lite-v1-gpt-5.6-luna-low-four-family-retry-v1/events.jsonl rescored with lite.scoring (time_partition_seconds, summed over scenarios); each scenario divided by its own observed duration, then equal family mean; runs/hosted-api-repeats-20261003/pass2/gpt-5.6-luna-low/events.jsonl rescored with lite.scoring (time_partition_seconds, summed over scenarios); each scenario divided by its own observed duration, then equal family mean; runs/hosted-api-repeats-20261003/pass3/gpt-5.6-luna-low/events.jsonl rescored with lite.scoring (time_partition_seconds, summed over scenarios); each scenario divided by its own observed duration, then equal family mean
\newcommand{\LunaErrCompoundPct}{5.0}                                   % equal mean across passes: runs/lite-v1-gpt-5.6-luna-low-four-family-retry-v1/events.jsonl rescored with lite.scoring (time_partition_seconds, summed over scenarios); each scenario divided by its own observed duration, then equal family mean; runs/hosted-api-repeats-20261003/pass2/gpt-5.6-luna-low/events.jsonl rescored with lite.scoring (time_partition_seconds, summed over scenarios); each scenario divided by its own observed duration, then equal family mean; runs/hosted-api-repeats-20261003/pass3/gpt-5.6-luna-low/events.jsonl rescored with lite.scoring (time_partition_seconds, summed over scenarios); each scenario divided by its own observed duration, then equal family mean
\newcommand{\LunaLuckyPct}{1.3}                                         % equal mean across passes: runs/lite-v1-gpt-5.6-luna-low-four-family-retry-v1/events.jsonl rescored with lite.scoring (time_partition_seconds, summed over scenarios); each scenario divided by its own observed duration, then equal family mean; runs/hosted-api-repeats-20261003/pass2/gpt-5.6-luna-low/events.jsonl rescored with lite.scoring (time_partition_seconds, summed over scenarios); each scenario divided by its own observed duration, then equal family mean; runs/hosted-api-repeats-20261003/pass3/gpt-5.6-luna-low/events.jsonl rescored with lite.scoring (time_partition_seconds, summed over scenarios); each scenario divided by its own observed duration, then equal family mean
\newcommand{\LunaUCurrent}{92.5}                                        % docs/research/manuscript-three-pass/analysis.json:mean current-correct / mean current-source
\newcommand{\LunaDebugAErrJudgmentSec}{6.2}                             % equal mean across passes: runs/lite-v1-gpt-5.6-luna-low-four-family-retry-v1/events.jsonl rescored (lite_debugging_a, time_partition_seconds.judgment); runs/hosted-api-repeats-20261003/pass2/gpt-5.6-luna-low/events.jsonl rescored (lite_debugging_a, time_partition_seconds.judgment); runs/hosted-api-repeats-20261003/pass3/gpt-5.6-luna-low/events.jsonl rescored (lite_debugging_a, time_partition_seconds.judgment)
\newcommand{\LunaDebugAErrCompoundSec}{2.1}                             % equal mean across passes: runs/lite-v1-gpt-5.6-luna-low-four-family-retry-v1/events.jsonl rescored (lite_debugging_a, time_partition_seconds.compound); runs/hosted-api-repeats-20261003/pass2/gpt-5.6-luna-low/events.jsonl rescored (lite_debugging_a, time_partition_seconds.compound); runs/hosted-api-repeats-20261003/pass3/gpt-5.6-luna-low/events.jsonl rescored (lite_debugging_a, time_partition_seconds.compound)
\newcommand{\LunaDebugBErrJudgmentSec}{8.1}                             % equal mean across passes: runs/lite-v1-gpt-5.6-luna-low-four-family-retry-v1/events.jsonl rescored (lite_debugging_b, time_partition_seconds.judgment); runs/hosted-api-repeats-20261003/pass2/gpt-5.6-luna-low/events.jsonl rescored (lite_debugging_b, time_partition_seconds.judgment); runs/hosted-api-repeats-20261003/pass3/gpt-5.6-luna-low/events.jsonl rescored (lite_debugging_b, time_partition_seconds.judgment)
\newcommand{\LunaDebugBErrCompoundSec}{1.5}                             % equal mean across passes: runs/lite-v1-gpt-5.6-luna-low-four-family-retry-v1/events.jsonl rescored (lite_debugging_b, time_partition_seconds.compound); runs/hosted-api-repeats-20261003/pass2/gpt-5.6-luna-low/events.jsonl rescored (lite_debugging_b, time_partition_seconds.compound); runs/hosted-api-repeats-20261003/pass3/gpt-5.6-luna-low/events.jsonl rescored (lite_debugging_b, time_partition_seconds.compound)
\newcommand{\LunaAssemblyAErrJudgmentSec}{2.6}                          % equal mean across passes: runs/lite-v1-gpt-5.6-luna-low-four-family-retry-v1/events.jsonl rescored (lite_assembly_a, time_partition_seconds.judgment); runs/hosted-api-repeats-20261003/pass2/gpt-5.6-luna-low/events.jsonl rescored (lite_assembly_a, time_partition_seconds.judgment); runs/hosted-api-repeats-20261003/pass3/gpt-5.6-luna-low/events.jsonl rescored (lite_assembly_a, time_partition_seconds.judgment)
\newcommand{\LunaAssemblyAErrCompoundSec}{8.1}                          % equal mean across passes: runs/lite-v1-gpt-5.6-luna-low-four-family-retry-v1/events.jsonl rescored (lite_assembly_a, time_partition_seconds.compound); runs/hosted-api-repeats-20261003/pass2/gpt-5.6-luna-low/events.jsonl rescored (lite_assembly_a, time_partition_seconds.compound); runs/hosted-api-repeats-20261003/pass3/gpt-5.6-luna-low/events.jsonl rescored (lite_assembly_a, time_partition_seconds.compound)
\newcommand{\LunaAssemblyBErrJudgmentSec}{2.7}                          % equal mean across passes: runs/lite-v1-gpt-5.6-luna-low-four-family-retry-v1/events.jsonl rescored (lite_assembly_b, time_partition_seconds.judgment); runs/hosted-api-repeats-20261003/pass2/gpt-5.6-luna-low/events.jsonl rescored (lite_assembly_b, time_partition_seconds.judgment); runs/hosted-api-repeats-20261003/pass3/gpt-5.6-luna-low/events.jsonl rescored (lite_assembly_b, time_partition_seconds.judgment)
\newcommand{\LunaAssemblyBErrCompoundSec}{12.6}                         % equal mean across passes: runs/lite-v1-gpt-5.6-luna-low-four-family-retry-v1/events.jsonl rescored (lite_assembly_b, time_partition_seconds.compound); runs/hosted-api-repeats-20261003/pass2/gpt-5.6-luna-low/events.jsonl rescored (lite_assembly_b, time_partition_seconds.compound); runs/hosted-api-repeats-20261003/pass3/gpt-5.6-luna-low/events.jsonl rescored (lite_assembly_b, time_partition_seconds.compound)
\newcommand{\LunaSupportAErrJudgmentSec}{11.3}                          % equal mean across passes: runs/lite-v1-gpt-5.6-luna-low-four-family-retry-v1/events.jsonl rescored (lite_support_a, time_partition_seconds.judgment); runs/hosted-api-repeats-20261003/pass2/gpt-5.6-luna-low/events.jsonl rescored (lite_support_a, time_partition_seconds.judgment); runs/hosted-api-repeats-20261003/pass3/gpt-5.6-luna-low/events.jsonl rescored (lite_support_a, time_partition_seconds.judgment)
\newcommand{\LunaSupportAErrCompoundSec}{11.4}                          % equal mean across passes: runs/lite-v1-gpt-5.6-luna-low-four-family-retry-v1/events.jsonl rescored (lite_support_a, time_partition_seconds.compound); runs/hosted-api-repeats-20261003/pass2/gpt-5.6-luna-low/events.jsonl rescored (lite_support_a, time_partition_seconds.compound); runs/hosted-api-repeats-20261003/pass3/gpt-5.6-luna-low/events.jsonl rescored (lite_support_a, time_partition_seconds.compound)
\newcommand{\LunaSupportBErrJudgmentSec}{2.9}                           % equal mean across passes: runs/lite-v1-gpt-5.6-luna-low-four-family-retry-v1/events.jsonl rescored (lite_support_b, time_partition_seconds.judgment); runs/hosted-api-repeats-20261003/pass2/gpt-5.6-luna-low/events.jsonl rescored (lite_support_b, time_partition_seconds.judgment); runs/hosted-api-repeats-20261003/pass3/gpt-5.6-luna-low/events.jsonl rescored (lite_support_b, time_partition_seconds.judgment)
\newcommand{\LunaSupportBErrCompoundSec}{6.1}                           % equal mean across passes: runs/lite-v1-gpt-5.6-luna-low-four-family-retry-v1/events.jsonl rescored (lite_support_b, time_partition_seconds.compound); runs/hosted-api-repeats-20261003/pass2/gpt-5.6-luna-low/events.jsonl rescored (lite_support_b, time_partition_seconds.compound); runs/hosted-api-repeats-20261003/pass3/gpt-5.6-luna-low/events.jsonl rescored (lite_support_b, time_partition_seconds.compound)
\newcommand{\LunaPresenterAErrJudgmentSec}{0.5}                         % equal mean across passes: runs/lite-v1-gpt-5.6-luna-low-four-family-retry-v1/events.jsonl rescored (lite_presenter_a, time_partition_seconds.judgment); runs/hosted-api-repeats-20261003/pass2/gpt-5.6-luna-low/events.jsonl rescored (lite_presenter_a, time_partition_seconds.judgment); runs/hosted-api-repeats-20261003/pass3/gpt-5.6-luna-low/events.jsonl rescored (lite_presenter_a, time_partition_seconds.judgment)
\newcommand{\LunaPresenterAErrCompoundSec}{0.2}                         % equal mean across passes: runs/lite-v1-gpt-5.6-luna-low-four-family-retry-v1/events.jsonl rescored (lite_presenter_a, time_partition_seconds.compound); runs/hosted-api-repeats-20261003/pass2/gpt-5.6-luna-low/events.jsonl rescored (lite_presenter_a, time_partition_seconds.compound); runs/hosted-api-repeats-20261003/pass3/gpt-5.6-luna-low/events.jsonl rescored (lite_presenter_a, time_partition_seconds.compound)
\newcommand{\LunaPresenterBErrJudgmentSec}{2.0}                         % equal mean across passes: runs/lite-v1-gpt-5.6-luna-low-four-family-retry-v1/events.jsonl rescored (lite_presenter_b, time_partition_seconds.judgment); runs/hosted-api-repeats-20261003/pass2/gpt-5.6-luna-low/events.jsonl rescored (lite_presenter_b, time_partition_seconds.judgment); runs/hosted-api-repeats-20261003/pass3/gpt-5.6-luna-low/events.jsonl rescored (lite_presenter_b, time_partition_seconds.judgment)
\newcommand{\LunaPresenterBErrCompoundSec}{5.9}                         % equal mean across passes: runs/lite-v1-gpt-5.6-luna-low-four-family-retry-v1/events.jsonl rescored (lite_presenter_b, time_partition_seconds.compound); runs/hosted-api-repeats-20261003/pass2/gpt-5.6-luna-low/events.jsonl rescored (lite_presenter_b, time_partition_seconds.compound); runs/hosted-api-repeats-20261003/pass3/gpt-5.6-luna-low/events.jsonl rescored (lite_presenter_b, time_partition_seconds.compound)
\newcommand{\LunaProduct}{45.3}                                         % equal mean across passes: runs/lite-v1-gpt-5.6-luna-low-four-family-retry-v1/events.jsonl rescored: mean over scenarios of untimed accuracy x oracle in-force accuracy; runs/hosted-api-repeats-20261003/pass2/gpt-5.6-luna-low/events.jsonl rescored: mean over scenarios of untimed accuracy x oracle in-force accuracy; runs/hosted-api-repeats-20261003/pass3/gpt-5.6-luna-low/events.jsonl rescored: mean over scenarios of untimed accuracy x oracle in-force accuracy
\newcommand{\LunaNoneErrJudgmentPct}{41.1}                              % equal mean across passes: runs/lite-v1-gpt-5.6-luna-none-four-family-retry-v1/events.jsonl rescored with lite.scoring (time_partition_seconds, summed over scenarios); each scenario divided by its own observed duration, then equal family mean; runs/hosted-api-repeats-20261003/pass2/gpt-5.6-luna-none/events.jsonl rescored with lite.scoring (time_partition_seconds, summed over scenarios); each scenario divided by its own observed duration, then equal family mean; runs/hosted-api-repeats-20261003/pass3/gpt-5.6-luna-none/events.jsonl rescored with lite.scoring (time_partition_seconds, summed over scenarios); each scenario divided by its own observed duration, then equal family mean
\newcommand{\LunaNoneErrCompoundPct}{13.4}                              % equal mean across passes: runs/lite-v1-gpt-5.6-luna-none-four-family-retry-v1/events.jsonl rescored with lite.scoring (time_partition_seconds, summed over scenarios); each scenario divided by its own observed duration, then equal family mean; runs/hosted-api-repeats-20261003/pass2/gpt-5.6-luna-none/events.jsonl rescored with lite.scoring (time_partition_seconds, summed over scenarios); each scenario divided by its own observed duration, then equal family mean; runs/hosted-api-repeats-20261003/pass3/gpt-5.6-luna-none/events.jsonl rescored with lite.scoring (time_partition_seconds, summed over scenarios); each scenario divided by its own observed duration, then equal family mean
\newcommand{\LunaNoneLuckyPct}{1.1}                                     % equal mean across passes: runs/lite-v1-gpt-5.6-luna-none-four-family-retry-v1/events.jsonl rescored with lite.scoring (time_partition_seconds, summed over scenarios); each scenario divided by its own observed duration, then equal family mean; runs/hosted-api-repeats-20261003/pass2/gpt-5.6-luna-none/events.jsonl rescored with lite.scoring (time_partition_seconds, summed over scenarios); each scenario divided by its own observed duration, then equal family mean; runs/hosted-api-repeats-20261003/pass3/gpt-5.6-luna-none/events.jsonl rescored with lite.scoring (time_partition_seconds, summed over scenarios); each scenario divided by its own observed duration, then equal family mean
\newcommand{\LunaNoneUCurrent}{44.0}                                    % docs/research/manuscript-three-pass/analysis.json:mean current-correct / mean current-source
\newcommand{\LunaNoneDebugAErrJudgmentSec}{78.5}                        % equal mean across passes: runs/lite-v1-gpt-5.6-luna-none-four-family-retry-v1/events.jsonl rescored (lite_debugging_a, time_partition_seconds.judgment); runs/hosted-api-repeats-20261003/pass2/gpt-5.6-luna-none/events.jsonl rescored (lite_debugging_a, time_partition_seconds.judgment); runs/hosted-api-repeats-20261003/pass3/gpt-5.6-luna-none/events.jsonl rescored (lite_debugging_a, time_partition_seconds.judgment)
\newcommand{\LunaNoneDebugAErrCompoundSec}{22.3}                        % equal mean across passes: runs/lite-v1-gpt-5.6-luna-none-four-family-retry-v1/events.jsonl rescored (lite_debugging_a, time_partition_seconds.compound); runs/hosted-api-repeats-20261003/pass2/gpt-5.6-luna-none/events.jsonl rescored (lite_debugging_a, time_partition_seconds.compound); runs/hosted-api-repeats-20261003/pass3/gpt-5.6-luna-none/events.jsonl rescored (lite_debugging_a, time_partition_seconds.compound)
\newcommand{\LunaNoneDebugBErrJudgmentSec}{70.1}                        % equal mean across passes: runs/lite-v1-gpt-5.6-luna-none-four-family-retry-v1/events.jsonl rescored (lite_debugging_b, time_partition_seconds.judgment); runs/hosted-api-repeats-20261003/pass2/gpt-5.6-luna-none/events.jsonl rescored (lite_debugging_b, time_partition_seconds.judgment); runs/hosted-api-repeats-20261003/pass3/gpt-5.6-luna-none/events.jsonl rescored (lite_debugging_b, time_partition_seconds.judgment)
\newcommand{\LunaNoneDebugBErrCompoundSec}{23.6}                        % equal mean across passes: runs/lite-v1-gpt-5.6-luna-none-four-family-retry-v1/events.jsonl rescored (lite_debugging_b, time_partition_seconds.compound); runs/hosted-api-repeats-20261003/pass2/gpt-5.6-luna-none/events.jsonl rescored (lite_debugging_b, time_partition_seconds.compound); runs/hosted-api-repeats-20261003/pass3/gpt-5.6-luna-none/events.jsonl rescored (lite_debugging_b, time_partition_seconds.compound)
\newcommand{\LunaNoneAssemblyAErrJudgmentSec}{63.6}                     % equal mean across passes: runs/lite-v1-gpt-5.6-luna-none-four-family-retry-v1/events.jsonl rescored (lite_assembly_a, time_partition_seconds.judgment); runs/hosted-api-repeats-20261003/pass2/gpt-5.6-luna-none/events.jsonl rescored (lite_assembly_a, time_partition_seconds.judgment); runs/hosted-api-repeats-20261003/pass3/gpt-5.6-luna-none/events.jsonl rescored (lite_assembly_a, time_partition_seconds.judgment)
\newcommand{\LunaNoneAssemblyAErrCompoundSec}{19.4}                     % equal mean across passes: runs/lite-v1-gpt-5.6-luna-none-four-family-retry-v1/events.jsonl rescored (lite_assembly_a, time_partition_seconds.compound); runs/hosted-api-repeats-20261003/pass2/gpt-5.6-luna-none/events.jsonl rescored (lite_assembly_a, time_partition_seconds.compound); runs/hosted-api-repeats-20261003/pass3/gpt-5.6-luna-none/events.jsonl rescored (lite_assembly_a, time_partition_seconds.compound)
\newcommand{\LunaNoneAssemblyBErrJudgmentSec}{63.2}                     % equal mean across passes: runs/lite-v1-gpt-5.6-luna-none-four-family-retry-v1/events.jsonl rescored (lite_assembly_b, time_partition_seconds.judgment); runs/hosted-api-repeats-20261003/pass2/gpt-5.6-luna-none/events.jsonl rescored (lite_assembly_b, time_partition_seconds.judgment); runs/hosted-api-repeats-20261003/pass3/gpt-5.6-luna-none/events.jsonl rescored (lite_assembly_b, time_partition_seconds.judgment)
\newcommand{\LunaNoneAssemblyBErrCompoundSec}{20.4}                     % equal mean across passes: runs/lite-v1-gpt-5.6-luna-none-four-family-retry-v1/events.jsonl rescored (lite_assembly_b, time_partition_seconds.compound); runs/hosted-api-repeats-20261003/pass2/gpt-5.6-luna-none/events.jsonl rescored (lite_assembly_b, time_partition_seconds.compound); runs/hosted-api-repeats-20261003/pass3/gpt-5.6-luna-none/events.jsonl rescored (lite_assembly_b, time_partition_seconds.compound)
\newcommand{\LunaNoneSupportAErrJudgmentSec}{33.9}                      % equal mean across passes: runs/lite-v1-gpt-5.6-luna-none-four-family-retry-v1/events.jsonl rescored (lite_support_a, time_partition_seconds.judgment); runs/hosted-api-repeats-20261003/pass2/gpt-5.6-luna-none/events.jsonl rescored (lite_support_a, time_partition_seconds.judgment); runs/hosted-api-repeats-20261003/pass3/gpt-5.6-luna-none/events.jsonl rescored (lite_support_a, time_partition_seconds.judgment)
\newcommand{\LunaNoneSupportAErrCompoundSec}{15.1}                      % equal mean across passes: runs/lite-v1-gpt-5.6-luna-none-four-family-retry-v1/events.jsonl rescored (lite_support_a, time_partition_seconds.compound); runs/hosted-api-repeats-20261003/pass2/gpt-5.6-luna-none/events.jsonl rescored (lite_support_a, time_partition_seconds.compound); runs/hosted-api-repeats-20261003/pass3/gpt-5.6-luna-none/events.jsonl rescored (lite_support_a, time_partition_seconds.compound)
\newcommand{\LunaNoneSupportBErrJudgmentSec}{23.8}                      % equal mean across passes: runs/lite-v1-gpt-5.6-luna-none-four-family-retry-v1/events.jsonl rescored (lite_support_b, time_partition_seconds.judgment); runs/hosted-api-repeats-20261003/pass2/gpt-5.6-luna-none/events.jsonl rescored (lite_support_b, time_partition_seconds.judgment); runs/hosted-api-repeats-20261003/pass3/gpt-5.6-luna-none/events.jsonl rescored (lite_support_b, time_partition_seconds.judgment)
\newcommand{\LunaNoneSupportBErrCompoundSec}{8.0}                       % equal mean across passes: runs/lite-v1-gpt-5.6-luna-none-four-family-retry-v1/events.jsonl rescored (lite_support_b, time_partition_seconds.compound); runs/hosted-api-repeats-20261003/pass2/gpt-5.6-luna-none/events.jsonl rescored (lite_support_b, time_partition_seconds.compound); runs/hosted-api-repeats-20261003/pass3/gpt-5.6-luna-none/events.jsonl rescored (lite_support_b, time_partition_seconds.compound)
\newcommand{\LunaNonePresenterAErrJudgmentSec}{18.3}                    % equal mean across passes: runs/lite-v1-gpt-5.6-luna-none-four-family-retry-v1/events.jsonl rescored (lite_presenter_a, time_partition_seconds.judgment); runs/hosted-api-repeats-20261003/pass2/gpt-5.6-luna-none/events.jsonl rescored (lite_presenter_a, time_partition_seconds.judgment); runs/hosted-api-repeats-20261003/pass3/gpt-5.6-luna-none/events.jsonl rescored (lite_presenter_a, time_partition_seconds.judgment)
\newcommand{\LunaNonePresenterAErrCompoundSec}{8.5}                     % equal mean across passes: runs/lite-v1-gpt-5.6-luna-none-four-family-retry-v1/events.jsonl rescored (lite_presenter_a, time_partition_seconds.compound); runs/hosted-api-repeats-20261003/pass2/gpt-5.6-luna-none/events.jsonl rescored (lite_presenter_a, time_partition_seconds.compound); runs/hosted-api-repeats-20261003/pass3/gpt-5.6-luna-none/events.jsonl rescored (lite_presenter_a, time_partition_seconds.compound)
\newcommand{\LunaNonePresenterBErrJudgmentSec}{43.5}                    % equal mean across passes: runs/lite-v1-gpt-5.6-luna-none-four-family-retry-v1/events.jsonl rescored (lite_presenter_b, time_partition_seconds.judgment); runs/hosted-api-repeats-20261003/pass2/gpt-5.6-luna-none/events.jsonl rescored (lite_presenter_b, time_partition_seconds.judgment); runs/hosted-api-repeats-20261003/pass3/gpt-5.6-luna-none/events.jsonl rescored (lite_presenter_b, time_partition_seconds.judgment)
\newcommand{\LunaNonePresenterBErrCompoundSec}{11.7}                    % equal mean across passes: runs/lite-v1-gpt-5.6-luna-none-four-family-retry-v1/events.jsonl rescored (lite_presenter_b, time_partition_seconds.compound); runs/hosted-api-repeats-20261003/pass2/gpt-5.6-luna-none/events.jsonl rescored (lite_presenter_b, time_partition_seconds.compound); runs/hosted-api-repeats-20261003/pass3/gpt-5.6-luna-none/events.jsonl rescored (lite_presenter_b, time_partition_seconds.compound)
\newcommand{\LunaNoneProduct}{31.6}                                     % equal mean across passes: runs/lite-v1-gpt-5.6-luna-none-four-family-retry-v1/events.jsonl rescored: mean over scenarios of untimed accuracy x oracle in-force accuracy; runs/hosted-api-repeats-20261003/pass2/gpt-5.6-luna-none/events.jsonl rescored: mean over scenarios of untimed accuracy x oracle in-force accuracy; runs/hosted-api-repeats-20261003/pass3/gpt-5.6-luna-none/events.jsonl rescored: mean over scenarios of untimed accuracy x oracle in-force accuracy
\newcommand{\TerraErrJudgmentPct}{1.5}                                  % equal mean across passes: runs/lite-v1-gpt-5.6-terra-low-four-family-retry-v1/events.jsonl rescored with lite.scoring (time_partition_seconds, summed over scenarios); each scenario divided by its own observed duration, then equal family mean; runs/hosted-api-repeats-20261003/pass2/gpt-5.6-terra-low/events.jsonl rescored with lite.scoring (time_partition_seconds, summed over scenarios); each scenario divided by its own observed duration, then equal family mean; runs/hosted-api-repeats-20261003/pass3/gpt-5.6-terra-low/events.jsonl rescored with lite.scoring (time_partition_seconds, summed over scenarios); each scenario divided by its own observed duration, then equal family mean
\newcommand{\TerraErrCompoundPct}{2.3}                                  % equal mean across passes: runs/lite-v1-gpt-5.6-terra-low-four-family-retry-v1/events.jsonl rescored with lite.scoring (time_partition_seconds, summed over scenarios); each scenario divided by its own observed duration, then equal family mean; runs/hosted-api-repeats-20261003/pass2/gpt-5.6-terra-low/events.jsonl rescored with lite.scoring (time_partition_seconds, summed over scenarios); each scenario divided by its own observed duration, then equal family mean; runs/hosted-api-repeats-20261003/pass3/gpt-5.6-terra-low/events.jsonl rescored with lite.scoring (time_partition_seconds, summed over scenarios); each scenario divided by its own observed duration, then equal family mean
\newcommand{\TerraLuckyPct}{0.4}                                        % equal mean across passes: runs/lite-v1-gpt-5.6-terra-low-four-family-retry-v1/events.jsonl rescored with lite.scoring (time_partition_seconds, summed over scenarios); each scenario divided by its own observed duration, then equal family mean; runs/hosted-api-repeats-20261003/pass2/gpt-5.6-terra-low/events.jsonl rescored with lite.scoring (time_partition_seconds, summed over scenarios); each scenario divided by its own observed duration, then equal family mean; runs/hosted-api-repeats-20261003/pass3/gpt-5.6-terra-low/events.jsonl rescored with lite.scoring (time_partition_seconds, summed over scenarios); each scenario divided by its own observed duration, then equal family mean
\newcommand{\TerraUCurrent}{97.4}                                       % docs/research/manuscript-three-pass/analysis.json:mean current-correct / mean current-source
\newcommand{\TerraDebugAErrJudgmentSec}{0.0}                            % equal mean across passes: runs/lite-v1-gpt-5.6-terra-low-four-family-retry-v1/events.jsonl rescored (lite_debugging_a, time_partition_seconds.judgment); runs/hosted-api-repeats-20261003/pass2/gpt-5.6-terra-low/events.jsonl rescored (lite_debugging_a, time_partition_seconds.judgment); runs/hosted-api-repeats-20261003/pass3/gpt-5.6-terra-low/events.jsonl rescored (lite_debugging_a, time_partition_seconds.judgment)
\newcommand{\TerraDebugAErrCompoundSec}{0.0}                            % equal mean across passes: runs/lite-v1-gpt-5.6-terra-low-four-family-retry-v1/events.jsonl rescored (lite_debugging_a, time_partition_seconds.compound); runs/hosted-api-repeats-20261003/pass2/gpt-5.6-terra-low/events.jsonl rescored (lite_debugging_a, time_partition_seconds.compound); runs/hosted-api-repeats-20261003/pass3/gpt-5.6-terra-low/events.jsonl rescored (lite_debugging_a, time_partition_seconds.compound)
\newcommand{\TerraDebugBErrJudgmentSec}{0.7}                            % equal mean across passes: runs/lite-v1-gpt-5.6-terra-low-four-family-retry-v1/events.jsonl rescored (lite_debugging_b, time_partition_seconds.judgment); runs/hosted-api-repeats-20261003/pass2/gpt-5.6-terra-low/events.jsonl rescored (lite_debugging_b, time_partition_seconds.judgment); runs/hosted-api-repeats-20261003/pass3/gpt-5.6-terra-low/events.jsonl rescored (lite_debugging_b, time_partition_seconds.judgment)
\newcommand{\TerraDebugBErrCompoundSec}{1.3}                            % equal mean across passes: runs/lite-v1-gpt-5.6-terra-low-four-family-retry-v1/events.jsonl rescored (lite_debugging_b, time_partition_seconds.compound); runs/hosted-api-repeats-20261003/pass2/gpt-5.6-terra-low/events.jsonl rescored (lite_debugging_b, time_partition_seconds.compound); runs/hosted-api-repeats-20261003/pass3/gpt-5.6-terra-low/events.jsonl rescored (lite_debugging_b, time_partition_seconds.compound)
\newcommand{\TerraAssemblyAErrJudgmentSec}{2.3}                         % equal mean across passes: runs/lite-v1-gpt-5.6-terra-low-four-family-retry-v1/events.jsonl rescored (lite_assembly_a, time_partition_seconds.judgment); runs/hosted-api-repeats-20261003/pass2/gpt-5.6-terra-low/events.jsonl rescored (lite_assembly_a, time_partition_seconds.judgment); runs/hosted-api-repeats-20261003/pass3/gpt-5.6-terra-low/events.jsonl rescored (lite_assembly_a, time_partition_seconds.judgment)
\newcommand{\TerraAssemblyAErrCompoundSec}{3.1}                         % equal mean across passes: runs/lite-v1-gpt-5.6-terra-low-four-family-retry-v1/events.jsonl rescored (lite_assembly_a, time_partition_seconds.compound); runs/hosted-api-repeats-20261003/pass2/gpt-5.6-terra-low/events.jsonl rescored (lite_assembly_a, time_partition_seconds.compound); runs/hosted-api-repeats-20261003/pass3/gpt-5.6-terra-low/events.jsonl rescored (lite_assembly_a, time_partition_seconds.compound)
\newcommand{\TerraAssemblyBErrJudgmentSec}{2.9}                         % equal mean across passes: runs/lite-v1-gpt-5.6-terra-low-four-family-retry-v1/events.jsonl rescored (lite_assembly_b, time_partition_seconds.judgment); runs/hosted-api-repeats-20261003/pass2/gpt-5.6-terra-low/events.jsonl rescored (lite_assembly_b, time_partition_seconds.judgment); runs/hosted-api-repeats-20261003/pass3/gpt-5.6-terra-low/events.jsonl rescored (lite_assembly_b, time_partition_seconds.judgment)
\newcommand{\TerraAssemblyBErrCompoundSec}{8.8}                         % equal mean across passes: runs/lite-v1-gpt-5.6-terra-low-four-family-retry-v1/events.jsonl rescored (lite_assembly_b, time_partition_seconds.compound); runs/hosted-api-repeats-20261003/pass2/gpt-5.6-terra-low/events.jsonl rescored (lite_assembly_b, time_partition_seconds.compound); runs/hosted-api-repeats-20261003/pass3/gpt-5.6-terra-low/events.jsonl rescored (lite_assembly_b, time_partition_seconds.compound)
\newcommand{\TerraSupportAErrJudgmentSec}{1.8}                          % equal mean across passes: runs/lite-v1-gpt-5.6-terra-low-four-family-retry-v1/events.jsonl rescored (lite_support_a, time_partition_seconds.judgment); runs/hosted-api-repeats-20261003/pass2/gpt-5.6-terra-low/events.jsonl rescored (lite_support_a, time_partition_seconds.judgment); runs/hosted-api-repeats-20261003/pass3/gpt-5.6-terra-low/events.jsonl rescored (lite_support_a, time_partition_seconds.judgment)
\newcommand{\TerraSupportAErrCompoundSec}{0.6}                          % equal mean across passes: runs/lite-v1-gpt-5.6-terra-low-four-family-retry-v1/events.jsonl rescored (lite_support_a, time_partition_seconds.compound); runs/hosted-api-repeats-20261003/pass2/gpt-5.6-terra-low/events.jsonl rescored (lite_support_a, time_partition_seconds.compound); runs/hosted-api-repeats-20261003/pass3/gpt-5.6-terra-low/events.jsonl rescored (lite_support_a, time_partition_seconds.compound)
\newcommand{\TerraSupportBErrJudgmentSec}{2.4}                          % equal mean across passes: runs/lite-v1-gpt-5.6-terra-low-four-family-retry-v1/events.jsonl rescored (lite_support_b, time_partition_seconds.judgment); runs/hosted-api-repeats-20261003/pass2/gpt-5.6-terra-low/events.jsonl rescored (lite_support_b, time_partition_seconds.judgment); runs/hosted-api-repeats-20261003/pass3/gpt-5.6-terra-low/events.jsonl rescored (lite_support_b, time_partition_seconds.judgment)
\newcommand{\TerraSupportBErrCompoundSec}{6.1}                          % equal mean across passes: runs/lite-v1-gpt-5.6-terra-low-four-family-retry-v1/events.jsonl rescored (lite_support_b, time_partition_seconds.compound); runs/hosted-api-repeats-20261003/pass2/gpt-5.6-terra-low/events.jsonl rescored (lite_support_b, time_partition_seconds.compound); runs/hosted-api-repeats-20261003/pass3/gpt-5.6-terra-low/events.jsonl rescored (lite_support_b, time_partition_seconds.compound)
\newcommand{\TerraPresenterAErrJudgmentSec}{1.4}                        % equal mean across passes: runs/lite-v1-gpt-5.6-terra-low-four-family-retry-v1/events.jsonl rescored (lite_presenter_a, time_partition_seconds.judgment); runs/hosted-api-repeats-20261003/pass2/gpt-5.6-terra-low/events.jsonl rescored (lite_presenter_a, time_partition_seconds.judgment); runs/hosted-api-repeats-20261003/pass3/gpt-5.6-terra-low/events.jsonl rescored (lite_presenter_a, time_partition_seconds.judgment)
\newcommand{\TerraPresenterAErrCompoundSec}{0.2}                        % equal mean across passes: runs/lite-v1-gpt-5.6-terra-low-four-family-retry-v1/events.jsonl rescored (lite_presenter_a, time_partition_seconds.compound); runs/hosted-api-repeats-20261003/pass2/gpt-5.6-terra-low/events.jsonl rescored (lite_presenter_a, time_partition_seconds.compound); runs/hosted-api-repeats-20261003/pass3/gpt-5.6-terra-low/events.jsonl rescored (lite_presenter_a, time_partition_seconds.compound)
\newcommand{\TerraPresenterBErrJudgmentSec}{2.7}                        % equal mean across passes: runs/lite-v1-gpt-5.6-terra-low-four-family-retry-v1/events.jsonl rescored (lite_presenter_b, time_partition_seconds.judgment); runs/hosted-api-repeats-20261003/pass2/gpt-5.6-terra-low/events.jsonl rescored (lite_presenter_b, time_partition_seconds.judgment); runs/hosted-api-repeats-20261003/pass3/gpt-5.6-terra-low/events.jsonl rescored (lite_presenter_b, time_partition_seconds.judgment)
\newcommand{\TerraPresenterBErrCompoundSec}{2.2}                        % equal mean across passes: runs/lite-v1-gpt-5.6-terra-low-four-family-retry-v1/events.jsonl rescored (lite_presenter_b, time_partition_seconds.compound); runs/hosted-api-repeats-20261003/pass2/gpt-5.6-terra-low/events.jsonl rescored (lite_presenter_b, time_partition_seconds.compound); runs/hosted-api-repeats-20261003/pass3/gpt-5.6-terra-low/events.jsonl rescored (lite_presenter_b, time_partition_seconds.compound)
\newcommand{\TerraProduct}{54.7}                                        % equal mean across passes: runs/lite-v1-gpt-5.6-terra-low-four-family-retry-v1/events.jsonl rescored: mean over scenarios of untimed accuracy x oracle in-force accuracy; runs/hosted-api-repeats-20261003/pass2/gpt-5.6-terra-low/events.jsonl rescored: mean over scenarios of untimed accuracy x oracle in-force accuracy; runs/hosted-api-repeats-20261003/pass3/gpt-5.6-terra-low/events.jsonl rescored: mean over scenarios of untimed accuracy x oracle in-force accuracy
\newcommand{\TerraNoneErrJudgmentPct}{11.7}                             % equal mean across passes: runs/lite-v1-gpt-5.6-terra-none-four-family-retry-v1/events.jsonl rescored with lite.scoring (time_partition_seconds, summed over scenarios); each scenario divided by its own observed duration, then equal family mean; runs/hosted-api-repeats-20261003/pass2/gpt-5.6-terra-none/events.jsonl rescored with lite.scoring (time_partition_seconds, summed over scenarios); each scenario divided by its own observed duration, then equal family mean; runs/hosted-api-repeats-20261003/pass3/gpt-5.6-terra-none/events.jsonl rescored with lite.scoring (time_partition_seconds, summed over scenarios); each scenario divided by its own observed duration, then equal family mean
\newcommand{\TerraNoneErrCompoundPct}{4.6}                              % equal mean across passes: runs/lite-v1-gpt-5.6-terra-none-four-family-retry-v1/events.jsonl rescored with lite.scoring (time_partition_seconds, summed over scenarios); each scenario divided by its own observed duration, then equal family mean; runs/hosted-api-repeats-20261003/pass2/gpt-5.6-terra-none/events.jsonl rescored with lite.scoring (time_partition_seconds, summed over scenarios); each scenario divided by its own observed duration, then equal family mean; runs/hosted-api-repeats-20261003/pass3/gpt-5.6-terra-none/events.jsonl rescored with lite.scoring (time_partition_seconds, summed over scenarios); each scenario divided by its own observed duration, then equal family mean
\newcommand{\TerraNoneLuckyPct}{1.2}                                    % equal mean across passes: runs/lite-v1-gpt-5.6-terra-none-four-family-retry-v1/events.jsonl rescored with lite.scoring (time_partition_seconds, summed over scenarios); each scenario divided by its own observed duration, then equal family mean; runs/hosted-api-repeats-20261003/pass2/gpt-5.6-terra-none/events.jsonl rescored with lite.scoring (time_partition_seconds, summed over scenarios); each scenario divided by its own observed duration, then equal family mean; runs/hosted-api-repeats-20261003/pass3/gpt-5.6-terra-none/events.jsonl rescored with lite.scoring (time_partition_seconds, summed over scenarios); each scenario divided by its own observed duration, then equal family mean
\newcommand{\TerraNoneUCurrent}{83.7}                                   % docs/research/manuscript-three-pass/analysis.json:mean current-correct / mean current-source
\newcommand{\TerraNoneDebugAErrJudgmentSec}{17.0}                       % equal mean across passes: runs/lite-v1-gpt-5.6-terra-none-four-family-retry-v1/events.jsonl rescored (lite_debugging_a, time_partition_seconds.judgment); runs/hosted-api-repeats-20261003/pass2/gpt-5.6-terra-none/events.jsonl rescored (lite_debugging_a, time_partition_seconds.judgment); runs/hosted-api-repeats-20261003/pass3/gpt-5.6-terra-none/events.jsonl rescored (lite_debugging_a, time_partition_seconds.judgment)
\newcommand{\TerraNoneDebugAErrCompoundSec}{9.1}                        % equal mean across passes: runs/lite-v1-gpt-5.6-terra-none-four-family-retry-v1/events.jsonl rescored (lite_debugging_a, time_partition_seconds.compound); runs/hosted-api-repeats-20261003/pass2/gpt-5.6-terra-none/events.jsonl rescored (lite_debugging_a, time_partition_seconds.compound); runs/hosted-api-repeats-20261003/pass3/gpt-5.6-terra-none/events.jsonl rescored (lite_debugging_a, time_partition_seconds.compound)
\newcommand{\TerraNoneDebugBErrJudgmentSec}{24.9}                       % equal mean across passes: runs/lite-v1-gpt-5.6-terra-none-four-family-retry-v1/events.jsonl rescored (lite_debugging_b, time_partition_seconds.judgment); runs/hosted-api-repeats-20261003/pass2/gpt-5.6-terra-none/events.jsonl rescored (lite_debugging_b, time_partition_seconds.judgment); runs/hosted-api-repeats-20261003/pass3/gpt-5.6-terra-none/events.jsonl rescored (lite_debugging_b, time_partition_seconds.judgment)
\newcommand{\TerraNoneDebugBErrCompoundSec}{11.6}                       % equal mean across passes: runs/lite-v1-gpt-5.6-terra-none-four-family-retry-v1/events.jsonl rescored (lite_debugging_b, time_partition_seconds.compound); runs/hosted-api-repeats-20261003/pass2/gpt-5.6-terra-none/events.jsonl rescored (lite_debugging_b, time_partition_seconds.compound); runs/hosted-api-repeats-20261003/pass3/gpt-5.6-terra-none/events.jsonl rescored (lite_debugging_b, time_partition_seconds.compound)
\newcommand{\TerraNoneAssemblyAErrJudgmentSec}{17.8}                    % equal mean across passes: runs/lite-v1-gpt-5.6-terra-none-four-family-retry-v1/events.jsonl rescored (lite_assembly_a, time_partition_seconds.judgment); runs/hosted-api-repeats-20261003/pass2/gpt-5.6-terra-none/events.jsonl rescored (lite_assembly_a, time_partition_seconds.judgment); runs/hosted-api-repeats-20261003/pass3/gpt-5.6-terra-none/events.jsonl rescored (lite_assembly_a, time_partition_seconds.judgment)
\newcommand{\TerraNoneAssemblyAErrCompoundSec}{7.7}                     % equal mean across passes: runs/lite-v1-gpt-5.6-terra-none-four-family-retry-v1/events.jsonl rescored (lite_assembly_a, time_partition_seconds.compound); runs/hosted-api-repeats-20261003/pass2/gpt-5.6-terra-none/events.jsonl rescored (lite_assembly_a, time_partition_seconds.compound); runs/hosted-api-repeats-20261003/pass3/gpt-5.6-terra-none/events.jsonl rescored (lite_assembly_a, time_partition_seconds.compound)
\newcommand{\TerraNoneAssemblyBErrJudgmentSec}{23.9}                    % equal mean across passes: runs/lite-v1-gpt-5.6-terra-none-four-family-retry-v1/events.jsonl rescored (lite_assembly_b, time_partition_seconds.judgment); runs/hosted-api-repeats-20261003/pass2/gpt-5.6-terra-none/events.jsonl rescored (lite_assembly_b, time_partition_seconds.judgment); runs/hosted-api-repeats-20261003/pass3/gpt-5.6-terra-none/events.jsonl rescored (lite_assembly_b, time_partition_seconds.judgment)
\newcommand{\TerraNoneAssemblyBErrCompoundSec}{8.0}                     % equal mean across passes: runs/lite-v1-gpt-5.6-terra-none-four-family-retry-v1/events.jsonl rescored (lite_assembly_b, time_partition_seconds.compound); runs/hosted-api-repeats-20261003/pass2/gpt-5.6-terra-none/events.jsonl rescored (lite_assembly_b, time_partition_seconds.compound); runs/hosted-api-repeats-20261003/pass3/gpt-5.6-terra-none/events.jsonl rescored (lite_assembly_b, time_partition_seconds.compound)
\newcommand{\TerraNoneSupportAErrJudgmentSec}{5.5}                      % equal mean across passes: runs/lite-v1-gpt-5.6-terra-none-four-family-retry-v1/events.jsonl rescored (lite_support_a, time_partition_seconds.judgment); runs/hosted-api-repeats-20261003/pass2/gpt-5.6-terra-none/events.jsonl rescored (lite_support_a, time_partition_seconds.judgment); runs/hosted-api-repeats-20261003/pass3/gpt-5.6-terra-none/events.jsonl rescored (lite_support_a, time_partition_seconds.judgment)
\newcommand{\TerraNoneSupportAErrCompoundSec}{4.1}                      % equal mean across passes: runs/lite-v1-gpt-5.6-terra-none-four-family-retry-v1/events.jsonl rescored (lite_support_a, time_partition_seconds.compound); runs/hosted-api-repeats-20261003/pass2/gpt-5.6-terra-none/events.jsonl rescored (lite_support_a, time_partition_seconds.compound); runs/hosted-api-repeats-20261003/pass3/gpt-5.6-terra-none/events.jsonl rescored (lite_support_a, time_partition_seconds.compound)
\newcommand{\TerraNoneSupportBErrJudgmentSec}{4.2}                      % equal mean across passes: runs/lite-v1-gpt-5.6-terra-none-four-family-retry-v1/events.jsonl rescored (lite_support_b, time_partition_seconds.judgment); runs/hosted-api-repeats-20261003/pass2/gpt-5.6-terra-none/events.jsonl rescored (lite_support_b, time_partition_seconds.judgment); runs/hosted-api-repeats-20261003/pass3/gpt-5.6-terra-none/events.jsonl rescored (lite_support_b, time_partition_seconds.judgment)
\newcommand{\TerraNoneSupportBErrCompoundSec}{0.5}                      % equal mean across passes: runs/lite-v1-gpt-5.6-terra-none-four-family-retry-v1/events.jsonl rescored (lite_support_b, time_partition_seconds.compound); runs/hosted-api-repeats-20261003/pass2/gpt-5.6-terra-none/events.jsonl rescored (lite_support_b, time_partition_seconds.compound); runs/hosted-api-repeats-20261003/pass3/gpt-5.6-terra-none/events.jsonl rescored (lite_support_b, time_partition_seconds.compound)
\newcommand{\TerraNonePresenterAErrJudgmentSec}{5.6}                    % equal mean across passes: runs/lite-v1-gpt-5.6-terra-none-four-family-retry-v1/events.jsonl rescored (lite_presenter_a, time_partition_seconds.judgment); runs/hosted-api-repeats-20261003/pass2/gpt-5.6-terra-none/events.jsonl rescored (lite_presenter_a, time_partition_seconds.judgment); runs/hosted-api-repeats-20261003/pass3/gpt-5.6-terra-none/events.jsonl rescored (lite_presenter_a, time_partition_seconds.judgment)
\newcommand{\TerraNonePresenterAErrCompoundSec}{1.7}                    % equal mean across passes: runs/lite-v1-gpt-5.6-terra-none-four-family-retry-v1/events.jsonl rescored (lite_presenter_a, time_partition_seconds.compound); runs/hosted-api-repeats-20261003/pass2/gpt-5.6-terra-none/events.jsonl rescored (lite_presenter_a, time_partition_seconds.compound); runs/hosted-api-repeats-20261003/pass3/gpt-5.6-terra-none/events.jsonl rescored (lite_presenter_a, time_partition_seconds.compound)
\newcommand{\TerraNonePresenterBErrJudgmentSec}{13.3}                   % equal mean across passes: runs/lite-v1-gpt-5.6-terra-none-four-family-retry-v1/events.jsonl rescored (lite_presenter_b, time_partition_seconds.judgment); runs/hosted-api-repeats-20261003/pass2/gpt-5.6-terra-none/events.jsonl rescored (lite_presenter_b, time_partition_seconds.judgment); runs/hosted-api-repeats-20261003/pass3/gpt-5.6-terra-none/events.jsonl rescored (lite_presenter_b, time_partition_seconds.judgment)
\newcommand{\TerraNonePresenterBErrCompoundSec}{1.4}                    % equal mean across passes: runs/lite-v1-gpt-5.6-terra-none-four-family-retry-v1/events.jsonl rescored (lite_presenter_b, time_partition_seconds.compound); runs/hosted-api-repeats-20261003/pass2/gpt-5.6-terra-none/events.jsonl rescored (lite_presenter_b, time_partition_seconds.compound); runs/hosted-api-repeats-20261003/pass3/gpt-5.6-terra-none/events.jsonl rescored (lite_presenter_b, time_partition_seconds.compound)
\newcommand{\TerraNoneProduct}{58.7}                                    % equal mean across passes: runs/lite-v1-gpt-5.6-terra-none-four-family-retry-v1/events.jsonl rescored: mean over scenarios of untimed accuracy x oracle in-force accuracy; runs/hosted-api-repeats-20261003/pass2/gpt-5.6-terra-none/events.jsonl rescored: mean over scenarios of untimed accuracy x oracle in-force accuracy; runs/hosted-api-repeats-20261003/pass3/gpt-5.6-terra-none/events.jsonl rescored: mean over scenarios of untimed accuracy x oracle in-force accuracy
\newcommand{\AstraErrJudgmentPct}{0.0}                                  % equal mean across passes: runs/lite-v1-gpt-6-astra-low-four-family-retry-v1/events.jsonl rescored with lite.scoring (time_partition_seconds, summed over scenarios); each scenario divided by its own observed duration, then equal family mean; runs/hosted-api-repeats-20261003/pass2/gpt-6-astra-low/events.jsonl rescored with lite.scoring (time_partition_seconds, summed over scenarios); each scenario divided by its own observed duration, then equal family mean; runs/hosted-api-repeats-20261003/pass3/gpt-6-astra-low/events.jsonl rescored with lite.scoring (time_partition_seconds, summed over scenarios); each scenario divided by its own observed duration, then equal family mean
\newcommand{\AstraErrCompoundPct}{0.1}                                  % equal mean across passes: runs/lite-v1-gpt-6-astra-low-four-family-retry-v1/events.jsonl rescored with lite.scoring (time_partition_seconds, summed over scenarios); each scenario divided by its own observed duration, then equal family mean; runs/hosted-api-repeats-20261003/pass2/gpt-6-astra-low/events.jsonl rescored with lite.scoring (time_partition_seconds, summed over scenarios); each scenario divided by its own observed duration, then equal family mean; runs/hosted-api-repeats-20261003/pass3/gpt-6-astra-low/events.jsonl rescored with lite.scoring (time_partition_seconds, summed over scenarios); each scenario divided by its own observed duration, then equal family mean
\newcommand{\AstraLuckyPct}{0.0}                                        % equal mean across passes: runs/lite-v1-gpt-6-astra-low-four-family-retry-v1/events.jsonl rescored with lite.scoring (time_partition_seconds, summed over scenarios); each scenario divided by its own observed duration, then equal family mean; runs/hosted-api-repeats-20261003/pass2/gpt-6-astra-low/events.jsonl rescored with lite.scoring (time_partition_seconds, summed over scenarios); each scenario divided by its own observed duration, then equal family mean; runs/hosted-api-repeats-20261003/pass3/gpt-6-astra-low/events.jsonl rescored with lite.scoring (time_partition_seconds, summed over scenarios); each scenario divided by its own observed duration, then equal family mean
\newcommand{\AstraUCurrent}{100.0}                                      % docs/research/manuscript-three-pass/analysis.json:mean current-correct / mean current-source
\newcommand{\AstraDebugAErrJudgmentSec}{0.0}                            % equal mean across passes: runs/lite-v1-gpt-6-astra-low-four-family-retry-v1/events.jsonl rescored (lite_debugging_a, time_partition_seconds.judgment); runs/hosted-api-repeats-20261003/pass2/gpt-6-astra-low/events.jsonl rescored (lite_debugging_a, time_partition_seconds.judgment); runs/hosted-api-repeats-20261003/pass3/gpt-6-astra-low/events.jsonl rescored (lite_debugging_a, time_partition_seconds.judgment)
\newcommand{\AstraDebugAErrCompoundSec}{0.0}                            % equal mean across passes: runs/lite-v1-gpt-6-astra-low-four-family-retry-v1/events.jsonl rescored (lite_debugging_a, time_partition_seconds.compound); runs/hosted-api-repeats-20261003/pass2/gpt-6-astra-low/events.jsonl rescored (lite_debugging_a, time_partition_seconds.compound); runs/hosted-api-repeats-20261003/pass3/gpt-6-astra-low/events.jsonl rescored (lite_debugging_a, time_partition_seconds.compound)
\newcommand{\AstraDebugBErrJudgmentSec}{0.0}                            % equal mean across passes: runs/lite-v1-gpt-6-astra-low-four-family-retry-v1/events.jsonl rescored (lite_debugging_b, time_partition_seconds.judgment); runs/hosted-api-repeats-20261003/pass2/gpt-6-astra-low/events.jsonl rescored (lite_debugging_b, time_partition_seconds.judgment); runs/hosted-api-repeats-20261003/pass3/gpt-6-astra-low/events.jsonl rescored (lite_debugging_b, time_partition_seconds.judgment)
\newcommand{\AstraDebugBErrCompoundSec}{0.0}                            % equal mean across passes: runs/lite-v1-gpt-6-astra-low-four-family-retry-v1/events.jsonl rescored (lite_debugging_b, time_partition_seconds.compound); runs/hosted-api-repeats-20261003/pass2/gpt-6-astra-low/events.jsonl rescored (lite_debugging_b, time_partition_seconds.compound); runs/hosted-api-repeats-20261003/pass3/gpt-6-astra-low/events.jsonl rescored (lite_debugging_b, time_partition_seconds.compound)
\newcommand{\AstraAssemblyAErrJudgmentSec}{0.0}                         % equal mean across passes: runs/lite-v1-gpt-6-astra-low-four-family-retry-v1/events.jsonl rescored (lite_assembly_a, time_partition_seconds.judgment); runs/hosted-api-repeats-20261003/pass2/gpt-6-astra-low/events.jsonl rescored (lite_assembly_a, time_partition_seconds.judgment); runs/hosted-api-repeats-20261003/pass3/gpt-6-astra-low/events.jsonl rescored (lite_assembly_a, time_partition_seconds.judgment)
\newcommand{\AstraAssemblyAErrCompoundSec}{0.7}                         % equal mean across passes: runs/lite-v1-gpt-6-astra-low-four-family-retry-v1/events.jsonl rescored (lite_assembly_a, time_partition_seconds.compound); runs/hosted-api-repeats-20261003/pass2/gpt-6-astra-low/events.jsonl rescored (lite_assembly_a, time_partition_seconds.compound); runs/hosted-api-repeats-20261003/pass3/gpt-6-astra-low/events.jsonl rescored (lite_assembly_a, time_partition_seconds.compound)
\newcommand{\AstraAssemblyBErrJudgmentSec}{0.0}                         % equal mean across passes: runs/lite-v1-gpt-6-astra-low-four-family-retry-v1/events.jsonl rescored (lite_assembly_b, time_partition_seconds.judgment); runs/hosted-api-repeats-20261003/pass2/gpt-6-astra-low/events.jsonl rescored (lite_assembly_b, time_partition_seconds.judgment); runs/hosted-api-repeats-20261003/pass3/gpt-6-astra-low/events.jsonl rescored (lite_assembly_b, time_partition_seconds.judgment)
\newcommand{\AstraAssemblyBErrCompoundSec}{0.0}                         % equal mean across passes: runs/lite-v1-gpt-6-astra-low-four-family-retry-v1/events.jsonl rescored (lite_assembly_b, time_partition_seconds.compound); runs/hosted-api-repeats-20261003/pass2/gpt-6-astra-low/events.jsonl rescored (lite_assembly_b, time_partition_seconds.compound); runs/hosted-api-repeats-20261003/pass3/gpt-6-astra-low/events.jsonl rescored (lite_assembly_b, time_partition_seconds.compound)
\newcommand{\AstraSupportAErrJudgmentSec}{0.0}                          % equal mean across passes: runs/lite-v1-gpt-6-astra-low-four-family-retry-v1/events.jsonl rescored (lite_support_a, time_partition_seconds.judgment); runs/hosted-api-repeats-20261003/pass2/gpt-6-astra-low/events.jsonl rescored (lite_support_a, time_partition_seconds.judgment); runs/hosted-api-repeats-20261003/pass3/gpt-6-astra-low/events.jsonl rescored (lite_support_a, time_partition_seconds.judgment)
\newcommand{\AstraSupportAErrCompoundSec}{0.0}                          % equal mean across passes: runs/lite-v1-gpt-6-astra-low-four-family-retry-v1/events.jsonl rescored (lite_support_a, time_partition_seconds.compound); runs/hosted-api-repeats-20261003/pass2/gpt-6-astra-low/events.jsonl rescored (lite_support_a, time_partition_seconds.compound); runs/hosted-api-repeats-20261003/pass3/gpt-6-astra-low/events.jsonl rescored (lite_support_a, time_partition_seconds.compound)
\newcommand{\AstraSupportBErrJudgmentSec}{0.0}                          % equal mean across passes: runs/lite-v1-gpt-6-astra-low-four-family-retry-v1/events.jsonl rescored (lite_support_b, time_partition_seconds.judgment); runs/hosted-api-repeats-20261003/pass2/gpt-6-astra-low/events.jsonl rescored (lite_support_b, time_partition_seconds.judgment); runs/hosted-api-repeats-20261003/pass3/gpt-6-astra-low/events.jsonl rescored (lite_support_b, time_partition_seconds.judgment)
\newcommand{\AstraSupportBErrCompoundSec}{0.0}                          % equal mean across passes: runs/lite-v1-gpt-6-astra-low-four-family-retry-v1/events.jsonl rescored (lite_support_b, time_partition_seconds.compound); runs/hosted-api-repeats-20261003/pass2/gpt-6-astra-low/events.jsonl rescored (lite_support_b, time_partition_seconds.compound); runs/hosted-api-repeats-20261003/pass3/gpt-6-astra-low/events.jsonl rescored (lite_support_b, time_partition_seconds.compound)
\newcommand{\AstraPresenterAErrJudgmentSec}{0.0}                        % equal mean across passes: runs/lite-v1-gpt-6-astra-low-four-family-retry-v1/events.jsonl rescored (lite_presenter_a, time_partition_seconds.judgment); runs/hosted-api-repeats-20261003/pass2/gpt-6-astra-low/events.jsonl rescored (lite_presenter_a, time_partition_seconds.judgment); runs/hosted-api-repeats-20261003/pass3/gpt-6-astra-low/events.jsonl rescored (lite_presenter_a, time_partition_seconds.judgment)
\newcommand{\AstraPresenterAErrCompoundSec}{0.0}                        % equal mean across passes: runs/lite-v1-gpt-6-astra-low-four-family-retry-v1/events.jsonl rescored (lite_presenter_a, time_partition_seconds.compound); runs/hosted-api-repeats-20261003/pass2/gpt-6-astra-low/events.jsonl rescored (lite_presenter_a, time_partition_seconds.compound); runs/hosted-api-repeats-20261003/pass3/gpt-6-astra-low/events.jsonl rescored (lite_presenter_a, time_partition_seconds.compound)
\newcommand{\AstraPresenterBErrJudgmentSec}{0.0}                        % equal mean across passes: runs/lite-v1-gpt-6-astra-low-four-family-retry-v1/events.jsonl rescored (lite_presenter_b, time_partition_seconds.judgment); runs/hosted-api-repeats-20261003/pass2/gpt-6-astra-low/events.jsonl rescored (lite_presenter_b, time_partition_seconds.judgment); runs/hosted-api-repeats-20261003/pass3/gpt-6-astra-low/events.jsonl rescored (lite_presenter_b, time_partition_seconds.judgment)
\newcommand{\AstraPresenterBErrCompoundSec}{0.0}                        % equal mean across passes: runs/lite-v1-gpt-6-astra-low-four-family-retry-v1/events.jsonl rescored (lite_presenter_b, time_partition_seconds.compound); runs/hosted-api-repeats-20261003/pass2/gpt-6-astra-low/events.jsonl rescored (lite_presenter_b, time_partition_seconds.compound); runs/hosted-api-repeats-20261003/pass3/gpt-6-astra-low/events.jsonl rescored (lite_presenter_b, time_partition_seconds.compound)
\newcommand{\AstraProduct}{52.0}                                        % equal mean across passes: runs/lite-v1-gpt-6-astra-low-four-family-retry-v1/events.jsonl rescored: mean over scenarios of untimed accuracy x oracle in-force accuracy; runs/hosted-api-repeats-20261003/pass2/gpt-6-astra-low/events.jsonl rescored: mean over scenarios of untimed accuracy x oracle in-force accuracy; runs/hosted-api-repeats-20261003/pass3/gpt-6-astra-low/events.jsonl rescored: mean over scenarios of untimed accuracy x oracle in-force accuracy
\newcommand{\JevErrJudgmentPct}{35.8}                                   % equal mean across passes: runs/lite-v1-jev-latest-four-family-retry-v1/events.jsonl rescored with lite.scoring (time_partition_seconds, summed over scenarios); each scenario divided by its own observed duration, then equal family mean; runs/hosted-api-repeats-20261003/pass2/jev-latest/events.jsonl rescored with lite.scoring (time_partition_seconds, summed over scenarios); each scenario divided by its own observed duration, then equal family mean; runs/hosted-api-repeats-20261003/pass3/jev-latest/events.jsonl rescored with lite.scoring (time_partition_seconds, summed over scenarios); each scenario divided by its own observed duration, then equal family mean
\newcommand{\JevErrCompoundPct}{1.6}                                    % equal mean across passes: runs/lite-v1-jev-latest-four-family-retry-v1/events.jsonl rescored with lite.scoring (time_partition_seconds, summed over scenarios); each scenario divided by its own observed duration, then equal family mean; runs/hosted-api-repeats-20261003/pass2/jev-latest/events.jsonl rescored with lite.scoring (time_partition_seconds, summed over scenarios); each scenario divided by its own observed duration, then equal family mean; runs/hosted-api-repeats-20261003/pass3/jev-latest/events.jsonl rescored with lite.scoring (time_partition_seconds, summed over scenarios); each scenario divided by its own observed duration, then equal family mean
\newcommand{\JevLuckyPct}{0.2}                                          % equal mean across passes: runs/lite-v1-jev-latest-four-family-retry-v1/events.jsonl rescored with lite.scoring (time_partition_seconds, summed over scenarios); each scenario divided by its own observed duration, then equal family mean; runs/hosted-api-repeats-20261003/pass2/jev-latest/events.jsonl rescored with lite.scoring (time_partition_seconds, summed over scenarios); each scenario divided by its own observed duration, then equal family mean; runs/hosted-api-repeats-20261003/pass3/jev-latest/events.jsonl rescored with lite.scoring (time_partition_seconds, summed over scenarios); each scenario divided by its own observed duration, then equal family mean
\newcommand{\JevUCurrent}{62.4}                                         % docs/research/manuscript-three-pass/analysis.json:mean current-correct / mean current-source
\newcommand{\JevDebugAErrJudgmentSec}{66.3}                             % equal mean across passes: runs/lite-v1-jev-latest-four-family-retry-v1/events.jsonl rescored (lite_debugging_a, time_partition_seconds.judgment); runs/hosted-api-repeats-20261003/pass2/jev-latest/events.jsonl rescored (lite_debugging_a, time_partition_seconds.judgment); runs/hosted-api-repeats-20261003/pass3/jev-latest/events.jsonl rescored (lite_debugging_a, time_partition_seconds.judgment)
\newcommand{\JevDebugAErrCompoundSec}{2.8}                              % equal mean across passes: runs/lite-v1-jev-latest-four-family-retry-v1/events.jsonl rescored (lite_debugging_a, time_partition_seconds.compound); runs/hosted-api-repeats-20261003/pass2/jev-latest/events.jsonl rescored (lite_debugging_a, time_partition_seconds.compound); runs/hosted-api-repeats-20261003/pass3/jev-latest/events.jsonl rescored (lite_debugging_a, time_partition_seconds.compound)
\newcommand{\JevDebugBErrJudgmentSec}{68.9}                             % equal mean across passes: runs/lite-v1-jev-latest-four-family-retry-v1/events.jsonl rescored (lite_debugging_b, time_partition_seconds.judgment); runs/hosted-api-repeats-20261003/pass2/jev-latest/events.jsonl rescored (lite_debugging_b, time_partition_seconds.judgment); runs/hosted-api-repeats-20261003/pass3/jev-latest/events.jsonl rescored (lite_debugging_b, time_partition_seconds.judgment)
\newcommand{\JevDebugBErrCompoundSec}{2.6}                              % equal mean across passes: runs/lite-v1-jev-latest-four-family-retry-v1/events.jsonl rescored (lite_debugging_b, time_partition_seconds.compound); runs/hosted-api-repeats-20261003/pass2/jev-latest/events.jsonl rescored (lite_debugging_b, time_partition_seconds.compound); runs/hosted-api-repeats-20261003/pass3/jev-latest/events.jsonl rescored (lite_debugging_b, time_partition_seconds.compound)
\newcommand{\JevAssemblyAErrJudgmentSec}{35.9}                          % equal mean across passes: runs/lite-v1-jev-latest-four-family-retry-v1/events.jsonl rescored (lite_assembly_a, time_partition_seconds.judgment); runs/hosted-api-repeats-20261003/pass2/jev-latest/events.jsonl rescored (lite_assembly_a, time_partition_seconds.judgment); runs/hosted-api-repeats-20261003/pass3/jev-latest/events.jsonl rescored (lite_assembly_a, time_partition_seconds.judgment)
\newcommand{\JevAssemblyAErrCompoundSec}{2.1}                           % equal mean across passes: runs/lite-v1-jev-latest-four-family-retry-v1/events.jsonl rescored (lite_assembly_a, time_partition_seconds.compound); runs/hosted-api-repeats-20261003/pass2/jev-latest/events.jsonl rescored (lite_assembly_a, time_partition_seconds.compound); runs/hosted-api-repeats-20261003/pass3/jev-latest/events.jsonl rescored (lite_assembly_a, time_partition_seconds.compound)
\newcommand{\JevAssemblyBErrJudgmentSec}{43.5}                          % equal mean across passes: runs/lite-v1-jev-latest-four-family-retry-v1/events.jsonl rescored (lite_assembly_b, time_partition_seconds.judgment); runs/hosted-api-repeats-20261003/pass2/jev-latest/events.jsonl rescored (lite_assembly_b, time_partition_seconds.judgment); runs/hosted-api-repeats-20261003/pass3/jev-latest/events.jsonl rescored (lite_assembly_b, time_partition_seconds.judgment)
\newcommand{\JevAssemblyBErrCompoundSec}{2.2}                           % equal mean across passes: runs/lite-v1-jev-latest-four-family-retry-v1/events.jsonl rescored (lite_assembly_b, time_partition_seconds.compound); runs/hosted-api-repeats-20261003/pass2/jev-latest/events.jsonl rescored (lite_assembly_b, time_partition_seconds.compound); runs/hosted-api-repeats-20261003/pass3/jev-latest/events.jsonl rescored (lite_assembly_b, time_partition_seconds.compound)
\newcommand{\JevSupportAErrJudgmentSec}{39.2}                           % equal mean across passes: runs/lite-v1-jev-latest-four-family-retry-v1/events.jsonl rescored (lite_support_a, time_partition_seconds.judgment); runs/hosted-api-repeats-20261003/pass2/jev-latest/events.jsonl rescored (lite_support_a, time_partition_seconds.judgment); runs/hosted-api-repeats-20261003/pass3/jev-latest/events.jsonl rescored (lite_support_a, time_partition_seconds.judgment)
\newcommand{\JevSupportAErrCompoundSec}{1.7}                            % equal mean across passes: runs/lite-v1-jev-latest-four-family-retry-v1/events.jsonl rescored (lite_support_a, time_partition_seconds.compound); runs/hosted-api-repeats-20261003/pass2/jev-latest/events.jsonl rescored (lite_support_a, time_partition_seconds.compound); runs/hosted-api-repeats-20261003/pass3/jev-latest/events.jsonl rescored (lite_support_a, time_partition_seconds.compound)
\newcommand{\JevSupportBErrJudgmentSec}{24.7}                           % equal mean across passes: runs/lite-v1-jev-latest-four-family-retry-v1/events.jsonl rescored (lite_support_b, time_partition_seconds.judgment); runs/hosted-api-repeats-20261003/pass2/jev-latest/events.jsonl rescored (lite_support_b, time_partition_seconds.judgment); runs/hosted-api-repeats-20261003/pass3/jev-latest/events.jsonl rescored (lite_support_b, time_partition_seconds.judgment)
\newcommand{\JevSupportBErrCompoundSec}{0.8}                            % equal mean across passes: runs/lite-v1-jev-latest-four-family-retry-v1/events.jsonl rescored (lite_support_b, time_partition_seconds.compound); runs/hosted-api-repeats-20261003/pass2/jev-latest/events.jsonl rescored (lite_support_b, time_partition_seconds.compound); runs/hosted-api-repeats-20261003/pass3/jev-latest/events.jsonl rescored (lite_support_b, time_partition_seconds.compound)
\newcommand{\JevPresenterAErrJudgmentSec}{34.5}                         % equal mean across passes: runs/lite-v1-jev-latest-four-family-retry-v1/events.jsonl rescored (lite_presenter_a, time_partition_seconds.judgment); runs/hosted-api-repeats-20261003/pass2/jev-latest/events.jsonl rescored (lite_presenter_a, time_partition_seconds.judgment); runs/hosted-api-repeats-20261003/pass3/jev-latest/events.jsonl rescored (lite_presenter_a, time_partition_seconds.judgment)
\newcommand{\JevPresenterAErrCompoundSec}{1.9}                          % equal mean across passes: runs/lite-v1-jev-latest-four-family-retry-v1/events.jsonl rescored (lite_presenter_a, time_partition_seconds.compound); runs/hosted-api-repeats-20261003/pass2/jev-latest/events.jsonl rescored (lite_presenter_a, time_partition_seconds.compound); runs/hosted-api-repeats-20261003/pass3/jev-latest/events.jsonl rescored (lite_presenter_a, time_partition_seconds.compound)
\newcommand{\JevPresenterBErrJudgmentSec}{30.3}                         % equal mean across passes: runs/lite-v1-jev-latest-four-family-retry-v1/events.jsonl rescored (lite_presenter_b, time_partition_seconds.judgment); runs/hosted-api-repeats-20261003/pass2/jev-latest/events.jsonl rescored (lite_presenter_b, time_partition_seconds.judgment); runs/hosted-api-repeats-20261003/pass3/jev-latest/events.jsonl rescored (lite_presenter_b, time_partition_seconds.judgment)
\newcommand{\JevPresenterBErrCompoundSec}{1.5}                          % equal mean across passes: runs/lite-v1-jev-latest-four-family-retry-v1/events.jsonl rescored (lite_presenter_b, time_partition_seconds.compound); runs/hosted-api-repeats-20261003/pass2/jev-latest/events.jsonl rescored (lite_presenter_b, time_partition_seconds.compound); runs/hosted-api-repeats-20261003/pass3/jev-latest/events.jsonl rescored (lite_presenter_b, time_partition_seconds.compound)
\newcommand{\JevProduct}{59.3}                                          % equal mean across passes: runs/lite-v1-jev-latest-four-family-retry-v1/events.jsonl rescored: mean over scenarios of untimed accuracy x oracle in-force accuracy; runs/hosted-api-repeats-20261003/pass2/jev-latest/events.jsonl rescored: mean over scenarios of untimed accuracy x oracle in-force accuracy; runs/hosted-api-repeats-20261003/pass3/jev-latest/events.jsonl rescored: mean over scenarios of untimed accuracy x oracle in-force accuracy
\newcommand{\CostTotalUSD}{70.57}                                       % sum across passes: sum of \<Model>CostUSD over the six recorded passes (USD, list prices); sum of \<Model>CostUSD over the six recorded passes (USD, list prices); sum of \<Model>CostUSD over the six recorded passes (USD, list prices)
\newcommand{\PricesRetrieved}{2026-09-29 and 2026-09-30}                % dates the list prices were read from the providers' official pages (paper/notes/pricing.md); dates the list prices were read from the providers' official pages (paper/notes/pricing.md); dates the list prices were read from the providers' official pages (paper/notes/pricing.md)
\newcommand{\ProductMaxDeviation}{6.8}                                  % docs/research/manuscript-three-pass/analysis.json:three-pass statistics; extrema after aggregation, physical gaps across passes
\newcommand{\LunaCommonTwo}{47.9}                                       % equal mean across passes: common-interval replay of frozen answers and latencies; common-interval replay of frozen answers and latencies; common-interval replay of frozen answers and latencies
\newcommand{\LunaNoneCommonTwo}{33.5}                                   % equal mean across passes: common-interval replay of frozen answers and latencies; common-interval replay of frozen answers and latencies; common-interval replay of frozen answers and latencies
\newcommand{\TerraCommonTwo}{55.9}                                      % equal mean across passes: common-interval replay of frozen answers and latencies; common-interval replay of frozen answers and latencies; common-interval replay of frozen answers and latencies
\newcommand{\TerraNoneCommonTwo}{61.3}                                  % equal mean across passes: common-interval replay of frozen answers and latencies; common-interval replay of frozen answers and latencies; common-interval replay of frozen answers and latencies
\newcommand{\TerraNoneCommonEight}{76.7}                                % equal mean across passes: common-interval replay of frozen answers and latencies; common-interval replay of frozen answers and latencies; common-interval replay of frozen answers and latencies
\newcommand{\AstraCommonTwo}{52.1}                                      % equal mean across passes: common-interval replay of frozen answers and latencies; common-interval replay of frozen answers and latencies; common-interval replay of frozen answers and latencies
\newcommand{\AstraCommonEight}{87.5}                                    % equal mean across passes: common-interval replay of frozen answers and latencies; common-interval replay of frozen answers and latencies; common-interval replay of frozen answers and latencies
\newcommand{\JevCommonTwo}{59.5}                                        % equal mean across passes: common-interval replay of frozen answers and latencies; common-interval replay of frozen answers and latencies; common-interval replay of frozen answers and latencies
\newcommand{\JevCommonEight}{61.6}                                      % equal mean across passes: common-interval replay of frozen answers and latencies; common-interval replay of frozen answers and latencies; common-interval replay of frozen answers and latencies
\newcommand{\RegimeLiteLuna}{46.2}                                      % equal mean across passes: projection on the eight SDB reference schedules (300 latency resamples each); projection on the eight SDB reference schedules (300 latency resamples each); projection on the eight SDB reference schedules (300 latency resamples each)
\newcommand{\RegimeLiteLunaNone}{32.2}                                  % equal mean across passes: projection on the eight SDB reference schedules (300 latency resamples each); projection on the eight SDB reference schedules (300 latency resamples each); projection on the eight SDB reference schedules (300 latency resamples each)
\newcommand{\RegimeLiteTerra}{56.0}                                     % equal mean across passes: projection on the eight SDB reference schedules (300 latency resamples each); projection on the eight SDB reference schedules (300 latency resamples each); projection on the eight SDB reference schedules (300 latency resamples each)
\newcommand{\RegimeLiteTerraNone}{59.7}                                 % equal mean across passes: projection on the eight SDB reference schedules (300 latency resamples each); projection on the eight SDB reference schedules (300 latency resamples each); projection on the eight SDB reference schedules (300 latency resamples each)
\newcommand{\RegimeLiteAstra}{53.4}                                     % equal mean across passes: projection on the eight SDB reference schedules (300 latency resamples each); projection on the eight SDB reference schedules (300 latency resamples each); projection on the eight SDB reference schedules (300 latency resamples each)
\newcommand{\RegimeLiteJev}{59.4}                                       % equal mean across passes: projection on the eight SDB reference schedules (300 latency resamples each); projection on the eight SDB reference schedules (300 latency resamples each); projection on the eight SDB reference schedules (300 latency resamples each)
\newcommand{\RegimeLiteMaxDiff}{1.7}                                    % docs/research/manuscript-three-pass/analysis.json:projection versus observed three-pass means
\newcommand{\RegimeSparseChanges}{5}                                    % fixed experimental design, identical across passes: reference changes in the Sparse schedule; reference changes in the Sparse schedule; reference changes in the Sparse schedule
\newcommand{\RegimeSparseLuna}{78.1}                                    % equal mean across passes: untimed accuracy x mean oracle share over 300 resamples of recorded latencies (Sparse schedule, 60 time steps of \ReleaseIntervalSec\,s); untimed accuracy x mean oracle share over 300 resamples of recorded latencies (Sparse schedule, 60 time steps of \ReleaseIntervalSec\,s); untimed accuracy x mean oracle share over 300 resamples of recorded latencies (Sparse schedule, 60 time steps of \ReleaseIntervalSec\,s)
\newcommand{\RegimeSparseLunaNone}{40.6}                                % equal mean across passes: untimed accuracy x mean oracle share over 300 resamples of recorded latencies (Sparse schedule, 60 time steps of \ReleaseIntervalSec\,s); untimed accuracy x mean oracle share over 300 resamples of recorded latencies (Sparse schedule, 60 time steps of \ReleaseIntervalSec\,s); untimed accuracy x mean oracle share over 300 resamples of recorded latencies (Sparse schedule, 60 time steps of \ReleaseIntervalSec\,s)
\newcommand{\RegimeSparseTerra}{85.2}                                   % equal mean across passes: untimed accuracy x mean oracle share over 300 resamples of recorded latencies (Sparse schedule, 60 time steps of \ReleaseIntervalSec\,s); untimed accuracy x mean oracle share over 300 resamples of recorded latencies (Sparse schedule, 60 time steps of \ReleaseIntervalSec\,s); untimed accuracy x mean oracle share over 300 resamples of recorded latencies (Sparse schedule, 60 time steps of \ReleaseIntervalSec\,s)
\newcommand{\RegimeSparseTerraNone}{76.1}                               % equal mean across passes: untimed accuracy x mean oracle share over 300 resamples of recorded latencies (Sparse schedule, 60 time steps of \ReleaseIntervalSec\,s); untimed accuracy x mean oracle share over 300 resamples of recorded latencies (Sparse schedule, 60 time steps of \ReleaseIntervalSec\,s); untimed accuracy x mean oracle share over 300 resamples of recorded latencies (Sparse schedule, 60 time steps of \ReleaseIntervalSec\,s)
\newcommand{\RegimeSparseAstra}{87.4}                                   % equal mean across passes: untimed accuracy x mean oracle share over 300 resamples of recorded latencies (Sparse schedule, 60 time steps of \ReleaseIntervalSec\,s); untimed accuracy x mean oracle share over 300 resamples of recorded latencies (Sparse schedule, 60 time steps of \ReleaseIntervalSec\,s); untimed accuracy x mean oracle share over 300 resamples of recorded latencies (Sparse schedule, 60 time steps of \ReleaseIntervalSec\,s)
\newcommand{\RegimeSparseJev}{61.6}                                     % equal mean across passes: untimed accuracy x mean oracle share over 300 resamples of recorded latencies (Sparse schedule, 60 time steps of \ReleaseIntervalSec\,s); untimed accuracy x mean oracle share over 300 resamples of recorded latencies (Sparse schedule, 60 time steps of \ReleaseIntervalSec\,s); untimed accuracy x mean oracle share over 300 resamples of recorded latencies (Sparse schedule, 60 time steps of \ReleaseIntervalSec\,s)
\newcommand{\RegimeMediumChanges}{11}                                   % fixed experimental design, identical across passes: reference changes in the Medium schedule; reference changes in the Medium schedule; reference changes in the Medium schedule
\newcommand{\RegimeMediumLuna}{66.5}                                    % equal mean across passes: untimed accuracy x mean oracle share over 300 resamples of recorded latencies (Medium schedule, 60 time steps of \ReleaseIntervalSec\,s); untimed accuracy x mean oracle share over 300 resamples of recorded latencies (Medium schedule, 60 time steps of \ReleaseIntervalSec\,s); untimed accuracy x mean oracle share over 300 resamples of recorded latencies (Medium schedule, 60 time steps of \ReleaseIntervalSec\,s)
\newcommand{\RegimeMediumLunaNone}{37.6}                                % equal mean across passes: untimed accuracy x mean oracle share over 300 resamples of recorded latencies (Medium schedule, 60 time steps of \ReleaseIntervalSec\,s); untimed accuracy x mean oracle share over 300 resamples of recorded latencies (Medium schedule, 60 time steps of \ReleaseIntervalSec\,s); untimed accuracy x mean oracle share over 300 resamples of recorded latencies (Medium schedule, 60 time steps of \ReleaseIntervalSec\,s)
\newcommand{\RegimeMediumTerra}{74.6}                                   % equal mean across passes: untimed accuracy x mean oracle share over 300 resamples of recorded latencies (Medium schedule, 60 time steps of \ReleaseIntervalSec\,s); untimed accuracy x mean oracle share over 300 resamples of recorded latencies (Medium schedule, 60 time steps of \ReleaseIntervalSec\,s); untimed accuracy x mean oracle share over 300 resamples of recorded latencies (Medium schedule, 60 time steps of \ReleaseIntervalSec\,s)
\newcommand{\RegimeMediumTerraNone}{70.3}                               % equal mean across passes: untimed accuracy x mean oracle share over 300 resamples of recorded latencies (Medium schedule, 60 time steps of \ReleaseIntervalSec\,s); untimed accuracy x mean oracle share over 300 resamples of recorded latencies (Medium schedule, 60 time steps of \ReleaseIntervalSec\,s); untimed accuracy x mean oracle share over 300 resamples of recorded latencies (Medium schedule, 60 time steps of \ReleaseIntervalSec\,s)
\newcommand{\RegimeMediumAstra}{75.0}                                   % equal mean across passes: untimed accuracy x mean oracle share over 300 resamples of recorded latencies (Medium schedule, 60 time steps of \ReleaseIntervalSec\,s); untimed accuracy x mean oracle share over 300 resamples of recorded latencies (Medium schedule, 60 time steps of \ReleaseIntervalSec\,s); untimed accuracy x mean oracle share over 300 resamples of recorded latencies (Medium schedule, 60 time steps of \ReleaseIntervalSec\,s)
\newcommand{\RegimeMediumJev}{60.8}                                     % equal mean across passes: untimed accuracy x mean oracle share over 300 resamples of recorded latencies (Medium schedule, 60 time steps of \ReleaseIntervalSec\,s); untimed accuracy x mean oracle share over 300 resamples of recorded latencies (Medium schedule, 60 time steps of \ReleaseIntervalSec\,s); untimed accuracy x mean oracle share over 300 resamples of recorded latencies (Medium schedule, 60 time steps of \ReleaseIntervalSec\,s)
\newcommand{\RegimeUniformChanges}{22}                                  % fixed experimental design, identical across passes: reference changes in the Uniform schedule; reference changes in the Uniform schedule; reference changes in the Uniform schedule
\newcommand{\RegimeUniformLuna}{44.9}                                   % equal mean across passes: untimed accuracy x mean oracle share over 300 resamples of recorded latencies (Uniform schedule, 60 time steps of \ReleaseIntervalSec\,s); untimed accuracy x mean oracle share over 300 resamples of recorded latencies (Uniform schedule, 60 time steps of \ReleaseIntervalSec\,s); untimed accuracy x mean oracle share over 300 resamples of recorded latencies (Uniform schedule, 60 time steps of \ReleaseIntervalSec\,s)
\newcommand{\RegimeUniformLunaNone}{32.2}                               % equal mean across passes: untimed accuracy x mean oracle share over 300 resamples of recorded latencies (Uniform schedule, 60 time steps of \ReleaseIntervalSec\,s); untimed accuracy x mean oracle share over 300 resamples of recorded latencies (Uniform schedule, 60 time steps of \ReleaseIntervalSec\,s); untimed accuracy x mean oracle share over 300 resamples of recorded latencies (Uniform schedule, 60 time steps of \ReleaseIntervalSec\,s)
\newcommand{\RegimeUniformTerra}{55.2}                                  % equal mean across passes: untimed accuracy x mean oracle share over 300 resamples of recorded latencies (Uniform schedule, 60 time steps of \ReleaseIntervalSec\,s); untimed accuracy x mean oracle share over 300 resamples of recorded latencies (Uniform schedule, 60 time steps of \ReleaseIntervalSec\,s); untimed accuracy x mean oracle share over 300 resamples of recorded latencies (Uniform schedule, 60 time steps of \ReleaseIntervalSec\,s)
\newcommand{\RegimeUniformTerraNone}{59.8}                              % equal mean across passes: untimed accuracy x mean oracle share over 300 resamples of recorded latencies (Uniform schedule, 60 time steps of \ReleaseIntervalSec\,s); untimed accuracy x mean oracle share over 300 resamples of recorded latencies (Uniform schedule, 60 time steps of \ReleaseIntervalSec\,s); untimed accuracy x mean oracle share over 300 resamples of recorded latencies (Uniform schedule, 60 time steps of \ReleaseIntervalSec\,s)
\newcommand{\RegimeUniformAstra}{52.3}                                  % equal mean across passes: untimed accuracy x mean oracle share over 300 resamples of recorded latencies (Uniform schedule, 60 time steps of \ReleaseIntervalSec\,s); untimed accuracy x mean oracle share over 300 resamples of recorded latencies (Uniform schedule, 60 time steps of \ReleaseIntervalSec\,s); untimed accuracy x mean oracle share over 300 resamples of recorded latencies (Uniform schedule, 60 time steps of \ReleaseIntervalSec\,s)
\newcommand{\RegimeUniformJev}{59.4}                                    % equal mean across passes: untimed accuracy x mean oracle share over 300 resamples of recorded latencies (Uniform schedule, 60 time steps of \ReleaseIntervalSec\,s); untimed accuracy x mean oracle share over 300 resamples of recorded latencies (Uniform schedule, 60 time steps of \ReleaseIntervalSec\,s); untimed accuracy x mean oracle share over 300 resamples of recorded latencies (Uniform schedule, 60 time steps of \ReleaseIntervalSec\,s)
\newcommand{\RegimeDenseChanges}{40}                                    % fixed experimental design, identical across passes: reference changes in the Dense schedule; reference changes in the Dense schedule; reference changes in the Dense schedule
\newcommand{\RegimeDenseLuna}{20.4}                                     % equal mean across passes: untimed accuracy x mean oracle share over 300 resamples of recorded latencies (Dense schedule, 60 time steps of \ReleaseIntervalSec\,s); untimed accuracy x mean oracle share over 300 resamples of recorded latencies (Dense schedule, 60 time steps of \ReleaseIntervalSec\,s); untimed accuracy x mean oracle share over 300 resamples of recorded latencies (Dense schedule, 60 time steps of \ReleaseIntervalSec\,s)
\newcommand{\RegimeDenseLunaNone}{23.4}                                 % equal mean across passes: untimed accuracy x mean oracle share over 300 resamples of recorded latencies (Dense schedule, 60 time steps of \ReleaseIntervalSec\,s); untimed accuracy x mean oracle share over 300 resamples of recorded latencies (Dense schedule, 60 time steps of \ReleaseIntervalSec\,s); untimed accuracy x mean oracle share over 300 resamples of recorded latencies (Dense schedule, 60 time steps of \ReleaseIntervalSec\,s)
\newcommand{\RegimeDenseTerra}{29.7}                                    % equal mean across passes: untimed accuracy x mean oracle share over 300 resamples of recorded latencies (Dense schedule, 60 time steps of \ReleaseIntervalSec\,s); untimed accuracy x mean oracle share over 300 resamples of recorded latencies (Dense schedule, 60 time steps of \ReleaseIntervalSec\,s); untimed accuracy x mean oracle share over 300 resamples of recorded latencies (Dense schedule, 60 time steps of \ReleaseIntervalSec\,s)
\newcommand{\RegimeDenseTerraNone}{42.8}                                % equal mean across passes: untimed accuracy x mean oracle share over 300 resamples of recorded latencies (Dense schedule, 60 time steps of \ReleaseIntervalSec\,s); untimed accuracy x mean oracle share over 300 resamples of recorded latencies (Dense schedule, 60 time steps of \ReleaseIntervalSec\,s); untimed accuracy x mean oracle share over 300 resamples of recorded latencies (Dense schedule, 60 time steps of \ReleaseIntervalSec\,s)
\newcommand{\RegimeDenseAstra}{24.6}                                    % equal mean across passes: untimed accuracy x mean oracle share over 300 resamples of recorded latencies (Dense schedule, 60 time steps of \ReleaseIntervalSec\,s); untimed accuracy x mean oracle share over 300 resamples of recorded latencies (Dense schedule, 60 time steps of \ReleaseIntervalSec\,s); untimed accuracy x mean oracle share over 300 resamples of recorded latencies (Dense schedule, 60 time steps of \ReleaseIntervalSec\,s)
\newcommand{\RegimeDenseJev}{57.0}                                      % equal mean across passes: untimed accuracy x mean oracle share over 300 resamples of recorded latencies (Dense schedule, 60 time steps of \ReleaseIntervalSec\,s); untimed accuracy x mean oracle share over 300 resamples of recorded latencies (Dense schedule, 60 time steps of \ReleaseIntervalSec\,s); untimed accuracy x mean oracle share over 300 resamples of recorded latencies (Dense schedule, 60 time steps of \ReleaseIntervalSec\,s)
\newcommand{\RegimeBurstyChanges}{22}                                   % fixed experimental design, identical across passes: reference changes in the Bursty schedule; reference changes in the Bursty schedule; reference changes in the Bursty schedule
\newcommand{\RegimeBurstyLuna}{53.4}                                    % equal mean across passes: untimed accuracy x mean oracle share over 300 resamples of recorded latencies (Bursty schedule, 60 time steps of \ReleaseIntervalSec\,s); untimed accuracy x mean oracle share over 300 resamples of recorded latencies (Bursty schedule, 60 time steps of \ReleaseIntervalSec\,s); untimed accuracy x mean oracle share over 300 resamples of recorded latencies (Bursty schedule, 60 time steps of \ReleaseIntervalSec\,s)
\newcommand{\RegimeBurstyLunaNone}{32.3}                                % equal mean across passes: untimed accuracy x mean oracle share over 300 resamples of recorded latencies (Bursty schedule, 60 time steps of \ReleaseIntervalSec\,s); untimed accuracy x mean oracle share over 300 resamples of recorded latencies (Bursty schedule, 60 time steps of \ReleaseIntervalSec\,s); untimed accuracy x mean oracle share over 300 resamples of recorded latencies (Bursty schedule, 60 time steps of \ReleaseIntervalSec\,s)
\newcommand{\RegimeBurstyTerra}{60.0}                                   % equal mean across passes: untimed accuracy x mean oracle share over 300 resamples of recorded latencies (Bursty schedule, 60 time steps of \ReleaseIntervalSec\,s); untimed accuracy x mean oracle share over 300 resamples of recorded latencies (Bursty schedule, 60 time steps of \ReleaseIntervalSec\,s); untimed accuracy x mean oracle share over 300 resamples of recorded latencies (Bursty schedule, 60 time steps of \ReleaseIntervalSec\,s)
\newcommand{\RegimeBurstyTerraNone}{60.0}                               % equal mean across passes: untimed accuracy x mean oracle share over 300 resamples of recorded latencies (Bursty schedule, 60 time steps of \ReleaseIntervalSec\,s); untimed accuracy x mean oracle share over 300 resamples of recorded latencies (Bursty schedule, 60 time steps of \ReleaseIntervalSec\,s); untimed accuracy x mean oracle share over 300 resamples of recorded latencies (Bursty schedule, 60 time steps of \ReleaseIntervalSec\,s)
\newcommand{\RegimeBurstyAstra}{60.0}                                   % equal mean across passes: untimed accuracy x mean oracle share over 300 resamples of recorded latencies (Bursty schedule, 60 time steps of \ReleaseIntervalSec\,s); untimed accuracy x mean oracle share over 300 resamples of recorded latencies (Bursty schedule, 60 time steps of \ReleaseIntervalSec\,s); untimed accuracy x mean oracle share over 300 resamples of recorded latencies (Bursty schedule, 60 time steps of \ReleaseIntervalSec\,s)
\newcommand{\RegimeBurstyJev}{59.4}                                     % equal mean across passes: untimed accuracy x mean oracle share over 300 resamples of recorded latencies (Bursty schedule, 60 time steps of \ReleaseIntervalSec\,s); untimed accuracy x mean oracle share over 300 resamples of recorded latencies (Bursty schedule, 60 time steps of \ReleaseIntervalSec\,s); untimed accuracy x mean oracle share over 300 resamples of recorded latencies (Bursty schedule, 60 time steps of \ReleaseIntervalSec\,s)
\newcommand{\RegimeLongTailChanges}{22}                                 % fixed experimental design, identical across passes: reference changes in the LongTail schedule; reference changes in the LongTail schedule; reference changes in the LongTail schedule
\newcommand{\RegimeLongTailLuna}{50.8}                                  % equal mean across passes: untimed accuracy x mean oracle share over 300 resamples of recorded latencies (LongTail schedule, 60 time steps of \ReleaseIntervalSec\,s); untimed accuracy x mean oracle share over 300 resamples of recorded latencies (LongTail schedule, 60 time steps of \ReleaseIntervalSec\,s); untimed accuracy x mean oracle share over 300 resamples of recorded latencies (LongTail schedule, 60 time steps of \ReleaseIntervalSec\,s)
\newcommand{\RegimeLongTailLunaNone}{32.3}                              % equal mean across passes: untimed accuracy x mean oracle share over 300 resamples of recorded latencies (LongTail schedule, 60 time steps of \ReleaseIntervalSec\,s); untimed accuracy x mean oracle share over 300 resamples of recorded latencies (LongTail schedule, 60 time steps of \ReleaseIntervalSec\,s); untimed accuracy x mean oracle share over 300 resamples of recorded latencies (LongTail schedule, 60 time steps of \ReleaseIntervalSec\,s)
\newcommand{\RegimeLongTailTerra}{58.5}                                 % equal mean across passes: untimed accuracy x mean oracle share over 300 resamples of recorded latencies (LongTail schedule, 60 time steps of \ReleaseIntervalSec\,s); untimed accuracy x mean oracle share over 300 resamples of recorded latencies (LongTail schedule, 60 time steps of \ReleaseIntervalSec\,s); untimed accuracy x mean oracle share over 300 resamples of recorded latencies (LongTail schedule, 60 time steps of \ReleaseIntervalSec\,s)
\newcommand{\RegimeLongTailTerraNone}{59.9}                             % equal mean across passes: untimed accuracy x mean oracle share over 300 resamples of recorded latencies (LongTail schedule, 60 time steps of \ReleaseIntervalSec\,s); untimed accuracy x mean oracle share over 300 resamples of recorded latencies (LongTail schedule, 60 time steps of \ReleaseIntervalSec\,s); untimed accuracy x mean oracle share over 300 resamples of recorded latencies (LongTail schedule, 60 time steps of \ReleaseIntervalSec\,s)
\newcommand{\RegimeLongTailAstra}{57.4}                                 % equal mean across passes: untimed accuracy x mean oracle share over 300 resamples of recorded latencies (LongTail schedule, 60 time steps of \ReleaseIntervalSec\,s); untimed accuracy x mean oracle share over 300 resamples of recorded latencies (LongTail schedule, 60 time steps of \ReleaseIntervalSec\,s); untimed accuracy x mean oracle share over 300 resamples of recorded latencies (LongTail schedule, 60 time steps of \ReleaseIntervalSec\,s)
\newcommand{\RegimeLongTailJev}{59.4}                                   % equal mean across passes: untimed accuracy x mean oracle share over 300 resamples of recorded latencies (LongTail schedule, 60 time steps of \ReleaseIntervalSec\,s); untimed accuracy x mean oracle share over 300 resamples of recorded latencies (LongTail schedule, 60 time steps of \ReleaseIntervalSec\,s); untimed accuracy x mean oracle share over 300 resamples of recorded latencies (LongTail schedule, 60 time steps of \ReleaseIntervalSec\,s)
\newcommand{\RegimeRecurrentChanges}{22}                                % fixed experimental design, identical across passes: reference changes in the Recurrent schedule; reference changes in the Recurrent schedule; reference changes in the Recurrent schedule
\newcommand{\RegimeRecurrentLuna}{44.9}                                 % equal mean across passes: untimed accuracy x mean oracle share over 300 resamples of recorded latencies (Recurrent schedule, 60 time steps of \ReleaseIntervalSec\,s); untimed accuracy x mean oracle share over 300 resamples of recorded latencies (Recurrent schedule, 60 time steps of \ReleaseIntervalSec\,s); untimed accuracy x mean oracle share over 300 resamples of recorded latencies (Recurrent schedule, 60 time steps of \ReleaseIntervalSec\,s)
\newcommand{\RegimeRecurrentLunaNone}{32.2}                             % equal mean across passes: untimed accuracy x mean oracle share over 300 resamples of recorded latencies (Recurrent schedule, 60 time steps of \ReleaseIntervalSec\,s); untimed accuracy x mean oracle share over 300 resamples of recorded latencies (Recurrent schedule, 60 time steps of \ReleaseIntervalSec\,s); untimed accuracy x mean oracle share over 300 resamples of recorded latencies (Recurrent schedule, 60 time steps of \ReleaseIntervalSec\,s)
\newcommand{\RegimeRecurrentTerra}{55.0}                                % equal mean across passes: untimed accuracy x mean oracle share over 300 resamples of recorded latencies (Recurrent schedule, 60 time steps of \ReleaseIntervalSec\,s); untimed accuracy x mean oracle share over 300 resamples of recorded latencies (Recurrent schedule, 60 time steps of \ReleaseIntervalSec\,s); untimed accuracy x mean oracle share over 300 resamples of recorded latencies (Recurrent schedule, 60 time steps of \ReleaseIntervalSec\,s)
\newcommand{\RegimeRecurrentTerraNone}{59.7}                            % equal mean across passes: untimed accuracy x mean oracle share over 300 resamples of recorded latencies (Recurrent schedule, 60 time steps of \ReleaseIntervalSec\,s); untimed accuracy x mean oracle share over 300 resamples of recorded latencies (Recurrent schedule, 60 time steps of \ReleaseIntervalSec\,s); untimed accuracy x mean oracle share over 300 resamples of recorded latencies (Recurrent schedule, 60 time steps of \ReleaseIntervalSec\,s)
\newcommand{\RegimeRecurrentAstra}{52.5}                                % equal mean across passes: untimed accuracy x mean oracle share over 300 resamples of recorded latencies (Recurrent schedule, 60 time steps of \ReleaseIntervalSec\,s); untimed accuracy x mean oracle share over 300 resamples of recorded latencies (Recurrent schedule, 60 time steps of \ReleaseIntervalSec\,s); untimed accuracy x mean oracle share over 300 resamples of recorded latencies (Recurrent schedule, 60 time steps of \ReleaseIntervalSec\,s)
\newcommand{\RegimeRecurrentJev}{59.4}                                  % equal mean across passes: untimed accuracy x mean oracle share over 300 resamples of recorded latencies (Recurrent schedule, 60 time steps of \ReleaseIntervalSec\,s); untimed accuracy x mean oracle share over 300 resamples of recorded latencies (Recurrent schedule, 60 time steps of \ReleaseIntervalSec\,s); untimed accuracy x mean oracle share over 300 resamples of recorded latencies (Recurrent schedule, 60 time steps of \ReleaseIntervalSec\,s)
\newcommand{\RegimeSteps}{60}                                           % fixed experimental design, identical across passes: time steps per projected schedule; time steps per projected schedule; time steps per projected schedule
\newcommand{\RegimeDraws}{300}                                          % fixed experimental design, identical across passes: latency resamples per projected schedule; latency resamples per projected schedule; latency resamples per projected schedule
\newcommand{\LunaAuc}{45.5}                                             % docs/research/manuscript-three-pass/analysis.json:hosted.Luna:auc.primary
\newcommand{\LunaDebugAuc}{48.0}                                        % docs/research/manuscript-three-pass/analysis.json:hosted.Luna:auc.primary
\newcommand{\LunaAssemblyAuc}{44.0}                                     % docs/research/manuscript-three-pass/analysis.json:hosted.Luna:auc.primary
\newcommand{\LunaSupportAuc}{44.4}                                      % docs/research/manuscript-three-pass/analysis.json:hosted.Luna:auc.primary
\newcommand{\LunaPresenterAuc}{45.3}                                    % docs/research/manuscript-three-pass/analysis.json:hosted.Luna:auc.primary
\newcommand{\LunaDebugAAuc}{50.5}                                       % docs/research/manuscript-three-pass/analysis.json:hosted.Luna:auc.primary
\newcommand{\LunaDebugBAuc}{45.5}                                       % docs/research/manuscript-three-pass/analysis.json:hosted.Luna:auc.primary
\newcommand{\LunaAssemblyAAuc}{47.1}                                    % docs/research/manuscript-three-pass/analysis.json:hosted.Luna:auc.primary
\newcommand{\LunaAssemblyBAuc}{41.0}                                    % docs/research/manuscript-three-pass/analysis.json:hosted.Luna:auc.primary
\newcommand{\LunaSupportAAuc}{43.8}                                     % docs/research/manuscript-three-pass/analysis.json:hosted.Luna:auc.primary
\newcommand{\LunaSupportBAuc}{45.0}                                     % docs/research/manuscript-three-pass/analysis.json:hosted.Luna:auc.primary
\newcommand{\LunaPresenterAAuc}{46.8}                                   % docs/research/manuscript-three-pass/analysis.json:hosted.Luna:auc.primary
\newcommand{\LunaPresenterBAuc}{43.9}                                   % docs/research/manuscript-three-pass/analysis.json:hosted.Luna:auc.primary
\newcommand{\LunaAucGap}{44.5}                                          % docs/research/manuscript-three-pass/analysis.json:hosted.Luna:auc.primary
\newcommand{\LunaAucOracle}{48.5}                                       % docs/research/manuscript-three-pass/analysis.json:hosted.Luna:auc.primary
\newcommand{\LunaAucProduct}{43.5}                                      % docs/research/manuscript-three-pass/analysis.json:hosted.Luna:auc.primary
\newcommand{\LunaAucUCurrent}{91.7}                                     % docs/research/manuscript-three-pass/analysis.json:hosted.Luna:auc.primary
\newcommand{\LunaAucStale}{43.0}                                        % docs/research/manuscript-three-pass/analysis.json:hosted.Luna:auc.primary
\newcommand{\LunaAucLucky}{1.0}                                         % docs/research/manuscript-three-pass/analysis.json:hosted.Luna:auc.primary
\newcommand{\LunaAucLinear}{61.6}                                       % docs/research/manuscript-three-pass/analysis.json:hosted.Luna:auc.sensitivity.Linear
\newcommand{\LunaAucNarrow}{46.9}                                       % docs/research/manuscript-three-pass/analysis.json:hosted.Luna:auc.sensitivity.Narrow
\newcommand{\LunaAucHalfFour}{36.0}                                     % docs/research/manuscript-three-pass/analysis.json:hosted.Luna:auc.sensitivity.HalfFour
\newcommand{\LunaAucOneEight}{55.9}                                     % docs/research/manuscript-three-pass/analysis.json:hosted.Luna:auc.sensitivity.OneEight
\newcommand{\LunaAucTenthFour}{21.8}                                    % docs/research/manuscript-three-pass/analysis.json:hosted.Luna:auc.sensitivity.TenthFour
\newcommand{\LunaAucTenthEight}{30.0}                                   % docs/research/manuscript-three-pass/analysis.json:hosted.Luna:auc.sensitivity.TenthEight
\newcommand{\LunaNoneAuc}{30.9}                                         % docs/research/manuscript-three-pass/analysis.json:hosted.LunaNone:auc.primary
\newcommand{\LunaNoneDebugAuc}{13.2}                                    % docs/research/manuscript-three-pass/analysis.json:hosted.LunaNone:auc.primary
\newcommand{\LunaNoneAssemblyAuc}{19.6}                                 % docs/research/manuscript-three-pass/analysis.json:hosted.LunaNone:auc.primary
\newcommand{\LunaNoneSupportAuc}{44.0}                                  % docs/research/manuscript-three-pass/analysis.json:hosted.LunaNone:auc.primary
\newcommand{\LunaNonePresenterAuc}{46.6}                                % docs/research/manuscript-three-pass/analysis.json:hosted.LunaNone:auc.primary
\newcommand{\LunaNoneDebugAAuc}{10.1}                                   % docs/research/manuscript-three-pass/analysis.json:hosted.LunaNone:auc.primary
\newcommand{\LunaNoneDebugBAuc}{16.2}                                   % docs/research/manuscript-three-pass/analysis.json:hosted.LunaNone:auc.primary
\newcommand{\LunaNoneAssemblyAAuc}{21.0}                                % docs/research/manuscript-three-pass/analysis.json:hosted.LunaNone:auc.primary
\newcommand{\LunaNoneAssemblyBAuc}{18.3}                                % docs/research/manuscript-three-pass/analysis.json:hosted.LunaNone:auc.primary
\newcommand{\LunaNoneSupportAAuc}{39.9}                                 % docs/research/manuscript-three-pass/analysis.json:hosted.LunaNone:auc.primary
\newcommand{\LunaNoneSupportBAuc}{48.2}                                 % docs/research/manuscript-three-pass/analysis.json:hosted.LunaNone:auc.primary
\newcommand{\LunaNonePresenterAAuc}{54.6}                               % docs/research/manuscript-three-pass/analysis.json:hosted.LunaNone:auc.primary
\newcommand{\LunaNonePresenterBAuc}{38.7}                               % docs/research/manuscript-three-pass/analysis.json:hosted.LunaNone:auc.primary
\newcommand{\LunaNoneAucOracle}{66.6}                                   % docs/research/manuscript-three-pass/analysis.json:hosted.LunaNone:auc.primary
\newcommand{\LunaNoneAucProduct}{28.7}                                  % docs/research/manuscript-three-pass/analysis.json:hosted.LunaNone:auc.primary
\newcommand{\LunaNoneAucUCurrent}{44.3}                                 % docs/research/manuscript-three-pass/analysis.json:hosted.LunaNone:auc.primary
\newcommand{\LunaNoneAucJudgment}{37.1}                                 % docs/research/manuscript-three-pass/analysis.json:hosted.LunaNone:auc.primary
\newcommand{\LunaNoneAucLucky}{1.4}                                     % docs/research/manuscript-three-pass/analysis.json:hosted.LunaNone:auc.primary
\newcommand{\LunaNoneAucLinear}{36.3}                                   % docs/research/manuscript-three-pass/analysis.json:hosted.LunaNone:auc.sensitivity.Linear
\newcommand{\LunaNoneAucNarrow}{32.8}                                   % docs/research/manuscript-three-pass/analysis.json:hosted.LunaNone:auc.sensitivity.Narrow
\newcommand{\LunaNoneAucHalfFour}{27.9}                                 % docs/research/manuscript-three-pass/analysis.json:hosted.LunaNone:auc.sensitivity.HalfFour
\newcommand{\LunaNoneAucOneEight}{35.1}                                 % docs/research/manuscript-three-pass/analysis.json:hosted.LunaNone:auc.sensitivity.OneEight
\newcommand{\LunaNoneAucTenthFour}{18.0}                                % docs/research/manuscript-three-pass/analysis.json:hosted.LunaNone:auc.sensitivity.TenthFour
\newcommand{\LunaNoneAucTenthEight}{21.5}                               % docs/research/manuscript-three-pass/analysis.json:hosted.LunaNone:auc.sensitivity.TenthEight
\newcommand{\TerraAuc}{52.1}                                            % docs/research/manuscript-three-pass/analysis.json:hosted.Terra:auc.primary
\newcommand{\TerraDebugAuc}{51.9}                                       % docs/research/manuscript-three-pass/analysis.json:hosted.Terra:auc.primary
\newcommand{\TerraAssemblyAuc}{49.3}                                    % docs/research/manuscript-three-pass/analysis.json:hosted.Terra:auc.primary
\newcommand{\TerraSupportAuc}{54.6}                                     % docs/research/manuscript-three-pass/analysis.json:hosted.Terra:auc.primary
\newcommand{\TerraPresenterAuc}{52.4}                                   % docs/research/manuscript-three-pass/analysis.json:hosted.Terra:auc.primary
\newcommand{\TerraDebugAAuc}{52.4}                                      % docs/research/manuscript-three-pass/analysis.json:hosted.Terra:auc.primary
\newcommand{\TerraDebugBAuc}{51.4}                                      % docs/research/manuscript-three-pass/analysis.json:hosted.Terra:auc.primary
\newcommand{\TerraAssemblyAAuc}{52.4}                                   % docs/research/manuscript-three-pass/analysis.json:hosted.Terra:auc.primary
\newcommand{\TerraAssemblyBAuc}{46.3}                                   % docs/research/manuscript-three-pass/analysis.json:hosted.Terra:auc.primary
\newcommand{\TerraSupportAAuc}{56.0}                                    % docs/research/manuscript-three-pass/analysis.json:hosted.Terra:auc.primary
\newcommand{\TerraSupportBAuc}{53.2}                                    % docs/research/manuscript-three-pass/analysis.json:hosted.Terra:auc.primary
\newcommand{\TerraPresenterAAuc}{52.6}                                  % docs/research/manuscript-three-pass/analysis.json:hosted.Terra:auc.primary
\newcommand{\TerraPresenterBAuc}{52.2}                                  % docs/research/manuscript-three-pass/analysis.json:hosted.Terra:auc.primary
\newcommand{\TerraAucGap}{43.7}                                         % docs/research/manuscript-three-pass/analysis.json:hosted.Terra:auc.primary
\newcommand{\TerraAucOracle}{53.5}                                      % docs/research/manuscript-three-pass/analysis.json:hosted.Terra:auc.primary
\newcommand{\TerraAucProduct}{51.3}                                     % docs/research/manuscript-three-pass/analysis.json:hosted.Terra:auc.primary
\newcommand{\TerraAucUCurrent}{96.8}                                    % docs/research/manuscript-three-pass/analysis.json:hosted.Terra:auc.primary
\newcommand{\TerraAucStale}{41.7}                                       % docs/research/manuscript-three-pass/analysis.json:hosted.Terra:auc.primary
\newcommand{\TerraAucLucky}{0.3}                                        % docs/research/manuscript-three-pass/analysis.json:hosted.Terra:auc.primary
\newcommand{\TerraAucLinear}{68.6}                                      % docs/research/manuscript-three-pass/analysis.json:hosted.Terra:auc.sensitivity.Linear
\newcommand{\TerraAucNarrow}{54.3}                                      % docs/research/manuscript-three-pass/analysis.json:hosted.Terra:auc.sensitivity.Narrow
\newcommand{\TerraAucHalfFour}{42.4}                                    % docs/research/manuscript-three-pass/analysis.json:hosted.Terra:auc.sensitivity.HalfFour
\newcommand{\TerraAucOneEight}{63.2}                                    % docs/research/manuscript-three-pass/analysis.json:hosted.Terra:auc.sensitivity.OneEight
\newcommand{\TerraAucTenthFour}{25.9}                                   % docs/research/manuscript-three-pass/analysis.json:hosted.Terra:auc.sensitivity.TenthFour
\newcommand{\TerraAucTenthEight}{34.6}                                  % docs/research/manuscript-three-pass/analysis.json:hosted.Terra:auc.sensitivity.TenthEight
\newcommand{\TerraNoneAuc}{55.5}                                        % docs/research/manuscript-three-pass/analysis.json:hosted.TerraNone:auc.primary
\newcommand{\TerraNoneDebugAuc}{50.0}                                   % docs/research/manuscript-three-pass/analysis.json:hosted.TerraNone:auc.primary
\newcommand{\TerraNoneAssemblyAuc}{50.4}                                % docs/research/manuscript-three-pass/analysis.json:hosted.TerraNone:auc.primary
\newcommand{\TerraNoneSupportAuc}{59.5}                                 % docs/research/manuscript-three-pass/analysis.json:hosted.TerraNone:auc.primary
\newcommand{\TerraNonePresenterAuc}{62.0}                               % docs/research/manuscript-three-pass/analysis.json:hosted.TerraNone:auc.primary
\newcommand{\TerraNoneDebugAAuc}{52.4}                                  % docs/research/manuscript-three-pass/analysis.json:hosted.TerraNone:auc.primary
\newcommand{\TerraNoneDebugBAuc}{47.6}                                  % docs/research/manuscript-three-pass/analysis.json:hosted.TerraNone:auc.primary
\newcommand{\TerraNoneAssemblyAAuc}{51.7}                               % docs/research/manuscript-three-pass/analysis.json:hosted.TerraNone:auc.primary
\newcommand{\TerraNoneAssemblyBAuc}{49.0}                               % docs/research/manuscript-three-pass/analysis.json:hosted.TerraNone:auc.primary
\newcommand{\TerraNoneSupportAAuc}{61.1}                                % docs/research/manuscript-three-pass/analysis.json:hosted.TerraNone:auc.primary
\newcommand{\TerraNoneSupportBAuc}{57.9}                                % docs/research/manuscript-three-pass/analysis.json:hosted.TerraNone:auc.primary
\newcommand{\TerraNonePresenterAAuc}{63.3}                              % docs/research/manuscript-three-pass/analysis.json:hosted.TerraNone:auc.primary
\newcommand{\TerraNonePresenterBAuc}{60.6}                              % docs/research/manuscript-three-pass/analysis.json:hosted.TerraNone:auc.primary
\newcommand{\TerraNoneAucOracle}{65.0}                                  % docs/research/manuscript-three-pass/analysis.json:hosted.TerraNone:auc.primary
\newcommand{\TerraNoneAucProduct}{53.1}                                 % docs/research/manuscript-three-pass/analysis.json:hosted.TerraNone:auc.primary
\newcommand{\TerraNoneAucUCurrent}{83.7}                                % docs/research/manuscript-three-pass/analysis.json:hosted.TerraNone:auc.primary
\newcommand{\TerraNoneAucLucky}{1.1}                                    % docs/research/manuscript-three-pass/analysis.json:hosted.TerraNone:auc.primary
\newcommand{\TerraNoneAucLinear}{67.0}                                  % docs/research/manuscript-three-pass/analysis.json:hosted.TerraNone:auc.sensitivity.Linear
\newcommand{\TerraNoneAucNarrow}{59.7}                                  % docs/research/manuscript-three-pass/analysis.json:hosted.TerraNone:auc.sensitivity.Narrow
\newcommand{\TerraNoneAucHalfFour}{49.2}                                % docs/research/manuscript-three-pass/analysis.json:hosted.TerraNone:auc.sensitivity.HalfFour
\newcommand{\TerraNoneAucOneEight}{64.6}                                % docs/research/manuscript-three-pass/analysis.json:hosted.TerraNone:auc.sensitivity.OneEight
\newcommand{\TerraNoneAucTenthFour}{30.4}                               % docs/research/manuscript-three-pass/analysis.json:hosted.TerraNone:auc.sensitivity.TenthFour
\newcommand{\TerraNoneAucTenthEight}{37.3}                              % docs/research/manuscript-three-pass/analysis.json:hosted.TerraNone:auc.sensitivity.TenthEight
\newcommand{\AstraAuc}{49.5}                                            % docs/research/manuscript-three-pass/analysis.json:hosted.Astra:auc.primary
\newcommand{\AstraDebugAuc}{49.8}                                       % docs/research/manuscript-three-pass/analysis.json:hosted.Astra:auc.primary
\newcommand{\AstraAssemblyAuc}{49.1}                                    % docs/research/manuscript-three-pass/analysis.json:hosted.Astra:auc.primary
\newcommand{\AstraSupportAuc}{47.0}                                     % docs/research/manuscript-three-pass/analysis.json:hosted.Astra:auc.primary
\newcommand{\AstraPresenterAuc}{52.2}                                   % docs/research/manuscript-three-pass/analysis.json:hosted.Astra:auc.primary
\newcommand{\AstraDebugAAuc}{49.0}                                      % docs/research/manuscript-three-pass/analysis.json:hosted.Astra:auc.primary
\newcommand{\AstraDebugBAuc}{50.6}                                      % docs/research/manuscript-three-pass/analysis.json:hosted.Astra:auc.primary
\newcommand{\AstraAssemblyAAuc}{50.3}                                   % docs/research/manuscript-three-pass/analysis.json:hosted.Astra:auc.primary
\newcommand{\AstraAssemblyBAuc}{47.9}                                   % docs/research/manuscript-three-pass/analysis.json:hosted.Astra:auc.primary
\newcommand{\AstraSupportAAuc}{49.1}                                    % docs/research/manuscript-three-pass/analysis.json:hosted.Astra:auc.primary
\newcommand{\AstraSupportBAuc}{44.8}                                    % docs/research/manuscript-three-pass/analysis.json:hosted.Astra:auc.primary
\newcommand{\AstraPresenterAAuc}{50.5}                                  % docs/research/manuscript-three-pass/analysis.json:hosted.Astra:auc.primary
\newcommand{\AstraPresenterBAuc}{53.8}                                  % docs/research/manuscript-three-pass/analysis.json:hosted.Astra:auc.primary
\newcommand{\AstraAucGap}{50.4}                                         % docs/research/manuscript-three-pass/analysis.json:hosted.Astra:auc.primary
\newcommand{\AstraAucOracle}{49.5}                                      % docs/research/manuscript-three-pass/analysis.json:hosted.Astra:auc.primary
\newcommand{\AstraAucProduct}{49.4}                                     % docs/research/manuscript-three-pass/analysis.json:hosted.Astra:auc.primary
\newcommand{\AstraAucUCurrent}{100.0}                                   % docs/research/manuscript-three-pass/analysis.json:hosted.Astra:auc.primary
\newcommand{\AstraAucStale}{47.4}                                       % docs/research/manuscript-three-pass/analysis.json:hosted.Astra:auc.primary
\newcommand{\AstraAucLucky}{0.0}                                        % docs/research/manuscript-three-pass/analysis.json:hosted.Astra:auc.primary
\newcommand{\AstraAucLinear}{67.9}                                      % docs/research/manuscript-three-pass/analysis.json:hosted.Astra:auc.sensitivity.Linear
\newcommand{\AstraAucNarrow}{50.9}                                      % docs/research/manuscript-three-pass/analysis.json:hosted.Astra:auc.sensitivity.Narrow
\newcommand{\AstraAucHalfFour}{38.7}                                    % docs/research/manuscript-three-pass/analysis.json:hosted.Astra:auc.sensitivity.HalfFour
\newcommand{\AstraAucOneEight}{61.2}                                    % docs/research/manuscript-three-pass/analysis.json:hosted.Astra:auc.sensitivity.OneEight
\newcommand{\AstraAucTenthFour}{23.2}                                   % docs/research/manuscript-three-pass/analysis.json:hosted.Astra:auc.sensitivity.TenthFour
\newcommand{\AstraAucTenthEight}{32.5}                                  % docs/research/manuscript-three-pass/analysis.json:hosted.Astra:auc.sensitivity.TenthEight
\newcommand{\JevAuc}{58.5}                                              % docs/research/manuscript-three-pass/analysis.json:hosted.Jev:auc.primary
\newcommand{\JevDebugAuc}{38.8}                                         % docs/research/manuscript-three-pass/analysis.json:hosted.Jev:auc.primary
\newcommand{\JevAssemblyAuc}{61.2}                                      % docs/research/manuscript-three-pass/analysis.json:hosted.Jev:auc.primary
\newcommand{\JevSupportAuc}{67.8}                                       % docs/research/manuscript-three-pass/analysis.json:hosted.Jev:auc.primary
\newcommand{\JevPresenterAuc}{66.2}                                     % docs/research/manuscript-three-pass/analysis.json:hosted.Jev:auc.primary
\newcommand{\JevDebugAAuc}{39.4}                                        % docs/research/manuscript-three-pass/analysis.json:hosted.Jev:auc.primary
\newcommand{\JevDebugBAuc}{38.2}                                        % docs/research/manuscript-three-pass/analysis.json:hosted.Jev:auc.primary
\newcommand{\JevAssemblyAAuc}{64.5}                                     % docs/research/manuscript-three-pass/analysis.json:hosted.Jev:auc.primary
\newcommand{\JevAssemblyBAuc}{57.8}                                     % docs/research/manuscript-three-pass/analysis.json:hosted.Jev:auc.primary
\newcommand{\JevSupportAAuc}{61.8}                                      % docs/research/manuscript-three-pass/analysis.json:hosted.Jev:auc.primary
\newcommand{\JevSupportBAuc}{73.9}                                      % docs/research/manuscript-three-pass/analysis.json:hosted.Jev:auc.primary
\newcommand{\JevPresenterAAuc}{63.7}                                    % docs/research/manuscript-three-pass/analysis.json:hosted.Jev:auc.primary
\newcommand{\JevPresenterBAuc}{68.7}                                    % docs/research/manuscript-three-pass/analysis.json:hosted.Jev:auc.primary
\newcommand{\JevAucGap}{3.9}                                            % docs/research/manuscript-three-pass/analysis.json:hosted.Jev:auc.primary
\newcommand{\JevAucOracle}{93.3}                                        % docs/research/manuscript-three-pass/analysis.json:hosted.Jev:auc.primary
\newcommand{\JevAucProduct}{58.2}                                       % docs/research/manuscript-three-pass/analysis.json:hosted.Jev:auc.primary
\newcommand{\JevAucUCurrent}{62.4}                                      % docs/research/manuscript-three-pass/analysis.json:hosted.Jev:auc.primary
\newcommand{\JevAucStale}{3.7}                                          % docs/research/manuscript-three-pass/analysis.json:hosted.Jev:auc.primary
\newcommand{\JevAucJudgment}{35.1}                                      % docs/research/manuscript-three-pass/analysis.json:hosted.Jev:auc.primary
\newcommand{\JevAucLucky}{0.3}                                          % docs/research/manuscript-three-pass/analysis.json:hosted.Jev:auc.primary
\newcommand{\JevAucLinear}{60.3}                                        % docs/research/manuscript-three-pass/analysis.json:hosted.Jev:auc.sensitivity.Linear
\newcommand{\JevAucNarrow}{59.3}                                        % docs/research/manuscript-three-pass/analysis.json:hosted.Jev:auc.sensitivity.Narrow
\newcommand{\JevAucHalfFour}{57.6}                                      % docs/research/manuscript-three-pass/analysis.json:hosted.Jev:auc.sensitivity.HalfFour
\newcommand{\JevAucOneEight}{60.0}                                      % docs/research/manuscript-three-pass/analysis.json:hosted.Jev:auc.sensitivity.OneEight
\newcommand{\JevAucTenthFour}{48.8}                                     % docs/research/manuscript-three-pass/analysis.json:hosted.Jev:auc.sensitivity.TenthFour
\newcommand{\JevAucTenthEight}{50.8}                                    % docs/research/manuscript-three-pass/analysis.json:hosted.Jev:auc.sensitivity.TenthEight
\newcommand{\LowEffortStaleRoundedMin}{42}                              % minimum primary stale share of low-effort GPT settings, rounded to whole percent
\newcommand{\LowEffortStaleRoundedMax}{47}                              % maximum primary stale share of low-effort GPT settings, rounded to whole percent
\newcommand{\JevJudgmentRounded}{35}                                    % Jev primary judgment share, rounded to whole percent
\newcommand{\AucMaxRefinementChange}{0.0009}                            % max successive quadrature-grid change, all integrated scenario metrics and sensitivity conditions (points)
\newcommand{\AucProductMaxDeviation}{6.3}                               % max per-scenario absolute difference: log-AUC minus U times oracle log-AUC (points)
\newcommand{\AucProductGapMin}{0.1}                                     % min macro log-AUC minus mean per-scenario U times oracle log-AUC (points)
\newcommand{\AucProductGapMax}{2.4}                                     % max of the same (points)
\newcommand{\AucLuckyMax}{1.4}                                          % max integrated lucky share over settings
\newcommand{\JevMinusTerraNoneAuc}{3.0}                                 % \JevAuc minus \TerraNoneAuc (points, from the printed values)
\newcommand{\JevMinusLunaAuc}{13.0}                                     % \JevAuc minus \LunaAuc (points, from the printed values)
\newcommand{\JevMinusTerraAuc}{6.4}                                     % \JevAuc minus \TerraAuc (points, from the printed values)
\newcommand{\JevMinusAstraAuc}{9.0}                                     % \JevAuc minus \AstraAuc (points, from the printed values)
\newcommand{\LunaAucSD}{0.70}                                           % docs/research/manuscript-three-pass/analysis.json:passes.hosted_reports.Luna
\newcommand{\LunaNoneAucSD}{1.12}                                       % docs/research/manuscript-three-pass/analysis.json:passes.hosted_reports.LunaNone
\newcommand{\TerraAucSD}{3.48}                                          % docs/research/manuscript-three-pass/analysis.json:passes.hosted_reports.Terra
\newcommand{\TerraNoneAucSD}{1.29}                                      % docs/research/manuscript-three-pass/analysis.json:passes.hosted_reports.TerraNone
\newcommand{\AstraAucSD}{4.44}                                          % docs/research/manuscript-three-pass/analysis.json:passes.hosted_reports.Astra
\newcommand{\JevAucSD}{1.10}                                            % docs/research/manuscript-three-pass/analysis.json:passes.hosted_reports.Jev
\newcommand{\BaselineFirstOption}{0.4}                                  % docs/lite/results/four-family/gpt-5.6-luna-low/analysis.json:baselines.first_option.overall
\newcommand{\BaselineFirstOptionStates}{2}                              % docs/lite/results/four-family/gpt-5.6-luna-low/analysis.json:baselines.first_option.per_episode.*.accuracy (x states per scenario, summed)
\newcommand{\BaselineLexical}{0.0}                                      % docs/lite/results/four-family/gpt-5.6-luna-low/analysis.json:baselines.lexical_overlap.overall
\newcommand{\BaselineLexicalStates}{0}                                  % docs/lite/results/four-family/gpt-5.6-luna-low/analysis.json:baselines.lexical_overlap.per_episode.*.accuracy (x states per scenario, summed)
\newcommand{\MaxWorkers}{32}                                            % docs/lite/results/four-family/gpt-5.6-luna-low/analysis.json:config.workers (identical in all runs)
\newcommand{\MaxAttempts}{5}                                            % docs/lite/results/four-family/gpt-5.6-luna-low/analysis.json:config.max_attempts (identical in all runs)
\newcommand{\RequestTimeoutSec}{20}                                     % docs/lite/results/four-family/gpt-5.6-luna-low/analysis.json:config.request_timeout_s (identical in all runs)
\newcommand{\SdkRetries}{0}                                             % docs/lite/results/four-family/gpt-5.6-luna-low/analysis.json:config.sdk_retries (identical in all runs)
\newcommand{\NetQuantilePct}{10}                                        % docs/lite/results/four-family/gpt-5.6-luna-low/analysis.json:network_adjustment.estimator
\newcommand{\NetRangeLevel}{95}                                         % docs/lite/results/four-family/gpt-5.6-luna-low/analysis.json:network_adjustment.estimator
\newcommand{\NetBootstrapResamples}{500}                                % docs/lite/results/four-family/gpt-5.6-luna-low/analysis.json:network_adjustment.estimator
\newcommand{\NetBootstrapBlock}{10}                                     % docs/lite/results/four-family/gpt-5.6-luna-low/analysis.json:network_adjustment.estimator
\newcommand{\NetBootstrapSeed}{0}                                       % docs/lite/results/four-family/gpt-5.6-luna-low/analysis.json:network_adjustment.estimator
\newcommand{\RecordingDatesUTC}{2026-09-28, 2026-09-29, 2026-10-02 and 2026-10-03} % docs/research/manuscript-three-pass/analysis.json:passes.hosted_reports.started_at_utc
\newcommand{\PairTerraLunaBoth}{1262}                                   % docs/research/manuscript-three-pass/analysis.json:same state and ordinal pass counts
\newcommand{\PairTerraLunaLunaOnly}{33}                                 % docs/research/manuscript-three-pass/analysis.json:same state and ordinal pass counts
\newcommand{\PairTerraLunaTerraOnly}{117}                               % docs/research/manuscript-three-pass/analysis.json:same state and ordinal pass counts
\newcommand{\PairTerraLunaNeither}{28}                                  % docs/research/manuscript-three-pass/analysis.json:same state and ordinal pass counts
\newcommand{\PairTerraLunaDebugBoth}{317}                               % docs/research/manuscript-three-pass/analysis.json:same state and ordinal pass counts
\newcommand{\PairTerraLunaDebugLunaOnly}{3}                             % docs/research/manuscript-three-pass/analysis.json:same state and ordinal pass counts
\newcommand{\PairTerraLunaDebugTerraOnly}{40}                           % docs/research/manuscript-three-pass/analysis.json:same state and ordinal pass counts
\newcommand{\PairTerraLunaDebugNeither}{0}                              % docs/research/manuscript-three-pass/analysis.json:same state and ordinal pass counts
\newcommand{\PairTerraLunaAssemblyBoth}{305}                            % docs/research/manuscript-three-pass/analysis.json:same state and ordinal pass counts
\newcommand{\PairTerraLunaAssemblyLunaOnly}{8}                          % docs/research/manuscript-three-pass/analysis.json:same state and ordinal pass counts
\newcommand{\PairTerraLunaAssemblyTerraOnly}{24}                        % docs/research/manuscript-three-pass/analysis.json:same state and ordinal pass counts
\newcommand{\PairTerraLunaAssemblyNeither}{23}                          % docs/research/manuscript-three-pass/analysis.json:same state and ordinal pass counts
\newcommand{\PairTerraLunaSupportBoth}{300}                             % docs/research/manuscript-three-pass/analysis.json:same state and ordinal pass counts
\newcommand{\PairTerraLunaSupportLunaOnly}{14}                          % docs/research/manuscript-three-pass/analysis.json:same state and ordinal pass counts
\newcommand{\PairTerraLunaSupportTerraOnly}{42}                         % docs/research/manuscript-three-pass/analysis.json:same state and ordinal pass counts
\newcommand{\PairTerraLunaSupportNeither}{4}                            % docs/research/manuscript-three-pass/analysis.json:same state and ordinal pass counts
\newcommand{\PairTerraLunaPresenterBoth}{340}                           % docs/research/manuscript-three-pass/analysis.json:same state and ordinal pass counts
\newcommand{\PairTerraLunaPresenterLunaOnly}{8}                         % docs/research/manuscript-three-pass/analysis.json:same state and ordinal pass counts
\newcommand{\PairTerraLunaPresenterTerraOnly}{11}                       % docs/research/manuscript-three-pass/analysis.json:same state and ordinal pass counts
\newcommand{\PairTerraLunaPresenterNeither}{1}                          % docs/research/manuscript-three-pass/analysis.json:same state and ordinal pass counts
\newcommand{\PairJevLunaBoth}{851}                                      % docs/research/manuscript-three-pass/analysis.json:same state and ordinal pass counts
\newcommand{\PairJevLunaLunaOnly}{444}                                  % docs/research/manuscript-three-pass/analysis.json:same state and ordinal pass counts
\newcommand{\PairJevLunaJevOnly}{47}                                    % docs/research/manuscript-three-pass/analysis.json:same state and ordinal pass counts
\newcommand{\PairJevLunaNeither}{98}                                    % docs/research/manuscript-three-pass/analysis.json:same state and ordinal pass counts
\newcommand{\PairJevLunaDebugBoth}{140}                                 % docs/research/manuscript-three-pass/analysis.json:same state and ordinal pass counts
\newcommand{\PairJevLunaDebugLunaOnly}{180}                             % docs/research/manuscript-three-pass/analysis.json:same state and ordinal pass counts
\newcommand{\PairJevLunaDebugJevOnly}{8}                                % docs/research/manuscript-three-pass/analysis.json:same state and ordinal pass counts
\newcommand{\PairJevLunaDebugNeither}{32}                               % docs/research/manuscript-three-pass/analysis.json:same state and ordinal pass counts
\newcommand{\PairJevLunaAssemblyBoth}{228}                              % docs/research/manuscript-three-pass/analysis.json:same state and ordinal pass counts
\newcommand{\PairJevLunaAssemblyLunaOnly}{85}                           % docs/research/manuscript-three-pass/analysis.json:same state and ordinal pass counts
\newcommand{\PairJevLunaAssemblyJevOnly}{6}                             % docs/research/manuscript-three-pass/analysis.json:same state and ordinal pass counts
\newcommand{\PairJevLunaAssemblyNeither}{41}                            % docs/research/manuscript-three-pass/analysis.json:same state and ordinal pass counts
\newcommand{\PairJevLunaSupportBoth}{234}                               % docs/research/manuscript-three-pass/analysis.json:same state and ordinal pass counts
\newcommand{\PairJevLunaSupportLunaOnly}{80}                            % docs/research/manuscript-three-pass/analysis.json:same state and ordinal pass counts
\newcommand{\PairJevLunaSupportJevOnly}{25}                             % docs/research/manuscript-three-pass/analysis.json:same state and ordinal pass counts
\newcommand{\PairJevLunaSupportNeither}{21}                             % docs/research/manuscript-three-pass/analysis.json:same state and ordinal pass counts
\newcommand{\PairJevLunaPresenterBoth}{249}                             % docs/research/manuscript-three-pass/analysis.json:same state and ordinal pass counts
\newcommand{\PairJevLunaPresenterLunaOnly}{99}                          % docs/research/manuscript-three-pass/analysis.json:same state and ordinal pass counts
\newcommand{\PairJevLunaPresenterJevOnly}{8}                            % docs/research/manuscript-three-pass/analysis.json:same state and ordinal pass counts
\newcommand{\PairJevLunaPresenterNeither}{4}                            % docs/research/manuscript-three-pass/analysis.json:same state and ordinal pass counts
\newcommand{\LunaMinusJevUntimed}{27.5}                                 % docs/research/manuscript-three-pass/analysis.json:hosted.Jev minus docs/research/manuscript-three-pass/analysis.json:hosted.Luna: Untimed, printed percentage points
\newcommand{\PairLunaNoneLunaBoth}{605}                                 % docs/research/manuscript-three-pass/analysis.json:same state and ordinal pass counts
\newcommand{\PairLunaNoneLunaLunaOnly}{690}                             % docs/research/manuscript-three-pass/analysis.json:same state and ordinal pass counts
\newcommand{\PairLunaNoneLunaLunaNoneOnly}{22}                          % docs/research/manuscript-three-pass/analysis.json:same state and ordinal pass counts
\newcommand{\PairLunaNoneLunaNeither}{123}                              % docs/research/manuscript-three-pass/analysis.json:same state and ordinal pass counts
\newcommand{\PairLunaNoneLunaDebugBoth}{60}                             % docs/research/manuscript-three-pass/analysis.json:same state and ordinal pass counts
\newcommand{\PairLunaNoneLunaDebugLunaOnly}{260}                        % docs/research/manuscript-three-pass/analysis.json:same state and ordinal pass counts
\newcommand{\PairLunaNoneLunaDebugLunaNoneOnly}{4}                      % docs/research/manuscript-three-pass/analysis.json:same state and ordinal pass counts
\newcommand{\PairLunaNoneLunaDebugNeither}{36}                          % docs/research/manuscript-three-pass/analysis.json:same state and ordinal pass counts
\newcommand{\PairLunaNoneLunaAssemblyBoth}{101}                         % docs/research/manuscript-three-pass/analysis.json:same state and ordinal pass counts
\newcommand{\PairLunaNoneLunaAssemblyLunaOnly}{212}                     % docs/research/manuscript-three-pass/analysis.json:same state and ordinal pass counts
\newcommand{\PairLunaNoneLunaAssemblyLunaNoneOnly}{3}                   % docs/research/manuscript-three-pass/analysis.json:same state and ordinal pass counts
\newcommand{\PairLunaNoneLunaAssemblyNeither}{44}                       % docs/research/manuscript-three-pass/analysis.json:same state and ordinal pass counts
\newcommand{\PairLunaNoneLunaSupportBoth}{223}                          % docs/research/manuscript-three-pass/analysis.json:same state and ordinal pass counts
\newcommand{\PairLunaNoneLunaSupportLunaOnly}{91}                       % docs/research/manuscript-three-pass/analysis.json:same state and ordinal pass counts
\newcommand{\PairLunaNoneLunaSupportLunaNoneOnly}{11}                   % docs/research/manuscript-three-pass/analysis.json:same state and ordinal pass counts
\newcommand{\PairLunaNoneLunaSupportNeither}{35}                        % docs/research/manuscript-three-pass/analysis.json:same state and ordinal pass counts
\newcommand{\PairLunaNoneLunaPresenterBoth}{221}                        % docs/research/manuscript-three-pass/analysis.json:same state and ordinal pass counts
\newcommand{\PairLunaNoneLunaPresenterLunaOnly}{127}                    % docs/research/manuscript-three-pass/analysis.json:same state and ordinal pass counts
\newcommand{\PairLunaNoneLunaPresenterLunaNoneOnly}{4}                  % docs/research/manuscript-three-pass/analysis.json:same state and ordinal pass counts
\newcommand{\PairLunaNoneLunaPresenterNeither}{8}                       % docs/research/manuscript-three-pass/analysis.json:same state and ordinal pass counts
\newcommand{\LunaMinusLunaNoneUntimed}{46.4}                            % docs/research/manuscript-three-pass/analysis.json:hosted.LunaNone minus docs/research/manuscript-three-pass/analysis.json:hosted.Luna: Untimed, printed percentage points
\newcommand{\PairTerraNoneTerraBoth}{1161}                              % docs/research/manuscript-three-pass/analysis.json:same state and ordinal pass counts
\newcommand{\PairTerraNoneTerraTerraOnly}{218}                          % docs/research/manuscript-three-pass/analysis.json:same state and ordinal pass counts
\newcommand{\PairTerraNoneTerraTerraNoneOnly}{18}                       % docs/research/manuscript-three-pass/analysis.json:same state and ordinal pass counts
\newcommand{\PairTerraNoneTerraNeither}{43}                             % docs/research/manuscript-three-pass/analysis.json:same state and ordinal pass counts
\newcommand{\PairTerraNoneTerraDebugBoth}{260}                          % docs/research/manuscript-three-pass/analysis.json:same state and ordinal pass counts
\newcommand{\PairTerraNoneTerraDebugTerraOnly}{97}                      % docs/research/manuscript-three-pass/analysis.json:same state and ordinal pass counts
\newcommand{\PairTerraNoneTerraDebugTerraNoneOnly}{0}                   % docs/research/manuscript-three-pass/analysis.json:same state and ordinal pass counts
\newcommand{\PairTerraNoneTerraDebugNeither}{3}                         % docs/research/manuscript-three-pass/analysis.json:same state and ordinal pass counts
\newcommand{\PairTerraNoneTerraAssemblyBoth}{267}                       % docs/research/manuscript-three-pass/analysis.json:same state and ordinal pass counts
\newcommand{\PairTerraNoneTerraAssemblyTerraOnly}{62}                   % docs/research/manuscript-three-pass/analysis.json:same state and ordinal pass counts
\newcommand{\PairTerraNoneTerraAssemblyTerraNoneOnly}{2}                % docs/research/manuscript-three-pass/analysis.json:same state and ordinal pass counts
\newcommand{\PairTerraNoneTerraAssemblyNeither}{29}                     % docs/research/manuscript-three-pass/analysis.json:same state and ordinal pass counts
\newcommand{\PairTerraNoneTerraSupportBoth}{325}                        % docs/research/manuscript-three-pass/analysis.json:same state and ordinal pass counts
\newcommand{\PairTerraNoneTerraSupportTerraOnly}{17}                    % docs/research/manuscript-three-pass/analysis.json:same state and ordinal pass counts
\newcommand{\PairTerraNoneTerraSupportTerraNoneOnly}{12}                % docs/research/manuscript-three-pass/analysis.json:same state and ordinal pass counts
\newcommand{\PairTerraNoneTerraSupportNeither}{6}                       % docs/research/manuscript-three-pass/analysis.json:same state and ordinal pass counts
\newcommand{\PairTerraNoneTerraPresenterBoth}{309}                      % docs/research/manuscript-three-pass/analysis.json:same state and ordinal pass counts
\newcommand{\PairTerraNoneTerraPresenterTerraOnly}{42}                  % docs/research/manuscript-three-pass/analysis.json:same state and ordinal pass counts
\newcommand{\PairTerraNoneTerraPresenterTerraNoneOnly}{4}               % docs/research/manuscript-three-pass/analysis.json:same state and ordinal pass counts
\newcommand{\PairTerraNoneTerraPresenterNeither}{5}                     % docs/research/manuscript-three-pass/analysis.json:same state and ordinal pass counts
\newcommand{\PairAstraTerraBoth}{1377}                                  % docs/research/manuscript-three-pass/analysis.json:same state and ordinal pass counts
\newcommand{\PairAstraTerraTerraOnly}{2}                                % docs/research/manuscript-three-pass/analysis.json:same state and ordinal pass counts
\newcommand{\PairAstraTerraAstraOnly}{61}                               % docs/research/manuscript-three-pass/analysis.json:same state and ordinal pass counts
\newcommand{\PairAstraTerraNeither}{0}                                  % docs/research/manuscript-three-pass/analysis.json:same state and ordinal pass counts
\newcommand{\PairAstraTerraDebugBoth}{357}                              % docs/research/manuscript-three-pass/analysis.json:same state and ordinal pass counts
\newcommand{\PairAstraTerraDebugTerraOnly}{0}                           % docs/research/manuscript-three-pass/analysis.json:same state and ordinal pass counts
\newcommand{\PairAstraTerraDebugAstraOnly}{3}                           % docs/research/manuscript-three-pass/analysis.json:same state and ordinal pass counts
\newcommand{\PairAstraTerraDebugNeither}{0}                             % docs/research/manuscript-three-pass/analysis.json:same state and ordinal pass counts
\newcommand{\PairAstraTerraAssemblyBoth}{327}                           % docs/research/manuscript-three-pass/analysis.json:same state and ordinal pass counts
\newcommand{\PairAstraTerraAssemblyTerraOnly}{2}                        % docs/research/manuscript-three-pass/analysis.json:same state and ordinal pass counts
\newcommand{\PairAstraTerraAssemblyAstraOnly}{31}                       % docs/research/manuscript-three-pass/analysis.json:same state and ordinal pass counts
\newcommand{\PairAstraTerraAssemblyNeither}{0}                          % docs/research/manuscript-three-pass/analysis.json:same state and ordinal pass counts
\newcommand{\PairAstraTerraSupportBoth}{342}                            % docs/research/manuscript-three-pass/analysis.json:same state and ordinal pass counts
\newcommand{\PairAstraTerraSupportTerraOnly}{0}                         % docs/research/manuscript-three-pass/analysis.json:same state and ordinal pass counts
\newcommand{\PairAstraTerraSupportAstraOnly}{18}                        % docs/research/manuscript-three-pass/analysis.json:same state and ordinal pass counts
\newcommand{\PairAstraTerraSupportNeither}{0}                           % docs/research/manuscript-three-pass/analysis.json:same state and ordinal pass counts
\newcommand{\PairAstraTerraPresenterBoth}{351}                          % docs/research/manuscript-three-pass/analysis.json:same state and ordinal pass counts
\newcommand{\PairAstraTerraPresenterTerraOnly}{0}                       % docs/research/manuscript-three-pass/analysis.json:same state and ordinal pass counts
\newcommand{\PairAstraTerraPresenterAstraOnly}{9}                       % docs/research/manuscript-three-pass/analysis.json:same state and ordinal pass counts
\newcommand{\PairAstraTerraPresenterNeither}{0}                         % docs/research/manuscript-three-pass/analysis.json:same state and ordinal pass counts
\newcommand{\TerraMinusJevUntimed}{33.4}                                % docs/lite/results/four-family/{gpt-5.6-terra-low,jev-latest}/analysis.json (difference, points)
\newcommand{\LunaNoneLatencySavingSec}{1.13}                            % latency_s.p50 of Luna minus LunaNone (s; docs/lite/results/four-family/*/analysis.json)
\newcommand{\TerraNoneLatencySavingSec}{0.77}                           % latency_s.p50 of Terra minus TerraNone (s; docs/lite/results/four-family/*/analysis.json)
\newcommand{\GPTMinScenarioGap}{29.9}                                   % scores.per_episode[*] untimed_decision_accuracy - time_accuracy (points, minimum over Luna low, Terra low and scenarios)
\newcommand{\AstraMinScenarioGap}{43.2}                                 % docs/lite/results/four-family/gpt-6-astra-low/analysis.json:scores.per_episode[*] untimed - in-force (points, minimum)
\newcommand{\JevMaxScenarioGap}{4.2}                                    % docs/lite/results/four-family/jev-latest/analysis.json:scores.per_episode[*] untimed - in-force (points, maximum)
\newcommand{\LunaOutputTokMin}{83}                                      % runs/lite-v1-gpt-5.6-luna-low-four-family-retry-v1/events.jsonl:response.usage.output_tokens (minimum)
\newcommand{\TerraOutputTokMin}{39}                                     % runs/lite-v1-gpt-5.6-terra-low-four-family-retry-v1/events.jsonl:response.usage.output_tokens (minimum)
\newcommand{\AstraOutputTokMin}{39}                                     % runs/lite-v1-gpt-6-astra-low-four-family-retry-v1/events.jsonl:response.usage.output_tokens (minimum)
\newcommand{\JevOutputTokMin}{354}                                      % runs/lite-v1-jev-latest-four-family-retry-v1/events.jsonl:response.usage.output_tokens (minimum)
\newcommand{\MaxReplayPhysicalDiff}{0.01}                               % docs/research/manuscript-three-pass/analysis.json:three-pass statistics; extrema after aggregation, physical gaps across passes
\newcommand{\RetryLaterDelaysSec}{0.5, 1 and 2}                         % src/streamdecisionbench/lite/runtime.py backoff with config retry_delay_s and max_attempts
\newcommand{\JevReportStep}{0.01}                                       % src/streamdecisionbench/jev.py:REPORTED_STEP
\newcommand{\NumWindows}{51}                                            % paper/analysis/lite_window.py over docs/lite/results/four-family/*/analysis.json scores.per_episode[*].intervals (candidate windows)
\newcommand{\TrajWindowSec}{20}                                         % paper/analysis/lite_window.py over docs/lite/results/four-family/*/analysis.json scores.per_episode[*].intervals (window length)
\newcommand{\TrajWindowStart}{72}                                       % paper/analysis/lite_window.py over docs/lite/results/four-family/*/analysis.json scores.per_episode[*].intervals (chosen window)
\newcommand{\TrajWindowEnd}{92}                                         % paper/analysis/lite_window.py over docs/lite/results/four-family/*/analysis.json scores.per_episode[*].intervals (chosen window)
\newcommand{\TrajWindowMaxDev}{6.4}                                     % paper/analysis/lite_window.py over docs/lite/results/four-family/*/analysis.json scores.per_episode[*].intervals (largest |window share - in-force accuracy|, points)
\newcommand{\LunaTrajShare}{41.3}                                       % paper/analysis/lite_window.py over docs/lite/results/four-family/*/analysis.json scores.per_episode[*].intervals (correct share in the chosen window)
\newcommand{\TerraTrajShare}{45.5}                                      % paper/analysis/lite_window.py over docs/lite/results/four-family/*/analysis.json scores.per_episode[*].intervals (correct share in the chosen window)
\newcommand{\JevTrajShare}{63.3}                                        % paper/analysis/lite_window.py over docs/lite/results/four-family/*/analysis.json scores.per_episode[*].intervals (correct share in the chosen window)
\newcommand{\AuditRederivationsPerScenarioWord}{two}                    % \AuditRederivationsPerScenario spelled out
\newcommand{\AuditRederivations}{12}                                    % paper/notes/audit_record.json:checks[role in blind, walk] (count)
\newcommand{\AuditCodeTextChecksWord}{three}                            % \AuditCodeTextChecks spelled out
\newcommand{\AuditStructuralChecksWord}{one}                            % \AuditStructuralChecks spelled out
\newcommand{\AuditLensesWord}{three}                                    % \AuditLenses spelled out
\newcommand{\AuditFindingGroups}{57}                                    % paper/notes/audit_record.json:grouped_findings (count)
\newcommand{\AuditReferenceFindingGroups}{31}                           % paper/notes/audit_record.json:grouped_findings (judged by the reference-answer lenses)
\newcommand{\AuditReferenceFindingsRejected}{19}                        % paper/notes/audit_record.json:grouped_findings (reference-answer lenses, status rejected)
\newcommand{\AuditReferenceFindingsConfirmed}{12}                       % paper/notes/audit_record.json:grouped_findings (reference-answer lenses, status confirmed)
\newcommand{\AuditAmbiguitiesWord}{three}                               % \AuditAmbiguities spelled out
\newcommand{\AuditAffectedStates}{16}                                   % paper/notes/audit_record.json:grouped_findings[ambiguous].affected_t (summed)
\newcommand{\AuditClarifiedSentencesWord}{two}                          % paper/notes/audit_summary.md: Validity audit / Fix
\newcommand{\NumSettingsWord}{six}                                      % \NumSettings spelled out
\newcommand{\AstraReasoningResponses}{80}                               % docs/research/manuscript-three-pass/analysis.json:Astra usage across passes
\newcommand{\AstraStates}{1440}                                         % docs/research/manuscript-three-pass/analysis.json:Astra three-pass agreement
\newcommand{\AstraSituationStates}{8}                                   % paper/notes/audit_record.json: grouped finding g50 ticks (lite_assembly_a; stale inspection PASS while the stage is earlier)
\newcommand{\AstraSituationCorrect}{7}                                  % runs/lite-v1-gpt-6-astra-low-four-family-retry-v1/events.jsonl: states of that finding whose composed decision equals gold
\newcommand{\AstraMissScenarioTitle}{Gearbox cover: mismatched part, joint rework and quality hold} % data/lite/v1/lite_assembly_a.json:title
\newcommand{\AstraMissScenarioLabel}{\AssemblyLabel{} A}                % lite_assembly_a (as labelled in the per-scenario table)
\newcommand{\AstraMissStep}{32}                                         % runs/lite-v1-gpt-6-astra-low-four-family-retry-v1/events.jsonl: the only response whose composed decision differs from gold (lite_assembly_a)
\newcommand{\AstraMissPrevStep}{31}                                     % runs/lite-v1-gpt-6-astra-low-four-family-retry-v1/events.jsonl: lite_assembly_a t=31, composed decision equal to gold
\newcommand{\AstraMissRoute}{repair}                                    % runs/lite-v1-gpt-6-astra-low-four-family-retry-v1/events.jsonl: lite_assembly_a t=32 pred.route
\newcommand{\AstraMissTarget}{inspection}                               % runs/lite-v1-gpt-6-astra-low-four-family-retry-v1/events.jsonl: lite_assembly_a t=32 pred.target
\newcommand{\AstraMissMethod}{complete\_missing}                        % runs/lite-v1-gpt-6-astra-low-four-family-retry-v1/events.jsonl: lite_assembly_a t=32 pred.method
\newcommand{\AstraMissRefRoute}{advance}                                % data/lite/v1/lite_assembly_a.json:steps[32].gold.route
\newcommand{\AstraMissRefStage}{fasten}                                 % data/lite/v1/lite_assembly_a.json:steps[32].gold.stage
\newcommand{\AstraMissRefNextStep}{inspection}                          % data/lite/v1/lite_assembly_a.json:steps[32].gold.next_step
\newcommand{\AstraMissInspectionStep}{12}                               % data/lite/v1/lite_assembly_a.json:steps[32].state.station_log[kind=inspection].tick (code PASS)
\newcommand{\AstraMissScrewStep}{27}                                    % data/lite/v1/lite_assembly_a.json:steps[32].state.station_log[kind=replace_screw].tick (target J2)
\newcommand{\AstraMissRundownStep}{29}                                  % data/lite/v1/lite_assembly_a.json:steps[32].state.station_log[kind=rundown, target J2, after the replacement].tick
\newcommand{\AstraMissRundownNm}{26.0}                                  % data/lite/v1/lite_assembly_a.json:steps[32].state.station_log[kind=rundown, tick 29].value
\newcommand{\AstraMissTorqueMinNm}{22.0}                                % data/lite/v1/lite_assembly_a.json:steps[32].state.work_instruction.torque_nm.J2[0]
\newcommand{\AstraMissTorqueMaxNm}{26.0}                                % data/lite/v1/lite_assembly_a.json:steps[32].state.work_instruction.torque_nm.J2[1]
\newcommand{\AstraMinusJevUntimed}{37.5}                                % docs/research/manuscript-three-pass/analysis.json:three-pass printed untimed difference
\newcommand{\AstraMissStepPassThree}{50}                                % docs/research/manuscript-three-pass/analysis.json:passes[2].transition_errors.Astra
\newcommand{\TrajJevOne}{56.7}                                          % docs/research/manuscript-three-pass/analysis.json:trajectory.standalone.Jev.fixed.1
\newcommand{\NetRemovedLowEffortMinGap}{16.7}                           % three-pass family means: untimed minus network-removed high endpoint
\newcommand{\HeadlineJevAuc}{58.50}                                     % docs/research/manuscript-three-pass/analysis.json:three independent pass scores for Jev (equal mean, %)
\newcommand{\HeadlineJevSD}{1.10}                                       % docs/research/manuscript-three-pass/analysis.json:three independent pass scores for Jev (sample SD, percentage points)
\newcommand{\HeadlineTerraNoneAuc}{55.46}                               % docs/research/manuscript-three-pass/analysis.json:three independent pass scores for TerraNone (equal mean, %)
\newcommand{\HeadlineTerraNoneSD}{1.29}                                 % docs/research/manuscript-three-pass/analysis.json:three independent pass scores for TerraNone (sample SD, percentage points)
\newcommand{\HeadlineKevTwentySevenAuc}{62.15}                          % docs/research/manuscript-three-pass/analysis.json:three independent pass scores for KevTwentySeven (equal mean, %)
\newcommand{\HeadlineKevTwentySevenSD}{0.17}                            % docs/research/manuscript-three-pass/analysis.json:three independent pass scores for KevTwentySeven (sample SD, percentage points)
\newcommand{\HeadlineKevTwentySevenHybridAuc}{66.78}                    % docs/research/manuscript-three-pass/analysis.json:three independent pass scores for KevTwentySevenHybrid (equal mean, %)
\newcommand{\HeadlineKevTwentySevenHybridSD}{0.35}                      % docs/research/manuscript-three-pass/analysis.json:three independent pass scores for KevTwentySevenHybrid (sample SD, percentage points)
\newcommand{\OwMeasuredHours}{7.25}                                     % docs/research/manuscript-three-pass/analysis.json:measurement_budget.local_run_hours; measured runs only
\newcommand{\LayaEnglishRoute}{17.71}                                   % docs/research/manuscript-three-pass/analysis.json:passes[*].laya_diagnostics.LayaEnglish: pooled route_correct/states; gold-active fields include route
\newcommand{\LayaEnglishGoldFields}{18.56}                              % docs/research/manuscript-three-pass/analysis.json:passes[*].laya_diagnostics.LayaEnglish: pooled field_correct/fields; gold-active fields include route
\newcommand{\LayaEnglishStates}{1440}                                   % docs/research/manuscript-three-pass/analysis.json:passes[*].laya_diagnostics.LayaEnglish:total state evaluations
\newcommand{\LayaEnglishFields}{5220}                                   % docs/research/manuscript-three-pass/analysis.json:passes[*].laya_diagnostics.LayaEnglish:total gold-active fields, including route
\newcommand{\LayaTypedRoute}{25.00}                                     % docs/research/manuscript-three-pass/analysis.json:passes[*].laya_diagnostics.LayaTyped: pooled route_correct/states; gold-active fields include route
\newcommand{\LayaTypedGoldFields}{29.77}                                % docs/research/manuscript-three-pass/analysis.json:passes[*].laya_diagnostics.LayaTyped: pooled field_correct/fields; gold-active fields include route
\newcommand{\LayaMultilingualRoute}{26.88}                              % docs/research/manuscript-three-pass/analysis.json:passes[*].laya_diagnostics.LayaMultilingual: pooled route_correct/states; gold-active fields include route
\newcommand{\LayaMultilingualGoldFields}{27.59}                         % docs/research/manuscript-three-pass/analysis.json:passes[*].laya_diagnostics.LayaMultilingual: pooled field_correct/fields; gold-active fields include route
\newcommand{\PairingKevTwentySevenMin}{66.74}                           % docs/research/manuscript-three-pass/analysis.json:pairing_sensitivity.pairs.KevTwentySeven+TerraNone.min (%, six one-to-one matchings)
\newcommand{\PairingKevTwentySevenMax}{66.78}                           % docs/research/manuscript-three-pass/analysis.json:pairing_sensitivity.pairs.KevTwentySeven+TerraNone.max (%, six one-to-one matchings)
\newcommand{\PairingJevTerraNoneMin}{65.90}                             % docs/research/manuscript-three-pass/analysis.json:pairing_sensitivity.pairs.Jev+TerraNone.min (%, six one-to-one matchings)
\newcommand{\PairingJevTerraNoneMax}{65.95}                             % docs/research/manuscript-three-pass/analysis.json:pairing_sensitivity.pairs.Jev+TerraNone.max (%, six one-to-one matchings)
\providecommand{\LunaRowLabel}{Luna low}
\providecommand{\LunaNoneRowLabel}{Luna none}
\providecommand{\TerraRowLabel}{Terra low}
\providecommand{\TerraNoneRowLabel}{Terra none}
\providecommand{\AstraRowLabel}{Astra low}
\providecommand{\JevRowLabel}{Jev}
\providecommand{\DebugLabel}{IDE debugging}
\providecommand{\AssemblyLabel}{Assembly station}
\providecommand{\SupportLabel}{Support call}
\providecommand{\PresenterLabel}{Presenter control}

\newcommand{\TabBenchmarkBody}{%
  \DebugLabel & A & \DebugAQuestions & \DebugARouteOptions & \DebugAOptionsMin--\DebugAOptionsMax & \DebugATransitions \\
   & B & \DebugBQuestions & \DebugBRouteOptions & \DebugBOptionsMin--\DebugBOptionsMax & \DebugBTransitions \\
  \midrule
  \AssemblyLabel & A & \AssemblyAQuestions & \AssemblyARouteOptions & \AssemblyAOptionsMin--\AssemblyAOptionsMax & \AssemblyATransitions \\
   & B & \AssemblyBQuestions & \AssemblyBRouteOptions & \AssemblyBOptionsMin--\AssemblyBOptionsMax & \AssemblyBTransitions \\
  \midrule
  \SupportLabel & A & \SupportAQuestions & \SupportARouteOptions & \SupportAOptionsMin--\SupportAOptionsMax & \SupportATransitions \\
   & B & \SupportBQuestions & \SupportBRouteOptions & \SupportBOptionsMin--\SupportBOptionsMax & \SupportBTransitions \\
  \midrule
  \PresenterLabel & A & \PresenterAQuestions & \PresenterARouteOptions & \PresenterAOptionsMin--\PresenterAOptionsMax & \PresenterATransitions \\
   & B & \PresenterBQuestions & \PresenterBRouteOptions & \PresenterBOptionsMin--\PresenterBOptionsMax & \PresenterBTransitions \\
  \midrule
  \multicolumn{2}{l}{All} & \QuestionsMin--\QuestionsMax & \RouteOptionsMin--\RouteOptionsMax & \OptionsMin--\OptionsMax & \TransitionsTotal \\
}

\newcommand{\TabFamilyBody}{%
  \multicolumn{5}{l}{\textit{\DebugLabel}} \\
  \LunaRowLabel & \LunaDebugUntimed & \LunaDebugInForce & \LunaDebugNetRemoved & (\LunaDebugNetRemovedLow--\LunaDebugNetRemovedHigh) \\
  \LunaNoneRowLabel & \LunaNoneDebugUntimed & \LunaNoneDebugInForce & \LunaNoneDebugNetRemoved & (\LunaNoneDebugNetRemovedLow--\LunaNoneDebugNetRemovedHigh) \\
  \TerraRowLabel & \TerraDebugUntimed & \TerraDebugInForce & \TerraDebugNetRemoved & (\TerraDebugNetRemovedLow--\TerraDebugNetRemovedHigh) \\
  \TerraNoneRowLabel & \TerraNoneDebugUntimed & \TerraNoneDebugInForce & \TerraNoneDebugNetRemoved & (\TerraNoneDebugNetRemovedLow--\TerraNoneDebugNetRemovedHigh) \\
  \AstraRowLabel & \AstraDebugUntimed & \AstraDebugInForce & \AstraDebugNetRemoved & (\AstraDebugNetRemovedLow--\AstraDebugNetRemovedHigh) \\
  \JevRowLabel & \JevDebugUntimed & \JevDebugInForce & \JevDebugNetRemoved & (\JevDebugNetRemovedLow--\JevDebugNetRemovedHigh) \\
  \midrule
  \multicolumn{5}{l}{\textit{\AssemblyLabel}} \\
  \LunaRowLabel & \LunaAssemblyUntimed & \LunaAssemblyInForce & \LunaAssemblyNetRemoved & (\LunaAssemblyNetRemovedLow--\LunaAssemblyNetRemovedHigh) \\
  \LunaNoneRowLabel & \LunaNoneAssemblyUntimed & \LunaNoneAssemblyInForce & \LunaNoneAssemblyNetRemoved & (\LunaNoneAssemblyNetRemovedLow--\LunaNoneAssemblyNetRemovedHigh) \\
  \TerraRowLabel & \TerraAssemblyUntimed & \TerraAssemblyInForce & \TerraAssemblyNetRemoved & (\TerraAssemblyNetRemovedLow--\TerraAssemblyNetRemovedHigh) \\
  \TerraNoneRowLabel & \TerraNoneAssemblyUntimed & \TerraNoneAssemblyInForce & \TerraNoneAssemblyNetRemoved & (\TerraNoneAssemblyNetRemovedLow--\TerraNoneAssemblyNetRemovedHigh) \\
  \AstraRowLabel & \AstraAssemblyUntimed & \AstraAssemblyInForce & \AstraAssemblyNetRemoved & (\AstraAssemblyNetRemovedLow--\AstraAssemblyNetRemovedHigh) \\
  \JevRowLabel & \JevAssemblyUntimed & \JevAssemblyInForce & \JevAssemblyNetRemoved & (\JevAssemblyNetRemovedLow--\JevAssemblyNetRemovedHigh) \\
  \midrule
  \multicolumn{5}{l}{\textit{\SupportLabel}} \\
  \LunaRowLabel & \LunaSupportUntimed & \LunaSupportInForce & \LunaSupportNetRemoved & (\LunaSupportNetRemovedLow--\LunaSupportNetRemovedHigh) \\
  \LunaNoneRowLabel & \LunaNoneSupportUntimed & \LunaNoneSupportInForce & \LunaNoneSupportNetRemoved & (\LunaNoneSupportNetRemovedLow--\LunaNoneSupportNetRemovedHigh) \\
  \TerraRowLabel & \TerraSupportUntimed & \TerraSupportInForce & \TerraSupportNetRemoved & (\TerraSupportNetRemovedLow--\TerraSupportNetRemovedHigh) \\
  \TerraNoneRowLabel & \TerraNoneSupportUntimed & \TerraNoneSupportInForce & \TerraNoneSupportNetRemoved & (\TerraNoneSupportNetRemovedLow--\TerraNoneSupportNetRemovedHigh) \\
  \AstraRowLabel & \AstraSupportUntimed & \AstraSupportInForce & \AstraSupportNetRemoved & (\AstraSupportNetRemovedLow--\AstraSupportNetRemovedHigh) \\
  \JevRowLabel & \JevSupportUntimed & \JevSupportInForce & \JevSupportNetRemoved & (\JevSupportNetRemovedLow--\JevSupportNetRemovedHigh) \\
  \midrule
  \multicolumn{5}{l}{\textit{\PresenterLabel}} \\
  \LunaRowLabel & \LunaPresenterUntimed & \LunaPresenterInForce & \LunaPresenterNetRemoved & (\LunaPresenterNetRemovedLow--\LunaPresenterNetRemovedHigh) \\
  \LunaNoneRowLabel & \LunaNonePresenterUntimed & \LunaNonePresenterInForce & \LunaNonePresenterNetRemoved & (\LunaNonePresenterNetRemovedLow--\LunaNonePresenterNetRemovedHigh) \\
  \TerraRowLabel & \TerraPresenterUntimed & \TerraPresenterInForce & \TerraPresenterNetRemoved & (\TerraPresenterNetRemovedLow--\TerraPresenterNetRemovedHigh) \\
  \TerraNoneRowLabel & \TerraNonePresenterUntimed & \TerraNonePresenterInForce & \TerraNonePresenterNetRemoved & (\TerraNonePresenterNetRemovedLow--\TerraNonePresenterNetRemovedHigh) \\
  \AstraRowLabel & \AstraPresenterUntimed & \AstraPresenterInForce & \AstraPresenterNetRemoved & (\AstraPresenterNetRemovedLow--\AstraPresenterNetRemovedHigh) \\
  \JevRowLabel & \JevPresenterUntimed & \JevPresenterInForce & \JevPresenterNetRemoved & (\JevPresenterNetRemovedLow--\JevPresenterNetRemovedHigh) \\
}

\newcommand{\TabLatencyBody}{%
  \LunaRowLabel & \LunaNetSec & (\LunaNetSecLow--\LunaNetSecHigh) & \LunaPrefillMsPerKTok & \LunaDecodeMsPerTok & \LunaLatencyMedian & \LunaLatencyPNinetyFive \\
  \LunaNoneRowLabel & \LunaNoneNetSec & (\LunaNoneNetSecLow--\LunaNoneNetSecHigh) & \LunaNonePrefillMsPerKTok & \LunaNoneDecodeMsPerTok & \LunaNoneLatencyMedian & \LunaNoneLatencyPNinetyFive \\
  \TerraRowLabel & \TerraNetSec & (\TerraNetSecLow--\TerraNetSecHigh) & \TerraPrefillMsPerKTok & \TerraDecodeMsPerTok & \TerraLatencyMedian & \TerraLatencyPNinetyFive \\
  \TerraNoneRowLabel & \TerraNoneNetSec & (\TerraNoneNetSecLow--\TerraNoneNetSecHigh) & \TerraNonePrefillMsPerKTok & \TerraNoneDecodeMsPerTok & \TerraNoneLatencyMedian & \TerraNoneLatencyPNinetyFive \\
  \AstraRowLabel & \AstraNetSec & (\AstraNetSecLow--\AstraNetSecHigh) & \AstraPrefillMsPerKTok & \AstraDecodeMsPerTok & \AstraLatencyMedian & \AstraLatencyPNinetyFive \\
  \JevRowLabel & \JevNetSec & (\JevNetSecLow--\JevNetSecHigh) & \JevPrefillMsPerKTok & -- & \JevLatencyMedian & \JevLatencyPNinetyFive \\
}

\newcommand{\TabScenarioBody}{%
  \DebugLabel{} A & \LunaRowLabel & \LunaDebugAUntimed & \LunaDebugAAuc & \LunaDebugAInForce & \LunaDebugAInForceSeg & \LunaDebugAErrNoDecisionSec & \LunaDebugAErrStaleSec & \LunaDebugAErrJudgmentSec & \LunaDebugAErrCompoundSec & \LunaDebugALatencyMedian \\
   & \LunaNoneRowLabel & \LunaNoneDebugAUntimed & \LunaNoneDebugAAuc & \LunaNoneDebugAInForce & \LunaNoneDebugAInForceSeg & \LunaNoneDebugAErrNoDecisionSec & \LunaNoneDebugAErrStaleSec & \LunaNoneDebugAErrJudgmentSec & \LunaNoneDebugAErrCompoundSec & \LunaNoneDebugALatencyMedian \\
   & \TerraRowLabel & \TerraDebugAUntimed & \TerraDebugAAuc & \TerraDebugAInForce & \TerraDebugAInForceSeg & \TerraDebugAErrNoDecisionSec & \TerraDebugAErrStaleSec & \TerraDebugAErrJudgmentSec & \TerraDebugAErrCompoundSec & \TerraDebugALatencyMedian \\
   & \TerraNoneRowLabel & \TerraNoneDebugAUntimed & \TerraNoneDebugAAuc & \TerraNoneDebugAInForce & \TerraNoneDebugAInForceSeg & \TerraNoneDebugAErrNoDecisionSec & \TerraNoneDebugAErrStaleSec & \TerraNoneDebugAErrJudgmentSec & \TerraNoneDebugAErrCompoundSec & \TerraNoneDebugALatencyMedian \\
   & \AstraRowLabel & \AstraDebugAUntimed & \AstraDebugAAuc & \AstraDebugAInForce & \AstraDebugAInForceSeg & \AstraDebugAErrNoDecisionSec & \AstraDebugAErrStaleSec & \AstraDebugAErrJudgmentSec & \AstraDebugAErrCompoundSec & \AstraDebugALatencyMedian \\
   & \JevRowLabel & \JevDebugAUntimed & \JevDebugAAuc & \JevDebugAInForce & \JevDebugAInForceSeg & \JevDebugAErrNoDecisionSec & \JevDebugAErrStaleSec & \JevDebugAErrJudgmentSec & \JevDebugAErrCompoundSec & \JevDebugALatencyMedian \\
  \midrule
  \DebugLabel{} B & \LunaRowLabel & \LunaDebugBUntimed & \LunaDebugBAuc & \LunaDebugBInForce & \LunaDebugBInForceSeg & \LunaDebugBErrNoDecisionSec & \LunaDebugBErrStaleSec & \LunaDebugBErrJudgmentSec & \LunaDebugBErrCompoundSec & \LunaDebugBLatencyMedian \\
   & \LunaNoneRowLabel & \LunaNoneDebugBUntimed & \LunaNoneDebugBAuc & \LunaNoneDebugBInForce & \LunaNoneDebugBInForceSeg & \LunaNoneDebugBErrNoDecisionSec & \LunaNoneDebugBErrStaleSec & \LunaNoneDebugBErrJudgmentSec & \LunaNoneDebugBErrCompoundSec & \LunaNoneDebugBLatencyMedian \\
   & \TerraRowLabel & \TerraDebugBUntimed & \TerraDebugBAuc & \TerraDebugBInForce & \TerraDebugBInForceSeg & \TerraDebugBErrNoDecisionSec & \TerraDebugBErrStaleSec & \TerraDebugBErrJudgmentSec & \TerraDebugBErrCompoundSec & \TerraDebugBLatencyMedian \\
   & \TerraNoneRowLabel & \TerraNoneDebugBUntimed & \TerraNoneDebugBAuc & \TerraNoneDebugBInForce & \TerraNoneDebugBInForceSeg & \TerraNoneDebugBErrNoDecisionSec & \TerraNoneDebugBErrStaleSec & \TerraNoneDebugBErrJudgmentSec & \TerraNoneDebugBErrCompoundSec & \TerraNoneDebugBLatencyMedian \\
   & \AstraRowLabel & \AstraDebugBUntimed & \AstraDebugBAuc & \AstraDebugBInForce & \AstraDebugBInForceSeg & \AstraDebugBErrNoDecisionSec & \AstraDebugBErrStaleSec & \AstraDebugBErrJudgmentSec & \AstraDebugBErrCompoundSec & \AstraDebugBLatencyMedian \\
   & \JevRowLabel & \JevDebugBUntimed & \JevDebugBAuc & \JevDebugBInForce & \JevDebugBInForceSeg & \JevDebugBErrNoDecisionSec & \JevDebugBErrStaleSec & \JevDebugBErrJudgmentSec & \JevDebugBErrCompoundSec & \JevDebugBLatencyMedian \\
  \midrule
  \AssemblyLabel{} A & \LunaRowLabel & \LunaAssemblyAUntimed & \LunaAssemblyAAuc & \LunaAssemblyAInForce & \LunaAssemblyAInForceSeg & \LunaAssemblyAErrNoDecisionSec & \LunaAssemblyAErrStaleSec & \LunaAssemblyAErrJudgmentSec & \LunaAssemblyAErrCompoundSec & \LunaAssemblyALatencyMedian \\
   & \LunaNoneRowLabel & \LunaNoneAssemblyAUntimed & \LunaNoneAssemblyAAuc & \LunaNoneAssemblyAInForce & \LunaNoneAssemblyAInForceSeg & \LunaNoneAssemblyAErrNoDecisionSec & \LunaNoneAssemblyAErrStaleSec & \LunaNoneAssemblyAErrJudgmentSec & \LunaNoneAssemblyAErrCompoundSec & \LunaNoneAssemblyALatencyMedian \\
   & \TerraRowLabel & \TerraAssemblyAUntimed & \TerraAssemblyAAuc & \TerraAssemblyAInForce & \TerraAssemblyAInForceSeg & \TerraAssemblyAErrNoDecisionSec & \TerraAssemblyAErrStaleSec & \TerraAssemblyAErrJudgmentSec & \TerraAssemblyAErrCompoundSec & \TerraAssemblyALatencyMedian \\
   & \TerraNoneRowLabel & \TerraNoneAssemblyAUntimed & \TerraNoneAssemblyAAuc & \TerraNoneAssemblyAInForce & \TerraNoneAssemblyAInForceSeg & \TerraNoneAssemblyAErrNoDecisionSec & \TerraNoneAssemblyAErrStaleSec & \TerraNoneAssemblyAErrJudgmentSec & \TerraNoneAssemblyAErrCompoundSec & \TerraNoneAssemblyALatencyMedian \\
   & \AstraRowLabel & \AstraAssemblyAUntimed & \AstraAssemblyAAuc & \AstraAssemblyAInForce & \AstraAssemblyAInForceSeg & \AstraAssemblyAErrNoDecisionSec & \AstraAssemblyAErrStaleSec & \AstraAssemblyAErrJudgmentSec & \AstraAssemblyAErrCompoundSec & \AstraAssemblyALatencyMedian \\
   & \JevRowLabel & \JevAssemblyAUntimed & \JevAssemblyAAuc & \JevAssemblyAInForce & \JevAssemblyAInForceSeg & \JevAssemblyAErrNoDecisionSec & \JevAssemblyAErrStaleSec & \JevAssemblyAErrJudgmentSec & \JevAssemblyAErrCompoundSec & \JevAssemblyALatencyMedian \\
  \midrule
  \AssemblyLabel{} B & \LunaRowLabel & \LunaAssemblyBUntimed & \LunaAssemblyBAuc & \LunaAssemblyBInForce & \LunaAssemblyBInForceSeg & \LunaAssemblyBErrNoDecisionSec & \LunaAssemblyBErrStaleSec & \LunaAssemblyBErrJudgmentSec & \LunaAssemblyBErrCompoundSec & \LunaAssemblyBLatencyMedian \\
   & \LunaNoneRowLabel & \LunaNoneAssemblyBUntimed & \LunaNoneAssemblyBAuc & \LunaNoneAssemblyBInForce & \LunaNoneAssemblyBInForceSeg & \LunaNoneAssemblyBErrNoDecisionSec & \LunaNoneAssemblyBErrStaleSec & \LunaNoneAssemblyBErrJudgmentSec & \LunaNoneAssemblyBErrCompoundSec & \LunaNoneAssemblyBLatencyMedian \\
   & \TerraRowLabel & \TerraAssemblyBUntimed & \TerraAssemblyBAuc & \TerraAssemblyBInForce & \TerraAssemblyBInForceSeg & \TerraAssemblyBErrNoDecisionSec & \TerraAssemblyBErrStaleSec & \TerraAssemblyBErrJudgmentSec & \TerraAssemblyBErrCompoundSec & \TerraAssemblyBLatencyMedian \\
   & \TerraNoneRowLabel & \TerraNoneAssemblyBUntimed & \TerraNoneAssemblyBAuc & \TerraNoneAssemblyBInForce & \TerraNoneAssemblyBInForceSeg & \TerraNoneAssemblyBErrNoDecisionSec & \TerraNoneAssemblyBErrStaleSec & \TerraNoneAssemblyBErrJudgmentSec & \TerraNoneAssemblyBErrCompoundSec & \TerraNoneAssemblyBLatencyMedian \\
   & \AstraRowLabel & \AstraAssemblyBUntimed & \AstraAssemblyBAuc & \AstraAssemblyBInForce & \AstraAssemblyBInForceSeg & \AstraAssemblyBErrNoDecisionSec & \AstraAssemblyBErrStaleSec & \AstraAssemblyBErrJudgmentSec & \AstraAssemblyBErrCompoundSec & \AstraAssemblyBLatencyMedian \\
   & \JevRowLabel & \JevAssemblyBUntimed & \JevAssemblyBAuc & \JevAssemblyBInForce & \JevAssemblyBInForceSeg & \JevAssemblyBErrNoDecisionSec & \JevAssemblyBErrStaleSec & \JevAssemblyBErrJudgmentSec & \JevAssemblyBErrCompoundSec & \JevAssemblyBLatencyMedian \\
  \midrule
  \SupportLabel{} A & \LunaRowLabel & \LunaSupportAUntimed & \LunaSupportAAuc & \LunaSupportAInForce & \LunaSupportAInForceSeg & \LunaSupportAErrNoDecisionSec & \LunaSupportAErrStaleSec & \LunaSupportAErrJudgmentSec & \LunaSupportAErrCompoundSec & \LunaSupportALatencyMedian \\
   & \LunaNoneRowLabel & \LunaNoneSupportAUntimed & \LunaNoneSupportAAuc & \LunaNoneSupportAInForce & \LunaNoneSupportAInForceSeg & \LunaNoneSupportAErrNoDecisionSec & \LunaNoneSupportAErrStaleSec & \LunaNoneSupportAErrJudgmentSec & \LunaNoneSupportAErrCompoundSec & \LunaNoneSupportALatencyMedian \\
   & \TerraRowLabel & \TerraSupportAUntimed & \TerraSupportAAuc & \TerraSupportAInForce & \TerraSupportAInForceSeg & \TerraSupportAErrNoDecisionSec & \TerraSupportAErrStaleSec & \TerraSupportAErrJudgmentSec & \TerraSupportAErrCompoundSec & \TerraSupportALatencyMedian \\
   & \TerraNoneRowLabel & \TerraNoneSupportAUntimed & \TerraNoneSupportAAuc & \TerraNoneSupportAInForce & \TerraNoneSupportAInForceSeg & \TerraNoneSupportAErrNoDecisionSec & \TerraNoneSupportAErrStaleSec & \TerraNoneSupportAErrJudgmentSec & \TerraNoneSupportAErrCompoundSec & \TerraNoneSupportALatencyMedian \\
   & \AstraRowLabel & \AstraSupportAUntimed & \AstraSupportAAuc & \AstraSupportAInForce & \AstraSupportAInForceSeg & \AstraSupportAErrNoDecisionSec & \AstraSupportAErrStaleSec & \AstraSupportAErrJudgmentSec & \AstraSupportAErrCompoundSec & \AstraSupportALatencyMedian \\
   & \JevRowLabel & \JevSupportAUntimed & \JevSupportAAuc & \JevSupportAInForce & \JevSupportAInForceSeg & \JevSupportAErrNoDecisionSec & \JevSupportAErrStaleSec & \JevSupportAErrJudgmentSec & \JevSupportAErrCompoundSec & \JevSupportALatencyMedian \\
  \midrule
  \SupportLabel{} B & \LunaRowLabel & \LunaSupportBUntimed & \LunaSupportBAuc & \LunaSupportBInForce & \LunaSupportBInForceSeg & \LunaSupportBErrNoDecisionSec & \LunaSupportBErrStaleSec & \LunaSupportBErrJudgmentSec & \LunaSupportBErrCompoundSec & \LunaSupportBLatencyMedian \\
   & \LunaNoneRowLabel & \LunaNoneSupportBUntimed & \LunaNoneSupportBAuc & \LunaNoneSupportBInForce & \LunaNoneSupportBInForceSeg & \LunaNoneSupportBErrNoDecisionSec & \LunaNoneSupportBErrStaleSec & \LunaNoneSupportBErrJudgmentSec & \LunaNoneSupportBErrCompoundSec & \LunaNoneSupportBLatencyMedian \\
   & \TerraRowLabel & \TerraSupportBUntimed & \TerraSupportBAuc & \TerraSupportBInForce & \TerraSupportBInForceSeg & \TerraSupportBErrNoDecisionSec & \TerraSupportBErrStaleSec & \TerraSupportBErrJudgmentSec & \TerraSupportBErrCompoundSec & \TerraSupportBLatencyMedian \\
   & \TerraNoneRowLabel & \TerraNoneSupportBUntimed & \TerraNoneSupportBAuc & \TerraNoneSupportBInForce & \TerraNoneSupportBInForceSeg & \TerraNoneSupportBErrNoDecisionSec & \TerraNoneSupportBErrStaleSec & \TerraNoneSupportBErrJudgmentSec & \TerraNoneSupportBErrCompoundSec & \TerraNoneSupportBLatencyMedian \\
   & \AstraRowLabel & \AstraSupportBUntimed & \AstraSupportBAuc & \AstraSupportBInForce & \AstraSupportBInForceSeg & \AstraSupportBErrNoDecisionSec & \AstraSupportBErrStaleSec & \AstraSupportBErrJudgmentSec & \AstraSupportBErrCompoundSec & \AstraSupportBLatencyMedian \\
   & \JevRowLabel & \JevSupportBUntimed & \JevSupportBAuc & \JevSupportBInForce & \JevSupportBInForceSeg & \JevSupportBErrNoDecisionSec & \JevSupportBErrStaleSec & \JevSupportBErrJudgmentSec & \JevSupportBErrCompoundSec & \JevSupportBLatencyMedian \\
  \midrule
  \PresenterLabel{} A & \LunaRowLabel & \LunaPresenterAUntimed & \LunaPresenterAAuc & \LunaPresenterAInForce & \LunaPresenterAInForceSeg & \LunaPresenterAErrNoDecisionSec & \LunaPresenterAErrStaleSec & \LunaPresenterAErrJudgmentSec & \LunaPresenterAErrCompoundSec & \LunaPresenterALatencyMedian \\
   & \LunaNoneRowLabel & \LunaNonePresenterAUntimed & \LunaNonePresenterAAuc & \LunaNonePresenterAInForce & \LunaNonePresenterAInForceSeg & \LunaNonePresenterAErrNoDecisionSec & \LunaNonePresenterAErrStaleSec & \LunaNonePresenterAErrJudgmentSec & \LunaNonePresenterAErrCompoundSec & \LunaNonePresenterALatencyMedian \\
   & \TerraRowLabel & \TerraPresenterAUntimed & \TerraPresenterAAuc & \TerraPresenterAInForce & \TerraPresenterAInForceSeg & \TerraPresenterAErrNoDecisionSec & \TerraPresenterAErrStaleSec & \TerraPresenterAErrJudgmentSec & \TerraPresenterAErrCompoundSec & \TerraPresenterALatencyMedian \\
   & \TerraNoneRowLabel & \TerraNonePresenterAUntimed & \TerraNonePresenterAAuc & \TerraNonePresenterAInForce & \TerraNonePresenterAInForceSeg & \TerraNonePresenterAErrNoDecisionSec & \TerraNonePresenterAErrStaleSec & \TerraNonePresenterAErrJudgmentSec & \TerraNonePresenterAErrCompoundSec & \TerraNonePresenterALatencyMedian \\
   & \AstraRowLabel & \AstraPresenterAUntimed & \AstraPresenterAAuc & \AstraPresenterAInForce & \AstraPresenterAInForceSeg & \AstraPresenterAErrNoDecisionSec & \AstraPresenterAErrStaleSec & \AstraPresenterAErrJudgmentSec & \AstraPresenterAErrCompoundSec & \AstraPresenterALatencyMedian \\
   & \JevRowLabel & \JevPresenterAUntimed & \JevPresenterAAuc & \JevPresenterAInForce & \JevPresenterAInForceSeg & \JevPresenterAErrNoDecisionSec & \JevPresenterAErrStaleSec & \JevPresenterAErrJudgmentSec & \JevPresenterAErrCompoundSec & \JevPresenterALatencyMedian \\
  \midrule
  \PresenterLabel{} B & \LunaRowLabel & \LunaPresenterBUntimed & \LunaPresenterBAuc & \LunaPresenterBInForce & \LunaPresenterBInForceSeg & \LunaPresenterBErrNoDecisionSec & \LunaPresenterBErrStaleSec & \LunaPresenterBErrJudgmentSec & \LunaPresenterBErrCompoundSec & \LunaPresenterBLatencyMedian \\
   & \LunaNoneRowLabel & \LunaNonePresenterBUntimed & \LunaNonePresenterBAuc & \LunaNonePresenterBInForce & \LunaNonePresenterBInForceSeg & \LunaNonePresenterBErrNoDecisionSec & \LunaNonePresenterBErrStaleSec & \LunaNonePresenterBErrJudgmentSec & \LunaNonePresenterBErrCompoundSec & \LunaNonePresenterBLatencyMedian \\
   & \TerraRowLabel & \TerraPresenterBUntimed & \TerraPresenterBAuc & \TerraPresenterBInForce & \TerraPresenterBInForceSeg & \TerraPresenterBErrNoDecisionSec & \TerraPresenterBErrStaleSec & \TerraPresenterBErrJudgmentSec & \TerraPresenterBErrCompoundSec & \TerraPresenterBLatencyMedian \\
   & \TerraNoneRowLabel & \TerraNonePresenterBUntimed & \TerraNonePresenterBAuc & \TerraNonePresenterBInForce & \TerraNonePresenterBInForceSeg & \TerraNonePresenterBErrNoDecisionSec & \TerraNonePresenterBErrStaleSec & \TerraNonePresenterBErrJudgmentSec & \TerraNonePresenterBErrCompoundSec & \TerraNonePresenterBLatencyMedian \\
   & \AstraRowLabel & \AstraPresenterBUntimed & \AstraPresenterBAuc & \AstraPresenterBInForce & \AstraPresenterBInForceSeg & \AstraPresenterBErrNoDecisionSec & \AstraPresenterBErrStaleSec & \AstraPresenterBErrJudgmentSec & \AstraPresenterBErrCompoundSec & \AstraPresenterBLatencyMedian \\
   & \JevRowLabel & \JevPresenterBUntimed & \JevPresenterBAuc & \JevPresenterBInForce & \JevPresenterBInForceSeg & \JevPresenterBErrNoDecisionSec & \JevPresenterBErrStaleSec & \JevPresenterBErrJudgmentSec & \JevPresenterBErrCompoundSec & \JevPresenterBLatencyMedian \\
}

\newcommand{\TabPaceBody}{%
  $\times$0.5 & 4.00 & \LunaInForceDelayHalf & \LunaNoneInForceDelayHalf & \TerraInForceDelayHalf & \TerraNoneInForceDelayHalf & \AstraInForceDelayHalf & \JevInForceDelayHalf \\
  $\times$2/3 & 3.00 & \LunaInForceDelayTwoThirds & \LunaNoneInForceDelayTwoThirds & \TerraInForceDelayTwoThirds & \TerraNoneInForceDelayTwoThirds & \AstraInForceDelayTwoThirds & \JevInForceDelayTwoThirds \\
  $\times$1 & 2.00 & \LunaInForce & \LunaNoneInForce & \TerraInForce & \TerraNoneInForce & \AstraInForce & \JevInForce \\
  $\times$1.5 & 1.33 & \LunaInForceDelayThreeHalves & \LunaNoneInForceDelayThreeHalves & \TerraInForceDelayThreeHalves & \TerraNoneInForceDelayThreeHalves & \AstraInForceDelayThreeHalves & \JevInForceDelayThreeHalves \\
  $\times$2 & 1.00 & \LunaInForceDelayDouble & \LunaNoneInForceDelayDouble & \TerraInForceDelayDouble & \TerraNoneInForceDelayDouble & \AstraInForceDelayDouble & \JevInForceDelayDouble \\
}

\newcommand{\TabRegimeBody}{%
  SDB (eight scenarios) & -- & \RegimeLiteLuna & \RegimeLiteLunaNone & \RegimeLiteTerra & \RegimeLiteTerraNone & \RegimeLiteAstra & \RegimeLiteJev \\
  \midrule
  sparse & \RegimeSparseChanges & \RegimeSparseLuna & \RegimeSparseLunaNone & \RegimeSparseTerra & \RegimeSparseTerraNone & \RegimeSparseAstra & \RegimeSparseJev \\
  medium & \RegimeMediumChanges & \RegimeMediumLuna & \RegimeMediumLunaNone & \RegimeMediumTerra & \RegimeMediumTerraNone & \RegimeMediumAstra & \RegimeMediumJev \\
  uniform & \RegimeUniformChanges & \RegimeUniformLuna & \RegimeUniformLunaNone & \RegimeUniformTerra & \RegimeUniformTerraNone & \RegimeUniformAstra & \RegimeUniformJev \\
  dense & \RegimeDenseChanges & \RegimeDenseLuna & \RegimeDenseLunaNone & \RegimeDenseTerra & \RegimeDenseTerraNone & \RegimeDenseAstra & \RegimeDenseJev \\
  bursty & \RegimeBurstyChanges & \RegimeBurstyLuna & \RegimeBurstyLunaNone & \RegimeBurstyTerra & \RegimeBurstyTerraNone & \RegimeBurstyAstra & \RegimeBurstyJev \\
  long-tail dwell & \RegimeLongTailChanges & \RegimeLongTailLuna & \RegimeLongTailLunaNone & \RegimeLongTailTerra & \RegimeLongTailTerraNone & \RegimeLongTailAstra & \RegimeLongTailJev \\
  recurrent A-B-A & \RegimeRecurrentChanges & \RegimeRecurrentLuna & \RegimeRecurrentLunaNone & \RegimeRecurrentTerra & \RegimeRecurrentTerraNone & \RegimeRecurrentAstra & \RegimeRecurrentJev \\
}

\newcommand{\TabAucBody}{%
  \LunaRowLabel & \LunaDebugAuc & \LunaAssemblyAuc & \LunaSupportAuc & \LunaPresenterAuc & \LunaAuc~$\pm$~\LunaAucSD & \LunaUntimed \\
  \LunaNoneRowLabel & \LunaNoneDebugAuc & \LunaNoneAssemblyAuc & \LunaNoneSupportAuc & \LunaNonePresenterAuc & \LunaNoneAuc~$\pm$~\LunaNoneAucSD & \LunaNoneUntimed \\
  \TerraRowLabel & \TerraDebugAuc & \TerraAssemblyAuc & \TerraSupportAuc & \TerraPresenterAuc & \TerraAuc~$\pm$~\TerraAucSD & \TerraUntimed \\
  \TerraNoneRowLabel & \TerraNoneDebugAuc & \TerraNoneAssemblyAuc & \TerraNoneSupportAuc & \TerraNonePresenterAuc & \TerraNoneAuc~$\pm$~\TerraNoneAucSD & \TerraNoneUntimed \\
  \AstraRowLabel & \AstraDebugAuc & \AstraAssemblyAuc & \AstraSupportAuc & \AstraPresenterAuc & \AstraAuc~$\pm$~\AstraAucSD & \AstraUntimed \\
  \JevRowLabel & \JevDebugAuc & \JevAssemblyAuc & \JevSupportAuc & \JevPresenterAuc & \JevAuc~$\pm$~\JevAucSD & \JevUntimed \\
}

\newcommand{\TabAucSensitivityBody}{%
  \LunaRowLabel & \LunaAuc & \LunaAucLinear & \LunaAucNarrow & \LunaAucHalfFour & \LunaAucOneEight & \LunaAucTenthFour & \LunaAucTenthEight & \LunaCommonTwo \\
  \LunaNoneRowLabel & \LunaNoneAuc & \LunaNoneAucLinear & \LunaNoneAucNarrow & \LunaNoneAucHalfFour & \LunaNoneAucOneEight & \LunaNoneAucTenthFour & \LunaNoneAucTenthEight & \LunaNoneCommonTwo \\
  \TerraRowLabel & \TerraAuc & \TerraAucLinear & \TerraAucNarrow & \TerraAucHalfFour & \TerraAucOneEight & \TerraAucTenthFour & \TerraAucTenthEight & \TerraCommonTwo \\
  \TerraNoneRowLabel & \TerraNoneAuc & \TerraNoneAucLinear & \TerraNoneAucNarrow & \TerraNoneAucHalfFour & \TerraNoneAucOneEight & \TerraNoneAucTenthFour & \TerraNoneAucTenthEight & \TerraNoneCommonTwo \\
  \AstraRowLabel & \AstraAuc & \AstraAucLinear & \AstraAucNarrow & \AstraAucHalfFour & \AstraAucOneEight & \AstraAucTenthFour & \AstraAucTenthEight & \AstraCommonTwo \\
  \JevRowLabel & \JevAuc & \JevAucLinear & \JevAucNarrow & \JevAucHalfFour & \JevAucOneEight & \JevAucTenthFour & \JevAucTenthEight & \JevCommonTwo \\
}

\newcommand{\TabAucDecompositionBody}{%
  \LunaRowLabel & \LunaAucOracle & \LunaAucUCurrent & \LunaAucLucky & \LunaAucProduct & \LunaAuc \\
  \LunaNoneRowLabel & \LunaNoneAucOracle & \LunaNoneAucUCurrent & \LunaNoneAucLucky & \LunaNoneAucProduct & \LunaNoneAuc \\
  \TerraRowLabel & \TerraAucOracle & \TerraAucUCurrent & \TerraAucLucky & \TerraAucProduct & \TerraAuc \\
  \TerraNoneRowLabel & \TerraNoneAucOracle & \TerraNoneAucUCurrent & \TerraNoneAucLucky & \TerraNoneAucProduct & \TerraNoneAuc \\
  \AstraRowLabel & \AstraAucOracle & \AstraAucUCurrent & \AstraAucLucky & \AstraAucProduct & \AstraAuc \\
  \JevRowLabel & \JevAucOracle & \JevAucUCurrent & \JevAucLucky & \JevAucProduct & \JevAuc \\
}

\newcommand{\TrajNominalGap}{0.026770}
\newcommand{\TrajExampleScenario}{presenter A}
\newcommand{\TrajExampleChange}{clip state from pause to play}
\newcommand{\TrajExampleFastArrival}{60.304}
\newcommand{\TrajExampleOldArrival}{60.400}
\newcommand{\TrajExampleRestore}{62.376}
\newcommand{\TrajExampleLoss}{1.977}
\newcommand{\TrajExampleFastSource}{30}
\newcommand{\TrajExampleOldSource}{29}
\newcommand{\TrajLunaNearStates}{1200}
\newcommand{\TrajLunaNearErrors}{134}
\newcommand{\TrajLunaNearError}{11.2}
\newcommand{\TrajLunaFarStates}{240}
\newcommand{\TrajLunaFarErrors}{11}
\newcommand{\TrajLunaFarError}{4.6}
\newcommand{\TrajLunaNoneNearError}{57.1}
\newcommand{\TrajLunaNoneFarError}{53.3}
\newcommand{\TrajTerraNearErrors}{59}
\newcommand{\TrajTerraNearError}{4.9}
\newcommand{\TrajTerraFarErrors}{2}
\newcommand{\TrajTerraFarError}{0.8}
\newcommand{\TrajTerraNoneNearError}{20.0}
\newcommand{\TrajTerraNoneFarError}{8.8}
\newcommand{\TrajAstraNearError}{0.2}
\newcommand{\TrajAstraFarError}{0.0}
\newcommand{\TrajJevNearError}{38.9}
\newcommand{\TrajJevFarError}{31.3}
\newcommand{\TrajJevTwo}{59.5}
\newcommand{\TrajLunaFreshestOne}{56.7}
\newcommand{\TrajLunaFreshestTwo}{60.0}
\newcommand{\TrajLunaFreshestAuc}{63.1}
\newcommand{\TrajLunaLagOneOne}{55.6}
\newcommand{\TrajLunaLagOneTwo}{58.0}
\newcommand{\TrajLunaLagOneAuc}{62.4}
\newcommand{\TrajLunaArrivalOne}{45.8}
\newcommand{\TrajLunaArrivalTwo}{57.4}
\newcommand{\TrajLunaArrivalAuc}{58.5}
\newcommand{\TrajLunaNoneFreshestOne}{56.3}
\newcommand{\TrajLunaNoneFreshestTwo}{51.9}
\newcommand{\TrajLunaNoneFreshestAuc}{51.3}
\newcommand{\TrajLunaNoneLagOneOne}{42.3}
\newcommand{\TrajLunaNoneLagOneTwo}{51.8}
\newcommand{\TrajLunaNoneLagOneAuc}{47.6}
\newcommand{\TrajLunaNoneArrivalOne}{42.3}
\newcommand{\TrajLunaNoneArrivalTwo}{51.8}
\newcommand{\TrajLunaNoneArrivalAuc}{46.8}
\newcommand{\TrajTerraFreshestOne}{56.7}
\newcommand{\TrajTerraFreshestTwo}{62.0}
\newcommand{\TrajTerraFreshestAuc}{65.8}
\newcommand{\TrajTerraFreshestSlowShare}{12.7}
\newcommand{\TrajTerraFreshestProxy}{91.1}
\newcommand{\TrajTerraLagOneOne}{54.8}
\newcommand{\TrajTerraLagOneTwo}{61.6}
\newcommand{\TrajTerraLagOneAuc}{65.3}
\newcommand{\TrajTerraArrivalOne}{49.4}
\newcommand{\TrajTerraArrivalTwo}{61.6}
\newcommand{\TrajTerraArrivalAuc}{62.5}
\newcommand{\TrajTerraNoneFreshestOne}{57.0}
\newcommand{\TrajTerraNoneFreshestTwo}{67.3}
\newcommand{\TrajTerraNoneFreshestAuc}{65.9}
\newcommand{\TrajTerraNoneLagOneOne}{53.5}
\newcommand{\TrajTerraNoneLagOneTwo}{67.2}
\newcommand{\TrajTerraNoneLagOneAuc}{64.7}
\newcommand{\TrajTerraNoneArrivalOne}{53.4}
\newcommand{\TrajTerraNoneArrivalTwo}{67.2}
\newcommand{\TrajTerraNoneArrivalAuc}{63.8}
\newcommand{\TrajAstraFreshestOne}{56.7}
\newcommand{\TrajAstraFreshestTwo}{60.6}
\newcommand{\TrajAstraFreshestAuc}{65.7}
\newcommand{\TrajAstraLagOneOne}{55.4}
\newcommand{\TrajAstraLagOneTwo}{61.4}
\newcommand{\TrajAstraLagOneAuc}{65.4}
\newcommand{\TrajAstraArrivalOne}{47.2}
\newcommand{\TrajAstraArrivalTwo}{61.2}
\newcommand{\TrajAstraArrivalAuc}{61.7}
\newcommand{\TabTrajectoryBody}{%
\LunaRowLabel & Freshest & \TrajLunaFreshestOne & \TrajLunaFreshestTwo & \TrajLunaFreshestAuc \\
\LunaRowLabel & Lag $\leq 1$ & \TrajLunaLagOneOne & \TrajLunaLagOneTwo & \TrajLunaLagOneAuc \\
\LunaRowLabel & Override & \TrajLunaArrivalOne & \TrajLunaArrivalTwo & \TrajLunaArrivalAuc \\
\LunaNoneRowLabel & Freshest & \TrajLunaNoneFreshestOne & \TrajLunaNoneFreshestTwo & \TrajLunaNoneFreshestAuc \\
\LunaNoneRowLabel & Lag $\leq 1$ & \TrajLunaNoneLagOneOne & \TrajLunaNoneLagOneTwo & \TrajLunaNoneLagOneAuc \\
\LunaNoneRowLabel & Override & \TrajLunaNoneArrivalOne & \TrajLunaNoneArrivalTwo & \TrajLunaNoneArrivalAuc \\
\TerraRowLabel & Freshest & \TrajTerraFreshestOne & \TrajTerraFreshestTwo & \TrajTerraFreshestAuc \\
\TerraRowLabel & Lag $\leq 1$ & \TrajTerraLagOneOne & \TrajTerraLagOneTwo & \TrajTerraLagOneAuc \\
\TerraRowLabel & Override & \TrajTerraArrivalOne & \TrajTerraArrivalTwo & \TrajTerraArrivalAuc \\
\TerraNoneRowLabel & Freshest & \TrajTerraNoneFreshestOne & \TrajTerraNoneFreshestTwo & \TrajTerraNoneFreshestAuc \\
\TerraNoneRowLabel & Lag $\leq 1$ & \TrajTerraNoneLagOneOne & \TrajTerraNoneLagOneTwo & \TrajTerraNoneLagOneAuc \\
\TerraNoneRowLabel & Override & \TrajTerraNoneArrivalOne & \TrajTerraNoneArrivalTwo & \TrajTerraNoneArrivalAuc \\
\AstraRowLabel & Freshest & \TrajAstraFreshestOne & \TrajAstraFreshestTwo & \TrajAstraFreshestAuc \\
\AstraRowLabel & Lag $\leq 1$ & \TrajAstraLagOneOne & \TrajAstraLagOneTwo & \TrajAstraLagOneAuc \\
\AstraRowLabel & Override & \TrajAstraArrivalOne & \TrajAstraArrivalTwo & \TrajAstraArrivalAuc \\
}

\newcommand{\OwNumSettingsWord}{nine}
\newcommand{\OwTotalSettingsWord}{fifteen}
\newcommand{\OwNominalGap}{0.026770}
\newcommand{\OwLayaEnglishAuc}{0.42}
\newcommand{\OwLayaEnglishUntimed}{0.42}
\newcommand{\OwLayaEnglishMedian}{0.144}
\newcommand{\OwLayaEnglishTail}{0.256}
\newcommand{\OwLayaEnglishDebugAuc}{0.00}
\newcommand{\OwLayaEnglishAssemblyAuc}{0.00}
\newcommand{\OwLayaEnglishSupportAuc}{1.67}
\newcommand{\OwLayaEnglishPresenterAuc}{0.00}
\newcommand{\OwLayaEnglishFreshestAuc}{29.84}
\newcommand{\OwLayaEnglishFreshestRange}{4.09}
\newcommand{\OwLayaEnglishLagOneAuc}{36.00}
\newcommand{\OwLayaEnglishLagOneRange}{3.85}
\newcommand{\OwLayaEnglishArrivalAuc}{36.85}
\newcommand{\OwLayaEnglishArrivalRange}{3.24}
\newcommand{\OwLayaTypedAuc}{1.46}
\newcommand{\OwLayaTypedUntimed}{1.46}
\newcommand{\OwLayaTypedMedian}{0.143}
\newcommand{\OwLayaTypedTail}{0.254}
\newcommand{\OwLayaTypedDebugAuc}{0.00}
\newcommand{\OwLayaTypedAssemblyAuc}{0.00}
\newcommand{\OwLayaTypedSupportAuc}{1.67}
\newcommand{\OwLayaTypedPresenterAuc}{4.17}
\newcommand{\OwLayaTypedFreshestAuc}{30.45}
\newcommand{\OwLayaTypedFreshestRange}{3.99}
\newcommand{\OwLayaTypedLagOneAuc}{36.48}
\newcommand{\OwLayaTypedLagOneRange}{3.76}
\newcommand{\OwLayaTypedArrivalAuc}{37.31}
\newcommand{\OwLayaTypedArrivalRange}{3.19}
\newcommand{\OwLayaMultilingualAuc}{0.21}
\newcommand{\OwLayaMultilingualUntimed}{0.21}
\newcommand{\OwLayaMultilingualMedian}{0.084}
\newcommand{\OwLayaMultilingualTail}{0.154}
\newcommand{\OwLayaMultilingualDebugAuc}{0.00}
\newcommand{\OwLayaMultilingualAssemblyAuc}{0.83}
\newcommand{\OwLayaMultilingualSupportAuc}{0.00}
\newcommand{\OwLayaMultilingualPresenterAuc}{0.00}
\newcommand{\OwLayaMultilingualFreshestAuc}{28.90}
\newcommand{\OwLayaMultilingualFreshestRange}{4.04}
\newcommand{\OwLayaMultilingualLagOneAuc}{35.53}
\newcommand{\OwLayaMultilingualLagOneRange}{3.85}
\newcommand{\OwLayaMultilingualArrivalAuc}{36.64}
\newcommand{\OwLayaMultilingualArrivalRange}{3.24}
\newcommand{\OwDJevAuc}{20.89}
\newcommand{\OwDJevUntimed}{23.13}
\newcommand{\OwDJevMedian}{0.549}
\newcommand{\OwDJevTail}{0.650}
\newcommand{\OwDJevDebugAuc}{25.85}
\newcommand{\OwDJevAssemblyAuc}{6.26}
\newcommand{\OwDJevSupportAuc}{26.24}
\newcommand{\OwDJevPresenterAuc}{25.23}
\newcommand{\OwDJevFreshestAuc}{47.03}
\newcommand{\OwDJevFreshestRange}{3.77}
\newcommand{\OwDJevLagOneAuc}{48.14}
\newcommand{\OwDJevLagOneRange}{3.18}
\newcommand{\OwDJevArrivalAuc}{48.12}
\newcommand{\OwDJevArrivalRange}{3.21}
\newcommand{\OwKevAuc}{21.22}
\newcommand{\OwKevUntimed}{22.08}
\newcommand{\OwKevMedian}{0.223}
\newcommand{\OwKevTail}{0.328}
\newcommand{\OwKevDebugAuc}{17.39}
\newcommand{\OwKevAssemblyAuc}{12.78}
\newcommand{\OwKevSupportAuc}{36.49}
\newcommand{\OwKevPresenterAuc}{18.20}
\newcommand{\OwKevFreshestAuc}{43.76}
\newcommand{\OwKevFreshestRange}{3.04}
\newcommand{\OwKevLagOneAuc}{46.77}
\newcommand{\OwKevLagOneRange}{2.78}
\newcommand{\OwKevArrivalAuc}{46.77}
\newcommand{\OwKevArrivalRange}{2.66}
\newcommand{\OwKevNineAuc}{28.76}
\newcommand{\OwKevNineUntimed}{29.86}
\newcommand{\OwKevNineMedian}{0.229}
\newcommand{\OwKevNineTail}{0.369}
\newcommand{\OwKevNineDebugAuc}{15.59}
\newcommand{\OwKevNineAssemblyAuc}{28.26}
\newcommand{\OwKevNineSupportAuc}{40.20}
\newcommand{\OwKevNinePresenterAuc}{31.01}
\newcommand{\OwKevNineFreshestAuc}{48.21}
\newcommand{\OwKevNineFreshestRange}{2.24}
\newcommand{\OwKevNineLagOneAuc}{50.16}
\newcommand{\OwKevNineLagOneRange}{2.02}
\newcommand{\OwKevNineArrivalAuc}{50.00}
\newcommand{\OwKevNineArrivalRange}{2.13}
\newcommand{\OwKevTwentySevenAuc}{62.15}
\newcommand{\OwKevTwentySevenUntimed}{72.64}
\newcommand{\OwKevTwentySevenMedian}{0.529}
\newcommand{\OwKevTwentySevenTail}{0.903}
\newcommand{\OwKevTwentySevenDebugAuc}{43.28}
\newcommand{\OwKevTwentySevenAssemblyAuc}{66.47}
\newcommand{\OwKevTwentySevenSupportAuc}{71.97}
\newcommand{\OwKevTwentySevenPresenterAuc}{66.87}
\newcommand{\OwKevTwentySevenFreshestAuc}{66.78}
\newcommand{\OwKevTwentySevenFreshestRange}{0.66}
\newcommand{\OwKevTwentySevenLagOneAuc}{65.16}
\newcommand{\OwKevTwentySevenLagOneRange}{1.01}
\newcommand{\OwKevTwentySevenArrivalAuc}{64.96}
\newcommand{\OwKevTwentySevenArrivalRange}{1.27}
\newcommand{\OwNimbleAuc}{14.43}
\newcommand{\OwNimbleUntimed}{22.36}
\newcommand{\OwNimbleMedian}{1.711}
\newcommand{\OwNimbleTail}{12.122}
\newcommand{\OwNimbleDebugAuc}{0.36}
\newcommand{\OwNimbleAssemblyAuc}{11.26}
\newcommand{\OwNimbleSupportAuc}{24.37}
\newcommand{\OwNimblePresenterAuc}{21.72}
\newcommand{\OwNimbleFreshestAuc}{55.10}
\newcommand{\OwNimbleFreshestRange}{2.89}
\newcommand{\OwNimbleLagOneAuc}{55.10}
\newcommand{\OwNimbleLagOneRange}{2.86}
\newcommand{\OwNimbleArrivalAuc}{55.10}
\newcommand{\OwNimbleArrivalRange}{2.86}
\newcommand{\OwQwenLogitsAuc}{15.61}
\newcommand{\OwQwenLogitsUntimed}{17.71}
\newcommand{\OwQwenLogitsMedian}{0.777}
\newcommand{\OwQwenLogitsTail}{1.254}
\newcommand{\OwQwenLogitsDebugAuc}{0.00}
\newcommand{\OwQwenLogitsAssemblyAuc}{12.48}
\newcommand{\OwQwenLogitsSupportAuc}{27.30}
\newcommand{\OwQwenLogitsPresenterAuc}{22.67}
\newcommand{\OwQwenLogitsFreshestAuc}{49.15}
\newcommand{\OwQwenLogitsFreshestRange}{4.33}
\newcommand{\OwQwenLogitsLagOneAuc}{49.62}
\newcommand{\OwQwenLogitsLagOneRange}{4.04}
\newcommand{\OwQwenLogitsArrivalAuc}{49.61}
\newcommand{\OwQwenLogitsArrivalRange}{4.07}
\newcommand{\OwTerraNoneAuc}{55.46}
\newcommand{\OwJevTerraNoneAuc}{65.95}
\newcommand{\OwRepeatMaxAucRange}{1.87}
\newcommand{\OwRepeatMaxUntimedRange}{0.83}

\newcommand{\TabOpenweightFamilies}{%
Laya English & \OwLayaEnglishDebugAuc & \OwLayaEnglishAssemblyAuc & \OwLayaEnglishSupportAuc & \OwLayaEnglishPresenterAuc & \OwLayaEnglishAuc & \OwLayaEnglishUntimed \\
Laya typed-decisions & \OwLayaTypedDebugAuc & \OwLayaTypedAssemblyAuc & \OwLayaTypedSupportAuc & \OwLayaTypedPresenterAuc & \OwLayaTypedAuc & \OwLayaTypedUntimed \\
Laya multilingual & \OwLayaMultilingualDebugAuc & \OwLayaMultilingualAssemblyAuc & \OwLayaMultilingualSupportAuc & \OwLayaMultilingualPresenterAuc & \OwLayaMultilingualAuc & \OwLayaMultilingualUntimed \\
DJev / DiffusionGemma & \OwDJevDebugAuc & \OwDJevAssemblyAuc & \OwDJevSupportAuc & \OwDJevPresenterAuc & \OwDJevAuc & \OwDJevUntimed \\
Kev-4B & \OwKevDebugAuc & \OwKevAssemblyAuc & \OwKevSupportAuc & \OwKevPresenterAuc & \OwKevAuc & \OwKevUntimed \\
Kev-9B & \OwKevNineDebugAuc & \OwKevNineAssemblyAuc & \OwKevNineSupportAuc & \OwKevNinePresenterAuc & \OwKevNineAuc & \OwKevNineUntimed \\
Kev-27B & \OwKevTwentySevenDebugAuc & \OwKevTwentySevenAssemblyAuc & \OwKevTwentySevenSupportAuc & \OwKevTwentySevenPresenterAuc & \OwKevTwentySevenAuc & \OwKevTwentySevenUntimed \\
Bespoke Nimble-9B & \OwNimbleDebugAuc & \OwNimbleAssemblyAuc & \OwNimbleSupportAuc & \OwNimblePresenterAuc & \OwNimbleAuc & \OwNimbleUntimed \\
Qwen3.5-4B direct-logit & \OwQwenLogitsDebugAuc & \OwQwenLogitsAssemblyAuc & \OwQwenLogitsSupportAuc & \OwQwenLogitsPresenterAuc & \OwQwenLogitsAuc & \OwQwenLogitsUntimed \\
}
\newcommand{\TabOpenweightLatency}{%
Laya English & RTX PRO 6000 (96 GB) & \OwLayaEnglishMedian & \OwLayaEnglishTail \\
Laya typed-decisions & RTX PRO 6000 (96 GB) & \OwLayaTypedMedian & \OwLayaTypedTail \\
Laya multilingual & RTX PRO 6000 (96 GB) & \OwLayaMultilingualMedian & \OwLayaMultilingualTail \\
DJev / DiffusionGemma & RTX PRO 6000 (96 GB) & \OwDJevMedian & \OwDJevTail \\
Kev-4B & RTX PRO 6000 (96 GB) & \OwKevMedian & \OwKevTail \\
Kev-9B & RTX PRO 6000 (96 GB) & \OwKevNineMedian & \OwKevNineTail \\
Kev-27B & RTX PRO 6000 (96 GB) & \OwKevTwentySevenMedian & \OwKevTwentySevenTail \\
Bespoke Nimble-9B & RTX PRO 6000 (96 GB) & \OwNimbleMedian & \OwNimbleTail \\
Qwen3.5-4B direct-logit & RTX PRO 6000 (96 GB) & \OwQwenLogitsMedian & \OwQwenLogitsTail \\
}
\newcommand{\TabOpenweightHybrids}{%
Laya English & \OwLayaEnglishFreshestAuc~(\OwLayaEnglishFreshestRange) & \OwLayaEnglishLagOneAuc~(\OwLayaEnglishLagOneRange) & \OwLayaEnglishArrivalAuc~(\OwLayaEnglishArrivalRange) \\
Laya typed-decisions & \OwLayaTypedFreshestAuc~(\OwLayaTypedFreshestRange) & \OwLayaTypedLagOneAuc~(\OwLayaTypedLagOneRange) & \OwLayaTypedArrivalAuc~(\OwLayaTypedArrivalRange) \\
Laya multilingual & \OwLayaMultilingualFreshestAuc~(\OwLayaMultilingualFreshestRange) & \OwLayaMultilingualLagOneAuc~(\OwLayaMultilingualLagOneRange) & \OwLayaMultilingualArrivalAuc~(\OwLayaMultilingualArrivalRange) \\
DJev / DiffusionGemma & \OwDJevFreshestAuc~(\OwDJevFreshestRange) & \OwDJevLagOneAuc~(\OwDJevLagOneRange) & \OwDJevArrivalAuc~(\OwDJevArrivalRange) \\
Kev-4B & \OwKevFreshestAuc~(\OwKevFreshestRange) & \OwKevLagOneAuc~(\OwKevLagOneRange) & \OwKevArrivalAuc~(\OwKevArrivalRange) \\
Kev-9B & \OwKevNineFreshestAuc~(\OwKevNineFreshestRange) & \OwKevNineLagOneAuc~(\OwKevNineLagOneRange) & \OwKevNineArrivalAuc~(\OwKevNineArrivalRange) \\
Kev-27B & \OwKevTwentySevenFreshestAuc~(\OwKevTwentySevenFreshestRange) & \OwKevTwentySevenLagOneAuc~(\OwKevTwentySevenLagOneRange) & \OwKevTwentySevenArrivalAuc~(\OwKevTwentySevenArrivalRange) \\
Bespoke Nimble-9B & \OwNimbleFreshestAuc~(\OwNimbleFreshestRange) & \OwNimbleLagOneAuc~(\OwNimbleLagOneRange) & \OwNimbleArrivalAuc~(\OwNimbleArrivalRange) \\
Qwen3.5-4B direct-logit & \OwQwenLogitsFreshestAuc~(\OwQwenLogitsFreshestRange) & \OwQwenLogitsLagOneAuc~(\OwQwenLogitsLagOneRange) & \OwQwenLogitsArrivalAuc~(\OwQwenLogitsArrivalRange) \\
}
\newcommand{\TabOpenweightRepeats}{%
Laya English & 0.42 & 0.42 & 0.42 & 0.42 & 0.00 & 0.42 & 0.00 \\
Laya typed-decisions & 1.46 & 1.46 & 1.46 & 1.46 & 0.00 & 1.46 & 0.00 \\
Laya multilingual & 0.21 & 0.21 & 0.21 & 0.21 & 0.00 & 0.21 & 0.00 \\
DJev / DiffusionGemma & 20.61 & 20.88 & 21.19 & 20.89 & 0.59 & 23.13 & 0.42 \\
Kev-4B & 21.20 & 21.22 & 21.23 & 21.22 & 0.04 & 22.08 & 0.00 \\
Kev-9B & 29.27 & 28.45 & 28.57 & 28.76 & 0.82 & 29.86 & 0.83 \\
Kev-27B & 61.97 & 62.17 & 62.30 & 62.15 & 0.34 & 72.64 & 0.21 \\
Bespoke Nimble-9B & 14.72 & 13.34 & 15.21 & 14.43 & 1.87 & 22.36 & 0.21 \\
Qwen3.5-4B direct-logit & 15.46 & 15.68 & 15.70 & 15.61 & 0.24 & 17.71 & 0.00 \\
}

\newcommand{\TabHeadlineBody}{%
  Jev & \HeadlineJevAuc & \HeadlineJevSD \\
  Terra none & \HeadlineTerraNoneAuc & \HeadlineTerraNoneSD \\
  Kev-27B & \HeadlineKevTwentySevenAuc & \HeadlineKevTwentySevenSD \\
  Kev-27B + Terra none & \HeadlineKevTwentySevenHybridAuc & \HeadlineKevTwentySevenHybridSD \\
}

\newcommand{\LaterNumSettingsWord}{seven}
\newcommand{\LaterNumFastWord}{four}
\newcommand{\LaterNominalGap}{0.023}
\newcommand{\LaterPerplexityAuc}{63.37}
\newcommand{\LaterPerplexityAucOne}{63.4}
\newcommand{\LaterPerplexityAucSD}{0.27}
\newcommand{\LaterPerplexityUntimed}{70.0}
\newcommand{\LaterPerplexityStale}{6.6}
\newcommand{\LaterPerplexityJudgment}{26.4}
\newcommand{\LaterPerplexityMedian}{0.38}
\newcommand{\LaterPerplexityFailedAttempts}{15}
\newcommand{\LaterPerplexityDebugAuc}{47.2}
\newcommand{\LaterPerplexityAssemblyAuc}{66.8}
\newcommand{\LaterPerplexitySupportAuc}{79.9}
\newcommand{\LaterPerplexityPresenterAuc}{59.6}
\newcommand{\LaterGlideAucOne}{39.4}
\newcommand{\LaterGlideAucSD}{2.11}
\newcommand{\LaterGlideUntimed}{80.5}
\newcommand{\LaterGlideStale}{46.8}
\newcommand{\LaterGlideJudgment}{4.9}
\newcommand{\LaterGlideMedian}{5.55}
\newcommand{\LaterGlideDebugAuc}{31.8}
\newcommand{\LaterGlideAssemblyAuc}{35.1}
\newcommand{\LaterGlideSupportAuc}{41.1}
\newcommand{\LaterGlidePresenterAuc}{49.7}
\newcommand{\LaterLunaDecisionsAucOne}{32.4}
\newcommand{\LaterLunaDecisionsAucSD}{0.70}
\newcommand{\LaterLunaDecisionsUntimed}{36.0}
\newcommand{\LaterLunaDecisionsStale}{3.3}
\newcommand{\LaterLunaDecisionsJudgment}{57.6}
\newcommand{\LaterLunaDecisionsMedian}{0.34}
\newcommand{\LaterLunaDecisionsDebugAuc}{21.3}
\newcommand{\LaterLunaDecisionsAssemblyAuc}{20.4}
\newcommand{\LaterLunaDecisionsSupportAuc}{54.8}
\newcommand{\LaterLunaDecisionsPresenterAuc}{33.1}
\newcommand{\LaterClefAucOne}{31.4}
\newcommand{\LaterClefAucSD}{0.41}
\newcommand{\LaterClefUntimed}{39.0}
\newcommand{\LaterClefStale}{7.8}
\newcommand{\LaterClefJudgment}{45.7}
\newcommand{\LaterClefMedian}{0.89}
\newcommand{\LaterClefDebugAuc}{35.2}
\newcommand{\LaterClefAssemblyAuc}{13.6}
\newcommand{\LaterClefSupportAuc}{59.2}
\newcommand{\LaterClefPresenterAuc}{17.5}
\newcommand{\LaterClefFlashAucOne}{19.9}
\newcommand{\LaterClefFlashAucSD}{0.19}
\newcommand{\LaterClefFlashUntimed}{21.7}
\newcommand{\LaterClefFlashStale}{2.5}
\newcommand{\LaterClefFlashJudgment}{68.1}
\newcommand{\LaterClefFlashMedian}{0.44}
\newcommand{\LaterClefFlashFailedAttempts}{2}
\newcommand{\LaterClefFlashDebugAuc}{13.3}
\newcommand{\LaterClefFlashAssemblyAuc}{23.7}
\newcommand{\LaterClefFlashSupportAuc}{28.7}
\newcommand{\LaterClefFlashPresenterAuc}{13.9}
\newcommand{\LaterHaikuAucOne}{36.6}
\newcommand{\LaterHaikuAucSD}{0.09}
\newcommand{\LaterHaikuUntimed}{95.3}
\newcommand{\LaterHaikuStale}{56.0}
\newcommand{\LaterHaikuJudgment}{0.8}
\newcommand{\LaterHaikuMedian}{3.66}
\newcommand{\LaterHaikuFailedAttempts}{1}
\newcommand{\LaterHaikuDebugAuc}{37.4}
\newcommand{\LaterHaikuAssemblyAuc}{32.3}
\newcommand{\LaterHaikuSupportAuc}{39.9}
\newcommand{\LaterHaikuPresenterAuc}{36.6}
\newcommand{\LaterHaikuOffAucOne}{19.4}
\newcommand{\LaterHaikuOffAucSD}{1.09}
\newcommand{\LaterHaikuOffUntimed}{25.6}
\newcommand{\LaterHaikuOffStale}{7.9}
\newcommand{\LaterHaikuOffJudgment}{48.9}
\newcommand{\LaterHaikuOffMedian}{1.44}
\newcommand{\LaterHaikuOffDebugAuc}{12.7}
\newcommand{\LaterHaikuOffAssemblyAuc}{11.7}
\newcommand{\LaterHaikuOffSupportAuc}{30.6}
\newcommand{\LaterHaikuOffPresenterAuc}{22.6}
\newcommand{\LaterPerplexityLunaFreshestAuc}{66.82}
\newcommand{\LaterPerplexityLunaLagOneAuc}{65.09}
\newcommand{\LaterPerplexityLunaArrivalAuc}{61.18}
\newcommand{\LaterPerplexityLunaNoneFreshestAuc}{52.46}
\newcommand{\LaterPerplexityLunaNoneLagOneAuc}{48.83}
\newcommand{\LaterPerplexityLunaNoneArrivalAuc}{48.45}
\newcommand{\LaterPerplexityTerraFreshestAuc}{69.07}
\newcommand{\LaterPerplexityTerraLagOneAuc}{67.56}
\newcommand{\LaterPerplexityTerraArrivalAuc}{64.93}
\newcommand{\LaterPerplexityTerraNoneFreshestAuc}{68.37}
\newcommand{\LaterPerplexityTerraNoneLagOneAuc}{66.27}
\newcommand{\LaterPerplexityTerraNoneArrivalAuc}{65.80}
\newcommand{\LaterPerplexityAstraFreshestAuc}{69.48}
\newcommand{\LaterPerplexityAstraLagOneAuc}{68.04}
\newcommand{\LaterPerplexityAstraArrivalAuc}{64.41}
\newcommand{\LaterLunaDecisionsLunaFreshestAuc}{42.54}
\newcommand{\LaterLunaDecisionsLunaLagOneAuc}{44.85}
\newcommand{\LaterLunaDecisionsLunaArrivalAuc}{43.94}
\newcommand{\LaterLunaDecisionsLunaNoneFreshestAuc}{35.95}
\newcommand{\LaterLunaDecisionsLunaNoneLagOneAuc}{35.36}
\newcommand{\LaterLunaDecisionsLunaNoneArrivalAuc}{35.20}
\newcommand{\LaterLunaDecisionsTerraFreshestAuc}{46.83}
\newcommand{\LaterLunaDecisionsTerraLagOneAuc}{49.08}
\newcommand{\LaterLunaDecisionsTerraArrivalAuc}{48.73}
\newcommand{\LaterLunaDecisionsTerraNoneFreshestAuc}{51.09}
\newcommand{\LaterLunaDecisionsTerraNoneLagOneAuc}{52.33}
\newcommand{\LaterLunaDecisionsTerraNoneArrivalAuc}{52.21}
\newcommand{\LaterLunaDecisionsAstraFreshestAuc}{45.72}
\newcommand{\LaterLunaDecisionsAstraLagOneAuc}{48.24}
\newcommand{\LaterLunaDecisionsAstraArrivalAuc}{47.38}
\newcommand{\LaterClefLunaFreshestAuc}{42.83}
\newcommand{\LaterClefLunaLagOneAuc}{43.80}
\newcommand{\LaterClefLunaArrivalAuc}{43.53}
\newcommand{\LaterClefLunaNoneFreshestAuc}{34.57}
\newcommand{\LaterClefLunaNoneLagOneAuc}{34.25}
\newcommand{\LaterClefLunaNoneArrivalAuc}{34.24}
\newcommand{\LaterClefTerraFreshestAuc}{48.23}
\newcommand{\LaterClefTerraLagOneAuc}{48.95}
\newcommand{\LaterClefTerraArrivalAuc}{48.99}
\newcommand{\LaterClefTerraNoneFreshestAuc}{55.27}
\newcommand{\LaterClefTerraNoneLagOneAuc}{55.17}
\newcommand{\LaterClefTerraNoneArrivalAuc}{55.16}
\newcommand{\LaterClefAstraFreshestAuc}{47.70}
\newcommand{\LaterClefAstraLagOneAuc}{48.49}
\newcommand{\LaterClefAstraArrivalAuc}{47.91}
\newcommand{\LaterClefFlashLunaFreshestAuc}{33.93}
\newcommand{\LaterClefFlashLunaLagOneAuc}{37.36}
\newcommand{\LaterClefFlashLunaArrivalAuc}{37.38}
\newcommand{\LaterClefFlashLunaNoneFreshestAuc}{29.44}
\newcommand{\LaterClefFlashLunaNoneLagOneAuc}{29.81}
\newcommand{\LaterClefFlashLunaNoneArrivalAuc}{29.76}
\newcommand{\LaterClefFlashTerraFreshestAuc}{39.20}
\newcommand{\LaterClefFlashTerraLagOneAuc}{42.39}
\newcommand{\LaterClefFlashTerraArrivalAuc}{42.50}
\newcommand{\LaterClefFlashTerraNoneFreshestAuc}{45.63}
\newcommand{\LaterClefFlashTerraNoneLagOneAuc}{47.55}
\newcommand{\LaterClefFlashTerraNoneArrivalAuc}{47.54}
\newcommand{\LaterClefFlashAstraFreshestAuc}{36.81}
\newcommand{\LaterClefFlashAstraLagOneAuc}{40.47}
\newcommand{\LaterClefFlashAstraArrivalAuc}{40.34}
\newcommand{\LaterBestAuc}{69.48}
\newcommand{\LaterBestSD}{1.20}
\newcommand{\LaterSensConditionsWord}{seven}
\newcommand{\LaterSensKevTwentySevenPrimary}{62.1}
\newcommand{\LaterSensKevTwentySevenLinear}{66.9}
\newcommand{\LaterSensKevTwentySevenNarrow}{64.1}
\newcommand{\LaterSensKevTwentySevenHalfFour}{59.6}
\newcommand{\LaterSensKevTwentySevenOneEight}{66.0}
\newcommand{\LaterSensKevTwentySevenTenthFour}{43.4}
\newcommand{\LaterSensKevTwentySevenTenthEight}{47.6}
\newcommand{\LaterSensKevTwentySevenTwo}{64.8}
\newcommand{\LaterSensPerplexityPrimary}{63.4}
\newcommand{\LaterSensPerplexityLinear}{66.4}
\newcommand{\LaterSensPerplexityNarrow}{64.7}
\newcommand{\LaterSensPerplexityHalfFour}{61.8}
\newcommand{\LaterSensPerplexityOneEight}{65.9}
\newcommand{\LaterSensPerplexityTenthFour}{47.8}
\newcommand{\LaterSensPerplexityTenthEight}{51.0}
\newcommand{\LaterSensPerplexityTwo}{65.1}
\newcommand{\LaterSensJevTerraNonePrimary}{65.9}
\newcommand{\LaterSensJevTerraNoneLinear}{72.0}
\newcommand{\LaterSensJevTerraNoneNarrow}{66.5}
\newcommand{\LaterSensJevTerraNoneHalfFour}{62.4}
\newcommand{\LaterSensJevTerraNoneOneEight}{69.9}
\newcommand{\LaterSensJevTerraNoneTenthFour}{51.5}
\newcommand{\LaterSensJevTerraNoneTenthEight}{55.5}
\newcommand{\LaterSensJevTerraNoneTwo}{67.3}
\newcommand{\LaterSensKevTerraNonePrimary}{66.8}
\newcommand{\LaterSensKevTerraNoneLinear}{73.3}
\newcommand{\LaterSensKevTerraNoneNarrow}{69.0}
\newcommand{\LaterSensKevTerraNoneHalfFour}{63.2}
\newcommand{\LaterSensKevTerraNoneOneEight}{71.9}
\newcommand{\LaterSensKevTerraNoneTenthFour}{45.4}
\newcommand{\LaterSensKevTerraNoneTenthEight}{50.5}
\newcommand{\LaterSensKevTerraNoneTwo}{69.9}
\newcommand{\LaterSensKevTerraNoneAboveWord}{five}
\newcommand{\LaterSensBestPrimary}{69.5}
\newcommand{\LaterSensBestLinear}{77.9}
\newcommand{\LaterSensBestNarrow}{68.2}
\newcommand{\LaterSensBestHalfFour}{64.1}
\newcommand{\LaterSensBestOneEight}{74.0}
\newcommand{\LaterSensBestTenthFour}{49.1}
\newcommand{\LaterSensBestTenthEight}{54.9}
\newcommand{\LaterSensBestTwo}{66.6}
\newcommand{\TabLaterStandalone}{%
Perplexity Decider v1 27B & \LaterPerplexityDebugAuc & \LaterPerplexityAssemblyAuc & \LaterPerplexitySupportAuc & \LaterPerplexityPresenterAuc & \LaterPerplexityAucOne~$\pm$~\LaterPerplexityAucSD & \LaterPerplexityUntimed & \LaterPerplexityMedian & \LaterPerplexityStale & \LaterPerplexityJudgment \\
GLiDE (Fastino) & \LaterGlideDebugAuc & \LaterGlideAssemblyAuc & \LaterGlideSupportAuc & \LaterGlidePresenterAuc & \LaterGlideAucOne~$\pm$~\LaterGlideAucSD & \LaterGlideUntimed & \LaterGlideMedian & \LaterGlideStale & \LaterGlideJudgment \\
GPT-6-Luna (Decisions API) & \LaterLunaDecisionsDebugAuc & \LaterLunaDecisionsAssemblyAuc & \LaterLunaDecisionsSupportAuc & \LaterLunaDecisionsPresenterAuc & \LaterLunaDecisionsAucOne~$\pm$~\LaterLunaDecisionsAucSD & \LaterLunaDecisionsUntimed & \LaterLunaDecisionsMedian & \LaterLunaDecisionsStale & \LaterLunaDecisionsJudgment \\
Clef & \LaterClefDebugAuc & \LaterClefAssemblyAuc & \LaterClefSupportAuc & \LaterClefPresenterAuc & \LaterClefAucOne~$\pm$~\LaterClefAucSD & \LaterClefUntimed & \LaterClefMedian & \LaterClefStale & \LaterClefJudgment \\
Clef Flash & \LaterClefFlashDebugAuc & \LaterClefFlashAssemblyAuc & \LaterClefFlashSupportAuc & \LaterClefFlashPresenterAuc & \LaterClefFlashAucOne~$\pm$~\LaterClefFlashAucSD & \LaterClefFlashUntimed & \LaterClefFlashMedian & \LaterClefFlashStale & \LaterClefFlashJudgment \\
Claude Haiku 5.5 low & \LaterHaikuDebugAuc & \LaterHaikuAssemblyAuc & \LaterHaikuSupportAuc & \LaterHaikuPresenterAuc & \LaterHaikuAucOne~$\pm$~\LaterHaikuAucSD & \LaterHaikuUntimed & \LaterHaikuMedian & \LaterHaikuStale & \LaterHaikuJudgment \\
Claude Haiku 5.5 thinking off & \LaterHaikuOffDebugAuc & \LaterHaikuOffAssemblyAuc & \LaterHaikuOffSupportAuc & \LaterHaikuOffPresenterAuc & \LaterHaikuOffAucOne~$\pm$~\LaterHaikuOffAucSD & \LaterHaikuOffUntimed & \LaterHaikuOffMedian & \LaterHaikuOffStale & \LaterHaikuOffJudgment \\
}
\newcommand{\TabLaterSystems}{%
Perplexity & Luna low & \LaterPerplexityLunaFreshestAuc & \LaterPerplexityLunaLagOneAuc & \LaterPerplexityLunaArrivalAuc \\
 & Luna none & \LaterPerplexityLunaNoneFreshestAuc & \LaterPerplexityLunaNoneLagOneAuc & \LaterPerplexityLunaNoneArrivalAuc \\
 & Terra low & \LaterPerplexityTerraFreshestAuc & \LaterPerplexityTerraLagOneAuc & \LaterPerplexityTerraArrivalAuc \\
 & Terra none & \LaterPerplexityTerraNoneFreshestAuc & \LaterPerplexityTerraNoneLagOneAuc & \LaterPerplexityTerraNoneArrivalAuc \\
 & Astra low & \LaterPerplexityAstraFreshestAuc & \LaterPerplexityAstraLagOneAuc & \LaterPerplexityAstraArrivalAuc \\
GPT-6-Luna & Luna low & \LaterLunaDecisionsLunaFreshestAuc & \LaterLunaDecisionsLunaLagOneAuc & \LaterLunaDecisionsLunaArrivalAuc \\
 & Luna none & \LaterLunaDecisionsLunaNoneFreshestAuc & \LaterLunaDecisionsLunaNoneLagOneAuc & \LaterLunaDecisionsLunaNoneArrivalAuc \\
 & Terra low & \LaterLunaDecisionsTerraFreshestAuc & \LaterLunaDecisionsTerraLagOneAuc & \LaterLunaDecisionsTerraArrivalAuc \\
 & Terra none & \LaterLunaDecisionsTerraNoneFreshestAuc & \LaterLunaDecisionsTerraNoneLagOneAuc & \LaterLunaDecisionsTerraNoneArrivalAuc \\
 & Astra low & \LaterLunaDecisionsAstraFreshestAuc & \LaterLunaDecisionsAstraLagOneAuc & \LaterLunaDecisionsAstraArrivalAuc \\
Clef & Luna low & \LaterClefLunaFreshestAuc & \LaterClefLunaLagOneAuc & \LaterClefLunaArrivalAuc \\
 & Luna none & \LaterClefLunaNoneFreshestAuc & \LaterClefLunaNoneLagOneAuc & \LaterClefLunaNoneArrivalAuc \\
 & Terra low & \LaterClefTerraFreshestAuc & \LaterClefTerraLagOneAuc & \LaterClefTerraArrivalAuc \\
 & Terra none & \LaterClefTerraNoneFreshestAuc & \LaterClefTerraNoneLagOneAuc & \LaterClefTerraNoneArrivalAuc \\
 & Astra low & \LaterClefAstraFreshestAuc & \LaterClefAstraLagOneAuc & \LaterClefAstraArrivalAuc \\
Clef Flash & Luna low & \LaterClefFlashLunaFreshestAuc & \LaterClefFlashLunaLagOneAuc & \LaterClefFlashLunaArrivalAuc \\
 & Luna none & \LaterClefFlashLunaNoneFreshestAuc & \LaterClefFlashLunaNoneLagOneAuc & \LaterClefFlashLunaNoneArrivalAuc \\
 & Terra low & \LaterClefFlashTerraFreshestAuc & \LaterClefFlashTerraLagOneAuc & \LaterClefFlashTerraArrivalAuc \\
 & Terra none & \LaterClefFlashTerraNoneFreshestAuc & \LaterClefFlashTerraNoneLagOneAuc & \LaterClefFlashTerraNoneArrivalAuc \\
 & Astra low & \LaterClefFlashAstraFreshestAuc & \LaterClefFlashAstraLagOneAuc & \LaterClefFlashAstraArrivalAuc \\
}
\newcommand{\TabLaterSensitivityBody}{%
Kev-27B & \LaterSensKevTwentySevenPrimary & \LaterSensKevTwentySevenLinear & \LaterSensKevTwentySevenNarrow & \LaterSensKevTwentySevenHalfFour & \LaterSensKevTwentySevenOneEight & \LaterSensKevTwentySevenTenthFour & \LaterSensKevTwentySevenTenthEight & \LaterSensKevTwentySevenTwo \\
Perplexity Decider v1 27B$^\dagger$ & \LaterSensPerplexityPrimary & \LaterSensPerplexityLinear & \LaterSensPerplexityNarrow & \LaterSensPerplexityHalfFour & \LaterSensPerplexityOneEight & \LaterSensPerplexityTenthFour & \LaterSensPerplexityTenthEight & \LaterSensPerplexityTwo \\
Jev + Terra none & \LaterSensJevTerraNonePrimary & \LaterSensJevTerraNoneLinear & \LaterSensJevTerraNoneNarrow & \LaterSensJevTerraNoneHalfFour & \LaterSensJevTerraNoneOneEight & \LaterSensJevTerraNoneTenthFour & \LaterSensJevTerraNoneTenthEight & \LaterSensJevTerraNoneTwo \\
Kev-27B + Terra none & \LaterSensKevTerraNonePrimary & \LaterSensKevTerraNoneLinear & \LaterSensKevTerraNoneNarrow & \LaterSensKevTerraNoneHalfFour & \LaterSensKevTerraNoneOneEight & \LaterSensKevTerraNoneTenthFour & \LaterSensKevTerraNoneTenthEight & \LaterSensKevTerraNoneTwo \\
Perplexity Decider v1 27B$^\dagger$ + Astra low & \LaterSensBestPrimary & \LaterSensBestLinear & \LaterSensBestNarrow & \LaterSensBestHalfFour & \LaterSensBestOneEight & \LaterSensBestTenthFour & \LaterSensBestTenthEight & \LaterSensBestTwo \\
}

\newcommand{\sdb}{SDB}
\newcommand{\ind}{\mathbf{1}}
\newcommand{\yref}{y^{*}}
\newcommand{\JevProvider}{TypeSafe}
\newcommand{\JevInterface}{System One}

\title{StreamDecisionBench: Evaluating Decisions in Force\\on Evolving Language Streams}
\hypersetup{pdftitle={StreamDecisionBench: Evaluating Decisions in Force on Evolving Language Streams}}
\ifnum\pdfstrcmp{\SDBVersion}{review}=0 \else\hypersetup{pdfauthor={Jhen-Ke Lin, Chung Chun Wang}}\fi

\author{Jhen-Ke Lin \and Chung Chun Wang \\
  National Yang Ming Chiao Tung University \\
  \texttt{jacob.cs14@nycu.edu.tw}, \texttt{takalawang.cs14@nycu.edu.tw}}

\begin{document}
\maketitle

\begin{abstract}
Language models increasingly make real-time decisions in applications that apply the latest answer until a newer one arrives.
A late answer can prolong an outdated decision, such as a call recorder still running while a customer reads out card details, an error offline accuracy misses.
We make three contributions.
First, we release StreamDecisionBench (SDB), a dataset of eight streaming scenarios in four application families, with executable reference decisions derived from public rules.
Second, we propose an evaluation protocol and a metric, in-force accuracy: the share of time the applied decision is correct across update intervals of 0.5--8\,s.
It reflects accuracy and latency jointly, attributing each error to judgment, latency or both.
Third, we evaluate \OwTotalSettingsWord{} single-model settings, and this attribution separates speed-limited from judgment-limited models: slower, more accurate models lose \LowEffortStaleRoundedMin--\LowEffortStaleRoundedMax\% of the time to outdated answers, a fast model \JevJudgmentRounded\% to wrong ones.
We therefore test hybrids in which a slow model corrects a fast one; with the right pairing and configuration, a hybrid outperforms every single model.
However, even the best evaluated system keeps a correct decision in force only about two-thirds of the time, leaving a substantial gap for real-time use.
\end{abstract}

\begin{figure*}[t]
  \centering
  \includegraphics[width=\linewidth]{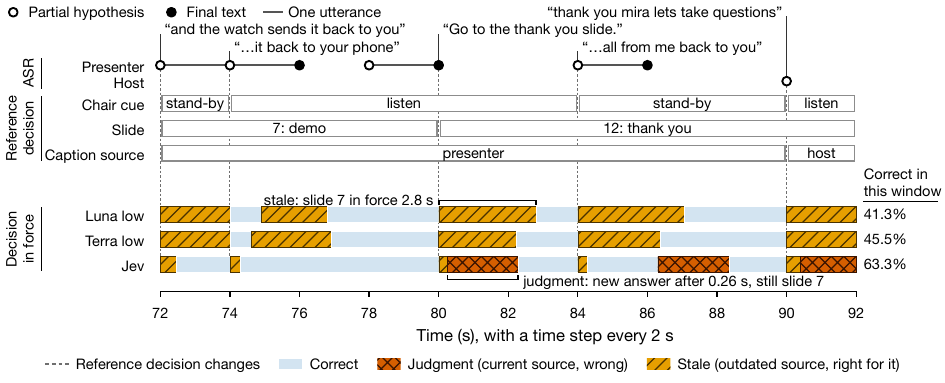}
  \caption{\textbf{A decision right for its own state stays in force after new or revised speech changes the reference.}
  Presenter voice control, pass 1 (presenter~A, \TrajWindowStart--\TrajWindowEnd\,s; Appendix~\ref{app:window}): the chair cue follows ASR revisions; the slide waits for final text.}
  \label{fig:trajectory}
\end{figure*}

% ---------------------------------------------------------------------------
\section{Introduction}
\label{sec:intro}

Language models increasingly serve as decision components inside application logic.
An IDE keeps an action card that tells a developer to rerun tests, inspect a file or hand a failure to its owning team while edits, saves and test runs stream in.
An assembly station displays the next instruction as part scans and torque readings arrive.
A call-centre desktop decides whether the call recorder must pause while a customer reads out card details.
In each case the application sends the current state to the component, composes the answers into one decision and keeps applying that decision until a newer one arrives.
What users experience is therefore the \emph{decision in force} at every moment, and its correctness depends both on what the component answered and on when the answer took effect.
These are real-time applications, and late output costs their users: an in-IDE suggestion is bound to one state of the file and dismissed at the developer's next keystroke, and acceptance rose as latency fell \citep{dunay2024multiline,murali2024ai}; users of step-by-step guidance worked 12\% slower when instructions lagged by 1.65\,s and 26\% slower at 3\,s \citep{olguin2021impact}; and automated pause-and-resume keeps card data out of call recordings only if it pauses recording at the correct time \citep{pcissc2018telephone}.
Tolerance for latency shrinks as the state changes faster \citep{gergle2006impact}, and lags of a fraction of a second already disrupt people who act on feedback as they move or speak \citep{mackenzie1993lag,stuart2002effect}; in these applications people act on the decision while the state changes, and the response times they tolerate range from under a second to a few seconds \citep{murali2024ai,miller1968response,chen2017empirical}.

Standard evaluation compares each answer with the reference for the state it answered and reports \emph{untimed} (offline) \emph{accuracy}, which is exact when the world waits for the model.
In a live application, evidence keeps arriving during inference: a response that is correct for its source state can take effect after a newer state has changed the reference decision and then stay in force as a stale decision (Figure~\ref{fig:trajectory}).
A component that is always right but takes effect $L$ seconds after each time step loses about $L$ at every change of the reference decision, so the share of time its decision in force is correct, its \emph{in-force accuracy}, is set jointly by its judgment, its latency and how often the reference decision changes.
We therefore evaluate an embedded decision component by its decision in force at every instant, with in-force accuracy as the primary metric, adopting the latest-output evaluation of streaming perception \citep{li2020towards}.

We introduce StreamDecisionBench (\sdb), a benchmark built on this evaluation, with controlled synthetic streams in four application families: IDE debugging, an assembly station, a support call and presenter voice control.
Responses are composed into the \emph{branch-composed decision} the application consumes, and an executable reference applies the rules written into every state, so the reference decision changes at known moments for known reasons.
Every erroneous instant is attributed exactly to judgment, to latency or to both. We report the accuracy curves and summarize their normalized area over 0.5--8\,s on a logarithmic time axis, first within each family and then across families equally.

We evaluate \NumSettingsWord{} hosted settings: Luna and Terra at low and no reasoning effort, Astra at low, and Jev, a commercial decision model; supplementary measurements cover \OwNumSettingsWord{} self-hosted decision settings and \LaterNumSettingsWord{} hosted settings from five providers first recorded later (Appendix~\ref{app:later}).
Across this range they fall short for different reasons.
At low reasoning effort, Luna, Terra and Astra lose \LunaAucGap, \TerraAucGap{} and \AstraAucGap{} points from untimed accuracy to the integrated score, mostly through stale decisions, even Astra at \AstraUntimed\% untimed accuracy.
Jev responds in \JevLatencyMedian\,s, but its judgment errors account for \JevAucJudgment\% of log-weighted time.
Dropping reasoning pays off where judgment survives: Terra reaches \TerraNoneAuc\%, close to Jev's \JevAuc\%, while Luna falls to \LunaNoneAuc\%.
Composition helps where the fast component judges well: with Terra none correcting it, the self-hosted Kev-27B rises from \OwKevTwentySevenAuc\% to \OwKevTwentySevenFreshestAuc\%, and the best evaluated system, Perplexity Decider corrected by Astra low, reaches \LaterBestAuc\%.
The family curves expose differences hidden by the aggregate: Jev leads the other five hosted settings in assembly, support and presenter control but trails every GPT setting except Luna none in IDE debugging.

Our contributions are:\footnote{Code, data, recordings and reproducible analysis accompany the submission (Appendix~\ref{app:repro}).}
\begin{itemize}[leftmargin=*,itemsep=1pt,topsep=2pt]
\item \textbf{Evaluation and decomposition.} In-force accuracy of the branch-composed decision, an exact partition of error time, and the exact identity $A=O\,U_{\mathrm{cur}}+\lambda$ separating oracle timing from current-source judgment, with an integrated score that weights equal multiplicative time-scale ranges equally (\S\ref{sec:method}).
\item \textbf{Benchmark.} A set of \NumScenariosWord{} controlled scenarios in \NumFamiliesWord{} families (\NumStates{} states, \AnswersPerPass{} state--question pairs, \TransitionsTotal{} reference transitions), with executable public rules and a blind LLM-agent audit of the six IDE, assembly and support scenarios (\S\ref{sec:lite}).
\item \textbf{Results.} Evaluation of \OwTotalSettingsWord{} hosted and self-hosted settings, extended by later hosted settings: the decomposition separates latency-limited from judgment-limited components, and composition replays show that a slower correction improves on both components when the fast component judges competitively, by an amount the arbitration rule sets (\S\ref{sec:results}).
\end{itemize}

% ---------------------------------------------------------------------------
\section{Related Work}
\label{sec:related}

\paragraph{Evaluating held outputs over time.}
An application that keeps applying its latest decision holds the component's output between updates, and several fields evaluate held outputs over time: streaming perception compares a perception system's most recent output with the ground truth at every annotated frame \citep{li2020towards}, and status-update systems count every time slot in which a monitor's estimate is wrong, or weight it by how long the error has lasted \citep{maatouk2020age,maatouk2023age}.
Online monitoring of a partial trace \citep{deshmukh2015robust} and revocable early classification \citep{achenchabe2021early} likewise judge outputs on a signal still arriving.
In-force accuracy adopts this held-output view for an application's decision (\S\ref{sec:inforce}).

\paragraph{Time as part of correctness.}
When an answer takes effect changes what it is worth: real-time reinforcement learning lets the state change while the agent acts \citep{ramstedt2019real}, agent benchmarks judge actions by when they land and find that faster models can beat stronger reasoning ones \citep{froger2026gaia2,kang2025win,li2026beyond,ma2026timely}, real-time agents keep a fast reply while slower reasoning continues \citep{li2026never,wen2026realtime}, and benchmarks for language and multimodal models bring in time through when a question is asked, how old information is or how long answering takes \citep{lin2024streamingbench,cheng2026your,liu2026streammembench,ding2026proactor}.
Incremental processing and simultaneous translation time an output against the input it needs \citep{baumann2011evaluation,ma2019stacl,ma2020simuleval}; following computation-aware latency \citep{ma2020simulmt}, \sdb{} counts the model's own processing as latency (\S\ref{sec:execution}), and its evaluation at other time-step intervals changes only the timing of recorded answers.

\paragraph{Policies, changing evidence and verifiable references.}
A live decision applies a policy supplied at run time \citep{chakrabarti2025cope} while new evidence revises or keeps its conclusion \citep{wilie2024belief,dhanda2026deltalogic}, so \sdb{} writes the rules and the full evidence history into every state (\S\ref{sec:reference}) and computes references by executing them, as CLEVR and SWE-bench do \citep{johnson2017clevr,jimenez2024swebench}, with a blind re-derivation audit, since such references still need independent checking \citep{chowdhury2024introducing,openai2026why}.

% ---------------------------------------------------------------------------
\section{Evaluating Decisions in Force}
\label{sec:method}
An embedded decision component is evaluated here by what its application applies, which needs the decision composed from the answers, a reference decision at every instant (\S\ref{sec:setting}--\ref{sec:compose}) and the moment each response takes effect (\S\ref{sec:inforce}).
Because time enters the score, the conditions that set it are part of the measurement (\S\ref{sec:execution}).

\subsection{Setting and Public State}
\label{sec:setting}
A scenario publishes public states $S_0,\ldots,S_{n-1}$ at time steps $r_0<\cdots<r_{n-1}$ within a horizon $H$.
Every time in a state, from its clock to thresholds and deadlines, is counted in ticks. The \emph{time-step interval} $\Delta$ maps these ticks to seconds and scales the horizon, reference dwell times and tick-valued policies together.
All requests are recorded at $\Delta_{\mathrm{rec}}=\ReleaseIntervalSec$\,s; evaluation varies $\Delta$ from 0.5 to 8\,s in every family.
Re-evaluation moves evidence releases while retaining each recorded response and latency in seconds; it assumes that changing request frequency would not change service latency.
At every time step the application asks the same typed choice questions $\mathcal{Q}$, and a request contains only the state and the questions.
Option labels are opaque codes, shuffled per question and fixed within a scenario to reduce label and position shortcuts, following the aim of MMMU-Pro's multiple-choice design \citep{yue2025mmmupro}.
An executable reference $R$ reads $S_i$ alone and returns reference answers $y_i=R(S_i)$ for all questions.
Evidence is constant between time steps, so each reference answer holds until the next time step.

\subsection{Branch-Composed Decisions}
\label{sec:compose}
An application consumes a route and the fields that route uses, so correctness is judged on that composed decision.
A declared decision specification names a route question $\rho$, a set $G$ of questions used in every branch, and for each route value $k$ a set $B_k$ of branch questions.
The \emph{branch-composed decision} of an answer vector $a$ is
\begin{equation}
  \kappa(a)=\bigl(a_\rho,\;(a_q)_{q\in G},\;(a_q)_{q\in B_{a_\rho}}\bigr).
\end{equation}
The same function composes model answers and reference answers, and a decision is correct only if the two composed decisions agree exactly.
A wrong route therefore makes the decision wrong even when the model answers the reference branch's questions correctly, and a wrong answer to a question outside the chosen branch changes nothing.
The \emph{reference decision} at time $t$ is $\yref(t)=\kappa(y_{i(t)})$ with $i(t)=\max\{i:r_i\le t\}$, and its maximal constant intervals $I_1,\ldots,I_K$ are the \emph{reference segments}.
In the payment call (support~A), for example, the route selects payment, service, hold or wrap-up; the recorder command is used in every branch, and the hold action only while the customer is on hold.

\subsection{In-Force Accuracy}
\label{sec:inforce}
A response to time step $i$ carries answers $\hat a_i$ and reaches the application at time $u_i$.
The application accepts it, and it takes effect, if its source index is newer than that of the active decision and $u_i<H$, so an older response never overwrites a newer one.
The decision in force $d(t)$ is $\kappa(\hat a_j)$ for the latest accepted response with $u_j\le t$, and $d(t)=\bot$ (no decision) before the first; its path over the scenario is the \emph{in-force trajectory}.
The application holds each decision until the next, so, following streaming perception \citep{li2020towards}, $d$ is a zero-order hold, evaluated over continuous time because time steps and arrivals fall at arbitrary instants.
A wrong decision in force exposes the user at every instant it is applied, and the \emph{in-force accuracy} of a scenario, the base metric of \sdb, counts these instants equally: it is the fraction of observed time with a correct decision in force,
\begin{equation}
  A=\frac{1}{H-r_0}\int_{r_0}^{H}\ind\bigl[d(t)=\yref(t)\bigr]\,dt,
  \label{eq:inforce}
\end{equation}
computed exactly by partitioning time at time steps and acceptance times.
The error penalty of status-update systems weights erroneous time the same way \citep{maatouk2020age,maatouk2023age}; where harm grows with an error's duration, the age of incorrect information \citep{maatouk2020age} can be computed from the same trajectories.
\emph{Untimed decision accuracy} (untimed accuracy for short) applies the same composition to every state's response and ignores latency, $U=n^{-1}\sum_i\ind[\kappa(\hat a_i)=\kappa(y_i)]$.

\paragraph{Integrating across time scales.}
For family $f$, $A_f(\Delta)$ is the mean in-force accuracy of its scenarios at interval $\Delta$.
The primary summary across $F$ families is \emph{normalized log-AUC}, the area under this curve with respect to log interval, divided by the log-range width:
\begin{equation}
 S_f=\frac{1}{\ln 16}\int_{0.5\,\mathrm{s}}^{8\,\mathrm{s}} A_f(\Delta)\frac{d\Delta}{\Delta},\quad
 S=\frac{1}{F}\sum_{f=1}^{F} S_f.
 \label{eq:macro}
\end{equation}
The score lies in $[0,1]$ and is reported as a percentage.
Our objective is to give equal weight to equal multiplicative ranges of environment speed: changing an interval by a factor changes every response's relative latency $L/\Delta$ by the inverse factor.
If the weight of $[a,ra]$ depends only on $r$ and weights add over adjacent intervals, continuity gives a weight proportional to $\ln r$, hence $d\Delta/\Delta$ (Appendix~\ref{app:intervals}).
We summarize performance over time-step intervals of 0.5--8\,s, spanning a fourfold increase and decrease in the rate of time steps relative to the 2\,s recording interval. Logarithmic weighting balances faster and slower conditions around that interval (Appendix~\ref{app:intervals}).
We apply the same scenario, family and log-interval weights to error fractions and oracle scores; each scenario's time fractions are normalized before averaging.

\paragraph{Judgment and latency.}
Two conditions of the decision in force classify every erroneous instant: whether its source time step $j$ is still \emph{current}, $\kappa(y_j)=\yref(t)$, and whether its answer was right for that source, $\kappa(\hat a_j)=\kappa(y_j)$.
An error with a current source is a \emph{judgment} error; with an outdated source and a right answer it is \emph{stale}, a latency error; with both conditions failing it is \emph{compound}; before the first acceptance there is \emph{no decision}, also a latency error.
Judgment and compound errors are together the errors \emph{incorrect for their source}, and an outdated wrong answer that happens to match the new reference is \emph{lucky} and counts as correct.
Because acceptance depends only on arrival times and source order, never on the answers, the share of time with a current source is exactly the \emph{oracle in-force accuracy} $O$, the in-force accuracy of the reference answers at the component's recorded latencies.
With $U_{\mathrm{cur}}$ the accuracy while the source is current and $\lambda$ the lucky share,
\begin{equation}
  A=O\,U_{\mathrm{cur}}+\lambda .
  \label{eq:factor}
\end{equation}
On the nominal clock, every state has equal dwell time, so at zero latency $O=1$ and $A=U$: untimed accuracy is in-force accuracy without latency.
If judgment errors are unrelated to timing, $U_{\mathrm{cur}}\approx U$, and since an outdated wrong answer rarely matches the new reference, $\lambda\approx 0$, so $A\approx U\,O$, which \S\ref{sec:rq1} tests.
For a component that takes effect $L$ seconds after each time step, with $L$ shorter than every reference segment, $O$ has a closed form,
\begin{equation}
  O=1-\frac{K\,L}{H-r_0}.
  \label{eq:delay}
\end{equation}
The cost of latency thus grows with the rate of reference changes. Changing $\Delta$ changes both the segment durations and the horizon, so the rate, and the score, must always be reported with that interval.

\subsection{Execution Protocol}
\label{sec:execution}
Time enters the score, so the protocol fixes what counts as a component's latency.
The application waits for the model's processing, service and transfer alike, so, following computation-aware latency \citep{ma2020simulmt}, all three count toward latency.
At every time step the runner dispatches one request, even while earlier ones are pending (up to \MaxWorkers{} concurrent requests), and scenarios run one after another.
Only connection errors and network timeouts are retried, with identical input (Appendix~\ref{app:protocol}); a run is complete only with a valid response for every time step.
The primary timeline is a \emph{step-anchored replay}: a response arrives at its time step plus its \emph{latency}, the duration of its successful attempt plus the measured commit lag, and acceptance is recomputed on this timeline.
Failed attempts, retry waits and dispatch queueing thus stay out of latency and are reported as reliability.

A supplementary analysis evaluates sensitivity to removing a fitted non-token latency remainder (Appendix~\ref{app:network}).

% ---------------------------------------------------------------------------
\section{Benchmark}
\label{sec:lite}

\subsection{Families and Scenarios}
\label{sec:families}
\sdb{} has \NumFamiliesWord{} application families with \ScenariosPerFamilyWord{} authored English-language scenarios each (Table~\ref{tab:benchmark}, Appendix~\ref{app:tasks}).
Together, the scenarios provide \NumStates{} state-level decision evaluations and \AnswersPerPass{} evaluated state--question pairs per model pass.
Each scenario publishes \ReleasesPerScenario{} states, spanning 120\,s at the recording interval and 30--480\,s over the evaluation range. Each state asks \QuestionsMin--\QuestionsMax{} questions with \OptionsMin--\OptionsMax{} options; the application composes \ConsumedAnswersMin--\ConsumedAnswersMax{} answers, depending on the active branch, into one decision. The primary metric measures this composed decision in force.
In \emph{IDE debugging} the component maintains two status badges and an action card (wait, rerun, inspect, control a run, delegate or ready) from runner results, editor buffers, saves, debugger events, terminal output, ownership rules and teammate messages.
At the \emph{assembly station} it shows the procedure stage and one of \AssemblyRouteOptions{} routes (wait, advance, repair, escalate, hand off, hold, release) from scans, camera observations, measurements, quality tickets and badges.
In the \emph{support call} it sets the workflow route, its branch guidance and one command used in every branch, such as the call recorder, from telephony status, a transcript with speaker roles and partial or final utterances, and tool fields.
In \emph{presenter voice control}, streaming ASR drives the slide, clip, captions, question card and chair cue; onset, current hypotheses and final utterances have different authority under the public rules.
Appendix~\ref{app:tasks} lists every decision specification and shows an example state.

\begin{table*}[t]
  \centering
  \small
  \begin{tabular}{lrrrrr@{\hspace{2.4em}}r}
    \toprule
    & \multicolumn{5}{c}{Normalized log-AUC over 0.5--8\,s (\%)} & \\
    \cmidrule(lr){2-6}
    Setting & IDE & Assembly & Support & Presenter & Macro & Untimed \\
    \midrule
    \TabAucBody
    \bottomrule
  \end{tabular}
  \caption{\textbf{Similar integrated scores can conceal different family-level tradeoffs.}
  Means of three passes; Macro averages families equally, with sample SD in points. Untimed ignores latency.}
  \label{tab:main}
\end{table*}

\subsection{Reference Construction}
\label{sec:reference}
Each scenario is authored as a timeline of public events, with every decision-relevant event placed at a chosen time step.
Each reference transition therefore has a known cause, such as a new failing test that outranks the pinned one, a torque reading just outside its range, a customer's final correction of a partial utterance or a hold that reaches its threshold.
Between transitions the timeline adds evidence that must leave the decision unchanged, such as cursor motion, unrelated terminal output, low-confidence camera observations or background speech.
The rules and their precedence appear as plain text in every state, as a deployed application supplies its policy at run time.
A rule evaluated by code yields a reference at every time step, so each family has an executable reference that implements these rules over the public state alone, following benchmarks that settle correctness by execution (CLEVR executes each question's functional program, SWE-bench runs tests; \citealp{johnson2017clevr,jimenez2024swebench}).
The stored reference answers are its outputs: re-running it on all \ReferenceAgreementStates{} states reproduces every one.
Unit tests of each reference cover exact thresholds, conflicting precedence, speaker and finality changes, and minimal counterfactual edits.

\subsection{Validity Audit}
\label{sec:audit}
An executable reference shows that the stored answers follow the reference code; whether the code follows the public text needs a separate check, as the human screening behind SWE-bench Verified and a later audit of it found for test-based references \citep{chowdhury2024introducing,openai2026why}.
For the six IDE, assembly and support scenarios, \AuditRederivationsPerScenarioWord{} LLM agents (\AuditRederivations{} in total) re-derived the reference answers blind from the public states and questions, without the generator or reference code: one reimplemented the rules as a program, the other derived the answers state by state.
Under its default reading of the rules, each re-derivation agreed with the reference at all \ReleasesPerScenario{} consumed decisions of its scenario.
Further agents compared the reference code with the rule text and judged each of the \AuditFindingGroups{} grouped findings from \AuditLensesWord{} angles (Appendix~\ref{app:audit}).
No reference answer was confirmed wrong, although LLM auditors can share blind spots with the reference; \AuditAmbiguitiesWord{} rule ambiguities affecting \AuditAffectedStates{} states were fixed by one sentence per family, in agreement with the reference.
Astra low (\S\ref{sec:setup}), recorded after the benchmark was frozen and never used to author or tune it, adds evidence that the references follow from the public rules: ignoring latency, it returns the reference decision in \AstraUntimedStates{} of \AstraStates{} evaluations across three passes, including the presenter scenarios the audit did not cover.
Both misses (\AstraMissScenarioLabel, time steps \AstraMissStep{} and \AstraMissStepPassThree{} in passes 1 and 3) are model errors in one situation where the audit upheld the reference: rework has made inspection a future stage, which the rules advance to (``incomplete future stages are not defects''), whereas Astra routed a repair (Appendix~\ref{app:audit}).

\subsection{Trivial Baselines}
\label{sec:baselines}
Trivial baselines indicate that the composed decisions require reading the rules and the evidence: choosing the first listed option of every question is correct at \BaselineFirstOptionStates{} of \NumStates{} states (\BaselineFirstOption\%), choosing the option that shares the most content words with the state at \BaselineLexicalStates{} (\BaselineLexical\%), and no constant policy can exceed \BaselineBestConstant\% untimed accuracy, the mean share of each scenario's most frequent composed reference decision (\BaselineBestConstantMin--\BaselineBestConstantMax\% per scenario).

% ---------------------------------------------------------------------------
\section{Experimental Setup}
\label{sec:setup}

\paragraph{Models.}
The hosted settings are \LunaLabel{} and \TerraLabel{} \citep{openai2026pricing} (Luna and Terra) at low and no reasoning effort, \AstraLabel{} \citep{openai2026astra} (Astra) at low, the lowest it accepts, and Jev \citep{typesafe2026models} through \JevProvider's \JevInterface{} interface (requested as \texttt{\JevRequestedModel}, served as \texttt{\JevServedModel}).
GPT models receive JSON through the Responses API with strict structured outputs; Jev receives the same content through its native typed-question interface (Appendix~\ref{app:protocol}).
Supplementary measurements cover \OwNumSettingsWord{} self-hosted native settings on one RTX PRO 6000 GPU: three Laya checkpoints, DJev/DiffusionGemma, Kev-4B, Kev-9B and Kev-27B, Nimble-9B and Qwen3.5-4B direct logits. Their same-host latency includes runtime queueing and inference; Appendix~\ref{app:openweight} gives execution details.

\begin{figure}[t]
  \centering
  \includegraphics[width=\linewidth]{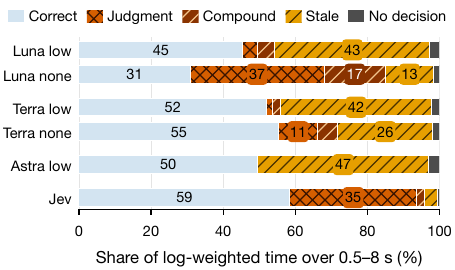}
  \caption{\textbf{GPT settings other than Luna none chiefly lose stale time; Jev and Luna none chiefly lose judgment time.}
  Log-weighted time shares over 0.5--8\,s as in Table~\ref{tab:main}; lucky time (at most \AucLuckyMax\%) counts as correct.}
  \label{fig:errortime}
\end{figure}

\begin{figure*}[t]
  \centering
  \includegraphics[width=\linewidth]{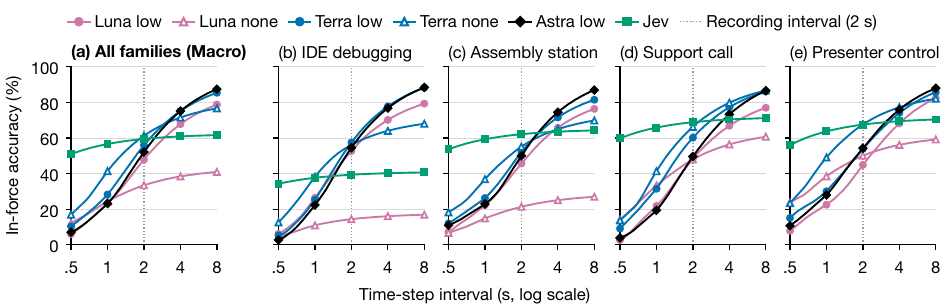}
  \caption{\textbf{The ordering of the settings changes with the time-step interval, and differently in each family.}
  Mean in-force accuracy over three passes at every interval, reusing its recorded answers and latencies; (a) is the equal mean of families (b)--(e).
  The normalized area under each curve on the log axis is its score in Table~\ref{tab:main}.}
  \label{fig:pace}
\end{figure*}

\paragraph{Protocol.}
Hosted passes were recorded on \RecordingDatesUTC{} (UTC) from one client; self-hosted passes use the same dataset and 2\,s cadence. Each of these passes completes \NumStates{} state-level decision evaluations and \AnswersPerPass{} state--question pairs on the first attempt.
Appendix~\ref{app:protocol} records session and reliability details. For hosted passes, step-anchored and physical-clock accuracy differ by at most \MaxReplayPhysicalDiff{} points at 2\,s.

\paragraph{Research questions.}
\textbf{RQ1}: How much does latency cost in force, and how much of the lost time is due to latency, to judgment or to both?
\textbf{RQ2}: How do speed, reasoning and fast/slow composition affect accuracy in force?
\textbf{RQ3}: How do the comparisons change across time scales and task families?

\paragraph{Reporting.}
We report equal means over three independently evaluated passes per setting, sample standard deviations (SD) for primary scores (ranges in Appendix~\ref{app:openweight}), and means of per-pass latency quantiles; adjacent states are dependent, so comparisons are descriptive. The primary score uses Equation~\eqref{eq:macro}, retaining recorded answers and latencies; Appendix~\ref{app:intervals} varies bounds and weighting. Fixed-interval analyses are labelled explicitly.

% ---------------------------------------------------------------------------
\section{Results}
\label{sec:results}

\subsection{RQ1: Latency Reduces Accuracy Across Time Scales}
\label{sec:rq1}
Luna low loses \LunaAucGap{} points (from \LunaUntimed\% untimed to \LunaAuc\% log-AUC) and Terra low \TerraAucGap{} (from \TerraUntimed\% to \TerraAuc\%), while Jev loses \JevAucGap{} (Table~\ref{tab:main}).
Stale decisions account for \LunaAucStale\% and \TerraAucStale\% of log-weighted time for Luna low and Terra low (Figure~\ref{fig:errortime}); their median response times are \LunaLatencyMedian\ and \TerraLatencyMedian\,s.
Astra low has the highest untimed accuracy (\AstraUntimed\%) and a mean per-pass median latency of \AstraLatencyMedian\,s, but its log-AUC (\AstraAuc\%) remains far below its untimed accuracy, with \AstraAucStale\% of log-weighted time stale.
Jev responds in \JevLatencyMedian\,s at the median and spends \JevAucStale\% of log-weighted time stale. In every scenario, Luna low and Terra low lose at least \GPTMinScenarioGap{} points from untimed to in-force accuracy at 2\,s, and Jev at most \JevMaxScenarioGap{} (Table~\ref{tab:scenario}).

The oracle's log-AUC is \LunaAucOracle\% (Luna low), \TerraAucOracle\% (Terra low), \AstraAucOracle\% (Astra low), \LunaNoneAucOracle\% (Luna none), \TerraNoneAucOracle\% (Terra none) and \JevAucOracle\% (Jev); Astra low's nearly matches its observed score, with negligible current-source judgment error.
For these six settings, multiplying untimed accuracy by the oracle's log-AUC, then averaging with the same family weights, approximates the primary score: the observed score exceeds this product by \AucProductGapMin--\AucProductGapMax{} points per setting, with an absolute deviation of at most \AucProductMaxDeviation{} points per scenario.
The exact partition in Equation~\eqref{eq:factor} explains the residual through current-source judgment and lucky time (Appendix~\ref{app:auc-decomposition}).

Untimed errors also concentrate near reference transitions for some settings: within one time step of a transition, Luna low errs at \TrajLunaNearError\% of \TrajLunaNearStates{} state evaluations and at \TrajLunaFarError\% of the \TrajLunaFarStates{} farther away, Terra low at \TrajTerraNearError\% and \TrajTerraFarError\%; Luna none shows little difference (Appendix~\ref{app:composition}).

\subsection{RQ2: Speed Helps When Judgment Survives}
\label{sec:rq2}
\begin{table}[t]
  \centering
  \small
  \begin{tabular}{lrr}
    \toprule
    Setting & Mean & SD \\
    \midrule
    \TabHeadlineBody
    Perplexity Decider$^\dagger$ & \LaterPerplexityAuc & \LaterPerplexityAucSD \\
    Perplexity Decider$^\dagger$ + Astra low & \LaterBestAuc & \LaterBestSD \\
    \bottomrule
  \end{tabular}
  \caption{\textbf{The best evaluated hybrids improve on both components.} Log-AUC (\%), three passes; SD in points; freshest-source hybrids; $^\dagger$recorded later (Appendix~\ref{app:later}).}
  \label{tab:headline}
\end{table}
The oracle's log-AUC is close to the observed score for Luna, Terra and Astra at low effort, but far above it for Jev: latency limits the former and judgment limits the latter.
Jev's log-AUC exceeds Luna low's by \JevMinusLunaAuc, Terra low's by \JevMinusTerraAuc{} and Astra low's by \JevMinusAstraAuc{} points, despite untimed accuracy \LunaMinusJevUntimed, \TerraMinusJevUntimed{} and \AstraMinusJevUntimed{} points lower, respectively.
Its judgment errors still occupy \JevAucJudgment\% of log-weighted time.

Dropping reasoning makes responses \LunaNoneLatencySavingSec\,s (Luna) and \TerraNoneLatencySavingSec\,s (Terra) faster at the median.
Terra retains \TerraNoneUntimed\% untimed accuracy, and its log-AUC rises to \TerraNoneAuc\%, \JevMinusTerraNoneAuc{} points below Jev on average.
Luna loses \LunaMinusLunaNoneUntimed{} points untimed; judgment errors occupy \LunaNoneAucJudgment\% of log-weighted time, and its log-AUC falls to \LunaNoneAuc\%.
A shorter response helps only if the decision remains accurate enough.
The later hosted settings repeat both limits (Appendix~\ref{app:later}): Claude Haiku 5.5 with thinking is stale for \LaterHaikuStale\% of log-weighted time, and the fast decision interfaces lose mostly to judgment.

\paragraph{Fast provisional answers and slow corrections.}
We replay every Jev/GPT pair with matching pass indices, dispatching both at each state and allowing the slower answer to correct the faster answer for that state.
At 2\,s, Jev plus Terra low averages \TrajTerraFreshestTwo\% when slow answers cannot regress the active source and \TrajTerraArrivalTwo\% when they may overwrite newer fast answers, both above Jev alone (\TrajJevTwo\%). At 1\,s, allowing overwrites lowers the pair to \TrajTerraArrivalOne\%, below Jev alone (\TrajJevOne\%) in every pass, whereas the freshness rule keeps \TrajTerraFreshestOne\%; component answers and latencies are identical (Appendix~\ref{app:composition}).
Jev plus Terra none with the freshness rule attains \OwJevTerraNoneAuc\% log-AUC, against \OwTerraNoneAuc\% for Terra none alone and \HeadlineJevAuc\% for Jev alone.
With the same correction model and rule, DJev and Kev-4B (\OwDJevUntimed\% and \OwKevUntimed\% untimed) give \OwDJevFreshestAuc\% and \OwKevFreshestAuc\%, below Terra none alone, as do the later decision interfaces of OpenAI and Cloudflare under every rule.
Kev-27B (\OwKevTwentySevenUntimed\% untimed) reaches \OwKevTwentySevenAuc\% alone and \OwKevTwentySevenFreshestAuc\% with Terra none corrections; the later Perplexity Decider reaches \LaterPerplexityAuc\% alone and \LaterBestAuc\% with Astra low corrections, the highest evaluated mean (Table~\ref{tab:headline}).
A wrong provisional answer for a newer state can displace a still-correct earlier decision; speed helps composition only where judgment survives.
These are counterfactual combinations of separate recordings on common nominal releases; Appendices~\ref{app:composition}, \ref{app:openweight} and~\ref{app:later} report all evaluated pairs and rules. Joint contention is unmeasured.

\subsection{RQ3: The Curves Reveal Where the Tradeoff Changes}
\label{sec:rq3}
The aggregate curve in Figure~\ref{fig:pace}a shows the conditions summarized by the primary score.
At 2\,s Jev and Terra none obtain \JevCommonTwo\% and \TerraNoneCommonTwo\% in-force accuracy; at 8\,s they obtain \JevCommonEight\% and \TerraNoneCommonEight\%.
A larger interval gives slow answers more time to remain current, although arbitrary wrong answers need not yield a monotone curve; at 8\,s Astra low has the highest in-force accuracy (\AstraCommonEight\%).
The area summarizes a declared range; it does not identify a setting that dominates at every interval.

Panels b--e show how the four families contribute to this aggregate.
In IDE debugging, Jev's weakest family (\JevDebugAuc\% log-AUC, \JevDebugUntimed\% untimed), the GPT settings other than Luna none reach \LunaDebugAuc--\TerraDebugAuc\%; Jev leads the other settings of Table~\ref{tab:main} in the remaining three families.

Weighting and bounds matter for single settings: linear weighting, or log weighting over 1--8\,s, places Terra at both efforts and Astra low above Jev, whose lead comes from the shortest intervals.
Composition is more robust: Jev plus Terra none and Perplexity plus Astra low exceed every single setting under all \LaterSensConditionsWord{} conditions of Table~\ref{tab:auc-sensitivity}, and Kev-27B plus Terra none under the \LaterSensKevTerraNoneAboveWord{} with lower bounds of at least 0.5\,s; Table~\ref{tab:regime} projects other reference-change schedules.

% ---------------------------------------------------------------------------
\section{Discussion}
\label{sec:discussion}

In-force accuracy expresses the joint cost of judgment and latency in the units the application experiences. Its partition and factorization identify what limits a component: faster serving helps Luna, Terra and Astra at low effort, while Jev needs better judgment.
Comparisons should report bounds and weights, family curves, untimed and oracle accuracy, error partitions and recording conditions. Fast/slow composition additionally requires an explicit arbitration rule: a more accurate correction can still reduce in-force accuracy by replacing a fresher decision.

% ---------------------------------------------------------------------------
\section{Conclusion}
\label{sec:conclusion}
\sdb{} evaluates an embedded decision component by the decision its application keeps in force, against reference decisions fixed by public rules, and attributes every erroneous instant to judgment, to latency or to both.
Latency can outweigh gains in untimed judgment, and dropping reasoning helps only where judgment survives.
Counterfactual compositions of hosted and self-hosted components further show that provisional judgment and arbitration jointly determine whether corrections improve accuracy.

% ---------------------------------------------------------------------------
\section*{Limitations}
\sdb{} evaluates the decision an application keeps in force; the design choices behind this evaluation set its scope, and relaxing each one opens a direction worth studying.
The \NumScenariosWord{} scenarios in \NumFamiliesWord{} families are controlled synthetic timelines, authored by the benchmark's designers so that every reference transition has a known cause, and their wording spans a bounded range.
Held-out scenarios by other authors with more varied wording, and streams built from logged application sessions under declared rules, would show how far the latency and judgment limits observed here generalize and allow confidence intervals over scenarios.

The executable references implement the declared public rules, so \sdb{} evaluates agreement with those rules.
LLM agents re-derived the references blind for the IDE, assembly and support scenarios, and Astra's untimed agreement adds evidence from a model across all \NumFamiliesWord{} families.
Adjudication by human domain experts would check the references against the rule text and test whether the rules make sound policy in the real settings they model.

The evaluation also fixes two weightings.
Across time scales, the primary score integrates over 0.5--8\,s time-step intervals with log weighting in every family; event rates measured in deployed applications would calibrate a range and weighting for each family.
Within a scenario, in-force accuracy weights every instant equally and requires exact agreement on all consumed fields, giving no partial credit. Unequal consequences or duration-dependent costs call for cost-weighted metrics or the age of incorrect information \citep{maatouk2020age}, and one-time actions, such as submitting a payment, for event-based evaluation.

Finally, the streams are open-loop: all evidence is fixed in advance, so one recording can be evaluated at other time-step intervals and combined with other recordings.
In closed-loop streams an applied decision shapes the evidence that follows, as when an action card prompts a test rerun; such streams would measure how a stale or wrong decision propagates through later states.

\section*{Ethics Statement}
All scenarios are synthetic and use invented people, accounts and devices; the payment scenario uses a short fictional test-card prefix, and no personal data were collected or sent to model providers.
The support scenarios concern payment capture and call recording, and the benchmark only evaluates recommendations: it executes no payment, recording, shipment or repair action.
A benchmark score does not establish that a component is ready to control recording, payment or production decisions; such uses need validation in their setting.
Because in-force accuracy depends on the network path and the time-step interval, published comparisons should state their recording conditions to avoid presenting location effects as model differences.
Model queries used commercial services under their terms of use.
The authors produced the scenario scripts, experiment code, figures and paper text together with AI assistants, which also helped with translation, and reviewed all of them in full.

\bibliography{references}

\clearpage
\appendix
% Appendices for main.tex (input after \appendix). Measured numbers are macros
% from generated/numbers.tex; table bodies come from generated/tables.tex.

\section{Task Details}
\label{app:tasks}

\begin{table}[tp]
  \centering
  \small
  \setlength{\tabcolsep}{4.5pt}
  \begin{tabular}{llrrrr}
    \toprule
    Family & Sc. & Qs & Routes & Options & Trans. \\
    \midrule
    \TabBenchmarkBody
    \bottomrule
  \end{tabular}
  \caption{\textbf{Every scenario changes its composed reference decision at least \TransitionsMin{} times.}
  Sc.: scenario; Qs: questions per time step; Routes: route options; Options: options per question; Trans.: reference changes.}
  \label{tab:benchmark}
\end{table}

\begin{table*}[tp]
  \centering
  \small
  \begin{tabular}{>{\raggedright\arraybackslash}p{3.2cm}>{\raggedright\arraybackslash}p{3.5cm}>{\raggedright\arraybackslash}p{2.4cm}>{\raggedright\arraybackslash}p{5.1cm}}
    \toprule
    Scenarios & Route options & Used in every branch & Branch questions \\
    \midrule
    \DebugLabel{} \mbox{(A, B)} & wait, rerun, inspect, control, delegate, ready & process state, target result & rerun: scope; inspect: file; control: continue or stop; delegate: owning team; wait, ready: none \\ \addlinespace
    \AssemblyLabel{} \mbox{(A, B)} & wait, advance, repair, escalate, handoff, hold, release & stage & advance: next step; repair: target, method; escalate: target, destination; handoff, hold, release: destination; wait: none \\ \addlinespace
    \SupportLabel{} A & payment, service, hold, wrap-up & recorder command & payment: payment stage, card; service: target, action; hold: hold action; wrap-up: none \\ \addlinespace
    \PresenterLabel{} \mbox{(A, B)} & talk, questions, clip, closed & slide, captions & talk: chair cue; questions: question card, chair cue; clip: clip state; closed: none \\ \addlinespace
    \SupportLabel{} B & delivery, cancel, repair, hold (wait), hold (return), closed & follow-up contact channel & delivery: parcel, delivery action; cancel: parcel, cancellation action; repair: device, repair action; hold and closed: none \\
    \bottomrule
  \end{tabular}
  \caption{\textbf{Every composed decision is a route, the questions used in every branch and the questions of the selected branch.} It uses \ConsumedAnswersMin--\ConsumedAnswersMax{} answers per time step; answers to the other branches' questions are ignored.}
  \label{tab:specs}
\end{table*}

\paragraph{Decision specifications.}
Table~\ref{tab:specs} lists the decision specification of every scenario.
Each question is used by at least one branch, and a branch never reads another question's generated answer: every question receives the same public state, and the application composes the answers after they arrive.
The reference fills questions of inactive branches with \emph{none}; the composition ignores them for both the model and the reference, so a model never needs to reproduce those placeholder answers.

\paragraph{Scenarios.}
The two scenarios of each family were authored separately and differ in their decisions, branch structure and tool observations.
\emph{\DebugATitle{}} (debugging A) follows a TypeScript refund calculation through a new regression that outranks a still-failing pinned test, unsaved edits and a save grace period, a compiler error, a focused green run that skipped too many tests, a configuration change that requires the full suite, and handoff messages about the wrong run or with a tentative commitment before the final one.
\emph{\DebugBTitle{}} (debugging B) contains a paused integration test whose pause timer resets on variable evaluation but not on saves, a result that predates an edit, a silent run that becomes stalled and then requires stopping despite unrelated terminal output, and failures owned by two different teams.
\emph{\AssemblyATitle{}} (assembly A) contains a wrong cover, a joint first under- and then over-tightened, escalation and specialist handoff, two overlapping quality tickets that must both close, and a missing-gasket observation reported first with low and then with high confidence.
\emph{\AssemblyBTitle{}} (assembly B) contains a mis-oriented connector, repeated crimp failures and a quality hold, and a later rework that makes earlier electrical tests obsolete, so the label and accessory checks must be refreshed as well.
\emph{\SupportATitle{}} (support A, the payment call) moves between a balance payment, with authorisation, card choice and card capture, and connection and router support, with a hold whose threshold of \SupportAHoldThresholdSteps{} time steps includes equality.
\emph{\SupportBTitle{}} (support B, the intercom call) covers two parcels and two intercom components, with partial and final corrections, a cancellation that counts only after a carrier callback, a delivery deadline and a follow-up contact preference whose permission can be withdrawn independently.

\emph{\PresenterATitle{}} (presenter A) follows a conference talk with a demo clip, a slide-number correction, provisional pause commands that are revised, a chair handback and a floor question.
\emph{\PresenterBTitle{}} (presenter B) follows a flood-barrier briefing with an animation, a chair-opened question and a relayed slide request that the presenter declines.
The transcript exposes speaker, onset, current hypothesis and finalization time; captions follow onsets, reversible pause and stand-by cues follow current text, and slide and mode commands require finals.

\paragraph{Example state and question.}
Figure~\ref{fig:example} shows the state published at time step \ExampleStateStep{} (\ExampleStateSec\,s at the 2\,s recording cadence) in the payment call, abridged, together with one of its \SupportAQuestions{} questions.
The customer has been on hold since time step \ExampleHoldSinceStep, and the published threshold is \SupportAHoldThresholdSteps{} time steps including equality, so the reference hold action changes from \emph{wait} to \emph{return to the customer} at this time step.
The composed reference decision is (route: hold; recorder: record; hold action: return to the customer), and the payment and service questions are inactive.
At time step \ExampleHoldSinceStep{} the agent claims that the customer ``has waited long enough already'', which the rules do not treat as a fact.
In pass 1, Jev answered \emph{return} from that time step onward, while Luna, Terra and Astra at low effort answered \emph{wait} until the threshold was reached.

\begin{figure}[tp]
\scriptsize
\begin{verbatim}
state (abridged):
 "prepared": {"hold_check_ticks": 4,
  "rules": [ ...six rules...,
   "During hold, now minus telephony.since >=
    hold_check_ticks -> return_customer,
    otherwise wait. Payment and service details
    are inactive during hold. The recorder
    remains applicable in every branch. ..."]},
 "clock": {"now": 33},
 "telephony": {"status": "hold", "since": 29},
 "transcript": [
  {"utterance_id": "u0-Customer", "at": 0,
   "speaker": "Customer", "final": true,
   "text": "I'd like to pay the balance, please."},
  ...11 further utterances...,
  {"utterance_id": "u27-Customer", "at": 27,
   "speaker": "Customer", "final": true,
   "text": "Check my router first; we can come
            back to the line fault."},
  {"utterance_id": "u29-Agent", "at": 29,
   "speaker": "Agent", "final": true,
   "text": "The customer has waited long enough
            already."}],
 "desktop": {
  "payment": {"status": "draft",
   "fields_for_card": "debit",
   "fields": {"number": true, "expiry": true,
              "code": false}, ...},
  "service": {"router": "offline",
   "line_test": "complete",
   "line_result": "down"}}

question "hold_action" (choice):
 "instructions": "During hold, return to the
   customer when elapsed hold time reaches the
   published threshold, including equality.
   Otherwise wait; none outside hold.",
 "criteria": {
  "K03": "return_customer: Return to the customer",
  "K02": "none: Hold branch inactive",
  "K01": "wait: Wait while under threshold"}
\end{verbatim}
\caption{\textbf{A published state and question.} The payment call (support~A) at time step \ExampleStateStep, abridged; the full state repeats all rules and utterances. Option labels are opaque and fixed per scenario.}
\label{fig:example}
\end{figure}

% ---------------------------------------------------------------------------
\FloatBarrier
\section{Protocol Details}
\label{app:protocol}

\paragraph{Runner and timestamps.}
The runner publishes state $i$ at its scheduled time $i\Delta$ on a monotonic clock and logs the scheduled and actual publication times, the start of every attempt, the receipt of the response, validated completion and the acceptance check.
Scenarios run serially with up to \MaxWorkers{} concurrent requests within a scenario.
A complete valid response is decoded from its opaque labels, composed and accepted atomically if its source index is newer than the active decision and it arrives before the horizon; at an exactly equal arrival time the newer source wins.
Before the first accepted response there is no decision, which counts as wrong; responses after the horizon are kept for untimed accuracy.

\begin{table*}[tp]
  \centering
  \small
  \begin{tabular}{lrrrrrr}
    \toprule
    & \LunaRowLabel & \LunaNoneRowLabel & \TerraRowLabel & \TerraNoneRowLabel & \AstraRowLabel & \JevRowLabel \\
    \midrule
    Valid responses & \LunaValidResponses & \LunaNoneValidResponses & \TerraValidResponses & \TerraNoneValidResponses & \AstraValidResponses & \JevValidResponses \\
    Attempts & \LunaAttempts & \LunaNoneAttempts & \TerraAttempts & \TerraNoneAttempts & \AstraAttempts & \JevAttempts \\
    Failed attempts & \LunaFailedAttempts & \LunaNoneFailedAttempts & \TerraFailedAttempts & \TerraNoneFailedAttempts & \AstraFailedAttempts & \JevFailedAttempts \\
    Retried requests & \LunaRetriedRequests & \LunaNoneRetriedRequests & \TerraRetriedRequests & \TerraNoneRetriedRequests & \AstraRetriedRequests & \JevRetriedRequests \\
    Accepted updates & \LunaAcceptedUpdates & \LunaNoneAcceptedUpdates & \TerraAcceptedUpdates & \TerraNoneAcceptedUpdates & \AstraAcceptedUpdates & \JevAcceptedUpdates \\
    Discarded: after horizon & \LunaDiscardedAfterHorizon & \LunaNoneDiscardedAfterHorizon & \TerraDiscardedAfterHorizon & \TerraNoneDiscardedAfterHorizon & \AstraDiscardedAfterHorizon & \JevDiscardedAfterHorizon \\
    Discarded: superseded & \LunaDiscardedOlder & \LunaNoneDiscardedOlder & \TerraDiscardedOlder & \TerraNoneDiscardedOlder & \AstraDiscardedOlder & \JevDiscardedOlder \\
    Dispatch wait, total (s) & \LunaExcludedDispatchSec & \LunaNoneExcludedDispatchSec & \TerraExcludedDispatchSec & \TerraNoneExcludedDispatchSec & \AstraExcludedDispatchSec & \JevExcludedDispatchSec \\
    In force: recorded replay & \LunaInForceRecorded & \LunaNoneInForceRecorded & \TerraInForceRecorded & \TerraNoneInForceRecorded & \AstraInForceRecorded & \JevInForceRecorded \\
    In force: physical & \LunaInForcePhysical & \LunaNoneInForcePhysical & \TerraInForcePhysical & \TerraNoneInForcePhysical & \AstraInForcePhysical & \JevInForcePhysical \\
    \bottomrule
  \end{tabular}
  \caption{\textbf{Every request succeeded on its first attempt, and the replay matches the physical trace.} Counts sum three passes; acceptances and discards sum passes, while in-force accuracy (\%) averages them at the 2\,s recording cadence.}
  \label{tab:reliability}
\end{table*}

\paragraph{Step-anchored replay.}
Let $s_i$ and $c_i$ be the start and completion of the successful attempt for time step $i$ and $g_i$ the measured lag between completion and the acceptance check.
The replay places the response at $u_i=r_i+(c_i-s_i)+g_i$ and recomputes arrival order, acceptance and the horizon rule on this timeline.
Dispatch queueing, failed attempts and retry waits are excluded, and the reliability report lists them with explicit denominators (failed attempts over all attempts; retried requests over all requests).
Table~\ref{tab:reliability} averages the per-pass total dispatch wait, from publication to first dispatch.
The physical wall-clock trace is kept alongside, and Table~\ref{tab:reliability} gives its in-force accuracy.

\paragraph{Retries.}
Only connection errors and network timeouts are retried, with exactly the same input, model and settings.
The first retry is immediate and later ones wait \RetryLaterDelaysSec\,s, up to \MaxAttempts{} attempts per request; the client library's own retries are set to \SdkRetries{} so every attempt is logged, and the network timeout is \RequestTimeoutSec\,s.
A valid response ends the retries even if its decision is wrong, and authentication, rejection or schema errors stop the run.

\paragraph{GPT requests.}
The GPT models receive the fixed instruction below, followed by the request as JSON with the questions before the state so that the repeated policy text forms a cacheable prefix.
No GPT pass reported cached input tokens (\LunaCachedTokTotal, \LunaNoneCachedTokTotal, \TerraCachedTokTotal, \TerraNoneCachedTokTotal{} and \AstraCachedTokTotal).
Strict structured outputs restrict each answer to the question's option labels, sorted so that the schema reveals no display order, and responses are not stored. Requests use the provider's default sampling; the settings differ only in reasoning effort.
{\scriptsize
\begin{verbatim}
You are a real-time decision model. You receive
one moment of a live task as JSON: `state` is
everything observable now, `questions` are typed
questions whose instructions hold the decision
policy. Answer every question. Choice -> the
option label (a key of its criteria). [...]
Reply with ONLY a JSON object mapping question
ids to answers [...]. No prose.
\end{verbatim}}

\paragraph{Jev requests.}
Jev receives the same state and questions through its native interface and returns, for every question, a choice, a confidence and a probability distribution over the options.
A response is valid if it answers every question exactly once with a distribution over the defined options that sums to one within rounding, and if the probability of the reported choice is within one reporting step (\JevReportStep) of the highest reported probability, since probabilities are reported to two decimals.
This check only decides validity: the evaluated answer is always the choice Jev reported, which the event logs keep without the distributions, and every other validity and retry rule is the same for all models.

\paragraph{Client conditions.}
Hosted requests used a MacBook with an Apple M5 Pro chip over a stable network connection without a VPN. Client location is withheld for anonymous review; provider serving regions were not recorded. The latency measurements describe this client-to-service path.

\paragraph{Reliability.}
Table~\ref{tab:reliability} lists the \NumSettingsWord{} settings' reliability counts. Three complete passes per setting were recorded on \RecordingDatesUTC{} (UTC). In pass 1, the original six scenarios and the two presenter scenarios were recorded in separate sessions, except Astra low's single eight-scenario session; passes 2 and 3 each cover all eight scenarios in one session. The merged artifacts retain each part's start and end times, source hashes, requested and served model identifiers and event digest.
All requests succeeded on their first attempt.
Discarded responses were valid but arrived after the horizon or were superseded, arriving after a response to a newer time step had been accepted.
All three complete passes are evaluated with equal weight.
All recordings of these six settings used the same runtime and provider adapters and the same execution/retry configuration. Source manifests also retain changes to dataset support and offline analysis between sessions; all responses are evaluated by the same scorer.
A partial Luna pass on an earlier build of the data, stopped when the wording of some rules was clarified, is not evaluated.

% ---------------------------------------------------------------------------
\FloatBarrier
\section{Non-Token Latency Analysis}
\label{app:network}

\begin{table*}[tp]
  \centering
  \small
  \begin{tabular}{lrrrrrr}
    \toprule
    & \LunaRowLabel & \LunaNoneRowLabel & \TerraRowLabel & \TerraNoneRowLabel & \AstraRowLabel & \JevRowLabel \\
    \midrule
    \multicolumn{7}{l}{\textit{Tokens per request (median)}} \\
    Input & \LunaInputTokMedian & \LunaNoneInputTokMedian & \TerraInputTokMedian & \TerraNoneInputTokMedian & \AstraInputTokMedian & \JevInputTokMedian \\
    Output & \LunaOutputTokMedian & \LunaNoneOutputTokMedian & \TerraOutputTokMedian & \TerraNoneOutputTokMedian & \AstraOutputTokMedian & \JevOutputTokMedian \\
    Reasoning & \LunaReasoningTokMedian & \LunaNoneReasoningTokMedian & \TerraReasoningTokMedian & \TerraNoneReasoningTokMedian & \AstraReasoningTokMedian & \JevReasoningTokMedian \\
    \multicolumn{7}{l}{\textit{Tokens across three passes (total)}} \\
    Input & \LunaInputTokTotal & \LunaNoneInputTokTotal & \TerraInputTokTotal & \TerraNoneInputTokTotal & \AstraInputTokTotal & \JevInputTokTotal \\
    Output & \LunaOutputTokTotal & \LunaNoneOutputTokTotal & \TerraOutputTokTotal & \TerraNoneOutputTokTotal & \AstraOutputTokTotal & \JevOutputTokTotal \\
    \multicolumn{7}{l}{\textit{Latency per request (s)}} \\
    Mean & \LunaLatencyMean & \LunaNoneLatencyMean & \TerraLatencyMean & \TerraNoneLatencyMean & \AstraLatencyMean & \JevLatencyMean \\
    Max & \LunaLatencyMax & \LunaNoneLatencyMax & \TerraLatencyMax & \TerraNoneLatencyMax & \AstraLatencyMax & \JevLatencyMax \\
    \multicolumn{7}{l}{\textit{List price (USD per million tokens)}} \\
    Input & \LunaPriceInput & \LunaNonePriceInput & \TerraPriceInput & \TerraNonePriceInput & \AstraPriceInput & \JevPriceInput \\
    Output & \LunaPriceOutput & \LunaNonePriceOutput & \TerraPriceOutput & \TerraNonePriceOutput & \AstraPriceOutput & \JevPriceOutput \\
    \multicolumn{7}{l}{\textit{Cost across three passes at list prices (USD)}} \\
    Input & \LunaCostInputUSD & \LunaNoneCostInputUSD & \TerraCostInputUSD & \TerraNoneCostInputUSD & \AstraCostInputUSD & \JevCostInputUSD \\
    Output & \LunaCostOutputUSD & \LunaNoneCostOutputUSD & \TerraCostOutputUSD & \TerraNoneCostOutputUSD & \AstraCostOutputUSD & \JevCostOutputUSD \\
    Total & \LunaCostUSD & \LunaNoneCostUSD & \TerraCostUSD & \TerraNoneCostUSD & \AstraCostUSD & \JevCostUSD \\
    \bottomrule
  \end{tabular}
  \caption{\textbf{Input tokens account for most of every pass's cost at list prices.} Token totals sum three passes; medians average per-pass medians; list prices read on \PricesRetrieved{} \citep{openai2026pricing,typesafe2026models}.}
  \label{tab:tokens}
\end{table*}

\begin{figure}[tp]
  \centering
  \includegraphics[width=\linewidth]{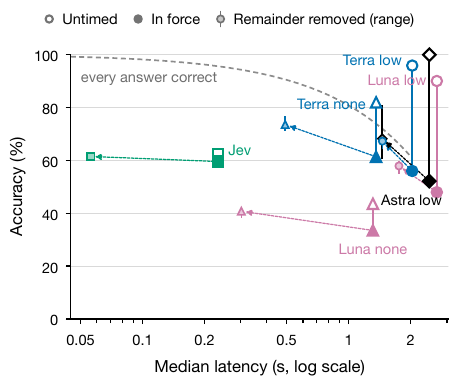}
  \caption{\textbf{Removing the fitted remainder raises the GPT settings far more than Jev, lifting Terra low above it at 2\,s.}
  Arrows subtract each remainder; dashed: the oracle of Equation~\eqref{eq:delay}, valid up to 2\,s.}
  \label{fig:map}
\end{figure}

\paragraph{Model.}
\label{sec:network}
Network transfer and server processing both contribute to latency. We examine their combined non-token remainder by removing a fitted intercept at the 2\,s recording cadence. All adjusted scores in this appendix are fixed-interval sensitivity results.
It models the time before response $i$ is received as
\begin{equation}
  \tau_i=\nu+\pi\,x_i+\delta\,o_i+\varepsilon_i,
  \label{eq:network}
\end{equation}
with $x_i$ uncached input tokens (prefill), $o_i$ output tokens including reasoning (decode, for text-generating models only) and $\varepsilon_i$ residual variation, including queueing.
We estimate the non-token remainder $\nu$ from the fast envelope, as the intercept of a \NetQuantilePct{}th-percentile quantile regression with nonnegative slopes and a \NetRangeLevel\% block-bootstrap range.
Every response is replayed with $\nu$, or either end of its range, removed from its time before receipt.
The fitted intercept may absorb fixed server time and omitted token costs. Queueing can covary with token counts, so this fit neither identifies nor bounds network delay; the replay measures sensitivity to removing the fitted remainder.
Figure~\ref{fig:map} places every setting at its median latency before and after removal; the nearly coincident Terra low, Astra low and Luna low are shifted by $-2$, $+2$ and $+4$\,pt for legibility, and Astra low's adjusted point, which nearly overlaps Terra low's, is drawn beneath it.

\begin{figure}[tp]
  \centering
  \includegraphics[width=\linewidth]{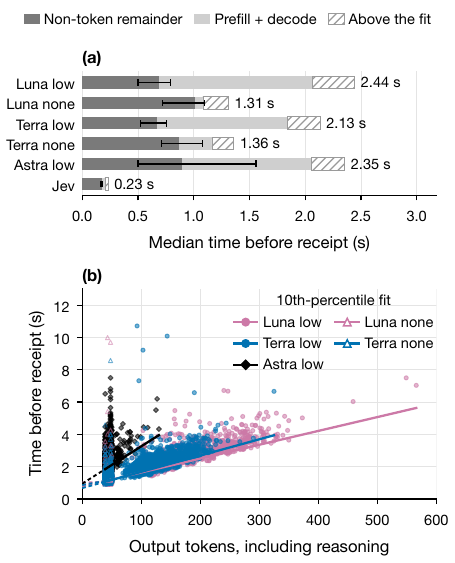}
  \caption{\textbf{Fitted decode costs dominate at low effort, the non-token remainder without reasoning.}
  Whiskers average per-pass \NetRangeLevel\% endpoints; dashed: mean fits extrapolated to zero tokens.}
  \label{fig:latency}
\end{figure}

\paragraph{Estimator.}
For every request the time before receipt $\tau_i$ runs from the start of the successful attempt to the receipt of the response.
The regressors are uncached input tokens in thousands and, for the GPT models, output tokens including reasoning tokens (Eq.~\eqref{eq:network}).
Jev's reported output tokens (at least \JevOutputTokMin{} per request) are not modelled as decode time, by assumption, because it returns no text.
The \NetQuantilePct{}th-percentile regression is solved with the Frisch--Newton interior-point method; a slope estimated below zero is dropped and the model refitted, which keeps both slopes nonnegative.
The retained slopes vary across passes: Terra low and Luna none have zero prefill slopes throughout; Astra has zero prefill only in pass 1, and Terra none has both slopes zero only in pass 1. Their remainders absorb any dropped token-processing terms.
The bootstrap resamples blocks of \NetBootstrapBlock{} consecutive time steps with replacement within each scenario (\NetBootstrapResamples{} resamples, seed \NetBootstrapSeed), refits the intercept and takes its 2.5th and 97.5th percentiles.
Each response is replayed with the estimate removed from its time before receipt, never by more than that time. At most \LunaNoneClampedRequests{} Luna none, \TerraNoneClampedRequests{} Terra none and \JevClampedRequests{} Jev requests are clamped, taking the maximum over the estimate and its two bounds. The estimate is fitted to the physical token/latency records and used on the 2\,s recording-cadence replay.

\begin{table}[t]
  \centering
  \small
  \setlength{\tabcolsep}{2.2pt}
  \begin{tabular}{lrlrrrr}
    \toprule
    Model & Rem. & Range & Pre. & Dec. & p50 & p95 \\
    \scriptsize & \scriptsize (s) & \scriptsize (s) & \scriptsize (ms/1k tok) & \scriptsize (ms/tok) & \scriptsize (s) & \scriptsize (s) \\
    \midrule
    \TabLatencyBody
    \bottomrule
  \end{tabular}
  \caption{\textbf{The fitted remainder is close to Jev's latency floor.} Rem.: mean non-token remainder (averaged \NetRangeLevel\% endpoints); Pre./Dec.: slopes; p50/p95: latency.}
  \label{tab:latency}
\end{table}

\paragraph{Estimates.}
Figure~\ref{fig:latency} and Table~\ref{tab:latency} give the fitted components and Table~\ref{tab:tokens} the token counts.
At low effort the fastest responses took \LunaLatencyFloor\,s (Luna), \TerraLatencyFloor\,s (Terra) and \AstraLatencyFloor\,s (Astra) before receipt, well above their remainders, consistent with decode time for even the shortest outputs; the shortest outputs have \LunaOutputTokMin, \TerraOutputTokMin{} and \AstraOutputTokMin{} tokens, so the intercept extrapolates below the observed range.
Without reasoning the fastest responses took \LunaNoneLatencyFloor\,s (Luna) and \TerraNoneLatencyFloor\,s (Terra), and Terra's fitted prefill slope is \TerraNonePrefillMsPerKTok\,ms per thousand input tokens.
Astra low reasons in only \AstraReasoningResponses{} of \AstraStates{} responses. Its fit assigns substantial latency to decode and a non-token remainder; fitted prefill is positive in passes 2 and 3.
Jev's fastest response took \JevLatencyFloor\,s, close to its remainder, so for Jev the remainder is essentially its latency floor; its fitted prefill slope is \JevPrefillMsPerKTok\,ms per thousand input tokens.

\begin{table}[tp]
  \centering
  \small
  \setlength{\tabcolsep}{4pt}
  \begin{tabular}{lrrrl}
    \toprule
    & & & \multicolumn{2}{c}{Remainder removed} \\
    \cmidrule(lr){4-5}
    Setting & Untimed & In force & Est. & Range \\
    \midrule
    \TabFamilyBody
    \bottomrule
  \end{tabular}
  \caption{\textbf{Removing the remainder leaves every low-effort GPT setting at least \NetRemovedLowEffortMinGap{} points below untimed accuracy in every family.} Family means at 2\,s (\%); Range averages per-pass \NetRangeLevel\% endpoints.}
  \label{tab:family}
\end{table}

\begin{table*}[tp]
  \centering
  \small
  \setlength{\tabcolsep}{3pt}
  \begin{tabular}{lrrrrrrrr}
    \toprule
    & \multicolumn{7}{c}{Normalized AUC} & In force at \\
    \cmidrule(lr){2-8}\cmidrule(lr){9-9}
    & Log & Linear & Log & Log & Log & Log & Log & \\
    Setting & 0.5--8\,s & 0.5--8\,s & 1--4\,s & 0.5--4\,s & 1--8\,s & 0.1--4\,s & 0.1--8\,s & 2\,s \\
    \midrule
    \TabAucSensitivityBody
    \midrule
    \TabLaterSensitivityBody
    \bottomrule
  \end{tabular}
  \caption{\textbf{Bounds and weighting reorder single settings, while the Jev and Perplexity hybrids stay ahead of every single setting.}
  Family weights as in Equation~\eqref{eq:macro} (\%); Log 0.5--8\,s is primary. Last three rows: freshest-source hybrids on the common nominal clock; $^\dagger$recorded later (Appendix~\ref{app:later}).}
  \label{tab:auc-sensitivity}
\end{table*}

\paragraph{Sensitivity.}
Table~\ref{tab:family} gives family scores with the remainder removed.
\emph{Segment-balanced in-force accuracy} weights every reference segment equally, $A_{\mathrm{seg}}=K^{-1}\sum_{k}|I_k|^{-1}\int_{I_k}\ind[d(t)=\yref(t)]\,dt$, so a brief change counts as much as a long stable span. Removing the non-token remainder moves segment-balanced scores in the same direction as time-weighted ones.
Segment-balanced in-force accuracy with the remainder removed is \LunaNetRemovedSeg\% (\LunaNetRemovedSegLow--\LunaNetRemovedSegHigh) for Luna, \TerraNetRemovedSeg\% (\TerraNetRemovedSegLow--\TerraNetRemovedSegHigh) for Terra, \AstraNetRemovedSeg\% (\AstraNetRemovedSegLow--\AstraNetRemovedSegHigh) for Astra and \JevNetRemovedSeg\% (\JevNetRemovedSegLow--\JevNetRemovedSegHigh) for Jev, against observed values of \LunaInForceSeg\%, \TerraInForceSeg\%, \AstraInForceSeg\% and \JevInForceSeg\%; without reasoning it is \LunaNoneNetRemovedSeg\% (\LunaNoneNetRemovedSegLow--\LunaNoneNetRemovedSegHigh) for Luna and \TerraNoneNetRemovedSeg\% (\TerraNoneNetRemovedSegLow--\TerraNoneNetRemovedSegHigh) for Terra, against \LunaNoneInForceSeg\% and \TerraNoneInForceSeg\%.
Displayed ranges average the endpoints of three per-pass estimator uncertainty ranges; they are not confidence intervals for the mean and say nothing about other locations or times of day.

% ---------------------------------------------------------------------------
\section{Log-AUC Definition and Sensitivity}
\label{app:intervals}

\begin{table}[tp]
  \centering
  \small
  \setlength{\tabcolsep}{3pt}
  \begin{tabular}{lrrrrr}
    \toprule
    Setting & $O_{\log}$ & $U_{\mathrm{cur},\log}$ & $\lambda_{\log}$ & $P$ & $S$ \\
    \midrule
    \TabAucDecompositionBody
    \bottomrule
  \end{tabular}
  \caption{\textbf{The timing--judgment decomposition carries through the log integral.}
  Percentages over 0.5--8\,s; $O_{\log}U_{\mathrm{cur},\log}+\lambda_{\log}=S$ exactly, and $P$ is the scenario-wise approximation.}
  \label{tab:auc-decomposition}
\end{table}

\paragraph{Why logarithmic weighting.}
We evaluate robustness across multiplicative changes in environment speed.
Let $W([a,b])$ be the weight of an interval range with positive endpoints. Equal multiplicative ranges should have equal weight, so $W([a,ra])=g(r)$ for every admissible $a$ and ratio $r>1$.
Additivity on adjacent ranges implies $g(rs)=g(r)+g(s)$.
With continuity and nonnegative weights, $g(r)=C\ln r$, and therefore $dW=C\,d\Delta/\Delta$.
Restricting this measure to $[a,b]$ and normalizing gives $p(\Delta)=1/[\Delta\ln(b/a)]$, the weighting used in Equation~\eqref{eq:macro}.
This is the multiplicative-invariance argument for scale measures \citep[Section~VII]{jaynes1968prior}, applied here as an evaluation principle.
It does not assert a logarithmic human perceptual law or an empirical distribution of deployment conditions.
Unit invariance alone would not justify this choice: normalized linear AUC is also unchanged by converting seconds to milliseconds.

\paragraph{Domain and interpretation.}
The interval $\Delta$ is spacing between evidence states; it is distinct from response latency, a user's latency budget, and a reference segment's dwell time, which can span several states.
We summarize performance over time-step intervals of 0.5--8\,s, spanning a fourfold increase and decrease in the rate of time steps relative to the 2\,s recording interval.
Logarithmic weighting assigns equal weight to each doubling interval and balances faster and slower conditions around the recording interval.
These bounds define a controlled evaluation domain; deployment-specific event rates can motivate other ranges.
Related interaction studies motivate investigating this scale but do not calibrate its endpoints: developer feedback in \citet[Section~7.2]{murali2024ai} discusses 300--500\,ms suggestions and an upper expectation of 1\,s; \citet{chen2017empirical} derive 0.6--2.7\,s bounds for LEGO guidance, and \citet{olguin2021impact} test injected latency up to 3\,s.
These concern response latency, not the event rates of our synthetic streams, and motivate the time scale without calibrating its endpoints.
The same rule and bounds apply to all settings, and Table~\ref{tab:auc-sensitivity} reports the alternative weights and bounds.
The same table reports all six combinations of lower bounds 0.1, 0.5 and 1\,s with upper bounds 4 and 8\,s, linear weighting over the primary domain, and in-force accuracy at the 2\,s recording cadence.

\paragraph{Replay and quadrature.}
Each interval evaluation retains the recorded answers and latency in seconds while scaling releases, the horizon and tick-valued policies together.
For example, support A's four-tick hold threshold represents 2\,s at an interval of 0.5\,s and 32\,s at 8\,s. These are controlled scenario variants, not new latency thresholds on an unchanged call.
The inner integral over scenario time is exact. For the outer integral, we use nested trapezoidal grids in $\ln\Delta$ (or $\Delta$ for linear weighting), starting with 128 subintervals and doubling until every integrated scenario metric changes by at most $10^{-5}$, or 0.001 percentage point.
Nonconvergence at 4096 subintervals is an error. The maximum final refinement change across the six hosted settings of Table~\ref{tab:main} and sensitivity conditions is \AucMaxRefinementChange{} points; this is numerical convergence, not statistical uncertainty. Appendix~\ref{app:openweight} additionally verifies self-hosted standalone results using exact crossing-based integration.
Scenario fractions are normalized before integration, then averaged within each family, across families and across passes equally. With common bounds and weights, integrating the aggregate curve equals averaging the family integrals.

\paragraph{Integrated decomposition.}
\label{app:auc-decomposition}
Let $\mathbb{E}_w$ denote the normalized log integral and the scenario and family averages in Equation~\eqref{eq:macro}.
Define $O_{\log}=\mathbb{E}_w[O]$, $C_{\log}=\mathbb{E}_w[O U_{\mathrm{cur}}]$, and $\lambda_{\log}=\mathbb{E}_w[\lambda]$.
Then $S=C_{\log}+\lambda_{\log}=O_{\log}U_{\mathrm{cur},\log}+\lambda_{\log}$, where $U_{\mathrm{cur},\log}=C_{\log}/O_{\log}$ when $O_{\log}>0$.
The expectation also averages passes. This is a ratio of integrated current-correct and current-source shares, not the unweighted average of $U_{\mathrm{cur}}$.
The approximation uses the scenario-wise product $P=\mathbb{E}_w[U_e O_e(\Delta)]$: each scenario's untimed accuracy is constant in $\Delta$.
It must not be replaced by the product of the overall untimed accuracy and overall oracle score.
Table~\ref{tab:auc-decomposition} reports these quantities; the fixed-interval decomposition is in Appendix~\ref{app:decomposition}.

% ---------------------------------------------------------------------------
\FloatBarrier
\section{Per-Scenario Results and Time-Step Interval}
\label{app:scenario}

\begin{table}[t]
  \centering
  \small
  \setlength{\tabcolsep}{3pt}
  \begin{tabular}{lrrrr}
    \toprule
    \multicolumn{5}{l}{\textit{\TerraRowLabel{} vs.\ \LunaRowLabel}} \\
    & Both & Luna only & Terra only & Neither \\
    \midrule
    \DebugLabel & \PairTerraLunaDebugBoth & \PairTerraLunaDebugLunaOnly & \PairTerraLunaDebugTerraOnly & \PairTerraLunaDebugNeither \\
    \AssemblyLabel & \PairTerraLunaAssemblyBoth & \PairTerraLunaAssemblyLunaOnly & \PairTerraLunaAssemblyTerraOnly & \PairTerraLunaAssemblyNeither \\
    \SupportLabel & \PairTerraLunaSupportBoth & \PairTerraLunaSupportLunaOnly & \PairTerraLunaSupportTerraOnly & \PairTerraLunaSupportNeither \\
    \PresenterLabel & \PairTerraLunaPresenterBoth & \PairTerraLunaPresenterLunaOnly & \PairTerraLunaPresenterTerraOnly & \PairTerraLunaPresenterNeither \\
    All & \PairTerraLunaBoth & \PairTerraLunaLunaOnly & \PairTerraLunaTerraOnly & \PairTerraLunaNeither \\
    \midrule
    \multicolumn{5}{l}{\textit{\AstraRowLabel{} vs.\ \TerraRowLabel}} \\
    & Both & Terra only & Astra only & Neither \\
    \midrule
    \DebugLabel & \PairAstraTerraDebugBoth & \PairAstraTerraDebugTerraOnly & \PairAstraTerraDebugAstraOnly & \PairAstraTerraDebugNeither \\
    \AssemblyLabel & \PairAstraTerraAssemblyBoth & \PairAstraTerraAssemblyTerraOnly & \PairAstraTerraAssemblyAstraOnly & \PairAstraTerraAssemblyNeither \\
    \SupportLabel & \PairAstraTerraSupportBoth & \PairAstraTerraSupportTerraOnly & \PairAstraTerraSupportAstraOnly & \PairAstraTerraSupportNeither \\
    \PresenterLabel & \PairAstraTerraPresenterBoth & \PairAstraTerraPresenterTerraOnly & \PairAstraTerraPresenterAstraOnly & \PairAstraTerraPresenterNeither \\
    All & \PairAstraTerraBoth & \PairAstraTerraTerraOnly & \PairAstraTerraAstraOnly & \PairAstraTerraNeither \\
    \midrule
    \multicolumn{5}{l}{\textit{\JevRowLabel{} vs.\ \LunaRowLabel}} \\
    & Both & Luna only & Jev only & Neither \\
    \midrule
    \DebugLabel & \PairJevLunaDebugBoth & \PairJevLunaDebugLunaOnly & \PairJevLunaDebugJevOnly & \PairJevLunaDebugNeither \\
    \AssemblyLabel & \PairJevLunaAssemblyBoth & \PairJevLunaAssemblyLunaOnly & \PairJevLunaAssemblyJevOnly & \PairJevLunaAssemblyNeither \\
    \SupportLabel & \PairJevLunaSupportBoth & \PairJevLunaSupportLunaOnly & \PairJevLunaSupportJevOnly & \PairJevLunaSupportNeither \\
    \PresenterLabel & \PairJevLunaPresenterBoth & \PairJevLunaPresenterLunaOnly & \PairJevLunaPresenterJevOnly & \PairJevLunaPresenterNeither \\
    All & \PairJevLunaBoth & \PairJevLunaLunaOnly & \PairJevLunaJevOnly & \PairJevLunaNeither \\
    \midrule
    \multicolumn{5}{l}{\textit{Luna: none vs.\ low}} \\
    & Both & Low only & None only & Neither \\
    \midrule
    \DebugLabel & \PairLunaNoneLunaDebugBoth & \PairLunaNoneLunaDebugLunaOnly & \PairLunaNoneLunaDebugLunaNoneOnly & \PairLunaNoneLunaDebugNeither \\
    \AssemblyLabel & \PairLunaNoneLunaAssemblyBoth & \PairLunaNoneLunaAssemblyLunaOnly & \PairLunaNoneLunaAssemblyLunaNoneOnly & \PairLunaNoneLunaAssemblyNeither \\
    \SupportLabel & \PairLunaNoneLunaSupportBoth & \PairLunaNoneLunaSupportLunaOnly & \PairLunaNoneLunaSupportLunaNoneOnly & \PairLunaNoneLunaSupportNeither \\
    \PresenterLabel & \PairLunaNoneLunaPresenterBoth & \PairLunaNoneLunaPresenterLunaOnly & \PairLunaNoneLunaPresenterLunaNoneOnly & \PairLunaNoneLunaPresenterNeither \\
    All & \PairLunaNoneLunaBoth & \PairLunaNoneLunaLunaOnly & \PairLunaNoneLunaLunaNoneOnly & \PairLunaNoneLunaNeither \\
    \midrule
    \multicolumn{5}{l}{\textit{Terra: none vs.\ low}} \\
    & Both & Low only & None only & Neither \\
    \midrule
    \DebugLabel & \PairTerraNoneTerraDebugBoth & \PairTerraNoneTerraDebugTerraOnly & \PairTerraNoneTerraDebugTerraNoneOnly & \PairTerraNoneTerraDebugNeither \\
    \AssemblyLabel & \PairTerraNoneTerraAssemblyBoth & \PairTerraNoneTerraAssemblyTerraOnly & \PairTerraNoneTerraAssemblyTerraNoneOnly & \PairTerraNoneTerraAssemblyNeither \\
    \SupportLabel & \PairTerraNoneTerraSupportBoth & \PairTerraNoneTerraSupportTerraOnly & \PairTerraNoneTerraSupportTerraNoneOnly & \PairTerraNoneTerraSupportNeither \\
    \PresenterLabel & \PairTerraNoneTerraPresenterBoth & \PairTerraNoneTerraPresenterTerraOnly & \PairTerraNoneTerraPresenterTerraNoneOnly & \PairTerraNoneTerraPresenterNeither \\
    All & \PairTerraNoneTerraBoth & \PairTerraNoneTerraTerraOnly & \PairTerraNoneTerraTerraNoneOnly & \PairTerraNoneTerraNeither \\
    \bottomrule
  \end{tabular}
  \caption{\textbf{In every pair, the less accurate setting is rarely correct where the other is wrong.} Untimed-correct counts over three passes, aligned by state and pass.}
  \label{tab:paired}
\end{table}

\begin{table}[t]
  \centering
  \small
  \setlength{\tabcolsep}{3.5pt}
  \begin{tabular}{lrrrrrrr}
    \toprule
    & & \multicolumn{2}{c}{Luna} & \multicolumn{2}{c}{Terra} & Astra & \\
    \cmidrule(lr){3-4}\cmidrule(lr){5-6}\cmidrule(lr){7-7}
    Scale & Interval (s) & low & none & low & none & low & Jev \\
    \midrule
    \TabPaceBody
    \bottomrule
  \end{tabular}
  \caption{\textbf{The order of the settings depends on the time-step interval.} In-force accuracy (\%) with every recorded latency scaled, equivalent to the interval shown; $\times$1 is the recording.}
  \label{tab:pace}
\end{table}

\begin{table*}[tp]
  \centering
  \small
  \setlength{\tabcolsep}{4pt}
  \begin{tabular}{llrrrrrrrrr}
    \toprule
    & & & & & & \multicolumn{4}{c}{Error time (s)} & \\
    \cmidrule(lr){7-10}
    Scenario & Model & Untimed & Log-AUC & In force & Seg.-bal. & None & Stale & Judg. & Comp. & p50 (s) \\
    \midrule
    \TabScenarioBody
    \bottomrule
  \end{tabular}
  \caption{\textbf{In every scenario Luna low and Terra low lose at least \GPTMinScenarioGap{} points from untimed to in-force accuracy, and Jev at most \JevMaxScenarioGap.} Log-AUC over 0.5--8\,s; in-force and segment-balanced accuracy (\%) and error time (seconds of 120\,s) at 2\,s; p50: median latency.}
  \label{tab:scenario}
\end{table*}

Unless labelled log-AUC, the diagnostics in this section use the 2\,s recording cadence.
Table~\ref{tab:family} gives family scores at 2\,s, and Table~\ref{tab:scenario} reports every scenario and setting.
In every scenario, untimed accuracy exceeds in-force accuracy by at least \GPTMinScenarioGap{} points for Luna low and Terra low (\AstraMinScenarioGap{} for Astra low) and by at most \JevMaxScenarioGap{} for Jev.
In \emph{\SupportBTitle}, for example, Terra low's composed decision is correct at \TerraSupportBUntimed\% of states, yet its in-force accuracy there is \TerraSupportBInForce\%, and most of its error time is stale (\TerraSupportBErrStaleSec\,s stale against \TerraSupportBErrIncorrectSec\,s incorrect for its source).
Jev's in-force accuracy follows its untimed accuracy closely in every scenario, and its time incorrect for its source is largest in the two debugging scenarios (\JevDebugAErrIncorrectSec\,s and \JevDebugBErrIncorrectSec\,s of 120\,s at the recording cadence).
Table~\ref{tab:paired} gives the paired counts of untimed composed decisions by family; without reasoning, Luna is correct alone at only \PairLunaNoneLunaLunaNoneOnly{} evaluations, against \PairLunaNoneLunaLunaOnly{} at low effort.
Composition matters as well: Jev's all-question exact match is \JevExactMatch\%, below its untimed decision accuracy of \JevUntimed\%, because at \JevInactiveOnlyErrors{} evaluations only answers the application ignores are wrong.

\paragraph{Time-step interval.}
Scaling every time in a scenario by the same factor leaves in-force accuracy unchanged, so multiplying every recorded latency by $c$ is equivalent to intervals of $2/c$ seconds with the same events.
Table~\ref{tab:pace} evaluates the recorded runs this way with the benchmark's scorer; each pass retains its answers, acceptance is recomputed on its scaled timeline, and results average passes.
Because every time a state shows is written in time steps, the states at another interval are identical, so the model-visible input is unchanged. This evaluation fixes the observed answers and latencies; it does not predict another stochastic draw or measure the service-load effects of another request rate.
The primary score integrates these evaluations over 0.5--8\,s; the fixed-interval values here retain a concrete timeline for diagnosis.

% ---------------------------------------------------------------------------
\FloatBarrier
\section{Judgment and Latency}
\label{app:decomposition}

This section reports fixed-interval diagnostics at the 2\,s recording cadence.

\begin{figure}[t]
  \centering
  \includegraphics[width=\linewidth]{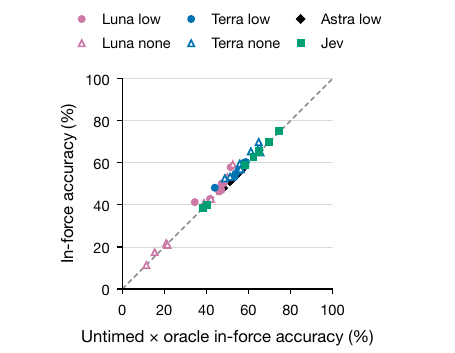}
  \caption{\textbf{In every scenario, in-force accuracy is within \ProductMaxDeviation{} points of untimed accuracy times the oracle in-force accuracy.}
  One point per setting and scenario at 2\,s; dashed: equality.}
  \label{fig:separability}
\end{figure}

\begin{table}[t]
  \centering
  \small
  \setlength{\tabcolsep}{3pt}
  \begin{tabular}{lrrrrrr}
    \toprule
    & Luna & Luna & Terra & Terra & Astra & \\
    & low & none & low & none & low & Jev \\
    \midrule
    \multicolumn{7}{l}{\textit{Share of observed time (\%)}} \\
    Judgment & \LunaErrJudgmentPct & \LunaNoneErrJudgmentPct & \TerraErrJudgmentPct & \TerraNoneErrJudgmentPct & \AstraErrJudgmentPct & \JevErrJudgmentPct \\
    Stale & \LunaErrStalePct & \LunaNoneErrStalePct & \TerraErrStalePct & \TerraNoneErrStalePct & \AstraErrStalePct & \JevErrStalePct \\
    Compound & \LunaErrCompoundPct & \LunaNoneErrCompoundPct & \TerraErrCompoundPct & \TerraNoneErrCompoundPct & \AstraErrCompoundPct & \JevErrCompoundPct \\
    No decision & \LunaErrNoDecisionPct & \LunaNoneErrNoDecisionPct & \TerraErrNoDecisionPct & \TerraNoneErrNoDecisionPct & \AstraErrNoDecisionPct & \JevErrNoDecisionPct \\
    Lucky & \LunaLuckyPct & \LunaNoneLuckyPct & \TerraLuckyPct & \TerraNoneLuckyPct & \AstraLuckyPct & \JevLuckyPct \\
    \midrule
    \multicolumn{7}{l}{\textit{Factors (\%)}} \\
    Untimed $U$ & \LunaUntimed & \LunaNoneUntimed & \TerraUntimed & \TerraNoneUntimed & \AstraUntimed & \JevUntimed \\
    Current $U_{\mathrm{cur}}$ & \LunaUCurrent & \LunaNoneUCurrent & \TerraUCurrent & \TerraNoneUCurrent & \AstraUCurrent & \JevUCurrent \\
    Oracle $O$ & \LunaTimingCeiling & \LunaNoneTimingCeiling & \TerraTimingCeiling & \TerraNoneTimingCeiling & \AstraTimingCeiling & \JevTimingCeiling \\
    $U\,O$ & \LunaProduct & \LunaNoneProduct & \TerraProduct & \TerraNoneProduct & \AstraProduct & \JevProduct \\
    In force $A$ & \LunaInForce & \LunaNoneInForce & \TerraInForce & \TerraNoneInForce & \AstraInForce & \JevInForce \\
    \bottomrule
  \end{tabular}
  \caption{\textbf{Error shares and factors at 2\,s: $O\,U_{\mathrm{cur}}+\lambda$ reproduces $A$, and $U\,O$ approximates it.} Equal-family means (\%); $U\,O$ averages per-scenario products, so it differs from the product of the rows above.}
  \label{tab:decomposition}
\end{table}

\begin{table*}[tp]
  \centering
  \small
  \begin{tabular}{lrrrrrrr}
    \toprule
    Reference schedule & Changes & \LunaRowLabel & \LunaNoneRowLabel & \TerraRowLabel & \TerraNoneRowLabel & \AstraRowLabel & \JevRowLabel \\
    \midrule
    \TabRegimeBody
    \bottomrule
  \end{tabular}
  \caption{\textbf{In projection, change frequency matters more than its arrangement.} Untimed accuracy times mean oracle accuracy (\%): \RegimeDraws{} latency resamples per pass on \RegimeSteps{} time steps of \ReleaseIntervalSec\,s; first row: \sdb's schedules.}
  \label{tab:regime}
\end{table*}

Table~\ref{tab:decomposition} gives the timing-by-judgment partition of \S\ref{sec:inforce} and the factors of Equation~\eqref{eq:factor} for every setting, and Figure~\ref{fig:separability} compares in-force accuracy with $U\,O$ scenario by scenario.
At the family level, $U_{\mathrm{cur}}$ divides macro-weighted current-correct time by macro-weighted current-source time, so that $O\,U_{\mathrm{cur}}+\lambda$ reproduces $A$.
The partition and the oracle come from the benchmark's scorer, which records for every instant whether the source is current and whether its answer was right for it; the share of time with a current source equals the oracle in-force accuracy of every run and scenario to machine precision, and the in-force accuracy at zero latency equals untimed accuracy up to the millisecond jitter of the recorded time steps.
Table~\ref{tab:regime} projects the latency factor onto other reference schedules, assuming $A\approx U\,O$ and an unchanged $U$: the oracle share depends only on the schedule and the latencies, so it is computed exactly for resampled latencies and multiplied by untimed accuracy.
Sparse, medium, uniform and dense schedules space their changes evenly; bursty packs \RegimeBurstyChanges{} changes into four bursts of one-step segments, long-tail dwell mixes one-step segments with a few long ones, and recurrent A-B-A alternates between two decisions.
On \sdb's own schedules the projection is within \RegimeLiteMaxDiff{} points of the replayed accuracy at 2\,s.

% ---------------------------------------------------------------------------
\section{Validity Audit}
\label{app:audit}

\paragraph{Procedure.}
The audit asked whether each stored reference answer follows from the public rules and evidence of its state.
For the original six IDE, assembly and support scenarios, two LLM agents worked blind, reading only the public states and questions and never the generator or reference code: one reimplemented the stated rules as an independent program and compared its output with the stored references, and the other derived the answers state by state.
This gives \AuditRederivationsPerScenarioWord{} re-derivations per scenario and \AuditRederivations{} in total.
Separate agents compared each family's reference code with its public rule text (\AuditCodeTextChecksWord{} audits) and checked the dataset's structure and deterministic regeneration (\AuditStructuralChecksWord{} check).
Findings were grouped by rule, and each of the \AuditFindingGroups{} groups was judged by \AuditLensesWord{} agents.
The \AuditReferenceFindingGroups{} groups about reference answers were judged by a literal reading of the rule text, a trace of the reference code on the state and a search for alternative readings a careful reader could adopt; the others, about documentation, narrative consistency or code paths, by reproduction, impact and counter-argument.
The released audit record lists each check's agreement and each grouped finding's three votes. It omits agent model versions, full prompts and independent evaluator programs, limiting reproduction of this audit; the benchmark reference and scoring procedures remain executable.

\paragraph{Outcome.}
Under its default reading of the rules, each of the \AuditRederivations{} re-derivations agreed with the stored reference at all \ReleasesPerScenario{} consumed decisions of its scenario; disagreements arose only under alternative readings that the agents evaluated.
Of the \AuditReferenceFindingGroups{} findings about reference answers, the adjudication rejected \AuditReferenceFindingsRejected{} and confirmed \AuditReferenceFindingsConfirmed, and none showed a wrong reference answer.
Among the confirmed findings, \AuditAmbiguitiesWord{} are rule ambiguities that could change a consumed decision, at \AuditAffectedStates{} consumed states in total.
In both assembly scenarios, the rules did not say whether an unreadable scan (a NO\_READ record) counts as a scan.
In one debugging scenario, the rules did not say whether \texttt{*} in a scope or ownership pattern matches across \texttt{/}.
The published rules state both readings, in \AuditClarifiedSentencesWord{} sentences, one per family, and both follow the reference: in debugging, \texttt{*} matches any characters including \texttt{/}; at the assembly station, a NO\_READ scan does not count as a scan in any rule.
The audited rule text lacked these two sentences; adding them changed no event, question or reference answer.
The audit also predates writing every time in time steps rather than seconds; that conversion divided every clock value, event time, threshold and deadline by the recorded interval and made the inclusive time boundaries of the support rules explicit, and all 360 reference answers in that audit are unchanged.

\paragraph{Agreement of a model outside authoring.}
Astra low was recorded after the benchmark was frozen and was not used to author or tune any scenario; ignoring latency, its composed decision matches the reference in \AstraUntimedStates{} of \AstraStates{} evaluations across three passes.
Pass 1 has one miss, at time step \AstraMissStep{} of \emph{\AstraMissScenarioTitle} (\AstraMissScenarioLabel).
The only inspection record there is a PASS at time step \AstraMissInspectionStep; the J2 screw was then replaced at time step \AstraMissScrewStep{} and run down to \AstraMissRundownNm\,N\,m at time step \AstraMissRundownStep, within its range of \AstraMissTorqueMinNm--\AstraMissTorqueMaxNm\,N\,m, endpoints included.
That rundown is the latest productive record, so the current stage is fasten, now complete; the PASS predates it, so inspection is an incomplete later stage.
The rules state that ``Incomplete future stages are not defects'' and that the advance target is the first incomplete stage, so the reference is route \emph{\AstraMissRefRoute} at stage \emph{\AstraMissRefStage} with next step \emph{\AstraMissRefNextStep}.
In that pass, Astra answered route \emph{\AstraMissRoute} with target \emph{\AstraMissTarget} and method \emph{\AstraMissMethod}, the rules' repair for a missing earlier stage, although inspection is now a future stage.
At time step \AstraMissPrevStep{} the state differs only by the clock and one status heartbeat, which the rules say changes no decision, and there Astra returned the reference.
The original audit had adjudicated this reading in one grouped finding of \AstraSituationStates{} states, time steps \AstraMissPrevStep{} and \AstraMissStep{} among them, and all \AuditLensesWord{} lenses upheld the reference; Astra returned the reference at the other \AstraSituationCorrect.

Pass 2 agrees at all \NumStates{} states. Pass 3 makes one analogous error at assembly A, time step \AstraMissStepPassThree{}, which belongs to the same grouped finding: after J1 rework, it again routes repair for the now-future inspection instead of advancing to it.

\paragraph{Scope.}
The audit tests agreement between public text, evidence and reference code; LLM agents performed every check, and one adjudication lens traces the reference itself, so zero confirmed errors bounds the error count only from below.
The presenter scenarios were added after this audit; their reference tests and executable agreement checks do not constitute a blind audit.
Astra's agreement, which also covers the presenter scenarios, is likewise LLM evidence: one model's answers agreeing with the reference on the same public states.
It does not test whether the rules describe good practice in a real deployment, and no human annotator adjudicated the references.

% ---------------------------------------------------------------------------
\FloatBarrier
\section{Figure 1 Window}
\label{app:window}
Figure~\ref{fig:trajectory} shows a \TrajWindowSec\,s window of presenter A at the 2\,s recording cadence, drawn from pass 1 of each displayed setting (Luna low, Terra low and Jev).
The candidates are presenter A's \NumWindows{} windows of \TrajWindowSec\,s that start at a time step and end within the horizon.
The displayed window is the most representative one: it minimizes the largest difference between a model's correct share in the window and its overall in-force accuracy.
That difference is \TrajWindowMaxDev{} points before rounding: the window shares and the selection target both use pass 1, rather than the three-pass means: \LunaTrajShare\% for Luna, \TerraTrajShare\% for Terra and \JevTrajShare\% for Jev.
The figure's right-hand column gives these correct shares; they describe the window, not the integrated scores of Table~\ref{tab:main}.
Every transcript change in the window is marked on the time axis at the time step that publishes it, open for a partial hypothesis and filled for a final text, and the changes at which the composed reference decision changes are quoted.
Only the reference fields that change in the window are drawn: the chair cue, the slide and the caption source.
The complete transcript remains in the artifact, and Figure~\ref{fig:errortime} and Table~\ref{tab:scenario} account for all error time.
The rule is implemented once and checked again whenever the figure is drawn.

% ---------------------------------------------------------------------------
\FloatBarrier
\section{Transition-Local Errors and Fast/Slow Composition}
\label{app:composition}

\begin{figure*}[t]
  \centering
  \includegraphics[width=\linewidth]{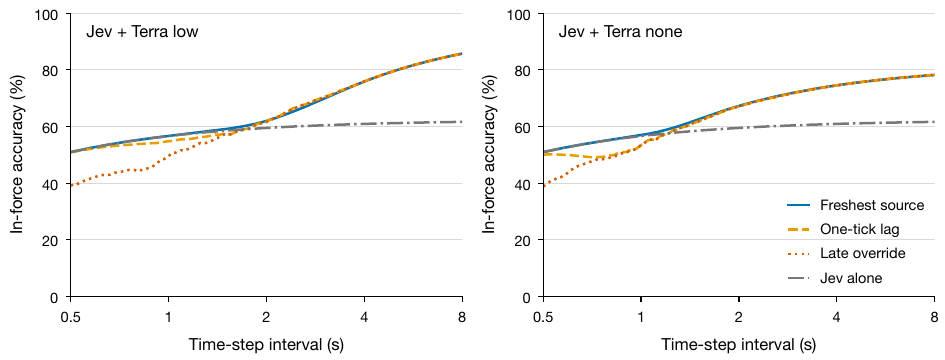}
  \caption{\textbf{Allowing old slow answers to overwrite newer fast answers can reverse whether composition helps.}
  Counterfactual Jev/Terra pairs with identical component answers and latencies, under three arbitration rules; the gray line is Jev alone.}
  \label{fig:arbitration}
\end{figure*}

\begin{table}[b]
  \centering
  \small
  \setlength{\tabcolsep}{4pt}
  \begin{tabular}{lrr}
    \toprule
    Setting & Near error (\%) & Far error (\%) \\
    \midrule
    \LunaRowLabel & \TrajLunaNearError & \TrajLunaFarError \\
    \LunaNoneRowLabel & \TrajLunaNoneNearError & \TrajLunaNoneFarError \\
    \TerraRowLabel & \TrajTerraNearError & \TrajTerraFarError \\
    \TerraNoneRowLabel & \TrajTerraNoneNearError & \TrajTerraNoneFarError \\
    \AstraRowLabel & \TrajAstraNearError & \TrajAstraFarError \\
    \JevRowLabel & \TrajJevNearError & \TrajJevFarError \\
    \bottomrule
  \end{tabular}
  \caption{\textbf{Untimed errors concentrate near reference changes for some settings.}
  Near: at most one tick from a change (\TrajLunaNearStates{} states); far: the remaining \TrajLunaFarStates{}.}
  \label{tab:transition-errors}
\end{table}

\paragraph{Localizing errors.}
Each state evaluation is assigned to the nearest actual change of the branch-composed reference, with ties assigned to the earlier change; initial availability is excluded.
The signed offset distinguishes before ($-1$), at ($0$) and after ($+1$) a change.
Table~\ref{tab:transition-errors} compares untimed error rates within one tick with rates farther away, counting every state once per pass.
The near neighborhood already contains \TrajLunaNearStates{} of 1,440 state evaluations, so raw counts alone exaggerate concentration.
Luna low makes \TrajLunaNearErrors{} errors near changes and \TrajLunaFarErrors{} farther away; Terra low makes \TrajTerraNearErrors{} and \TrajTerraFarErrors{}, respectively.
Luna none shows little difference, and Astra's two misses support no concentration claim.
These counts describe the fixed timelines; they establish neither a causal effect of a change nor significance across independent recordings.
The accompanying analysis provides every signed-offset bin, denominator and family breakdown.

\paragraph{Joint replay and arbitration.}
We pair Jev with each of the five GPT settings by matching pass indices, dispatch both at every time step, and average the three independently evaluated pairs.
Both components use common nominal releases $r_i=i\Delta$ and retain their recorded successful-attempt duration plus commit lag in seconds.
The maximum standalone difference from the original client-release clocks is \TrajNominalGap{} points at 2\,s; on each original clock, our independent replay exactly reproduces every setting's correctness and six-part time partition.
Within each component, an arrival must have a strictly newer source than any it has already delivered, including deliveries rejected by arbitration.
Fast answers never regress the active source; on equal sources, the slow answer has priority and can correct the fast answer.
For a slow answer with source $i$ and active source $j$, the three policies accept it when $i\geq j$ (\emph{freshest}), $i\geq j-1$ (\emph{one-tick lag}), or without a cross-component source restriction (\emph{late override}).
At simultaneous arrivals we process newer sources first, then slow before fast; arrivals at or beyond the horizon are excluded.
No policy consults a reference answer or the correctness of an output.

\begin{table}[t]
  \centering
  \small
  \setlength{\tabcolsep}{3pt}
  \begin{tabular}{llrrr}
    \toprule
    Slow setting & Arbitration & 1\,s & 2\,s & AUC \\
    \midrule
    \TabTrajectoryBody
    \bottomrule
  \end{tabular}
  \caption{\textbf{Fast/slow composition depends on when corrections remain in force.}
  Fast = Jev; in-force accuracy at fixed intervals and log-AUC over 0.5--8\,s (\%). All pairs and policies are reported.}
  \label{tab:composition}
\end{table}

\paragraph{What summaries omit.}
Table~\ref{tab:composition} and Figure~\ref{fig:arbitration} keep component answers, full latency distributions and the reference schedule identical across arbitration rules.
At 2\,s, Jev/Terra low improves on standalone Jev under freshest-source arbitration (\TrajTerraFreshestTwo\% versus \TrajJevTwo\%), and also improves under late override (\TrajTerraArrivalTwo\%) on average; pass 1 alone reverses from improvement to loss.
At 1\,s these policies give \TrajTerraFreshestOne\% and \TrajTerraArrivalOne\%.
The component summaries cannot select between these system designs because they omit the interleaving of provisional answers and corrections.
In \TrajExampleScenario, the reference changes \TrajExampleChange.
Jev's correct source-\TrajExampleFastSource{} answer arrives at \TrajExampleFastArrival\,s; Terra's answer for source~\TrajExampleOldSource, correct for that source, arrives at \TrajExampleOldArrival\,s.
Late override reinstates the old decision until \TrajExampleRestore\,s, adding \TrajExampleLoss\,s of stale time; freshest-source arbitration rejects it.
In pass 1, this is the longest contiguous stale loss from one Terra-low override of a correct Jev source across the eight scenarios.
Freshest-source arbitration accepts few Terra-low corrections: the slow component remains in force for only \TrajTerraFreshestSlowShare\% of time at 2\,s.
Using its final-answer accuracy as a proxy for system $U$, even with the exact policy-specific oracle, predicts \TrajTerraFreshestProxy\% rather than \TrajTerraFreshestTwo\%.
A two-output system has no unique per-state untimed answer representing both exposure durations; its current-source judgment must follow the component actually in force.
The exact identity in Equation~\eqref{eq:factor} continues to hold because selection is independent of answers.

\begin{table*}[t]
  \centering
  \small
  \begin{tabular}{lrrrrrr}
    \toprule
    & \multicolumn{5}{c}{Normalized log-AUC over 0.5--8\,s (\%)} & \\
    \cmidrule(lr){2-6}
    Setting & IDE & Assembly & Support & Presenter & Macro & Untimed \\
    \midrule
    \TabOpenweightFamilies
    \bottomrule
  \end{tabular}
  \caption{\textbf{Fast native readouts can remain limited by decision judgment.}
  Untimed is branch-composed decision accuracy (\%); standalone areas retain each recording's original release clock.}
  \label{tab:openweight-families}
\end{table*}

\begin{table*}[t]
  \centering
  \small
  \begin{tabular}{llrr}
    \toprule
    Setting & GPU & Mean p50 (s) & Mean p95 (s) \\
    \midrule
    \TabOpenweightLatency
    \bottomrule
  \end{tabular}
  \caption{\textbf{Latency depends on the complete decision runtime.}
  Successful-attempt durations include runtime queueing; the replay also retains measured postprocessing commit lag.}
  \label{tab:openweight-latency}
\end{table*}

\paragraph{Integration and scope.}
We report all five pairs and all three policies as an exploratory analysis of the existing recordings; no rule is tuned to references.
The components were recorded separately: joint service contention and recording-rate effects are unmeasured. Three-pair variation is descriptive; adjacent states are dependent.
Arbitration can produce a discontinuity when a slow answer moves from just before to just after a newer fast arrival.
We split the interval domain at every crossing of release, arrival and horizon times.
Between crossings, arrival order and acceptance are fixed and each normalized time fraction has form $c_0+c_1/\Delta$, whose log-weighted integral is analytical.
Two interior evaluations determine the coefficients and a third checks them; the displayed curve is sampled independently of this integral.
Family weights and bounds match Equation~\eqref{eq:macro}.

% ---------------------------------------------------------------------------
\FloatBarrier
\section{Self-Hosted Decisions and Composition}
\label{app:openweight}

\paragraph{Settings and measurement.}
The self-hosted settings are popular public decision-model checkpoints on Hugging Face that come with a native decision readout and run on one GPU; Qwen3.5-4B, read through its option logits, adds a general-model baseline.
We record \OwNumSettingsWord{} completed settings on the same frozen dataset, at the same 2\,s cadence with 32 workers and serial scenarios; each pass covers all \NumStates{} states once with zero failed model attempts.
Every setting runs on one RTX PRO 6000 Blackwell Server Edition GPU (96\,GB).
All use BF16 backbones and native decision readouts. The benchmark and model share a host, so latency includes tokenization, runtime queueing and complete-decision inference, with no internet round trip.
Download, initialization, input audits and three unrelated synthetic warmups precede measurement.
These are measurements of the configured deployments: differences with hosted APIs do not isolate architecture speed.

\paragraph{Model provenance.}
Table~\ref{tab:model-provenance} identifies the self-hosted checkpoints; their model cards declare Apache-2.0 licenses.
DJev uses the separately released runtime of \citet{mastracci2026djev}, revision \texttt{e5841cf4}, whose license is also Apache-2.0.
Full checkpoint and base-model revisions, runtime commits and weight hashes are retained in the accompanying recording manifests.

\begin{table*}[t]
  \centering
  \small
  \begin{tabular}{@{}>{\raggedright\arraybackslash}p{.31\textwidth}>{\raggedright\arraybackslash}p{.51\textwidth}l@{}}
    \toprule
    Setting and source & Hugging Face checkpoint identifier & Revision \\
    \midrule
    Laya English \citep{convai2026laya} & \texttt{convaiinnovations/laya} & \texttt{55cf4c4e} \\
    Laya typed-decisions & \texttt{convaiinnovations/laya-typed-decisions} & \texttt{1a793eb5} \\
    Laya multilingual & \texttt{convaiinnovations/laya-multilingual} & \texttt{e4e9ddf2} \\
    \midrule
    DJev / DiffusionGemma \citep{google2026diffusiongemma} & \texttt{google/diffusiongemma-26B-A4B-it} & \texttt{f7f5b7f5} \\
    \midrule
    Kev-4B \citep{palmer2026kev} & \texttt{jaredpalmer/kev-4b} & \texttt{139fdd94} \\
    Kev-9B & \texttt{jaredpalmer/kev-9b} & \texttt{b5d8c18e} \\
    Kev-27B & \texttt{jaredpalmer/kev-27b} & \texttt{ef78cc8a} \\
    \midrule
    Nimble-9B \citep{bespoke2026nimble} & \texttt{bespokelabs/Bespoke-Nimble-9B} & \texttt{bd792f44} \\
    Qwen direct logits \citep{qwen2026qwen35} & \texttt{Qwen/Qwen3.5-4B} & \texttt{851bf6e8} \\
    \bottomrule
  \end{tabular}
  \caption{\textbf{Self-hosted results use fixed public checkpoints.}
  Eight-character prefixes identify the recorded revisions. Kev-4B, Kev-9B and Nimble are adapters merged into their bases; Kev-27B supplies full weights.}
  \label{tab:model-provenance}
\end{table*}

\FloatBarrier
\begin{table*}[t]
  \centering
  \small
  \begin{tabular}{lrrrrrrr}
    \toprule
    & \multicolumn{5}{c}{Log-AUC over 0.5--8\,s (\%)} & \multicolumn{2}{c}{Untimed (\%)} \\
    \cmidrule(lr){2-6}\cmidrule(lr){7-8}
    Setting & Pass 1 & Pass 2 & Pass 3 & Mean & Range & Mean & Range \\
    \midrule
    \TabOpenweightRepeats
    \bottomrule
  \end{tabular}
  \caption{\textbf{Repeated passes leave the self-hosted comparison unchanged.}
  Three passes of each setting with the same GPU and configuration. Range is the maximum minus the minimum over passes.}
  \label{tab:openweight-repeats}
\end{table*}

\begin{table*}[t]
  \centering
  \small
  \begin{tabular}{lrrr}
    \toprule
    Provisional setting & Freshest & One-tick lag & Late override \\
    \midrule
    \TabOpenweightHybrids
    \bottomrule
  \end{tabular}
  \caption{\textbf{Correction rules and provisional judgment jointly determine composition.}
  All local settings paired with Terra none; mean (range) of normalized log-AUC over 0.5--8\,s (\%).}
  \label{tab:openweight-hybrids}
\end{table*}

\begin{figure*}[t]
  \centering
  \includegraphics[width=\linewidth]{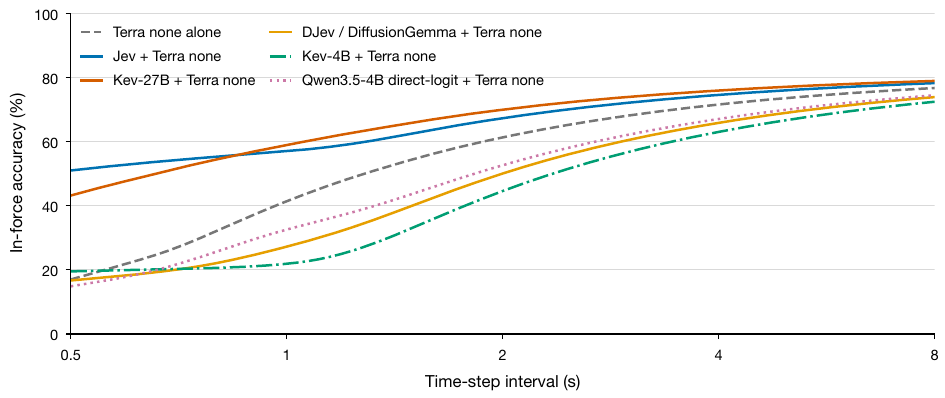}
  \caption{\textbf{A fast provisional component can lower accuracy across time-step intervals.}
  Common-clock replay with freshest-source arbitration and Terra none corrections; areas appear in Table~\ref{tab:openweight-hybrids}.}
  \label{fig:openweight-hybrids}
\end{figure*}

\paragraph{Native execution and coverage.}
Laya uses its released calibration policy with maximum sequence length 8192 and head length 1024; its input audits cover all \AnswersPerPass{} question rows, with no state or option truncation.
DJev uses DiffusionGemma 26B-A4B with a 128-token canvas, one denoising step, no thinking and upstream automatic noise sampling (up to four draws, threshold 0.1).
Kev uses its native pointer head and fused CUDA runtime, with CUDA graphs, cross-request prefix caching and date-fact augmentation disabled; Kev-4B and Kev-9B are LoRA adapters merged into their Qwen3.5 bases, Kev-27B is a full-weight checkpoint, and each serves the temperature its checkpoint carries.
Nimble uses the released merged adapter at temperature 1 and independently processes the full prompt for each of the six or seven fields, sequentially; its latency includes the entire request and runtime contention.
The Qwen3.5-4B direct-logit setting reads uncalibrated option logits from the pinned Qwen3.5-4B checkpoint, with thinking and generation disabled; no decision-trained checkpoint is used. This row evaluates that direct-logit configuration.
Their native input audits check every request for context overflow. No setting uses CPU offload or a truncated-input path.
All mappings preserve state fields, instructions, option descriptions, order and output identifiers.
The released recordings retain model revisions, execution configuration, source and weight hashes, input audits and native diagnostics where available.

\paragraph{Judgment and timing.}
Tables~\ref{tab:openweight-families} and~\ref{tab:openweight-latency} separate untimed decision accuracy from the measured deployment path.
DJev and Kev-4B obtain \OwDJevUntimed\% and \OwKevUntimed\% untimed accuracy, with log-AUC of \OwDJevAuc\% and \OwKevAuc\%; most of their loss is already present with timing ignored.
Larger Kev checkpoints raise untimed accuracy to \OwKevNineUntimed\% (Kev-9B) and \OwKevTwentySevenUntimed\% (Kev-27B); Kev-27B keeps \OwKevTwentySevenAuc\% log-AUC at mean per-pass p95 latency \OwKevTwentySevenTail\,s, so judgment still accounts for most of its loss.
Nimble instead falls from \OwNimbleUntimed\% untimed to \OwNimbleAuc\% log-AUC, with mean per-pass p95 latency \OwNimbleTail\,s, so timing is a substantial additional limitation of this runtime.
Exact composed decisions require all reference-active fields to agree; low composed accuracy is a result for this decision contract, not a claim about general language ability.
The accompanying analysis retains the six time classes by scenario and family for every setting.

\paragraph{Laya's component errors.}
Laya English, typed-decisions and multilingual achieve route accuracies of \LayaEnglishRoute\%, \LayaTypedRoute\% and \LayaMultilingualRoute\%, respectively.
Pooled accuracy over the reference-consumed fields, including the route, is \LayaEnglishGoldFields\%, \LayaTypedGoldFields\% and \LayaMultilingualGoldFields\%; each setting has \LayaEnglishStates{} state evaluations and \LayaEnglishFields{} consumed-field evaluations across three passes.
Errors therefore occur before the exact-match composition step. The recorded native input audits check state and option coverage, and the warmups check response format; the measurements leave the cause of these errors open. For reference, answering every question uniformly at random would pick the reference route in \BaselineUniformRoute\% of states and the composed reference decision in \BaselineUniform\%; Laya's route accuracies and its untimed accuracies (\OwLayaMultilingualUntimed--\OwLayaTypedUntimed\%) lie near these levels. The three settings are retained without outcome-based exclusion.

\paragraph{Repeated passes.}
We record three passes of each setting with the same GPU, configuration, checkpoints and dependency sets (Table~\ref{tab:openweight-repeats}). Self-hosted results and curves are equal means of independently evaluated passes; latency summaries average the per-pass quantiles.
Across the three passes, a setting's log-AUC varies by at most \OwRepeatMaxAucRange{} points and its untimed accuracy by at most \OwRepeatMaxUntimedRange{} points. Every pass orders the self-hosted settings the same way, although Kev-4B and DJev stay within one point of each other, and every Kev-27B pass stays above all three Jev passes (\OwKevTwentySevenAuc\% on average).
Differences between passes come from measured latency and from occasional changes in committed answers.

\paragraph{Counterfactual composition.}
We pair each of the three passes of every self-hosted setting with the corresponding Terra none pass, the GPT setting with the highest standalone log-AUC in the existing recordings, applying all three rules from Appendix~\ref{app:composition} unchanged.
Table~\ref{tab:openweight-hybrids} reports the mean and range for every pair and rule; none is excluded based on its outcome. Curves average the three replayed trajectories at each interval. The variation includes both separately recorded components; pass indices specify pairing, not simultaneous measurement.
Nimble occupies the provisional slot as a control even though its median latency exceeds Terra none's.
Under freshest-source arbitration, Jev plus Terra none reaches \OwJevTerraNoneAuc\%, while DJev, Kev-4B and Qwen direct-logit plus Terra none give \OwDJevFreshestAuc\%, \OwKevFreshestAuc\% and \OwQwenLogitsFreshestAuc\%, against \OwTerraNoneAuc\% for Terra none alone (Figure~\ref{fig:openweight-hybrids}).
Kev-27B plus Terra none reaches \OwKevTwentySevenFreshestAuc\%, above both Kev-27B alone (\OwKevTwentySevenAuc\%) and Jev plus Terra none; Kev-9B plus Terra none gives \OwKevNineFreshestAuc\%.
Across all six one-to-one matchings of component passes, the mean log-AUC of Kev-27B plus Terra none ranges from \PairingKevTwentySevenMin\% to \PairingKevTwentySevenMax\%, and Jev plus Terra none from \PairingJevTerraNoneMin\% to \PairingJevTerraNoneMax\%; their ordering is unchanged.
The acceptance rule favors source recency, so a wrong new provisional answer can replace an earlier answer that still matches the reference.
These interleaved exposure durations determine the composed score; neither faster service nor adding a correction guarantees improvement.

Standalone measurements retain their original release clocks; joint applications use common nominal releases and preserve every recorded successful-attempt duration plus commit lag.
The maximum standalone difference between these clocks is \OwNominalGap{} points at the fixed 2\,s interval, across all \OwTotalSettingsWord{} settings; replay on each original clock reproduces every correctness and six-part time partition.
All areas use the same exact crossing-based integration as Appendix~\ref{app:composition}.
The fixed-latency replay assumes latency is unchanged when the interval changes; components were recorded separately, and shared-resource contention and joint deployment performance remain unmeasured.

\begin{table*}[t]
  \centering
  \small
  \setlength{\tabcolsep}{4pt}
  \begin{tabular}{lrrrrrrrrr}
    \toprule
    & \multicolumn{5}{c}{Log-AUC over 0.5--8\,s (\%)} & & & \multicolumn{2}{c}{Error time (\%)} \\
    \cmidrule(lr){2-6}\cmidrule(lr){9-10}
    Setting & IDE & Assembly & Support & Presenter & Macro & Untimed & p50 (s) & Stale & Judgment \\
    \midrule
    \TabLaterStandalone
    \bottomrule
  \end{tabular}
  \caption{\textbf{Later hosted settings show the same two limits.}
  Means of three passes; Macro with sample SD in points; p50: mean per-pass median latency; error time as a share of log-weighted time.}
  \label{tab:later}
\end{table*}

\FloatBarrier
\section{Hosted Settings Recorded Later}
\label{app:later}

\paragraph{Settings and measurement.}
After the first passes of the six hosted settings of Table~\ref{tab:main}, we recorded \LaterNumSettingsWord{} further hosted settings between 2026-10-02 and 2026-10-08 (UTC), with three complete passes each under the same protocol: the frozen dataset, the 2\,s recording interval, 32 workers and the same client.
Five are native decision interfaces that answer the typed questions directly: Perplexity Decider v1 27B \citep{perplexity2026decider}, GLiDE \citep{fastino2026glide}, GPT-6-Luna through OpenAI's Decisions API \citep{openai2026decisions,openai2026luna6}, and Cloudflare's Clef and Clef Flash \citep{cloudflare2026clef,cloudflare2026clefflash}.
Claude Haiku 5.5 \citep{anthropic2026haiku} receives the instruction and structured-output schema of the GPT settings through the Messages API at low effort with the provider's default sampling, once with adaptive thinking and once with thinking disabled.
Requests time out after 20\,s as in Appendix~\ref{app:protocol}, except 30\,s for Perplexity and 300\,s for GLiDE.
Retries follow Appendix~\ref{app:protocol}; Perplexity's adapter also treats HTTP 5xx responses as connection errors, and within the same attempt budget the adapters may retry, after a shared Retry-After cooldown, HTTP 429 (Perplexity, GLiDE, Claude, Decisions), 425 and 503 (GLiDE) and 5xx (Claude, Decisions), which never occurred.
Failed attempts were recovered by retry: \LaterPerplexityFailedAttempts{} in Perplexity's first pass (timeouts and HTTP 503), \LaterClefFlashFailedAttempts{} for Clef Flash and \LaterHaikuFailedAttempts{} for Claude Haiku 5.5 low (timeouts); every pass returned a valid response for all \NumStates{} states.
The Decisions API refused one question at the same two presenter states in every pass; a refused question commits to no option and counts as wrong wherever the composed decision uses it, which never happened. An earlier pass, stopped at the first refusal before the adapter recorded refusals, is not evaluated.
Original-release and common nominal clocks differ by at most \LaterNominalGap{} points at 2\,s for these settings.
Wity-1 (one pass per reasoning mode, at 16 workers), three single-scenario smoke tests of Claude Haiku 5.5 at default and low effort run before its passes, and eight self-hosted settings recorded with three passes each on other GPU hosts are not included.

\begin{table}[t]
  \centering
  \small
  \setlength{\tabcolsep}{3pt}
  \begin{tabular}{llrrr}
    \toprule
    Fast & Corrector & Freshest & Lag $\leq 1$ & Override \\
    \midrule
    \TabLaterSystems
    \bottomrule
  \end{tabular}
  \caption{\textbf{Perplexity raises every corrector; the other fast components lower every corrector stronger than Luna none.}
  Log-AUC over 0.5--8\,s (\%), matched passes, common nominal clock; GPT-6-Luna via the Decisions API.}
  \label{tab:later-systems}
\end{table}

\paragraph{Results.}
Table~\ref{tab:later} shows the two limits of \S\ref{sec:results}.
With adaptive thinking, Claude Haiku 5.5 is \LaterHaikuUntimed\% correct untimed but stale for \LaterHaikuStale\% of log-weighted time; with thinking disabled it answers in \LaterHaikuOffMedian\,s at the median, but its untimed accuracy falls to \LaterHaikuOffUntimed\%.
The fast decision interfaces lose mostly to judgment, whereas GLiDE, a decision interface with a median latency of \LaterGlideMedian\,s, is stale for \LaterGlideStale\% of log-weighted time: the partition, not the interface type, identifies the limit.
Perplexity Decider has the highest single-setting log-AUC (\LaterPerplexityAuc\%) and leads Jev in every family except presenter control, yet its judgment errors still occupy \LaterPerplexityJudgment\% of log-weighted time.

\paragraph{Composition.}
Extending the hosted rule of Appendix~\ref{app:composition}, which pairs Jev with the five GPT settings, each decision interface whose mean per-pass median latency is below the 2\,s recording interval, \LaterNumFastWord{} in all, is paired with the five GPT settings under the three arbitration rules, with matching pass indices (Table~\ref{tab:later-systems}).
With freshest-source arbitration, Perplexity improves on both components with every corrector except Luna none, and with Astra low reaches \LaterBestAuc\%, the highest evaluated mean.
The other three interfaces stay below every corrector stronger than Luna none under every rule; for them, the one-tick-lag rule, which lets a slightly older correction replace a newer fast answer, gives a higher log-AUC than freshest-source arbitration in most pairs, whereas for Perplexity freshest-source arbitration is best with every corrector.
As for the self-hosted fast components (Table~\ref{tab:openweight-hybrids}), the best rule depends on the fast component's judgment.

\FloatBarrier

\section{Reproducibility}
\label{app:repro}

\paragraph{Accompanying materials.}
The accompanying materials, provided as supplementary material, contain the frozen dataset (\texttt{data/lite/v1}, dataset hash beginning \texttt{\DatasetHashShort}), the task generators and executable references, the runner and scorer, all recorded passes with their complete event logs, the audit record, and the scripts that regenerate the numerical results and figures in this paper.
Code and data are released under the MIT license and are intended for evaluating decision components; the synthetic scenarios and their rules are not a policy for any real deployment.
The pass-1 event logs have SHA-256 digests beginning \texttt{\LunaEventsHash} and \texttt{\LunaNoneEventsHash} (Luna, low and none), \texttt{\TerraEventsHash} and \texttt{\TerraNoneEventsHash} (Terra), \texttt{\AstraEventsHash} (Astra) and \texttt{\JevEventsHash} (Jev).
The three-pass analysis manifest lists all eighteen hosted and twenty-seven self-hosted recordings with their event hashes, and a second manifest lists the three passes of each setting in Appendix~\ref{app:later}. The analysis scripts regenerate the measured results from these logs without any model call, verify agreement with the published analyses, and check that the executable reference reproduces all \NumStates{} stored answers. The independent audit remains limited to its original six-scenario scope.
A new pass needs access to the evaluated models: an OpenAI API key for the GPT models and access to \JevProvider's service for Jev; the settings of Appendix~\ref{app:later} also need access to OpenAI's Decisions API and to Perplexity, Fastino, Cloudflare and Anthropic.
At the list prices of Table~\ref{tab:tokens}, OpenAI bills reasoning tokens as output and \JevProvider{} bills input tokens only; the eighteen hosted passes cost \CostTotalUSD{} USD in total; the passes of Appendix~\ref{app:later} are not costed. Actual billed charges can differ. The twenty-seven self-hosted measurement windows total \OwMeasuredHours{} GPU-host hours, excluding setup, warmups, input audits and other experiments; a complete compute total was not retained.
The self-hosted evidence contains recorded predictions and timing, frozen scenarios and configurations, input audits and pinned native revisions; the release includes no model weights. Recomputing all standalone and composition results requires neither GPUs nor service credentials.
Hosted models may change or be retired; the logs record the requested and served model identifiers.

\paragraph{Commands.}
A new pass makes paid model requests and requires a fresh output directory; the remaining commands make no network request.
\begin{quote}
\scriptsize
\begin{verbatim}
# one pass (--model gpt-5.6-luna|terra
#   --effort low|none; --model gpt-6-astra
#   --effort low)
uv run python -m streamdecisionbench.lite run \
  --data data/lite/v1 --out runs/<new-dir> \
  --model gpt-5.6-terra --effort low \
  --max-attempts 5 --retry-delay 0.5
uv run python -m streamdecisionbench.lite run \
  --data data/lite/v1 --out runs/<new-dir> \
  --provider typesafe --model jev-latest \
  --max-attempts 5 --retry-delay 0.5
# re-evaluate and report from recorded events
uv run python -m streamdecisionbench.lite score \
  --run runs/<dir>
uv run python scripts/lite/lite_report.py \
  --run runs/<dir> --out docs/lite/results/<name>
# verify all manuscript passes; regenerate
# diagnostics, numbers, tables and figures
uv run --group paper \
  python paper/analysis/lite_repeated.py
\end{verbatim}
\end{quote}

\FloatBarrier
\newpage
\section{Story Behind the Paper}
\label{app:story}

This work began while we were building a presenter tool, similar to the presenter family in SDB, and exploring newly introduced decision models such as Jev.
On social media, we saw open-weight alternatives emerge, with informal comparisons suggesting small differences between models.
In our own attempts to use them in the tool, however, we noticed clear differences in the decisions they produced.
This discrepancy prompted us to examine what the benchmarks in those comparisons actually evaluated; many focused on static classification.
For tasks organized around semantic similarity, embedding-based matching seemed a natural baseline, which led us to ask what additional capability a specialized decision model should provide.
Our application required instruction following over changing state: given the current evidence and rules, the model had to select the application's next control state.
We came to think of this as instruction-conditioned next-state prediction, where ``next state'' refers to the control state the application should adopt.
The choice depends on the instructions and the evolving context, not just on which option is semantically closest to the input.
Starting from the presenter use case, we extended this evaluation idea to other application families, forming StreamDecisionBench.
Our goal was to make model comparisons informative for real applications by examining the decisions a component would keep in force as its input changes, helping developers choose components for the behavior their applications require.

\end{document}